\documentclass[twoside,11pt]{article}
\usepackage[preprint]{jmlr2e}

\makeatletter
\@for\jmlrenv:=proof,example,theorem,lemma,proposition,remark,corollary,definition,conjecture,axiom\do{%
  \expandafter\let\csname\jmlrenv\endcsname\relax
  \expandafter\let\csname end\jmlrenv\endcsname\relax
  \expandafter\let\csname c@\jmlrenv\endcsname\relax
  \expandafter\let\csname the\jmlrenv\endcsname\relax
  \expandafter\let\csname theH\jmlrenv\endcsname\relax
}
\makeatother

\usepackage{float}
\usepackage{makecell}
\usepackage{pgfplots}
\pgfplotsset{compat=1.18}
\usepackage{snu-causal-thesis-2026}

\newcommand{\jmlrtheoremspacing}[1]{%
  \AtBeginEnvironment{#1}{\linespread{1}\selectfont}%
}
\forcsvlist{\jmlrtheoremspacing}{proof,theorem,lemma,proposition,corollary,claim,definition,example,conjecture,remark}

\let\jmlrsection\section
\let\section\subsection
\let\subsection\subsubsection
\let\subsubsection\paragraph

\hypersetup{
  pdftitle={Generalizing Causal Decision-Making: Markov Equivalence, Transportability and Counterfactuals},
  pdfauthor={Min Woo Park and Sanghack Lee},
  pdfkeywords={Causal Inference, Causal Decision-Making, Markov Equivalence, Transportability, Counterfactual Inference, Multi-Armed Bandits}
}

\ShortHeadings{Generalizing Causal Decision-Making}{Park and Lee}
\firstpageno{1}

\newif\ifjmlrpreprint
\jmlrpreprinttrue
\newcommand{\workname}{paper}
\newcommand{\unitname}{section}

\newcommand{\unitnames}{sections}

\begin{document}
\allowdisplaybreaks

\title{Generalizing Causal Decision-Making:\\
Markov Equivalence, Transportability and Counterfactuals}
\author{\name Min Woo Park \email alsdn0110@snu.ac.kr\\
  \name Sanghack Lee \email sanghack@snu.ac.kr\\
  \addr Seoul National University\\
  Seoul, Republic of Korea}
\maketitle
\thispagestyle{plain}

\begin{tikzpicture}[remember picture,overlay]
        \node[anchor=north east,yshift=-0.3in,xshift=-0.3in] (box) at (current page.north east) {
         \begin{minipage}[t!]{0.30\textwidth}
            \begin{flushright} {\fontfamily{qhv}\selectfont
                TECHNICAL REPORT\\
                First version: Nov, 2026\\
                Last version: Nov, 2026
            }
            \end{flushright}
          \end{minipage}
        };
\end{tikzpicture}

\begin{abstract}
    Structural causal bandits use causal knowledge to guide exploration when reward mechanisms are unknown. This \workname{} extends the approach to partially known causal structure, information from other environments, and realizable counterfactual actions. First, we develop graphical criteria for minimal intervention sets under Markov equivalence and characterize possibly-optimal minimal intervention sets for maximal ancestral graphs. A search algorithm finds sets that are possibly optimal in at least one graph represented by a partial ancestral graph. Second, we derive target reward bounds from source distributions through partial transportability, accounting for non-manipulable variables. Dominance relations provide further bounds for action pruning and restricted exploration, with a regret guarantee under valid prior bounds. Third, we formulate counterfactual structural causal bandits over actions supported by specified experimental capabilities. Graphical characterizations of minimality and possible optimality yield algorithms that remove equivalent or dominated choices while preserving an optimal action. Experiments with specified structural causal models examine how these reductions and transferred bounds affect learning. Together, these results clarify how causal knowledge and experimental capabilities determine which actions merit exploration and guide learning within the retained action space. 
\end{abstract}

\jmlrsection{Introduction}\label{chap:intro}
Sequential decision-making requires an agent to choose actions while learning their consequences. In a multi-armed bandit (MAB), the agent repeatedly selects an action and observes its reward, balancing the value of gathering information against the value of choosing a promising action \citep{thompson1933likelihood,robbins1952some,lai1985asymptotically,auer2002finite,lattimore2020bandit}. An unstructured bandit treats the reward distributions of different actions as unrelated. In many applications, however, actions intervene on variables in a common causal system, and their effects are therefore connected.

For example, a physician choosing treatments must consider how they affect the same underlying biological mechanisms. An education policy may change several factors that influence later employment, such as access to training and acquired skills. In both cases, treating every combination of interventions as an unrelated action can require unnecessary exploration. Causal knowledge helps determine which interventions can affect the outcome and which choices can be excluded without losing an optimal decision \citep{bareinboim2024introduction}.

Structural causal models (SCMs) describe such systems through mechanisms that generate each variable from its causes and unobserved inputs \citep{pearl:2k}. An intervention replaces selected mechanisms while leaving the others intact. This representation connects the effects of different actions and supports questions that a list of unrelated reward distributions does not answer. Causal decision-making uses these relationships to inform learning and action selection \citep{zhang2019near,zhang2020designing,zhang2020causal,kumor2021sequential,zhang2022online,ruan2024causal,zhang2024eligibility,richens2024robust,richens2025general,mueller2023personalized,bareinboim2024introduction}.

Early work on causal bandits demonstrated how causal information can improve exploration \citep{bareinboim2015bandits,lattimore2016causal,pmlr-v70-forney17a}. Subsequent studies considered additional information or restrictions, including access to observational and interventional distributions and assumptions on causal mechanisms \citep{zhang2017transfer,lu2020regret,bilodeau2022adaptively,de2022causal,feng2023combinatorial,varici2023causal}. \textbf{Structural causal bandits} (SCBs) \citep{lee2018structural,lee2019structural} provide the starting point of this \workname{}. In an SCB, an action specifies both the variables to intervene on and the values assigned to them. 


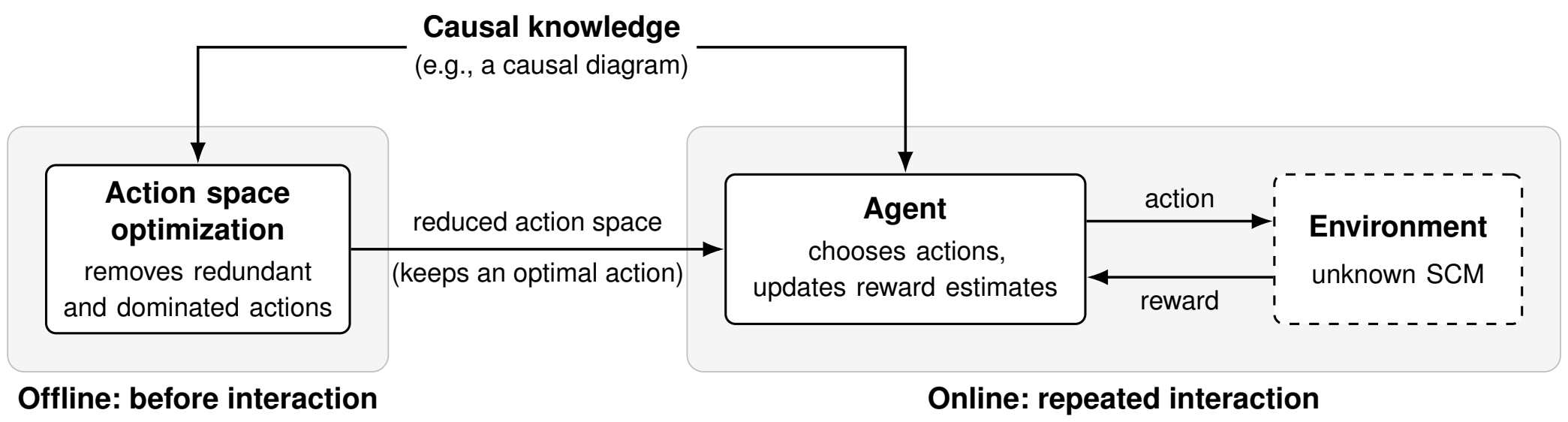
\begin{figure}[t]
    \centering
    \begingroup
    \renewcommand{\sfdefault}{phv}
    \renewcommand{\small}{\fontsize{9}{11}\selectfont}
    \renewcommand{\footnotesize}{\fontsize{8}{9.5}\selectfont}
    \linespread{1}\selectfont

    \pgfsetlayers{back,background,main}

    \resizebox{0.9\linewidth}{!}{%
    \begin{tikzpicture}[
        font=\sffamily\small,
        >={Latex[length=2.3mm,width=1.7mm]},
        box/.style={
            draw, rounded corners=3pt, line width=0.7pt,
            align=center, inner sep=5pt, fill=white
        },
        flow/.style={->, line width=0.8pt},
        note/.style={
            font=\sffamily\footnotesize, align=center
        },
        panel/.style={
            draw=black!25, fill=black!4,
            rounded corners=5pt, line width=0.4pt,
            inner sep=0pt
        },
        stage/.style={
            font=\sffamily\small\bfseries,
            anchor=north, inner sep=4pt
        }
        ]
        
        \node (algo) [
            box, text width=29mm, minimum height=16mm
        ] {
            \textbf{Action space optimization}\\[3pt]
            {\footnotesize removes redundant\\and dominated actions}
        };

        \node (agent) [
            box, right=40mm of algo,
            text width=35mm, minimum height=16mm
        ] {
            \textbf{Agent}\\[3pt]
            {\footnotesize chooses actions,\\updates reward estimates}
        };

        \node (env) [
            box, dashed, right=20mm of agent,
            text width=23mm, minimum height=16mm
        ] {
            \textbf{Environment}\\[3pt]
            {\footnotesize unknown SCM}
        };

        \node (knowledge) [align=center, inner sep=2pt]
            at ($(algo.north)!0.5!(agent.north) + (0,13mm)$)
        {
            \textbf{Causal knowledge}\\
            {\footnotesize (e.g., a causal diagram)}
        };

        \draw[flow] (knowledge.west) -| (algo.north);
        \draw[flow] (knowledge.east) -| (agent.north);

        \draw[flow] (algo.east)
            -- node[note, above] {reduced action space}
               node[note, below] {(keeps an optimal action)}
            (agent.west);

        \draw[flow] ([yshift=3mm]agent.east)
            -- node[note, above] {action}
            ([yshift=3mm]env.west);

        \draw[flow] ([yshift=-3mm]env.west)
            -- node[note, below] {reward}
            ([yshift=-3mm]agent.east);

        \coordinate (top) at ($(algo.north) + (0,4mm)$);
        \coordinate (bot) at ($(algo.south) + (0,-4mm)$);

        \begin{scope}[on background layer]
            \node (off) [
                panel,
                fit={(algo.west |- top) (algo.east |- bot)},
                inner xsep=4mm
            ] {};

            \node (on) [
                panel,
                fit={(agent.west |- top) (env.east |- bot)},
                inner xsep=4mm
            ] {};
        \end{scope}

        \node [stage] at (off.south)
            {Offline: before interaction};

        \node [stage] at (on.south)
            {Online: repeated interaction};

    \end{tikzpicture}%
    }
    \endgroup
    \caption{\small Structural causal bandits. Offline, the agent uses the causal knowledge (e.g., causal diagram) to reduce the action space while retaining at least one optimal action. Online, it selects interventions from this reduced space and uses rewards from the environment's SCM to guide subsequent choices.}
    \label{fig:intro_causal_bandit}
\end{figure}

\Cref{fig:intro_causal_bandit} separates the SCB framework into two stages. \textbf{Offline action space optimization} uses graphical equivalence and dominance relations in the known causal structure to remove unnecessary interventions while retaining at least one optimal action in every compatible SCM. This optimization requires neither the mechanisms nor the reward means to be known. During \textbf{online learning}, the agent selects an intervention from the reduced action space, observes the reward generated by the environment's SCM, and uses the accumulated observations to guide subsequent choices. 

\paragraph{Contributions.} The central question of this \workname{} is which actions remain worth exploring given the agent's causal knowledge and experimental capabilities. We extend SCBs in three complementary directions, each addressing a different limitation of the starting formulation. 
\begin{itemize}
    \item[$\triangleright$] \textbf{Markov equivalence.} Observational data generally do not determine a unique causal diagram. \Cref{chap:scb mec} develops action space optimizations when the agent knows only a Markov equivalence class, represented by a partial ancestral graph. We characterize minimal intervention sets for ancestral graphs, distinguish whether their minimality can be witnessed by one compatible model, and establish a graphical characterization of possibly-optimal minimal intervention sets for maximal ancestral graphs. A search algorithm identifies sets that are possibly optimal in at least one represented maximal ancestral graph \citep{park2025structural}.

    \item[$\triangleright$] \textbf{Transportability.} Data collected elsewhere can inform a target decision problem, but differences between environments and restrictions on manipulation limit what can be transferred. \Cref{chap:scb tr} first reviews the existing formulation of structural causal bandits with non-manipulable variables. It then develops partial transportability and dominance bounds for target rewards, combines information from available source experiments, and uses the resulting bounds to prune actions and restrict exploration. A regret guarantee describes the behavior of the proposed policy under valid prior bounds \citep{park2025transportability}.

    \item[$\triangleright$] \textbf{Counterfactuals.} Some systems allow different mechanisms to receive different versions of an input on the same unit. \Cref{chap:ctf scb} uses realizability to specify the counterfactual actions supported by these capabilities. We extend minimality and possible optimality to this action space, derive graphical characterizations, and construct a reduced collection that preserves an optimal action \citep{park2026counterfactual}. The required capabilities are stated explicitly, because a mathematically defined counterfactual need not be experimentally available.
\end{itemize}

\paragraph{Relation to earlier publications.}
This \workname{} brings together and revises our studies of structural causal bandits under Markov equivalence \citep{park2025structural}, transportability \citep{park2025transportability}, and counterfactual interventions \citep{park2026counterfactual}. The revisions include corrected graphical statements, algorithms, and proofs, together with new or recomputed experiments. The opening paragraphs of \Cref{chap:scb mec,chap:scb tr,chap:ctf scb} identify the changes relative to each publication; the statements and experimental results presented here are the revised versions.

\paragraph{Organization.} \Cref{chap:logical foundation} introduces the causal and graphical foundations. \Cref{chap:scb} develops the bandit objective and action space optimization. \Cref{chap:scb mec,chap:scb tr,chap:ctf scb} present the three extensions, and \Cref{chap:conclusion} brings the results together. The appendices provide supporting proofs and experimental model specifications.

\jmlrsection{Causal and Graphical Preliminaries}\label{chap:logical foundation}
This \unitname{} introduces the notation used throughout the \workname{}, the semantic framework represented by structural causal models (SCMs), and the necessary background results on which our work builds. A large part of this \unitname{} follows \citet{jaber2022thesis,bareinboim2022pearl,Zhang2023towards} and \citet{bareinboim2024introduction}. 

The SCM and intervention notation supports every subsequent \unitname{}. Readers can proceed to \Cref{chap:scb} after the material on SCMs and causal diagrams. The material on maximal ancestral graphs (MAGs), partial ancestral graphs (PAGs), and Fast Causal Inference (FCI) prepares the partial-structure setting in \Cref{chap:scb mec}. 

\section{Basic Notation}

Following standard conventions, we use a capital letter, such as $X$, to represent a variable, with its corresponding lowercase letter, $x$, denoting a realization of the variable. Boldface letters are used to represent sets of variables or values, denoted by $\mf{X}$ or $\mf{x}$. The domain $\Omega_X$ is the set of possible values of $X$, and $\Omega_{\mf{X}} = \bigtimes_{X \in \mf{X}}\Omega_X$. For a discrete distribution $P$, its support consists of the assignments in the domain that have positive probability; it may vary across environments and interventions. We consistently use $P(\mf{x})$ as an abbreviation for $P(\mf{X} = \mf{x})$. We denote by $\indicate\{\mf{X} = \mf{x}\}$ the indicator function. Two values $\mf{x}$ and $\mf{z}$ are \textit{consistent} if they share common values for $\mf{X} \cap \mf{Z}$. We denote by $\mf{x} \setminus \mf{Z}$ the value of $\mf{X} \setminus \mf{Z}$ consistent with $\mf{x}$ and by $\mf{x}\cap \mf{Z}$ the restriction of $\mf{x}$ to the variables in $\mf{Z}$. 

A graph $\mc{G} = \langle \mf{V}, \mf{E} \rangle$ is defined over a set of nodes (vertices) $\mf{V}$ and a set of edges $\mf{E}$. A graph is directed if all its edges are directed ($\to$). We use mixed graphs whose edges may be directed or bidirected ($\bedge$); either type may be absent. 

\section{Structural Causal Models}

A structural causal model (SCM) is a versatile and expressive framework that provides a principled way to represent and reason about causal relationships. Its generality makes it applicable across a wide range of domains. We use an SCM as the semantic framework to represent the environment in which the agent operates. 

Throughout this \workname{}, we consider recursive SCMs, so that each exogenous realization determines a unique value of every endogenous variable. Unless stated otherwise, all variables and mechanisms considered below satisfy this convention. 
\begin{definition}[Structural Causal Model \citep{pearl:2k}]
    An SCM $\mc{M}$ is a quadruple $\langle \mf{U}, \mf{V}, \mc{F}, P(\mf{U})\rangle$, where: 
    \begin{itemize}
        \item $\mf{U}$ is a set of exogenous variables that are determined by factors outside the model and follow a joint distribution $P(\mf{U})$;
        \item $\mf{V}$ is a set of endogenous variables representing observed variables inside the model;
        \item $\mc{F} = \{f_V\}_{V \in \mf{V}}$ is a set of functions (mechanisms) such that $v = f_V(\mf{pa}_V,\mf{u}_V)$ where $\mf{Pa}_V \subseteq \mf{V} \setminus \{V\}$ and $\mf{U}_V \subseteq \mf{U}$;
        \item $P(\mf{U})$ is a probability distribution over the exogenous variables. 
    \end{itemize}
\end{definition}
\section{Pearl's Causal Hierarchy}
\begin{table}[t]
\centering
\resizebox{\textwidth}{!}{
\begin{tabular}{lcclll}
\toprule
&\makecell{\bf Layer\\\bf (Symbolic)} & \makecell{\bf Typical\\\bf Activity} & \makecell{\bf Typical\\ \bf Question} & \makecell{\bf Example} & \makecell{\bf Machine\\\bf Learning} \\
\midrule

$\mc{L}_1$
&
\makecell{Observational \\ $P(y\mid x)$}
&
Seeing
&
\makecell{What is? \\ How would seeing $X$ \\ change my belief in $Y$?}
&
\makecell{What does a symptom tell us \\ about the disease?}
&
\makecell{Supervised \slash \\ Unsupervised\\Learning}
\\

\midrule

$\mc{L}_2$
&
\makecell{Interventional \\ $P(y \mid do(x))$}
&
Doing
&
\makecell{What if? \\ What if I do $X$?}
&
\makecell{If I take aspirin, \\ will my headache be cured?}
&
\makecell{Multi-armed\\bandits}
\\

\midrule
$\mc{L}_3$
&
\makecell{Counterfactual \\ $P(y_x\mid x',y')$}
&
Imagining
&
\makecell{Why? \\ What if I had acted differently?}
&
\makecell{Was it the aspirin that \\ stopped my headache?}
&
\\

\bottomrule
\end{tabular}
}
\caption{\small Pearl’s Causal Hierarchy \citep{bareinboim2022pearl}. }
\label{tab:pch}
\end{table}

An SCM induces all quantities within \textit{Pearl's Causal Hierarchy} (PCH) \citep{pearl2018book,bareinboim2022pearl}. The three layers of the PCH correspond to distinct regimes of reasoning about an environment: \textit{seeing}, \textit{doing}, and \textit{imagining} (\Cref{tab:pch}). 

The first layer $\mc{L}_1$ represents \textit{observational} distributions, such as $P(Y \mid x)$; the second layer $\mc{L}_2$ represents \textit{interventional} distributions, e.g., $P(Y\mid do(x))$, using the $do$-operator \citep{pearl:95}. The highest layer $\mc{L}_3$ represents \textit{counterfactual} distributions, such as $P(Y_x \mid X = x')$ \citep{heckman1985alternative,heckman1986alternative}: the distribution of $Y$ under intervention $do(X=x)$ among units whose observed treatment is $x'$. Such distributions underlie quantities such as the effect of treatment on the treated \citep{heckman1985alternative,heckman1986alternative}. 

\subsection{Observational Distributions}

An agent perceives the deployed environment through its sensors by ``seeing'', which can be formalized through the first layer of the PCH ($\mc{L}_1$). 
\begin{definition}[Observational distributions \citep{bareinboim2022pearl}]
    An SCM $\mc{M} = \langle \mf{U}, \mf{V}, \mc{F}, P(\mf{U}) \rangle$ defines a joint probability distribution $P(\mf{V})$ such that for each $\mf{Y} \subseteq \mf{V}$:
    \begin{align}
        P(\mf{y}) = \sum_{\mf{u}} \indicate\{\mf{Y}(\mf{u}) = \mf{y}\} P(\mf{u}),
    \end{align}
    where $\mf{Y}(\mf{u})$ is the solution for $\mf{Y}$ after evaluating $\mc{F}$ with $\mf{U} = \mf{u}$\footnote{We write sums for expectations over exogenous variables when these variables are discrete; for general exogenous distributions, the corresponding expressions are understood as integrals with respect to $P(\mf{U})$. The same sum-to-integral convention applies to expectations over endogenous variables. For non-discrete endogenous variables, distributions are defined analogously on measurable events rather than only on point events. }. 
\end{definition}
Therefore, observational distributions are computed using the following procedure: 
\begin{itemize}
    \item[1.]  For each exogenous realization $\mf{U} = \mf{u}$\footnote{One can view the realization as a situation or a unit \citep{pearl2016causal}.}, the environment evaluates functions in $\mc{F}$ according to a topological order (i.e., any variable on the left-hand side is evaluated after those on the right-hand side); and

    \item[2.] The probability $P(\mf{u})$ is accumulated for each realization $\mf{U} =\mf{u}$ consistent with $\mf{Y} = \mf{y}$. 
\end{itemize}

\begin{example}[{\bf Clinical system} \citep{Zhang2023towards,bareinboim2024introduction}]\label{ex:clinical system}
    In a clinic treating patients with a chronic disease, physicians must choose a treatment. There are two possible treatments, $X=0$ and $X=1$. The agent's goal is to identify the treatment that maximizes the probability of patient recovery ($Y$). Consider an SCM $\mc{M}$ that models this system:
    \begin{align}
        \mc{M}
    = \begin{cases}
            \mf{U} =   \{U\}\\
            \mf{V} = \{X,Y\} \\
            \mc{F} = 
            \begin{cases}
                f_X(u) = \indicate\{u < 0.8\}\\
                f_Y(x,u) = \indicate\{u < 0.4 - \eta \cdot x\}
            \end{cases}\\
            P(\mf{U}) = \{U \thicksim \Unif(0,1)\},
    \end{cases}\label{eq:clinical trial scm}
    \end{align}
    where the coefficient $\eta$ is a real number bounded in $(0, 0.4)$. 

    In words, the physician prescribes treatment $X = 0$ for senior patients ($U \geq 0.8$); otherwise, the alternative treatment $X = 1$ is prescribed. The recovery $Y$ depends on the treatment $X$ and the patient's age $U$. 

    By passively observing patient outcomes and collecting data, the agent can estimate the mean recovery rate as follows: 
    \begin{align}
        P(Y = 1) &= P(X = 0, Y = 1) + P(X = 1, Y = 1)\\
        &= P(X = 0, Y = 1) + P(Y = 1 \mid X = 1)P(X=1)\\
        &=\underbrace{P(U \geq 0.8, U < 0.4 - \eta \cdot X)}_{=0} + P(Y =1\mid X=1)P(U < 0.8)\\
        &= 0.8 \cdot P(Y =1\mid X=1). 
    \end{align}
    The agent may also estimate the recovery rate among patients who received each treatment. Among patients who received treatment $X = 0$, the recovery rate is: 
    \begin{align}
        P(Y = 1 \mid X = 0) & = P(U < 0.4 - \eta \cdot X \mid U \geq 0.8) \\
        &\leq P(U < 0.4 \mid U \geq 0.8)\\
        & = 0.
    \end{align}
    Similarly, the recovery rate for $X=1$ is: 
    \begin{align}
        P(Y = 1 \mid X = 1) & = P(U < 0.4 - \eta \cdot X \mid U < 0.8) \\
        & = 0.5 - 1.25 \eta.
    \end{align}
    Therefore, the mean recovery rate is given by:
    \begin{align}
        P(Y = 1) &= 0.8 \cdot P(Y =1\mid X=1)\\
        & = 0.4 - \eta. 
    \end{align}
    Since the coefficient $\eta$ lies in $(0,0.4)$, we can conclude:
    \begin{align}
        P(Y = 1 \mid X = 0) < P(Y = 1 \mid X = 1). \label{eq: clinic example observational result} 
    \end{align}
    This result from an observational study suggests that treatment $X=1$ is better than treatment $X=0$. 
\end{example}

\subsection{Interventional Distributions}

The agent interacts with the environment through the notion of ``doing''; that is, by intervening (acting) on the SCM (environment), thereby modifying the original system. We formally define the environment that arises under such interventions using submodels. 
\begin{definition}[Submodel \citep{bareinboim2024introduction}]
    Let $\mc{M} = \langle \mf{U}, \mf{V}, \mc{F}, P(\mf{U})\rangle$ be an SCM, and let $\mf{x}$ be a particular realization of a set of variables  $\mf{X} \subseteq \mf{V}$. A \textit{submodel} $\mc{M}_{\mf{x}}$ of $\mc{M}$ is an SCM $\langle \mf{U}, \mf{V}, \mc{F}_{\mf{x}}, P(\mf{U})\rangle$ where: 
    \begin{align}
        \mc{F}_{\mf{x}} = \{f_V : V \in \mf{V} \setminus \mf{X} \} 
        \cup \{f_X = x : X \in \mf{X}, x \text{~is the value of~} X \text{~in~} \mf{x}\}. 
    \end{align}
\end{definition}
In words, an intervention is modeled by replacing the original mechanisms associated with variables $\mf{X}$ with constant values $\mf{x}$, as represented by the \textit{do-operator}, i.e., $do(\mf{X} = \mf{x})$ (for short, $do(\mf{x})$). The outcome under the intervention is called the \textit{potential response} \citep{pearl:2k}. 
\begin{definition}[Potential response \citep{bareinboim2022pearl}]
    Given an SCM $\mc{M} = \langle \mf{U}, \mf{V}, \mc{F}, P(\mf{U})\rangle$, let $\mf{X}$ and $\mf{Y}$ be two sets of variables in $\mf{V}$, and let $\mf{u}$ be a realization of $\mf{U}$. The potential response $\mf{Y}_\mf{x}(\mf{u})$ is defined as the solution for $\mf{Y}$ obtained from the system $\mc{F}_{\mf{x}}$ with $\mf{U} = \mf{u}$ in $\mc{M}$. 
\end{definition}
For a fixed intervention $\mf{x}$, the potential response becomes a random variable by evaluating it at the random exogenous variables: $\mf{Y}_{\mf{x}}=\mf{Y}_{\mf{x}}(\mf{U})$, with $\mf{U} \sim P(\mf{U})$\footnote{The notation $\mf{Y}_{\mf{x}}(\mf{u})$ is borrowed from the potential-outcome framework \citep{neyman1923application,rubin1974estimating}.}. Its distribution is induced by $P(\mf{U})$; averaging a scalar potential response gives its expectation. We next define the resulting interventional distribution. 
\begin{definition}[Interventional distribution \citep{bareinboim2022pearl}]
    An SCM $\mc{M} = \langle \mf{U}, \mf{V}, \mc{F}, P(\mf{U})\rangle$ induces a collection of joint probability distributions over $\mf{V}$, one for each intervention $\mf{x}$. For each $\mf{Y}\subseteq \mf{V}$:
    \begin{align}
        P(\mf{y}_\mf{x}) = \sum_{\mf{u}} \indicate\{\mf{Y}_{\mf{x}}(\mf{u}) = \mf{y}\}P(\mf{u}).
    \end{align}
\end{definition}
Hence, the interventional distribution can be evaluated as follows:
\begin{itemize}
    \item[1.] Replace the mechanism of each $X \in \mf{X}$ with its value in $\mf{x}$; the resulting $\mc{F}_{\mf{x}}$ defines the submodel $\mc{M}_{\mf{x}}$. 

    \item[2.] For each realization $\mf{U} =\mf{u}$, the environment evaluates functions in $\mc{F}_\mf{x}$ according to a topological order.

    \item[3.] The probability $P(\mf{u})$ is accumulated for each realization $\mf{U} =\mf{u}$ consistent with $\mf{Y} = \mf{y}$ in the submodel $\mc{M}_{\mf{x}}$. 
\end{itemize}
In words, the interventional distributions induced by $do(\mf{x})$ in an SCM $\mc{M}$ are distributions over the endogenous variables in the submodel $\mc{M}_{\mf{x}}$. We denote the interventional distribution over $\mf{Y}$ induced by $do(\mf{x})$ interchangeably as $P_{\mf{x}}(\mf{y})$, $P(\mf{y}_{\mf{x}})$, or $P(\mf{y} \mid do(\mf{x}))$\footnote{We mainly use do-operator notation for $\mc{L}_2$ quantities.}. 
\begin{example}[\textbf{Continued}]
    Suppose that the patient's age $U$ is unavailable to the agent, so the agent observes only the treatment $X$ and the outcome $Y$. Consider an intervention $do(X = x)$ that sets treatment $X$ to a constant $x \in \{0,1\}$. The submodel $\mc{M}_x$ induced by $do(X = x)$ is given by: 
    \begin{align}
        \mc{M}_x
    = \begin{cases}
            \mf{U} =   \{U\}\\
            \mf{V} = \{X,Y\} \\
            \mc{F}_x = 
            \begin{cases}
                f_X = x\\
                f_Y(x,u) = \indicate\{u < 0.4 - \eta \cdot x\}
            \end{cases}\\
            P(\mf{U}) = \{U \thicksim \Unif(0,1)\}. 
    \end{cases}\label{eq:clinical trial submodel}
    \end{align}
    Evaluating the patient's recovery $Y$ in the submodel $\mc{M}_x$ gives:
    \begin{align}
        P_{X=0}(Y = 1) 
        &= P(Y = 1 \mid do(X=0))\\
        &= P(U < 0.4 - \eta \cdot X \mid do(X=0))\\
        &= P(U < 0.4)\\
        &= 0.4. \label{eq:clinical trial prob do x=0}
    \end{align}
    Similarly, the patient's recovery rate under treatment $do(X = 1)$ is:
    \begin{align}
        P_{X=1}(Y = 1) 
        &= P(Y = 1 \mid do(X=1))\\
        &= P(U < 0.4 - \eta \cdot X \mid do(X=1))\\
        &= P(U < 0.4 - \eta)\\
        &= 0.4 - \eta. \label{eq:clinical trial prob do x=1}
    \end{align}
    Given that $\eta > 0$, we can conclude: 
    \begin{align}
        P_{X=0}(Y = 1)  > P_{X=1}(Y = 1). 
    \end{align}
    Thus, treatment $X=0$ yields a higher recovery rate under intervention. This reverses the observational ranking in \Cref{eq: clinic example observational result}: patients receiving the two treatments differ in $U$, whereas an intervention fixes the treatment while preserving the population distribution of $U$. The example therefore distinguishes observational association from the effect of intervening on treatment.
\end{example}

\section{Causal Diagrams}

While SCMs provide precise semantics for the distributions underlying the PCH, they are typically not directly observable by the learning agent in practice. As a result, reasoning about the system requires explicit assumptions about the underlying causal model. These assumptions are commonly represented using a \textit{causal diagram}. 
\begin{definition}[Causal diagram \citep{pearl:95,bareinboim2022pearl}]\label{def: causal diagram}
    Consider an SCM $\mc{M} = \langle \mf{U}, \mf{V}, \mc{F}, P(\mf{U}) \rangle$. Then $\mc{G}$ is said to be a \textit{causal diagram} of $\mc{M}$ if it is constructed as follows: 
    \begin{itemize}
        \item[1.] Add a vertex for every endogenous variable in $\mf{V}$.
        \item[2.] Add a directed edge ($V \to W$) if $V$ appears as an argument of $f_W \in \mc{F}$. 
        \item[3.] Add a bidirected edge ($V \bedge W$) if the corresponding $\mf{U}_{V}, \mf{U}_W \subseteq \mf{U}$ are dependent\footnote{That is, there is no partition $\overline{\mf{U}}_V, \overline{\mf{U}}_W$ over $\mf{U}$ such that (1) $\mf{U}_V \subseteq \overline{\mf{U}}_V$ and $\mf{U}_W \subseteq \overline{\mf{U}}_W$; and (2) independence $\overline{\mf{U}}_V \Perp \overline{\mf{U}}_W$ holds in $P(\mf{U})$. Furthermore, we say that $V$ and $W$ are \textit{confounded}, and the underlying cause is referred to as an \textit{unobserved confounder}. } or the corresponding functions $f_V$,~$f_W$ share some $U_{VW} \in \mf{U}$ as an argument. 
    \end{itemize}
\end{definition}
We focus on recursive SCMs, i.e., \textit{acyclic} causal diagrams. Intuitively, there is a directed edge from an endogenous variable $V$ to $W$ ($V \to W$) whenever $W$ ``listens to'' $V$ in determining
its value. Similarly, a bidirected edge between $V$ and $W$ ($V \bedge W$) indicates the presence of shared unobserved factors that influence both variables. 

While an SCM $\mc{M} = \langle \mf{U}, \mf{V}, \mc{F}, P(\mf{U}) \rangle$ contains explicit quantitative information about the structural mechanisms $\mc{F}$ and the exogenous distribution $P(\mf{U})$, a causal diagram $\mc{G}$ encodes only qualitative information about which variables \textit{may}\footnote{A directed edge $V \to W$ encodes the possibility of a causal mechanism by which $W$ listens to $V$, but not its necessity. A similar interpretation holds for bidirected edges. More formally, the graphical constraints (assumptions) are not represented by the edges present in the diagram, but rather by the missing edges. In particular, a missing edge indicates that one variable cannot be an argument of the other, or that the corresponding exogenous inputs are independent. In summary, edges represent possible dependencies, while missing edges encode independence assumptions.} serve as inputs to the functions $f_V \in \mc{F}$ and how the exogenous variations are related. Hence, the diagram abstracts away the specific forms of the mechanisms $\mc{F}$ and the distribution $P(\mf{U})$, retaining only qualitative information about their arguments and independence structure.

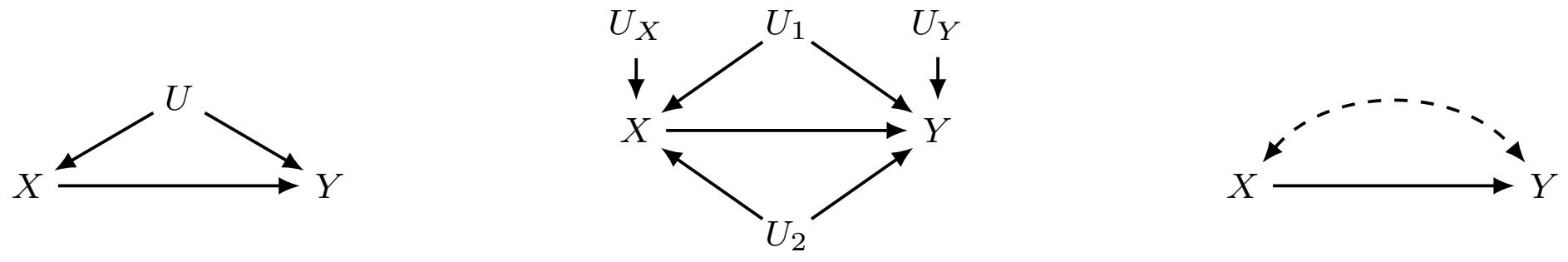
\begin{figure}[t]
    \hfill
    \begin{minipage}[b]{0.26\textwidth}\centering  
    \subfloat[]{
        \begin{tikzpicture}[x=1.2cm,y=1cm,>={Latex[width=1.4mm,length=1.7mm]},
        line width=0.25mm,
        R/.style={draw = none,circle,inner sep=0mm, minimum size=4.5mm},
        U/.style={draw = dashed,circle,inner sep=0mm, minimum size=4mm}
        ]
        \node[R] (U) at (1,0.7) {$U$};
        \node[R] (X) at (0,0) {$X$};
        \node[R] (Y) at (2,0) {$Y$};

        \draw[->] (X) -- (Y);
        \draw[->] (U) -- (X);
        \draw[->] (U) -- (Y);

        \node[R] () at (1,-0.4) {};
        \end{tikzpicture}\label{fig:foundation_causal_diagram_a}}
    \end{minipage}
    \hfill
    \begin{minipage}[b]{0.26\textwidth}\centering  
    \subfloat[]{
        \begin{tikzpicture}[x=1.2cm,y=1.2cm,>={Latex[width=1.4mm,length=1.7mm]},
        line width=0.25mm,
        R/.style={draw = none,circle,inner sep=0mm, minimum size=4.5mm}
        ]
        \node[R] (U1) at (1,0.7) {$U_1$};
        \node[R] (U2) at (1,-0.7) {$U_2$};
        \node[R] (Ux) at (0,0.7) {$U_X$};
        \node[R] (Uy) at (2,0.7) {$U_Y$};
        
        \node[R] (X) at (0,0) {$X$};
        \node[R] (Y) at (2,0) {$Y$};

        \draw[->] (Ux) -- (X);
        \draw[->] (Uy) -- (Y);
        
        \draw[->] (X) -- (Y);
        
        \draw[->] (U1) -- (X);
        \draw[->] (U1) -- (Y);

        \draw[->] (U2) -- (X);
        \draw[->] (U2) -- (Y);
        
        \end{tikzpicture}\label{fig:foundation_causal_diagram_b}}
    \end{minipage}
    \hfill
    \begin{minipage}[b]{0.26\textwidth}\centering  
    \subfloat[]{
        \begin{tikzpicture}[x=1.2cm,y=1cm,>={Latex[width=1.4mm,length=1.7mm]},
        line width=0.25mm,
        R/.style={draw = none,circle,inner sep=0mm, minimum size=4.5mm}]
        \node[R] (X) at (0,0) {$X$};
        \node[R] (Y) at (2,0) {$Y$};

        \draw[->] (X) -- (Y);
        \draw[dashed,<->] (X) to[bend left = 50] (Y);
        \node[R] () at (1,-0.4) {};
        \end{tikzpicture}\label{fig:foundation_causal_diagram_c}}
    \end{minipage}
    \hfill\null
    \hfill
    \caption{\small (a,~b) Exogenous-explicit diagrams, where each directed edge indicates the input-output relation of the mechanisms in $\mc{F}$; (c) causal diagram $\mc{G}$.} 
    \label{fig:foundation_causal_diagram}
\end{figure}
\begin{example}[\textbf{Continued}] Let us consider the running example $\mc{M}$. \Cref{fig:foundation_causal_diagram_a} illustrates an abstraction of $\mc{M}$. The causal mechanisms $\mc{F}$ are structural functions of the form: 
\begin{align}
    \mc{F} = 
        \begin{cases}
            x = f_X(u)\\
            y = f_Y(u,x). 
        \end{cases}
\end{align}
To apply the graphical construction, we start the process by examining each of the endogenous variables $\mf{V} = \{X,Y\}$ and adding them as nodes in the causal diagram $\mc{G}$. The corresponding diagram is illustrated in \Cref{fig:foundation_causal_diagram_c}\footnote{In causal diagrams, bidirected edges are drawn as dashed lines for clarity. }. 
\end{example}
For concreteness, we consider an additional instance of the following form:
\begin{align}
        \mc{M}'
    = \begin{cases}
            \mf{U} =   \{U_X, U_Y, U_1, U_2\}\\
            \mf{V} = \{X,Y\} \\
            \mc{F} = 
            \begin{cases}
                x = f_X(u_X, u_1, u_2) \\
                y = f_Y(x, u_Y, u_1, u_2)
            \end{cases}\\
            P(\mf{U}) = \{U \thicksim \Unif(0,1): \forall U \in \mf{U}\}. 
    \end{cases}
\end{align}
Take the exogenous variables to be mutually independent, so the displayed uniform marginals specify their joint distribution. In this setting, $\mc{M}'$ induces the same causal diagram $\mc{G}$, where $\mf{U}_X = \{U_X, U_1, U_2\}$ and $\mf{U}_Y = \{U_Y, U_1, U_2\}$. The diagram gives a compact summary of the possible directed and confounding relationships between $X$ and $Y$, while omitting the identities and number of exogenous variables and the quantitative forms of the mechanisms. 

\subsection{Graphical Notation}

If there is an edge between two vertices $X$ and $Y$ in $\mc{G}$, we say that the two vertices are \textit{adjacent} in $\mc{G}$, denoted by $Y \in \adj(X)_\mc{G}$ or $X \in \adj(Y)_\mc{G}$. A \textit{path} is a sequence of distinct vertices, with a specified edge connecting each consecutive pair. When two vertices are joined by both a directed and a bidirected edge, the path specifies which edge is traversed. If a path consists of directed edges with the same orientation, we say the path is \textit{directed}. A non-endpoint vertex is a \textit{collider} on a path when both adjacent path edges have arrowheads at that vertex, as in $A\to B\gets C$. Otherwise, it is a non-collider on that path. A path is a \textit{collider path} if all its non-endpoint vertices are colliders. 

We use kinship notation for graphical relationships: parents ($\Pa$), children ($\Ch$), ancestors ($\An$), and descendants ($\De$). 
\begin{itemize}
    \item $\Pa(X)_{\mc{G}}$ (parents of $X$ in $\mc{G}$): the set $\{Z_i\}$ such that there exists a directed edge $Z_i \to X$ in $\mc{G}$. 

    \item $\Ch(X)_{\mc{G}}$ (children of $X$ in $\mc{G}$): the set $\{Z_i\}$ such that there exists a directed edge $X \to Z_i$ in $\mc{G}$. 

    \item $\An(X)_{\mc{G}}$ (ancestors of $X$ in $\mc{G}$): the set $\{Z_i\}$ such that there exists a directed path $Z_i \to \cdots  \to X$ in $\mc{G}$. 

    \item $\De(X)_{\mc{G}}$ (descendants of $X$ in $\mc{G}$): the set $\{Z_i\}$ such that there exists a directed path $X \to \cdots  \to Z_i$ in $\mc{G}$. 
\end{itemize}
Ancestors and descendants include the variable itself. For a set of variables, we define $\An(\mf{X})_{\mc{G}} = \bigcup_{X\in\mf{X}} \An(X)_{\mc{G}}$, and similarly for other relationships.

The $\mf{X}$-lower-manipulation of $\mc{G}$ removes all directed edges out of variables in $\mf{X}$, denoted as $\mc{G}_{\underline{\mf{X}}}$, while the $\mf{X}$-upper-manipulation of $\mc{G}$ removes all edges with an arrowhead at a variable in $\mf{X}$, including bidirected edges, denoted as $\mc{G}_{\overline{\mf{X}}}$. We denote the set of variables in $\mc{G}$ by $\mf{V}(\mc{G})$. A subgraph $\mc{G}[\mf{V}']$, where $\mf{V}' \subseteq \mf{V}(\mc{G})$, is defined as a vertex-induced subgraph in which all edges among the vertices in $\mf{V}'$ are preserved. We define $\mc{G} \backslash \mf{X}$ as $\mc{G}[\mf{V}(\mc{G}) \setminus \mf{X}]$ for $\mf{X} \subseteq \mf{V}(\mc{G})$. 

\begin{definition}[Latent projection \citep{verma1990equivalence}]\label{def: latent projection}
    The \textit{latent projection} of a causal diagram $\mc{G}$ over $\mf{V}$ onto $\mf{X}\subseteq\mf{V}$ is a mixed graph over $\mf{X}$. It retains the edges in $\mc{G}[\mf{X}]$ and, for every pair of distinct vertices $V_i,V_j\in\mf{X}$, includes the following additional edges: 
    \begin{itemize}
        \item[1.] Add a directed edge $V_i \to V_j$ if there exists a directed path from $V_i$ to $V_j$ in $\mc{G}$ such that every non-endpoint vertex on the path is not in $\mf{X}$.

        \item[2.] Add a bidirected edge $V_i\bedge V_j$ if there exists a path between $V_i$ and $V_j$ in $\mc{G}$ with arrowheads at both endpoints, such that every non-endpoint vertex is a non-collider and lies outside $\mf{X}$. 
    \end{itemize}
    We denote by $\mc{G} \langle \mf{X} \rangle$ the latent projection of $\mc{G}$ onto $\mf{X}$. 
\end{definition}

\subsection{Graphical Constraints}

The structure of a causal diagram imposes constraints on the data generated by any causal model that is consistent with the diagram. One such constraint is conditional independence (CI), which can be inferred from the causal diagram via the \textit{d-separation} criterion. 
\begin{definition}[\textit{d}-separation \citep{pearl:2k}]
    In a causal diagram $\mc{G}$, let $X$ and $Y$ be distinct vertices and let $\mf{Z} \subseteq \mf{V} \setminus \{X,Y\}$. A path $p$ between vertices $X$ and $Y$ is a \textit{d-connecting} (\textit{active}) path relative to a set $\mf{Z}$ if: 
    \begin{itemize}
        \item[1.] every non-collider on $p$ is not a member of $\mf{Z}$; and
        \item[2.] every collider on $p$ is an ancestor of some member of $\mf{Z}$. 
    \end{itemize}   
    Two variables $X$ and $Y$ are said to be \textit{d-separated} by $\mf{Z}$ if there is no d-connecting path between $X$ and $Y$ relative to $\mf{Z}$. Furthermore, two disjoint sets $\mf{X}$ and $\mf{Y}$ are said to be d-separated by $\mf{Z}$ if every variable in $\mf{X}$ is d-separated from every variable in $\mf{Y}$ by $\mf{Z}$; this is denoted as $(\mf{X} \Perp \mf{Y} \mid \mf{Z})_{\mc{G}}$\footnote{Probabilistic independence statements are denoted as $(\mf{X} \Perp \mf{Y} \mid \mf{Z})_{P}$.}.
\end{definition}
We now present a key result that relates d-separation statements derived from a causal diagram $\mc{G}$ to constraints on the distribution $P(\mf{V})$. 
\begin{theorem}[Probabilistic implications of d-separation \citep{pearl:2k}]\label{thm: pearl2k d-sep means ind}
    If $\mf{X}$ and $\mf{Y}$ are d-separated by $\mf{Z}$ in a causal diagram $\mc{G}$, then $\mf{X}$ is independent of $\mf{Y}$ given $\mf{Z}$ in every distribution $P$ compatible with $\mc{G}$. Conversely, if $\mf{X}$ and $\mf{Y}$ are not d-separated by $\mf{Z}$ in $\mc{G}$, then $\mf{X}$ and $\mf{Y}$ are dependent given $\mf{Z}$ in at least one distribution $P$ compatible with $\mc{G}$. 
\end{theorem}
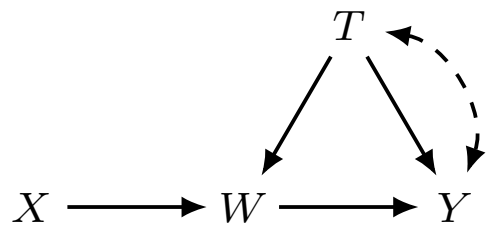
\begin{figure}[t]
    \centering
    \begin{minipage}[b]{0.26\textwidth}\centering  
        \begin{tikzpicture}[x=1.4cm,y=1.2cm,>={Latex[width=1.4mm,length=1.7mm]},
        line width=0.25mm,
        R/.style={draw = none,circle,inner sep=0mm, minimum size=4.5mm}]
        \node[R] (X) at (0,0) {$X$};
        \node[R] (W) at (1,0) {$W$};
        \node[R] (T) at (1.5,1) {$T$};
        \node[R] (Y) at (2,0) {$Y$};

        \draw[->] (X) -- (W);
        \draw[->] (W) -- (Y);
        
        \draw[->] (T) -- (W);
        \draw[->] (T) -- (Y);
        
        \draw[dashed,<->] (T) to[bend left = 50] (Y);
        \end{tikzpicture}
    \end{minipage}
    \caption{\small An example causal diagram for illustrating d-separation.}    \label{fig:foundation_d-sep}
\end{figure}
For concreteness, we consider the causal diagram in \Cref{fig:foundation_d-sep}. The path $p = \langle X,W,T,Y \rangle$ is active relative to $\mf{Z} = \{W\}$ since $T \notin \mf{Z}$ is a non-collider and $W \in \mf{Z}$ is a collider. Alternatively, the path is blocked relative to $\mf{Z}' = \{W,T\}$ since $T \in \mf{Z}'$ is a non-collider along $p$. Moreover, $p' = \langle X, W, Y \rangle $ is blocked relative to $\mf{Z}'$ since $W \in \mf{Z}'$ is a non-collider along $p'$. Hence, $X$ and $Y$ are d-separated relative to $\mf{Z}' = \{W,T\}$. Therefore, \Cref{thm: pearl2k d-sep means ind} states that:
\begin{align}
    (X \Perp Y \mid W,T)_{\mc{G}} \implies (X \Perp Y \mid W,T)_{P}.
\end{align}

\section{Ancestral Graphs and Causal Discovery}

\paragraph{Recovering causal structures.}

Causal discovery addresses the reverse direction: inferring from $P(\mf{V})$ which causal diagrams could have generated it \citep{pearl:2k,spirtes2000causation}. 

In practice, assumptions regarding model simplicity (à la Occam's razor) are used to avoid situations in which a saturated model would otherwise be preferred. A growing literature focuses on constructing equivalence classes of causal diagrams, acknowledging the difficulty of identifying a single true causal diagram from data \citep{Zhang2008b,hauser2012characterization,kocaoglu2017experimental,greenewald2019sample,squires2020active,choo2023adaptivity,zhou2025characterization}.

The simplest and most abundant form of data is observational, and conditional independence is a key property that is imprinted in the data by the underlying causal diagram via d-separation, as discussed earlier. Consequently, causal diagrams that encode the same set of d-separations imply the same CIs in the generated data. Hence, the goal is to leverage the CIs in the data to learn the equivalence class of the true, but unknown, causal diagram. 

\paragraph{Markov equivalence class.} The following definition formally introduces this equivalence class, known as the Markov equivalence class (MEC) \citep{richardson2002ancestral,zhang2006causal,Zhang2008b}. 
\begin{definition}[Markov equivalence]
    Two graphs are in the same \textit{Markov equivalence class} (MEC) if they imply the same set of conditional independence statements. 
\end{definition}
We use $\mc{E}$ to denote a Markov equivalence class. To denote a particular equivalence class to which a model $\mc{G}$ belongs, we write $\mc{E}(\mc{G})$.  

\paragraph{Ancestral graphs.} Ancestral graphs are designed to capture graph structures without explicitly modeling latent variables. While directed edges between vertices in a causal diagram represent a possible direct causal link between them, in ancestral graphs, directed edges instead represent ancestral relationships. Similar to the absence of directed cycles in causal diagrams, ancestral graphs do not permit an \textit{almost directed cycle}, which occurs when $X \leftrightarrow Y$ is present while $X$ is an ancestor of $Y$. 

\begin{figure}[t]
    \vspace{-1em}
    \begin{minipage}[b]{0.19\textwidth}\centering
    \subfloat[$\mc{G}_1$]{
        \begin{tikzpicture}[x=1cm,y=1.5cm,>={Latex[width=1.4mm,length=1.2mm]},
        line width=0.25mm,
        RR/.style={draw=none,circle,inner sep=0mm, minimum size=4mm}]
        \node[RR] (A) at (0,1.2) {$A$};
        \node[RR] (B) at (1.6,1.2) {$B$};
        \node[RR] (C) at (0.8,0.8) {$C$};
        \node[RR] (Y) at (0.8,0.2) {$Y$};

        \draw[->] (C) -- (Y);
        \draw[->] (B) -- (Y);
        
        \draw[dashed,<->] (A) to[bend left = 30] (C);
        \draw[dashed,<->] (B) to[bend right = 30] (C);
        \end{tikzpicture}\label{fig:foundation_ancestral_cd1}}
    \end{minipage}
    \hfill
    \begin{minipage}[b]{0.19\textwidth}\centering
    \subfloat[$\mc{G}_2$]{
        \begin{tikzpicture}[x=1cm,y=1.5cm,>={Latex[width=1.4mm,length=1.2mm]},
        line width=0.25mm,
        RR/.style={draw=none,circle,inner sep=0mm, minimum size=4mm}]
        \node[RR] (A) at (0,1.2) {$A$};
        \node[RR] (B) at (1.6,1.2) {$B$};
        \node[RR] (C) at (0.8,0.8) {$C$};
        \node[RR] (Y) at (0.8,0.2) {$Y$};

        \draw[->] (A) -- (C);
        \draw[->] (B) -- (C);
        \draw[->] (C) -- (Y);
        \draw[->] (B) -- (Y);

        \draw[dashed,<->] (A) to[bend left = 30] (C);
        \draw[dashed,<->] (B) to[bend left = 30] (Y);
    
        \end{tikzpicture}\label{fig:foundation_ancestral_cd2}}
    \end{minipage}
    \hfill
    \begin{minipage}[b]{0.19\textwidth}\centering
    \subfloat[$\mc{M}_1$]{
        \begin{tikzpicture}[x=1cm,y=1.5cm,>={Latex[width=1.4mm,length=1.2mm]},
        line width=0.25mm,
        RR/.style={draw=none,circle,inner sep=0mm, minimum size=4mm}]
        \node[RR] (A) at (0,1.2) {$A$};
        \node[RR] (B) at (1.6,1.2) {$B$};
        \node[RR] (C) at (0.8,0.8) {$C$};
        \node[RR] (Y) at (0.8,0.2) {$Y$};

        \draw[<->] (A) -- (C);
        \draw[<->] (B) -- (C);
        \draw[->, draw = betterpurple] (C) -- (Y);
        \draw[->, draw = betterpurple] (B) -- (Y);
    
        \end{tikzpicture}\label{fig:foundation_ancestral_mag1}}
    \end{minipage}
    \hfill
    \begin{minipage}[b]{0.19\textwidth}\centering
    \subfloat[$\mc{M}_2$]{
        \begin{tikzpicture}[x=1cm,y=1.5cm,>={Latex[width=1.4mm,length=1.2mm]},
        line width=0.25mm,
        RR/.style={draw=none,circle,inner sep=0mm, minimum size=4mm}]
        \node[RR] (A) at (0,1.2) {$A$};
        \node[RR] (B) at (1.6,1.2) {$B$};
        \node[RR] (C) at (0.8,0.8) {$C$};
        \node[RR] (Y) at (0.8,0.2) {$Y$};

        \draw[->] (A) -- (C);
        \draw[->] (B) -- (C);
        \draw[->, draw = betterpurple] (C) -- (Y);
        \draw[->] (B) -- (Y);
        \end{tikzpicture}\label{fig:foundation_ancestral_mag2}}
    \end{minipage}
    \hfill
    \begin{minipage}[b]{0.19\textwidth}\centering
    \subfloat[$\mc{P}$]{
        \begin{tikzpicture}[x=1cm,y=1.5cm,>={Latex[width=1.4mm,length=1.2mm]},
        line width=0.25mm,
        RR/.style={draw=none,circle,inner sep=0mm, minimum size=4mm},
        oo/.tip={Circle[open,length=3.5pt,width=3.5pt]},
        ]
        \node[RR] (A) at (0,1.2) {$A$};
        \node[RR] (B) at (1.6,1.2) {$B$};
        \node[RR] (C) at (0.8,0.8) {$C$};
        \node[RR] (Y) at (0.8,0.2) {$Y$};

        \draw[{oo}->] (A) -- (C);
        \draw[{oo}->] (B) -- (C);
        \draw[->, draw = betterpurple] (C) -- (Y);
        \draw[->] (B) -- (Y);
        \end{tikzpicture}\label{fig:foundation_ancestral_pag}}
    \end{minipage}
    \hfill\null
    \caption{\small Causal diagrams (a, b) with corresponding (c, d) MAGs and (e) a PAG. Purple directed edges are visible.}
    \label{fig:foundation_ancestral}
\end{figure}
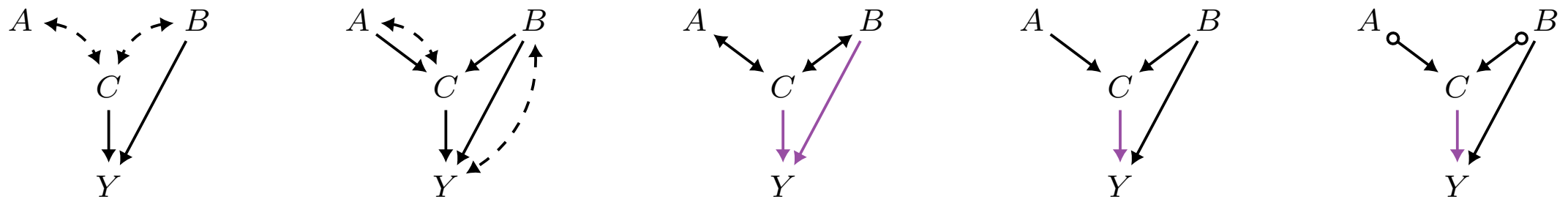

\paragraph{Maximal ancestral graphs.} We consider the case without selection variables, so undirected edges are excluded. A mixed graph is called a \textit{maximal ancestral graph} (MAG) if:
\begin{itemize}[leftmargin=5mm]
    \item[1]\textbf{(Ancestral)}. It does not contain any directed or almost directed cycles; and 

    \item[2]\textbf{(Maximal)}. There is no inducing path relative to the empty set\footnote{An \textit{inducing path} relative to $\mf{U}$ is defined as a path where (1) every non-endpoint vertex not in $\mf{U}$ is a collider, and (2) every collider is an ancestor of at least one endpoint. } between any two non-adjacent vertices. 
\end{itemize}
A MAG entails a conditional independence model via \textit{m-separation}, defined below. Its path-blocking conditions are the same as those used for causal diagrams above; the different name indicates that we now apply them to ancestral graphs\footnote{We use the notation $\mc{M}$ to denote a MAG when focusing on ancestral graphs, rather than an SCM.}.
\begin{definition}[\textit{m}-separation]
    Let $X$ and $Y$ be distinct vertices and let $\mf{Z} \subseteq \mf{V} \setminus \{X,Y\}$. A path $p$ between vertices $X$ and $Y$ is an \textit{m-connecting} (\textit{active}) path relative to a set $\mf{Z}$ if: 
    \begin{itemize}
        \item[1.] Every non-collider on $p$ is not a member of $\mf{Z}$; and
        \item[2.] Every collider on $p$ is an ancestor of some member of $\mf{Z}$. 
    \end{itemize}   
    Two variables $X$ and $Y$ are \textit{m-separated} by $\mf{Z}$ if there is no m-connecting path between $X$ and $Y$ relative to $\mf{Z}$. Two disjoint sets $\mf{X}$ and $\mf{Y}$ are said to be m-separated by $\mf{Z}$ if every variable in $\mf{X}$ is m-separated from every variable in $\mf{Y}$ by $\mf{Z}$.
\end{definition}
In general, a MAG represents a set of causal diagrams that entail the same conditional independence and ancestral relations among the observed variables. For each causal diagram, there is a unique corresponding MAG over the observed variables that preserves their ancestral relations and the conditional independence relations implied by the graph after marginalizing the latent variables. Distinct MAGs can imply the same conditional independence relations among the observed variables. To illustrate, consider the causal diagrams $\mc{G}_1$ and $\mc{G}_2$ in \Cref{fig:foundation_ancestral}. While they yield the same conditional independence relations: 
\begin{align}
        (A \Perp B)_P \text{~and~} (A \Perp Y \mid B, C )_P
\end{align}
they correspond to distinct MAGs, $\mc{M}_1$ and $\mc{M}_2$, respectively. The following graphical constructs are key to deriving a graphical characterization for when two MAGs are Markov equivalent. 

The \textit{skeleton} is obtained by ignoring all endpoint marks. A triple $\langle A,B,C\rangle$ is \textit{unshielded} if $A$ and $B$ are adjacent, $B$ and $C$ are adjacent, and $A$ and $C$ are not adjacent. A path is \textit{uncovered} if every consecutive triple on the path is unshielded. This condition concerns vertices two steps apart; it does not, by itself, prohibit every edge between nonconsecutive vertices. 
\begin{definition}[Discriminating path]
  A path $p = \langle X, \cdots, W,Z,Y \rangle$ is a \textit{discriminating path} for $Z$ if: 
  \begin{itemize}
    \item $p$ includes at least three edges;
    \item $Z$ is a non-endpoint vertex on $p$, and is adjacent to $Y$ on $p$; and
    \item $X$ is not adjacent to $Y$, and every vertex between $X$ and $Z$ is a collider on $p$ and is a parent of $Y$.   
  \end{itemize}
\end{definition}
For example, consider the MAG in \Cref{fig:foundation_ancestral_mag1}. The path $p= \langle A,C,B,Y \rangle$ is a discriminating path for $B$ between $A$ and $Y$. The path includes at least three edges, $A$ is not adjacent to $Y$, and $C$ is a collider along $p$ and a parent of $Y$. Using this structure, we have the following graphical characterization for when two MAGs are Markov equivalent.
\begin{lemma}[Graphical characterization of MEC \citep{pmlr-vR1-spirtes97b,zhang2012characterization}]\label{lem: MEC}
    Two MAGs over the same set of vertices are Markov equivalent \textit{if and only if}:
    \begin{itemize}
        \item[1.] they have the same adjacencies; 
        \item[2.] they have the same unshielded colliders; and
        \item[3.] for every path that is discriminating for the same vertex in both graphs, that vertex has the same collider status on the path in the two graphs.
    \end{itemize}
\end{lemma}
In words, two MAGs are Markov equivalent if they have the same skeleton and the same unshielded colliders (also known as \textit{v-structures}), and if every discriminating path common to both graphs identifies a collider in one graph if and only if it identifies a collider in the other.\footnote{Determining whether two MAGs are Markov equivalent is tractable, with a worst-case runtime of $\mc{O}(\lvert \mf{V} \rvert^3)$ \citep{zhao2005markov,pmlr-v180-wienobst22a}.}

For example, $\mc{M}_1$ and $\mc{M}_2$ in \Cref{fig:foundation_ancestral_mag1,fig:foundation_ancestral_mag2} are Markov equivalent because:
\begin{itemize}
    \item[1.] they have the same adjacencies $A \adedge C, B \adedge C, B \adedge Y$ and  $C \adedge Y$.
    
    \item[2.] they have the same unshielded collider $A \adedge C \daedge B$. 
    
    \item[3.] the path $p=\langle A,C,B,Y\rangle$ is a discriminating path for $B$ in both graphs, and $B$ is a non-collider on $p$ in both graphs. 
\end{itemize}

This characterization of equivalent MAGs motivates the formulation of a partial ancestral graph (PAG), which is a partial mixed graph (PMG) representing an equivalence class of MAGs. A PAG shares the same adjacencies as every MAG in $\mc{E}(\mc{M})$ and displays all and only the invariant edge marks. To make this precise, we first introduce PMGs and then PAGs. 

\paragraph{Partial mixed graphs.} A partial mixed graph (PMG) has endpoint marks drawn from  tails ($-$), arrowheads ($\mathord{>}$), and circles ($\circ$). A circle mark implies an undetermined mark that can be either an arrowhead or a tail. We use the $\boldsymbol{?}$ mark to denote a wildcard that represents either a tail ($-$) or a circle ($\circ$), but not an arrowhead ($\mathord{>}$). In addition, we use an asterisk ($*$) as a wildcard to denote any possible mark. 

In a PMG, if every edge mark on a path consists of circles, the path is called a \textit{circle path}, and each edge is called a \textit{circle edge} ($\,\cedge\,$). An edge is a \textit{partially directed edge} ($\,\cdedge$) if it has both a circle and an arrowhead. A \textit{circle component} is a maximal set of vertices connected by circle paths. A vertex incident to no circle edge forms a singleton circle component. Furthermore, $[\mc{Q}]$ denotes the set of MAGs represented by the PMG $\mc{Q}$, and similarly $[\mc{M}]$ denotes the set of causal diagrams conforming to the MAG $\mc{M}$.

A path $\langle V_0 (=X), V_1, \cdots, V_k (=Y) \rangle$ is \textit{possibly directed} from $X$ to $Y$ if, for every $i < k$, the edge between $V_i$ and $V_{i+1}$ has no arrowhead at $V_i$  and no tail at $V_{i+1}$. We use kinship notation for graphical relationships in PMGs, with the prefix $\Poss$: possible ancestors ($\Poss\An$) and possible descendants ($\Poss\De$). 
\begin{itemize}
    \item $\Poss\An(X)_{\mc{Q}}$ (possible ancestors of $X$ in $\mc{Q}$): the set $\{Z_i\}$ such that there exists a possibly directed path from $Z_i$ to $X$ in $\mc{Q}$. 

    \item $\Poss\De(X)_{\mc{Q}}$ (possible descendants of $X$ in $\mc{Q}$): the set $\{Z_i\}$ such that there exists a possibly directed path from $X$ to $Z_i$ in $\mc{Q}$. 
\end{itemize}
Both possible ancestors and possible descendants include the variable itself. For a set of variables $\mf{X}$, define:
\begin{align}
    \Poss\An(\mf{X})_{\mc{Q}} = \bigcup_{X \in \mf{X}} \Poss\An(X)_{\mc{Q}},
\end{align}
and define $\Poss\De(\mf{X})_{\mc{Q}}$ analogously. 

\paragraph{Partial ancestral graphs.} A \textit{partial ancestral graph} (PAG) \citep{richardson2002ancestral, zhang2006causal,Zhang2008b,zhang2008a,ali2005towards}, denoted by $\mc{P}$, is a PMG that represents a Markov equivalence class of MAGs. Every MAG $\mc{M}$ represented by a PAG has the same skeleton as $\mc{P}$, and the non-circle marks in $\mc{P}$ are identical to those in $\mc{M}$. Every circle in $\mc{P}$ corresponds to a non-invariant mark among the represented MAGs. The PAG $\mc{P}$ in \Cref{fig:foundation_ancestral_pag}, for instance, encodes every MAG obtained by orienting circle marks incident to $A$ and $B$ as either $\mathord{>}$ or $-$, including both $\mc{M}_{1}$ and $\mc{M}_{2}$. Therefore, $\{\mc{M}_1,\mc{M}_2\} \subseteq [\mc{P}] = \mc{E}(\mc{M}_1) = \mc{E}(\mc{M}_2)$. 

\begin{algorithm}[t]
\DontPrintSemicolon
\SetAlgoNoEnd
\SetKwInput{KwInput}{Input}
\SetKwInput{KwOutput}{Output}

\KwInput{Observational distribution $P(\mathbf{V})$}
\KwOutput{A PAG $\mathcal{P}$}

\caption{\small An oracle-level schematic of FCI for learning a PAG.}
\label{alg: fci}

\BlankLine
\tcp{Phase 1: Learn separating sets and adjacencies.}

Form a complete graph $\mathcal{P}$ over $\mathbf{V}$ composed of circle edges ($\,\cedge\,$)\;

\For{\rm each unordered pair of distinct vertices $X,Y \in \mathbf{V}$}{

    Set $\mathsf{Sepflag}(X,Y) = \mathsf{False}$. \;
    
    \For{$\mf{Z} \subseteq \mathbf{V}\setminus\{X,Y\}$}{
        \If{$(X \Perp Y \mid \mf{Z})_P$}{
            
            Set $\mathsf{Sepset}(X,Y) = \mathsf{Sepset}(Y,X) = \mf{Z}$. \;

            Set $\mathsf{Sepflag}(X,Y) = \mathsf{True}$. \;
            
            \textbf{break}\;
        }
    }
    \If{$\mathsf{Sepflag}(X,Y)$}{
        Remove the edge between $X$ and $Y$ from $\mathcal{P}$. \;
    }
}

\BlankLine
\tcp{Phase 2: Learn v-structures.}

Apply $\mc{R}_0$ until none applies. \;

\BlankLine
\tcp{Phase 3: Apply orientation rules.}

Apply the seven rules $\mc{R}_1$--$\mc{R}_4$, $\mc{R}_8$--$\mc{R}_{10}$ until none applies. \;

\end{algorithm}
\paragraph{Fast Causal Inference (FCI).} Assume the causal Markov condition and faithfulness with respect to the true MAG $\mc{M}^\ast$, with no selection bias. The causal Markov condition states that graphical separation implies conditional independence in the observational distribution; faithfulness requires the converse. For pairwise disjoint sets $\mf{X},\mf{Y},\mf{Z}\subseteq\mf{V}$, faithfulness states that: 
    \begin{align}
     (\mf{X}\Perp\mf{Y}\mid\mf{Z})_P
     \implies (\mf{X}\Perp\mf{Y}\mid\mf{Z})_{\mc{M}^\ast}.
    \end{align}
Given an oracle for these conditional independence relations, FCI \citep{spirtes2000causation,Zhang2008b} recovers the PAG representing the Markov equivalence class of $\mc{M}^\ast$. With finite data, conditional independence tests estimate this oracle information. 

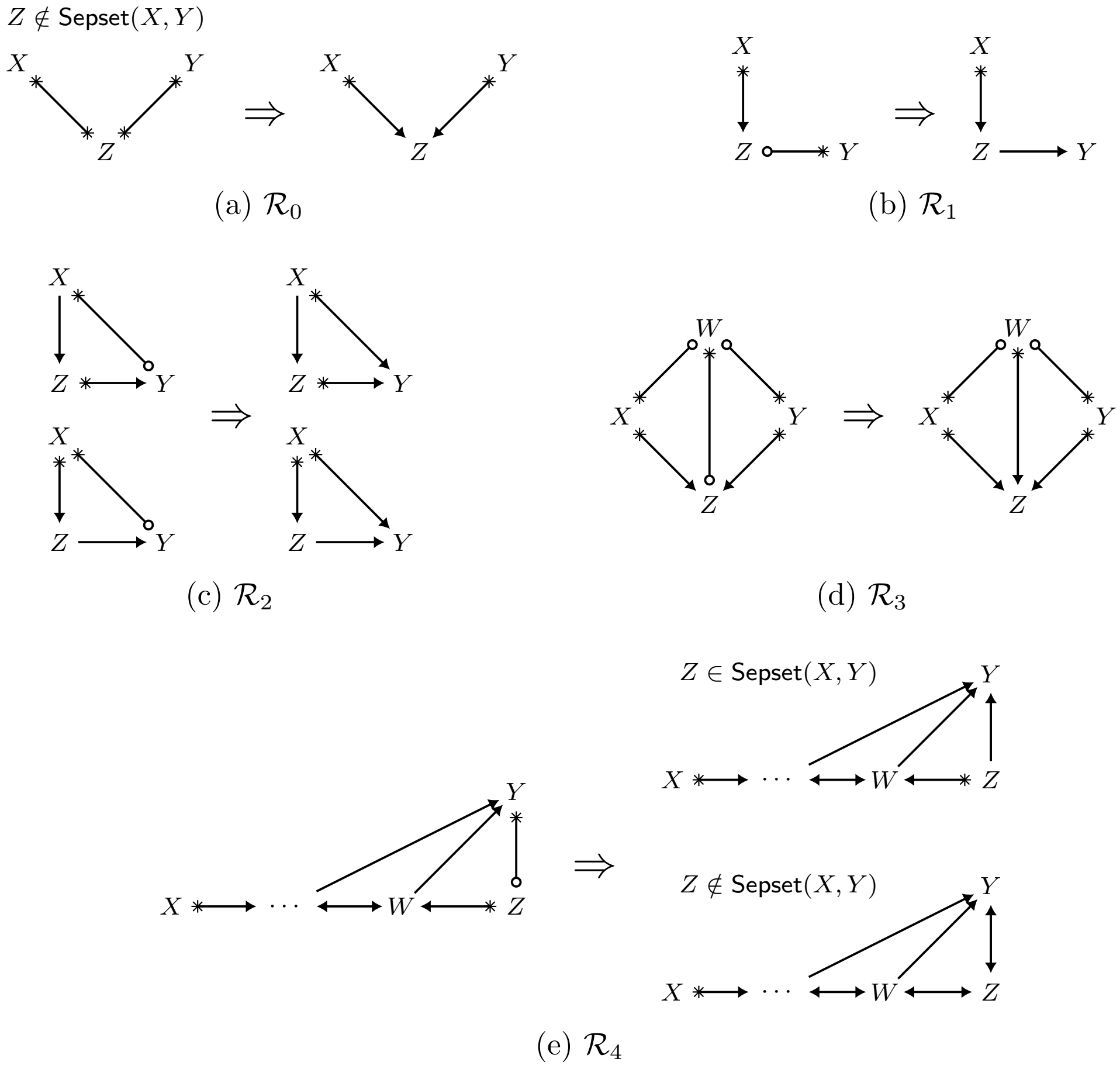
\begin{figure}[t]
    \hfill
    \begin{minipage}[b]{0.4\textwidth}
    \centering
    \subfloat[$\mc{R}_0$]{
        \begin{tikzpicture}[
            x=1cm,y=1cm,
            >={Latex[width=1.4mm,length=1.2mm]},
            line width=0.25mm,
            RR/.style={draw=none,circle,inner sep=0mm, minimum size=4mm}
        ]
        \node[] () at (1,1.5) {$Z \notin \textsf{Sepset}(X,Y)$};
        \node[RR] (X) at (0,1) {$X$};
        \node[RR] (Z) at (1,0) {$Z$};
        \node[RR] (Y) at (2,1) {$Y$};
    
        \draw[{ast}-{ast}] (Z) -- (X);
        \draw[{ast}-{ast}] (Z) -- (Y);
        \end{tikzpicture}
        \hspace{1pt}
        \raisebox{0.5cm}{\Large $\Rightarrow$}
        \hspace{1pt}
        \begin{tikzpicture}[
            x=1cm,y=1cm,
            >={Latex[width=1.4mm,length=1.2mm]},
            line width=0.25mm,
            RR/.style={draw=none,circle,inner sep=0mm, minimum size=4mm}
        ]
        \node[RR] (X) at (0,1) {$X$};
        \node[RR] (Z) at (1,0) {$Z$};
        \node[RR] (Y) at (2,1) {$Y$};
    
        \draw[<-{ast}] (Z) -- (X);
        \draw[<-{ast}] (Z) -- (Y);
        \end{tikzpicture}
    \label{fig:foundation_FCI_rules1_R0}
    }
    \end{minipage}
    \hspace{1.2cm}
    \begin{minipage}[b]{0.4\textwidth}
        \centering
        \subfloat[$\mc{R}_1$]{
            \begin{tikzpicture}[
                x=1.2cm,y=1.2cm,
                >={Latex[width=1.4mm,length=1.2mm]},
                line width=0.25mm,
                RR/.style={draw=none,circle,inner sep=0mm, minimum size=4mm}
            ]
            \node[RR] (X) at (0,1) {$X$};
            \node[RR] (Z) at (0,0) {$Z$};
            \node[RR] (Y) at (1,0) {$Y$};
        
            \draw[{ast}->] (X) -- (Z);
            \draw[{oo}-{ast}] (Z) -- (Y);
            \end{tikzpicture}
            \hspace{1pt}
            \raisebox{0.5cm}{\Large $\Rightarrow$}
            \hspace{1pt}
            \begin{tikzpicture}[
                x=1.2cm,y=1.2cm,
                >={Latex[width=1.4mm,length=1.2mm]},
                line width=0.25mm,
                RR/.style={draw=none,circle,inner sep=0mm, minimum size=4mm}
            ]
            \node[RR] (X) at (0,1) {$X$};
            \node[RR] (Z) at (0,0) {$Z$};
            \node[RR] (Y) at (1,0) {$Y$};
        
            \draw[{ast}->] (X) -- (Z);
            \draw[->] (Z) -- (Y);
            \end{tikzpicture}
        \label{fig:foundation_FCI_rules1_R1}
    }
    \end{minipage}
    \hfill\null
    \\[10pt]
    \null\hfill
    \begin{minipage}[b]{0.35\textwidth}
        \centering
        \subfloat[$\mc{R}_2$]{
            \begin{tikzpicture}[
                x=1.2cm,y=1.2cm,
                >={Latex[width=1.4mm,length=1.2mm]},
                line width=0.25mm,
                RR/.style={draw=none,circle,inner sep=0mm, minimum size=4mm}
            ]
            \node[RR] (X1) at (0,1) {$X$};
            \node[RR] (Z1) at (0,0) {$Z$};
            \node[RR] (Y1) at (1,0) {$Y$};
        
            \draw[->] (X1) -- (Z1);
            \draw[{ast}->] (Z1) -- (Y1);
            \draw[{ast}-{oo}] (X1) -- (Y1);

            \node[RR] (X2) at (0,-0.5) {$X$};
            \node[RR] (Z2) at (0,-1.5) {$Z$};
            \node[RR] (Y2) at (1,-1.5) {$Y$};
        
            \draw[{ast}->] (X2) -- (Z2);
            \draw[->] (Z2) -- (Y2);
            \draw[{ast}-{oo}] (X2) -- (Y2);
            
            \end{tikzpicture}
            \hspace{1pt}
            \raisebox{1.5cm}{\Large $\Rightarrow$}
            \hspace{1pt}
            \begin{tikzpicture}[
                x=1.2cm,y=1.2cm,
                >={Latex[width=1.4mm,length=1.2mm]},
                line width=0.25mm,
                RR/.style={draw=none,circle,inner sep=0mm, minimum size=4mm}
            ]
            \node[RR] (X1) at (0,1) {$X$};
            \node[RR] (Z1) at (0,0) {$Z$};
            \node[RR] (Y1) at (1,0) {$Y$};
        
            \draw[->] (X1) -- (Z1);
            \draw[{ast}->] (Z1) -- (Y1);
            \draw[{ast}->] (X1) -- (Y1);

            \node[RR] (X2) at (0,-0.5) {$X$};
            \node[RR] (Z2) at (0,-1.5) {$Z$};
            \node[RR] (Y2) at (1,-1.5) {$Y$};
        
            \draw[{ast}->] (X2) -- (Z2);
            \draw[->] (Z2) -- (Y2);
            \draw[{ast}->] (X2) -- (Y2);
            
            \end{tikzpicture}
        \label{fig:foundation_FCI_rules1_R2}
        }
        \end{minipage}
        \hfill
        \begin{minipage}[b]{0.45\textwidth}
        \centering
        \subfloat[$\mc{R}_3$]{
            \begin{tikzpicture}[
                x=1cm,y=1cm,
                >={Latex[width=1.4mm,length=1.2mm]},
                line width=0.25mm,
                RR/.style={draw=none,circle,inner sep=0mm, minimum size=4mm}
            ]
            \node[RR] (X) at (0,0) {$X$};
            \node[RR] (W) at (1,1) {$W$};
            \node[RR] (Z) at (1,-1) {$Z$};
            \node[RR] (Y) at (2,0) {$Y$};
        
            \draw[{ast}->] (X) -- (Z);
            \draw[<-{ast}] (Z) -- (Y);

            \draw[{ast}-{oo}] (X) -- (W);
            \draw[{oo}-{ast}] (W) -- (Y);

            \draw[{oo}-{ast}] (Z) -- (W);

            \node[] () at (0,-1.5){};
            \end{tikzpicture}
            \hspace{1pt}
            \raisebox{1.5cm}{\Large $\Rightarrow$}
            \hspace{1pt}
            \begin{tikzpicture}[
                x=1cm,y=1cm,
                >={Latex[width=1.4mm,length=1.2mm]},
                line width=0.25mm,
                RR/.style={draw=none,circle,inner sep=0mm, minimum size=4mm}
            ]
            \node[RR] (X) at (0,0) {$X$};
            \node[RR] (W) at (1,1) {$W$};
            \node[RR] (Z) at (1,-1) {$Z$};
            \node[RR] (Y) at (2,0) {$Y$};
        
            \draw[{ast}->] (X) -- (Z);
            \draw[<-{ast}] (Z) -- (Y);

            \draw[{ast}-{oo}] (X) -- (W);
            \draw[{oo}-{ast}] (W) -- (Y);

            \draw[<-{ast}] (Z) -- (W);

            \node[] () at (0,-1.5){};
            \end{tikzpicture}
        \label{fig:foundation_FCI_rules1_R3}
    }
    \end{minipage}
    \hfill\null
    \\[10pt]
    \null\hfill
    \begin{minipage}[b]{0.9\textwidth}
        \centering
        \subfloat[$\mc{R}_4$]{
            \begin{tikzpicture}[
                x=1.3cm,y=1.3cm,
                >={Latex[width=1.4mm,length=1.2mm]},
                line width=0.25mm,
                RR/.style={draw=none,circle,inner sep=0mm, minimum size=4mm}
            ]
            \node[RR] (X) at (0,1) {$X$};
            \node[] (cds1) at (1,1) {$\cdots$};
            \node[RR] (W) at (2,1) {$W$};
            \node[RR] (Z) at (3,1) {$Z$};
            \node[RR] (Y) at (3,2) {$Y$};
            
            \draw[{ast}->] (X) -- (cds1);
            \draw[<->] (cds1) -- (W);
            \draw[<-{ast}] (W) -- (Z);
            \draw[->] (cds1) -- (Y);
            \draw[->] (W) -- (Y);
            \draw[{oo}-{ast}] (Z) -- (Y);

            \node[] () at (0,0.2) {};
            \end{tikzpicture}
            \hspace{5pt}
            \raisebox{1.5cm}{\Large $\Rightarrow$}
            \hspace{5pt}
            \begin{tikzpicture}[
                x=1.2cm,y=1.2cm,
                >={Latex[width=1.4mm,length=1.2mm]},
                line width=0.25mm,
                RR/.style={draw=none,circle,inner sep=0mm, minimum size=4mm}
            ]
            \node[RR] (X1) at (0,1) {$X$};
            \node[] (cds1) at (1,1) {$\cdots$};
            \node[RR] (W1) at (2,1) {$W$};
            \node[RR] (Z1) at (3,1) {$Z$};
            \node[RR] (Y1) at (3,2) {$Y$};

            \node[] () at (1,2) {$Z \in \textsf{Sepset}(X,Y)$};

            \draw[{ast}->] (X1) -- (cds1);
            \draw[<->] (cds1) -- (W1);
            \draw[<-{ast}] (W1) -- (Z1);
            \draw[->] (cds1) -- (Y1);
            \draw[->] (W1) -- (Y1);
            \draw[->] (Z1) -- (Y1);

            \node[RR] (X2) at (0,-1) {$X$};
            \node[] (cds2) at (1,-1) {$\cdots$};
            \node[RR] (W2) at (2,-1) {$W$};
            \node[RR] (Z2) at (3,-1) {$Z$};
            \node[RR] (Y2) at (3,0) {$Y$};

            \node[] () at (1,0) {$Z \notin \textsf{Sepset}(X,Y)$};
            
            \draw[{ast}->] (X2) -- (cds2);
            \draw[<->] (cds2) -- (W2);
            \draw[<->] (W2) -- (Z2);
            \draw[->] (cds2) -- (Y2);
            \draw[->] (W2) -- (Y2);
            \draw[<->] (Z2) -- (Y2);
            
            \end{tikzpicture}
        \label{fig:foundation_FCI_rules1_R4}
        }
        \end{minipage}
    \hfill\null
    \caption{\small Illustration of the FCI arrowhead (\textgreater) orientation rules $\mc{R}_0$--$\mc{R}_4$.}
    \label{fig:foundation_FCI_rules1}
\end{figure}
\begin{figure}[t]
        \hfill
        \begin{minipage}[b]{0.4\textwidth}
        \centering
        \subfloat[$\mc{R}_8$]{
            \begin{tikzpicture}[
                x=1.2cm,y=1.2cm,
                >={Latex[width=1.4mm,length=1.2mm]},
                line width=0.25mm,
                RR/.style={draw=none,circle,inner sep=0mm, minimum size=4mm}
            ]
            \node[RR] (X) at (0,1) {$X$};
            \node[RR] (Z) at (0,0) {$Z$};
            \node[RR] (Y) at (1,0) {$Y$};
        
            \draw[->] (X) -- (Z);
            \draw[->] (Z) -- (Y);
            \draw[{oo}->] (X) -- (Y);
            
            \end{tikzpicture}
            \hspace{1pt}
            \raisebox{0.5cm}{\Large $\Rightarrow$}
            \hspace{1pt}
            \begin{tikzpicture}[
                x=1.2cm,y=1.2cm,
                >={Latex[width=1.4mm,length=1.2mm]},
                line width=0.25mm,
                RR/.style={draw=none,circle,inner sep=0mm, minimum size=4mm}
            ]
            \node[RR] (X) at (0,1) {$X$};
            \node[RR] (Z) at (0,0) {$Z$};
            \node[RR] (Y) at (1,0) {$Y$};
        
            \draw[->] (X) -- (Z);
            \draw[->] (Z) -- (Y);
            \draw[->] (X) -- (Y);
            \end{tikzpicture}
        \label{fig:foundation_FCI_rules2_R8}
    }
    \end{minipage}
    \hfill
    \begin{minipage}[b]{0.55\textwidth}
        \centering
        \subfloat[$\mc{R}_9$]{
            \begin{tikzpicture}[
                x=1.4cm,y=1.2cm,
                >={Latex[width=1.4mm,length=1.2mm]},
                line width=0.25mm,
                RR/.style={draw=none,circle,inner sep=0mm, minimum size=4mm}
            ]
            \node[RR] (X) at (0,0) {$X$};
            \node[RR] (Z) at (0,0.5) {$Z$};
            \node[RR] (Y) at (1.5,0) {$Y$};
        
            \draw[dashed,->] (X) to[out=75, in=105, looseness=1.5] (Y);
            \draw[{oo}->] (X) -- (Y);

            \node[font = \sffamily\tiny] () at (0,1.2){\shortstack{uncovered\\possibly directed}};
            
            \end{tikzpicture}
            \hspace{1pt}
            \raisebox{0.5cm}{\Large $\Rightarrow$}
            \hspace{1pt}
            \begin{tikzpicture}[
                x=1.4cm,y=1.2cm,
                >={Latex[width=1.4mm,length=1.2mm]},
                line width=0.25mm,
                RR/.style={draw=none,circle,inner sep=0mm, minimum size=4mm}
            ]
            \node[RR] (X) at (0,0) {$X$};
            \node[RR] (Z) at (0,0.5) {$Z$};
            \node[RR] (Y) at (1.5,0) {$Y$};
        
            \draw[dashed,->] (X) to[out=75, in=105, looseness=1.5] (Y);
            \draw[->] (X) -- (Y);

            \end{tikzpicture}
        \label{fig:foundation_FCI_rules2_R9}
    }
    \end{minipage}
    \hfill\null
    \\[12pt]
    \null\hfill
    \begin{minipage}[b]{0.5\textwidth}
        \centering
        \subfloat[$\mc{R}_{10}$]{
            \centering
            \begin{tikzpicture}[
                x=1.4cm,y=1.2cm,
                >={Latex[width=1.4mm,length=1.2mm]},
                line width=0.25mm,
                RR/.style={draw=none,circle,inner sep=0mm, minimum size=4mm}
            ]
            \node[RR] (X) at (0,0) {$X$};
            \node[RR] (Z) at (0.9,0.5) {$Z$};
            \node[RR] (U) at (0,0.4) {$U$};
            
            \node[RR] (Y) at (1.5,0) {$Y$};
            \node[RR] (W) at (0.9,-0.5) {$W$};
            \node[RR] (V) at (0,-0.4) {$V$};
        
            \draw[{oo}->] (X) -- (Y);
            \draw[dashed,->] (X) to [bend left = 50] (Z);
            \draw[->] (Z) -- (Y);

            \draw[dashed,->] (X) to [bend right = 50] (W);
            \draw[->] (W) -- (Y);

            \node[font = \sffamily\tiny] () at (0,1){\shortstack{uncovered\\possibly directed}};
            
            \end{tikzpicture}
            \hspace{5pt}
            \raisebox{0.7cm}{\Large $\Rightarrow$}
            \hspace{5pt}
            \begin{tikzpicture}[
                x=1.4cm,y=1.2cm,
                >={Latex[width=1.4mm,length=1.2mm]},
                line width=0.25mm,
                RR/.style={draw=none,circle,inner sep=0mm, minimum size=4mm}
            ]
            \node[RR] (X) at (0,0) {$X$};
            \node[RR] (Z) at (0.9,0.5) {$Z$};
            \node[RR] (U) at (0,0.4) {$U$};
            
            \node[RR] (Y) at (1.5,0) {$Y$};
            \node[RR] (W) at (0.9,-0.5) {$W$};
            \node[RR] (V) at (0,-0.4) {$V$};
        
            \draw[->] (X) -- (Y);
            \draw[dashed,->] (X) to [bend left = 50] (Z);
            \draw[->] (Z) -- (Y);

            \draw[dashed,->] (X) to [bend right = 50] (W);
            \draw[->] (W) -- (Y);

            \end{tikzpicture}
        \label{fig:foundation_FCI_rules2_R10}
    }
    \end{minipage}
    \hfill\null
    \caption{\small Illustration of the FCI tail ($-$) orientation rules $\mc{R}_8$--$\mc{R}_{10}$. In $\mc{R}_9$, $Z$ is the first vertex after $X$ on the displayed path and is not adjacent to $Y$. In $\mc{R}_{10}$, $U$ and $V$ are the first vertices after $X$ on the two displayed paths; they are distinct and nonadjacent. }
    \label{fig:foundation_FCI_rules2}
\end{figure}
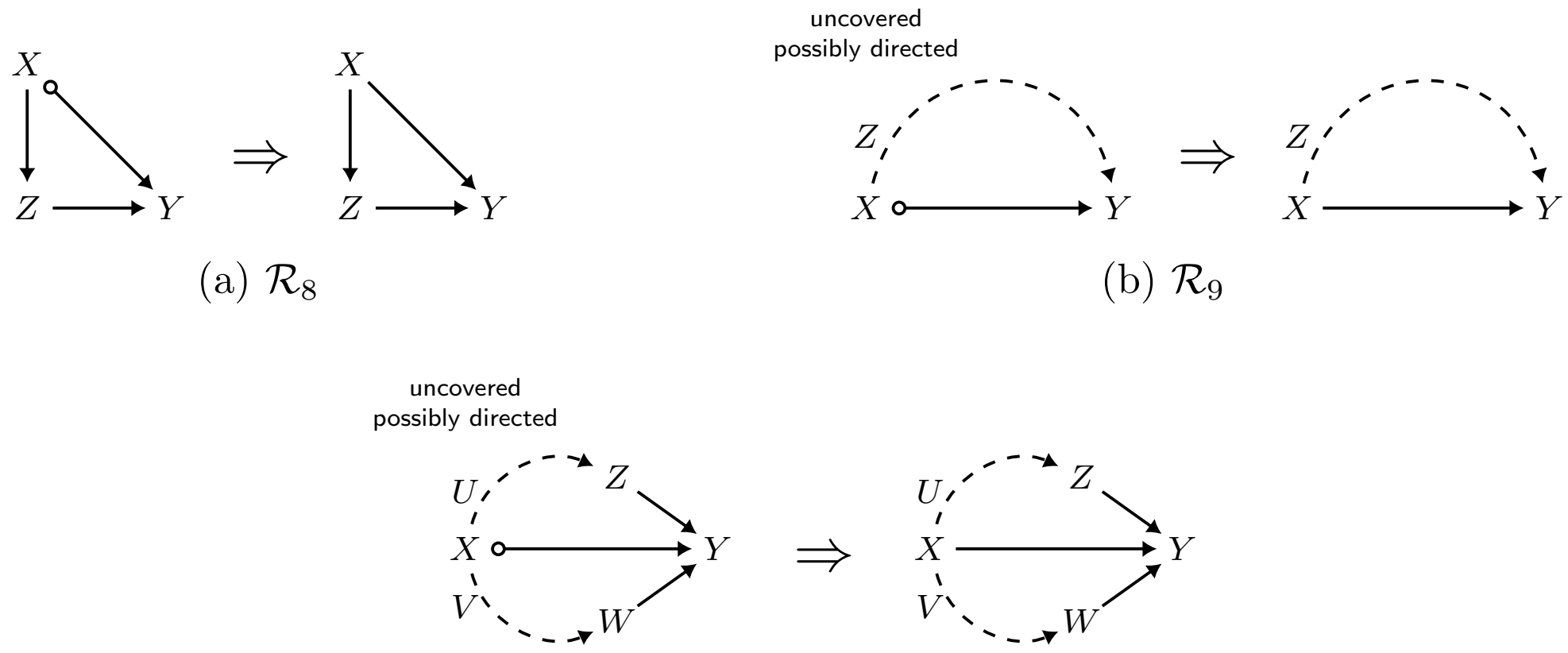
\bigskip
\Cref{alg: fci} gives an oracle-level schematic of FCI, using an exhaustive search for separating sets to simplify the presentation. Its three phases describe the logical steps of recovering adjacencies, identifying unshielded colliders, and completing invariant orientations. 

In Phase~1, we initialize a complete graph over the nodes in $\mf{V}$ consisting of circle edges ($\,\cedge\,$). For every pair of nodes $X$, $Y$, we remove the edge between $X$ and $Y$ if there exists a separating set $\mf{Z}$ such that $(X \Perp Y \mid \mf{Z})_{P}$. By the end of Phase~1, we obtain the skeleton of the resulting PAG $\mc{P}$. 

In Phase~2, we identify all v-structures by testing $\mc{R}_0$: 
{
\begin{itemize}[leftmargin=12mm]
    \item[$\mc{R}_0$.] For each unshielded triple $\langle X,Z,Y \rangle$ in $\mc{P}$, orient it as a collider $X \adedge Z\daedge Y$ \textit{if and only if} $Z$ is not in $\mathsf{Sepset}(X,Y)$. 
\end{itemize}}
\Cref{fig:foundation_FCI_rules1} illustrates $\mc{R}_0$ and the arrowhead orientation rules $\mc{R}_1$--$\mc{R}_4$. 

Finally, in Phase~3, we apply the orientation rules of \citet{Zhang2008b} other than $\mc{R}_5$--$\mc{R}_7$ until none applies. These three rules are omitted because selection bias is excluded. \Cref{fig:foundation_FCI_rules2} illustrates the tail orientation rules $\mc{R}_8$--$\mc{R}_{10}$. 

{
\begin{itemize}[leftmargin=12mm]
    \item[$\mc{R}_1$.]  If $X \adedge Z \caedge Y$, and $X$ and $Y$ are not adjacent, then orient $Z \caedge Y$ as $Z \to Y$.
    
    \item[$\mc{R}_2$.] If $X \to Z \adedge Y$  or $X \adedge Z \to Y$, and $X \acedge Y$, then orient $X \acedge Y$ as $X \adedge Y$.
    
    \item[$\mc{R}_3$.] If $X \adedge Z \daedge Y$, $X \acedge W \caedge Y$, $X$ and $Y$ are not adjacent, and $W \acedge Z$, then orient $W \acedge Z$ as $W \adedge Z$.
    
    \item[$\mc{R}_4$.] If $\langle X, \cdots, W,Z,Y \rangle$ is a discriminating path between $X$ and $Y$ for $Z$, and $Z \caedge Y$, then orient $Z \caedge Y$ as $Z \to Y$ if $Z \in \mathsf{Sepset}(X,Y)$; otherwise orient the triple $\langle W,Z,Y \rangle$ as $W \bedge Z \bedge Y$.
    
    \item[$\mc{R}_8$.] If $X \to Z \to Y$, and $X \cdedge Y$, orient $X \cdedge Y$ as $X \to Y$. 

    \item[$\mc{R}_9$.] If $X \cdedge Y$, and $p = \langle X, Z, W, \cdots, Y \rangle $ is an uncovered possibly directed path from $X$ to $Y$ such that $Z$ and $Y$ are not adjacent, then orient $X \cdedge Y$ as $X \to Y$.
    
    \item[$\mc{R}_{10}$.] Suppose $X \cdedge Y$, $Z \to Y \gets W$, $p_1$ is an uncovered possibly directed path from $X$ to $Z$, and $p_2$ is an uncovered possibly directed path from $X$ to $W$. Let $U$ be the vertex adjacent to $X$ on $p_1$ ($U$ could be $Z$), and $V$ be the vertex adjacent to $X$ on $p_2$ ($V$ could be $W$). If $U$ and $V$ are distinct, and not adjacent, then orient $X \cdedge Y$ as $X \to Y$.
\end{itemize}}
\begin{figure}[t]
    \vspace{-1em}
    \begin{minipage}[b]{0.19\textwidth}\centering
    \subfloat[]{
        \begin{tikzpicture}[x=1cm,y=1.5cm,>={Latex[width=1.4mm,length=1.2mm]},
        line width=0.25mm,
        RR/.style={draw=none,circle,inner sep=0mm, minimum size=4mm}]
        \node[RR] (A) at (0,1.2) {$A$};
        \node[RR] (B) at (1.6,1.2) {$B$};
        \node[RR] (C) at (0.8,0.8) {$C$};
        \node[RR] (Y) at (0.8,0.2) {$Y$};

        \draw[{oo}-{oo}] (A) -- (C);
        \draw[{oo}-{oo}] (A) -- (B);
        \draw[{oo}-{oo}] (A) -- (C);
        \draw[{oo}-{oo}] (A) -- (Y);
        
        \draw[{oo}-{oo}] (B) -- (C);
        \draw[{oo}-{oo}] (B) -- (Y);
        
        \draw[{oo}-{oo}] (C) -- (Y);
        \end{tikzpicture}\label{fig:foundation_fci_a}}
    \end{minipage}
    \hfill
    \begin{minipage}[b]{0.19\textwidth}\centering
    \subfloat[Phase~1]{
        \begin{tikzpicture}[x=1cm,y=1.5cm,>={Latex[width=1.4mm,length=1.2mm]},
        line width=0.25mm,
        RR/.style={draw=none,circle,inner sep=0mm, minimum size=4mm}]
        \node[RR] (A) at (0,1.2) {$A$};
        \node[RR] (B) at (1.6,1.2) {$B$};
        \node[RR] (C) at (0.8,0.8) {$C$};
        \node[RR] (Y) at (0.8,0.2) {$Y$};

        \draw[{oo}-{oo}] (A) -- (C);
        \draw[{oo}-{oo}] (B) -- (C);
        \draw[{oo}-{oo}] (C) -- (Y);
        \draw[{oo}-{oo}] (B) -- (Y);

        \end{tikzpicture}\label{fig:foundation_fci_b}}
    \end{minipage}
    \hfill
    \begin{minipage}[b]{0.19\textwidth}\centering
    \subfloat[Phase~2]{
        \begin{tikzpicture}[x=1cm,y=1.5cm,>={Latex[width=1.4mm,length=1.2mm]},
        line width=0.25mm,
        RR/.style={draw=none,circle,inner sep=0mm, minimum size=4mm}]
        \node[RR] (A) at (0,1.2) {$A$};
        \node[RR] (B) at (1.6,1.2) {$B$};
        \node[RR] (C) at (0.8,0.8) {$C$};
        \node[RR] (Y) at (0.8,0.2) {$Y$};

        \draw[{oo}->] (A) -- (C);
        \draw[{oo}->] (B) -- (C);
        \draw[{oo}-{oo}] (C) -- (Y);
        \draw[{oo}-{oo}] (B) -- (Y);
    
        \end{tikzpicture}\label{fig:foundation_fci_c}}
    \end{minipage}
    \hfill
    \begin{minipage}[b]{0.19\textwidth}\centering
    \subfloat[$\mc{R}_1$]{
        \begin{tikzpicture}[x=1cm,y=1.5cm,>={Latex[width=1.4mm,length=1.2mm]},
        line width=0.25mm,
        RR/.style={draw=none,circle,inner sep=0mm, minimum size=4mm}]
        \node[RR] (A) at (0,1.2) {$A$};
        \node[RR] (B) at (1.6,1.2) {$B$};
        \node[RR] (C) at (0.8,0.8) {$C$};
        \node[RR] (Y) at (0.8,0.2) {$Y$};

        \draw[{oo}->] (A) -- (C);
        \draw[{oo}->] (B) -- (C);
        \draw[->, draw = betterpurple] (C) -- (Y);
        \draw[{oo}-{oo}] (B) -- (Y);
        \end{tikzpicture}\label{fig:foundation_fci_d}}
    \end{minipage}
    \hfill
    \begin{minipage}[b]{0.19\textwidth}\centering
    \subfloat[$\mc{R}_4$]{
        \begin{tikzpicture}[x=1cm,y=1.5cm,>={Latex[width=1.4mm,length=1.2mm]},
        line width=0.25mm,
        RR/.style={draw=none,circle,inner sep=0mm, minimum size=4mm},
        oo/.tip={Circle[open,length=3.5pt,width=3.5pt]},
        ]
        \node[RR] (A) at (0,1.2) {$A$};
        \node[RR] (B) at (1.6,1.2) {$B$};
        \node[RR] (C) at (0.8,0.8) {$C$};
        \node[RR] (Y) at (0.8,0.2) {$Y$};

        \draw[{oo}->] (A) -- (C);
        \draw[{oo}->] (B) -- (C);
        \draw[->, draw = betterpurple] (C) -- (Y);
        \draw[->] (B) -- (Y);
        \end{tikzpicture}\label{fig:foundation_fci_e}}
    \end{minipage}
    \hfill\null
    \caption{\small Resulting graphs at each phase of FCI: (a) initial complete graph with circle edges; (b) after Phase~1; (c) after Phase~2; and (d, e) after applying $\mc{R}_1$ and $\mc{R}_4$ in Phase~3.}
    \label{fig:foundation_fci}
\end{figure}
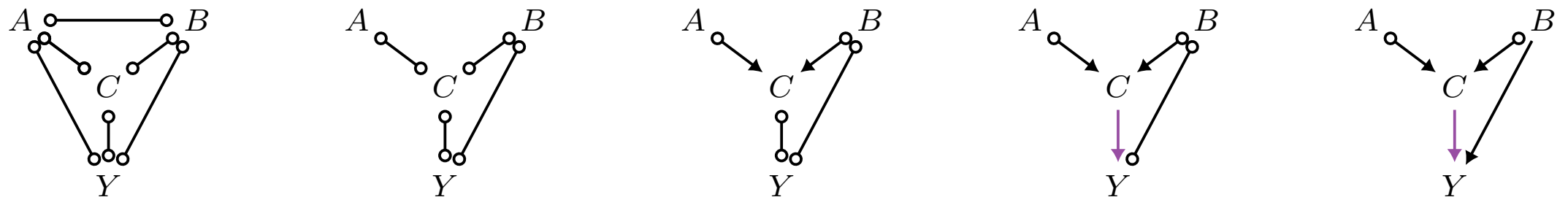
\Cref{fig:foundation_fci} illustrates how FCI works. We can learn the PAG $\mc{P}$ in \Cref{fig:foundation_fci_e} from observational data generated by the causal diagram in \Cref{fig:foundation_ancestral_cd1}. Nodes $A$ and $B$ are not adjacent in $\mc{P}$ because $(A \Perp B)_P$, and $A$ and $Y$ are not adjacent since $(A \Perp Y \mid B,C)_P$. Hence, Phase~1 results in the skeleton in \Cref{fig:foundation_fci_b}.   

In Phase~2, the v-structure $A \cdedge C \dcedge B$ is identified since $C \notin \mathsf{Sepset}(A, B)$, leading to \Cref{fig:foundation_fci_c}. In Phase~3, $C \cedge Y$ is oriented as $C \to Y$ by $\mc{R}_1$ since the triple $A \cdedge C \cedge Y$ is unshielded, which gives \Cref{fig:foundation_fci_d}. Moreover, $B \cedge Y$ is oriented as $B \to Y$ by $\mc{R}_4$ since $\langle A,C,B,Y \rangle$ is a discriminating path between $A$ and $Y$ for $B$, and $B \in \mathsf{Sepset}(A, Y)$. This yields the PAG in \Cref{fig:foundation_fci_e}. 

Under the oracle, Markov, and faithfulness assumptions stated above, FCI returns the PAG of the true MAG. Moreover, its orientation rules are sound and complete in the sense that they identify all and only the edge marks that are invariant within the represented Markov equivalence class. To state these properties precisely, we first define orientation soundness and completeness for a general PMG. 

\paragraph{Soundness and completeness of orientations.}
Fix a nonempty target class $\mathcal{J}$ of MAGs with a common skeleton, determined by the input PAG and any additional orientations. Let $\mc{Q}$ be a PMG on this skeleton. 
\begin{itemize}
    \item \textbf{Soundness.} Every tail or arrowhead in $\mc{Q}$ agrees with the corresponding mark in every MAG in $\mathcal{J}$.
    
    \item \textbf{Completeness.} Every endpoint mark that is invariant across $\mathcal{J}$ is displayed as a tail or arrowhead in $\mc{Q}$.
\end{itemize}
Thus, when orientations are sound and complete, each circle in $\mc{Q}$ has both a tail witness and an arrowhead witness in $\mathcal{J}$. For the PMGs used below, the represented class $[\mc{Q}]$ is understood relative to the input PAG and the incorporated additional orientations.

For example, consider the PAG $\mc{P}$ in \Cref{fig:foundation_ancestral_pag}, with target class $\mc{J} =[\mc{P}]$. The arrowheads at $C$ on $A \cdedge C$ and $B \cdedge C$ are invariant, as are both endpoint marks of $B \to Y$ and $C \to Y$. Soundness means that these displayed arrowheads and tails agree with every MAG in $\mc{J}$. Completeness means that all such invariant marks have been displayed. 

The MAGs $\mc{M}_1$ and $\mc{M}_2$ in \Cref{fig:foundation_ancestral_mag1,fig:foundation_ancestral_mag2} are two members of the class that witness the variation represented by the remaining circles. In $\mc{M}_1$, the edges $A \bedge C$ and $B \bedge C$ have arrowheads at $A$ and $B$, whereas in $\mc{M}_2$, the corresponding edges $A \to C$ and $B \to C$ have tails at those endpoints. Thus, neither endpoint mark is invariant, and the PAG correctly retains circles at $A$ and $B$. Its orientations are nevertheless complete: every remaining circle reflects variation within the target class rather than a missing invariant orientation. 

\paragraph{Visibility.}  In the context of ancestral graphs, it is crucial to be able to identify edges that are not confounded. A directed edge $X \to Y$ is \textit{visible} if no DAG whose corresponding MAG is the given MAG (or belongs to the class represented by the given PAG) contains an inducing path between $X$ and $Y$, relative to its latent vertices, that is into $X$. We refer to any edge that is not visible as \textit{invisible}. For induced subgraphs, we retain the model class specified by the full PMG $\mc{Q}$. Let $\mc{G}$ be a DAG over observed vertices $\mf{V}$ and latent vertices $\mf{U}$ whose corresponding MAG belongs to $[\mc{Q}]$. A directed edge $X\to Y$ in $\mc{Q}[\mf{A}]$ is visible relative to this class if no such DAG contains an inducing path into $X$ between $X$ and $Y$ in $\mc{G}[\mf{A}\cup\mf{U}]$, relative to $\mf{U}$. The latent vertices are retained when the observed vertex set is restricted to $\mf{A}$. 


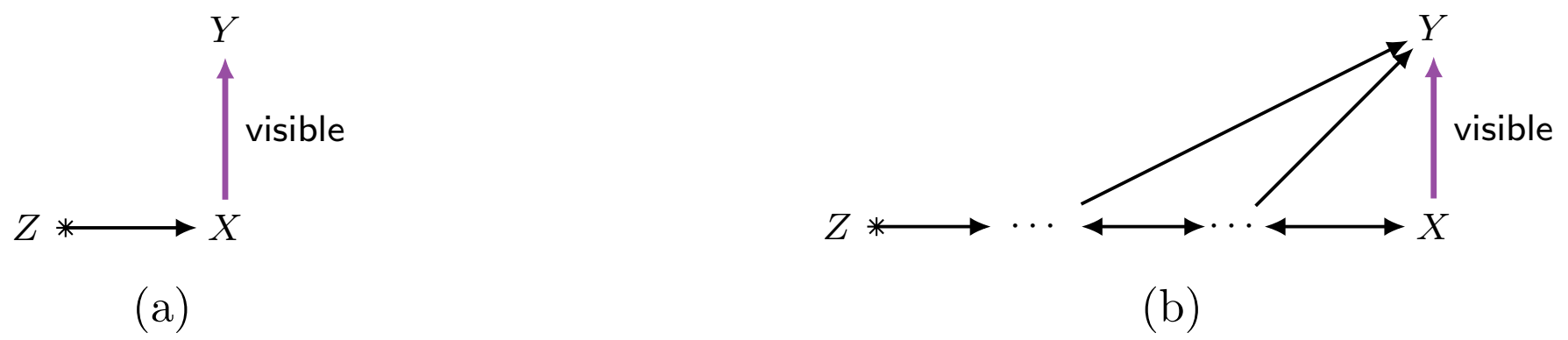
\begin{figure}[t]
    \begin{minipage}[b]{0.4\textwidth}\centering
    \subfloat[]{
        \begin{tikzpicture}[x=1.5cm,y=1.5cm,>={Latex[width=1.4mm,length=1.6mm]},
        line width=0.25mm,
        RR/.style={draw=none,circle,inner sep=0mm, minimum size=4mm}]
        \node[RR] (Z) at (0,0) {$Z$};
        \node[RR] (X) at (1,0) {$X$};
        \node[RR] (Y) at (1,1) {$Y$};
        
        \draw[{ast}->] (Z) -- (X);
        \draw[->, line width=0.45mm, draw = betterpurple] (X) -- node[right, font = \sffamily\scriptsize]{visible} (Y);
        \end{tikzpicture}\label{fig:foundation_visible_a}}
    \end{minipage}
    \hfill
    \begin{minipage}[b]{0.4\textwidth}\centering
    \subfloat[]{
        \begin{tikzpicture}[x=1.5cm,y=1.5cm,>={Latex[width=1.4mm,length=1.6mm]},
        line width=0.25mm,
        RR/.style={draw=none,circle,inner sep=0mm, minimum size=4mm},
        ]
        \node[RR] (Z) at (0,0) {$Z$};
        \node[] (cds1) at (1,0) {$\cdots$};
        \node[RR] (cds2) at (2,0) {$\cdots$};
        \node[RR] (X) at (3,0) {$X$};
        \node[RR] (Y) at (3,1) {$Y$};
        
        \draw[{ast}->] (Z) -- (cds1);
        \draw[<->] (cds1) -- (cds2);
        \draw[<->] (cds2) -- (X);
        \draw[->] (cds1) -- (Y);
        \draw[->] (cds2) -- (Y);
        \draw[->, line width=0.45mm, draw = betterpurple] (X) -- node[right, font = \sffamily\scriptsize]{visible} (Y);
        \end{tikzpicture}\label{fig:foundation_visible_b}}
    \end{minipage}
    \hfill\null
    \caption{\small Graphical characterization of visible edges.}
    \label{fig:foundation_visible}
\end{figure}
\begin{lemma}[Graphical characterization of visibility \citep{zhang2006causal, maathuis2015generalized}]\label{lem: visible}
A directed edge $X \to Y$ in a MAG or PAG is visible if and only if there exists a vertex $Z$ not adjacent to $Y$ for which at least one of the following conditions holds:
\begin{itemize}[leftmargin=10mm]
    \item[1.] $Z$ and $X$ are joined by an edge with an arrowhead at $X$. 

    \item[2.] A collider path from $Z$ to $X$ has an arrowhead at $X$, and all its non-endpoint vertices are parents of $Y$. 
\end{itemize}
\end{lemma} 
The criterion is evaluated in the full MAG or PAG, relative to its represented causal model class. For a PAG, a graphical visibility witness is preserved in every represented MAG. Conversely, if the criterion fails in the PAG, there is a represented MAG in which the directed edge is invisible (see \Cref{lem: perkovic 2018 lem48}). For an induced subgraph retaining the full graph's model class, a visibility witness may lie outside the retained observed vertices; the criterion should therefore not be recomputed using only the induced subgraph. 

\Cref{fig:foundation_visible} illustrates the conditions. Throughout this \workname{}, purple directed edges indicate visibility. For concreteness, $C \to Y$ in the PAG in \Cref{fig:foundation_fci_e} is visible, indicating that the two nodes are not confounded in any causal diagram in the MEC represented by the PAG. Note that $A \cdedge C \to Y$ satisfies the first condition of \Cref{lem: visible}. Moreover, consider the edge $B \to Y$ in the MAG of \Cref{fig:foundation_ancestral_mag1}, which is visible by the second condition of \Cref{lem: visible}. The path $\langle A, C, B \rangle$ forms a collider path into $B$, and $C$ on this path is a parent of $Y$. 

\jmlrsection{Structural Causal Bandits}\label{chap:scb}

Many canonical decision-making settings, such as multi-armed bandits (MABs), Markov decision processes (MDPs), and dynamic treatment regimes (DTRs), can be represented within the semantic framework of structural causal models (SCMs) \citep{Zhang2023towards,bareinboim2024introduction}. In this \unitname{}, we introduce the MAB framework, examine it through a causal lens, and generalize it within the SCM framework, resulting in what is known as structural causal bandits (SCBs) \citep{lee2018structural,lee2019structural}. 

\section{Multi-Armed Bandits}

The multi-armed bandit (MAB; simply, bandit) problem is a sequential decision-making problem. One can view this problem as a sequential game between an agent and an environment (i.e., a bandit instance). The game is played over $T$ rounds. In each round $t \in \{1,2,\cdots,T\}$, the agent first chooses an \textit{action} $A_t$ from a given set $\mc{A} = \{a_i\}_{i=1}^{\lvert \mc{A} \rvert}$, and the environment then reveals a \textit{reward} $Y_t$. The agent cannot access future information when choosing actions; its action
distribution may depend on the observed history: 
\begin{align}
    H_{t-1} = \langle A_1, Y_1, \cdots, A_{t-1}, Y_{t-1} \rangle.
\end{align}
A \textit{policy} $\pi = \{\pi_t\}_{t=1}^T$ assigns an action distribution $\pi_t(\cdot \mid H_{t-1})$ to each history. At round $t$, the agent samples $A_t$ from this distribution using internal randomness independent of the environment's randomness. Deterministic policies are included as a special case. The agent interacts with the environment by following a policy. The objective of the agent is to maximize the total reward, which is a random quantity depending on both the agent's actions and the rewards generated by the environment. Conditional on the selected action $A_t = a$ and the past history, $Y_t$ has distribution $\nu_a$, which does not change with $t$. All reward means are assumed finite. In the SCM formulation, each round uses a fresh independent draw of the exogenous variables from the same distribution $P(\mf{U})$. 

\paragraph{Unstructured vs. structured bandits.} 
The fundamental challenge in bandit problems is that the environment is unknown to the agent. The agent only knows that the true environment lies in some set $\mc{C}$, called the \textit{environment class}. 

An environment class $\mc{C}$ is called \textit{unstructured} if $\mc{A}$ is finite and playing an action provides no information about the reward distributions of other actions \citep{lattimore2020bandit}. Environment classes that are not unstructured are called \textit{structured}. An important feature of structured bandits is that the agent can often gain information about some actions without explicitly playing them. 
\paragraph{Cumulative regret.} 
Let $\nu = \{\nu_{a} : a \in \mc{A}\}$ be a stochastic bandit instance, and let $\mu_a$ denote the mean reward of action $a \in \mc{A}$. Let $a^\star$ be an optimal action, i.e., $\mu_{a^\star} = \max_{a \in \mc{A}} \mu_a$. The \textit{cumulative (pseudo-)regret} of a policy $\pi$ on instance $\nu$ is defined as:
\begin{align}
    R_T^{\mc{A}}(\pi,\nu) = T\mu_{a^\star} - \sum_{t=1}^T \mu_{A_t},
\end{align}
which compares the cumulative mean reward of always playing an optimal action with the sum of the mean rewards of the selected actions over $T$ rounds.  

Define the \textit{suboptimality gap} as $\Delta_a \triangleq \mu_{a^\star} - \mu_a$ for $a \in \mc{A}$. By the standard regret decomposition \citep{lattimore2020bandit}, we have:
\begin{align}
    R_T^{\mc{A}}(\pi,\nu) = \sum_{a \in \mc{A}} \Delta_a N_T(a),
\end{align}
where $N_T(a)$ denotes the number of times action $a$ is selected up to time $T$. This decomposition shows that minimizing expected cumulative regret amounts to controlling the expected number of pulls of suboptimal actions, weighted by their gaps: 
\begin{align}
    \mb{E}R_T^{\mc{A}}(\pi,\nu) = \sum_{a \in \mc{A}} \Delta_a \mb{E}N_T(a).
\end{align}
The expectation is taken with respect to the probability measure induced by the interaction between $\pi$ and $\nu$. When clear from context, we omit the dependence on $\pi$ and $\nu$.

\section{Structural Causal Bandits: Where to Intervene?}

\begin{figure}[t]
    \begin{minipage}[b]{0.3\textwidth}\centering
    \subfloat[]{
        \begin{tikzpicture}[x=1.2cm,y=1.3cm,>={Latex[width=1.4mm,length=1.7mm]},
        line width=0.25mm,
        R/.style={draw = none,circle,inner sep=0mm, minimum size=4.5mm}]
        \node[R] (U) at (1,0.7) {$U$};
        \node[R] (X) at (0,0) {$X$};
        \node[R] (Y) at (2,0) {$Y$};

        \draw[->] (X) -- (Y);
        \draw[->] (U) -- (X);
        \draw[->] (U) -- (Y);
        \end{tikzpicture}\label{fig:scb_standard_bandit_a}}
    \end{minipage}
    \hfill
    \begin{minipage}[b]{0.3\textwidth}\centering
    \subfloat[]{
        \begin{tikzpicture}[x=1.2cm,y=1.3cm,>={Latex[width=1.4mm,length=1.7mm]},
        line width=0.25mm,
        R/.style={draw = none,circle,inner sep=0mm, minimum size=4mm},
        transp/.style={regime, minimum size=1mm, draw = black,inner sep=0mm},
        regime line/.style={draw=black, thick}]
        \node[R] (U) at (1,0.7) {$U$};
        \node[R] (X) at (0,0) {$X$};
        \node[R] (Y) at (2,0) {$Y$};

        \node[transp, above = 2.5mm of X] (XT){};
        \node[right = -0.5mm of XT] (){$x$};
        \draw[->,regime line] (XT) -- (X);
        
        \draw[->] (X) -- (Y);
        \draw[->] (U) -- (Y);
        \end{tikzpicture}\label{fig:scb_standard_bandit_b}}
    \end{minipage}
    \hfill
    \begin{minipage}[b]{0.3\textwidth}\centering
    \subfloat[]{
        \begin{tikzpicture}[x=1cm,y=1.3cm,>={Latex[width=1.4mm,length=1.7mm]},
        line width=0.25mm,
        R/.style={draw = none,circle,inner sep=0mm, minimum size=4.5mm}]
        \node[R] (X1) at (-1,0.7) {$X_1$};
        \node[R] (X2) at (0,0) {$X_2$};
        \node[R] (Y) at (2,0) {$Y$};

        \draw[->] (X1) -- (X2);
        \draw[->] (X2) -- (Y);
        \draw[dashed,<->] (X2) to[bend left = 50] (Y);
        \end{tikzpicture}\label{fig:scb_standard_bandit_c}}
    \end{minipage}
    \hfill\null
    \caption{\small (a) A multi-armed bandit environment; (b) the submodel induced by the action $do(x)$, representing the post-action model perceived by the agent; and (c) a structural causal bandit environment. }
    \label{fig:scb_standard_bandit}
\end{figure}
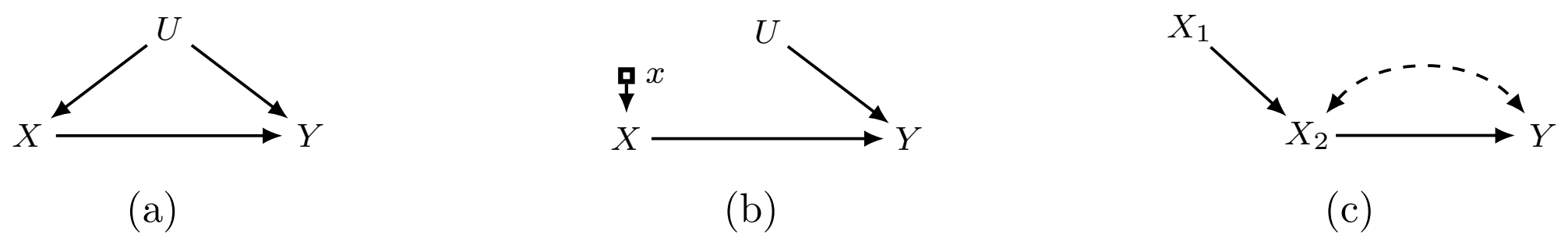
We begin by noting that a standard bandit can be viewed as the simple causal model shown in \Cref{fig:scb_standard_bandit_a}. Interventions on $X$ correspond to actions with $\Omega_X = \{x_1, x_2, \cdots, x_k\}$, and the action space $\mc{A} = \{a_i\}_{i=1}^k$ is defined as:
\begin{align}
    \mc{A} = \{a_i = do(x_i) : x_i \in \Omega_X\}.
\end{align}
The variable $Y$ denotes the reward, and $U$ represents an unobserved confounder. 
Then $\nu_{a_i}$ can be written as follows:
\begin{align}
    \nu_{a_i} = P(Y \mid do(x_i)) = P_{x_i}(Y). \label{eq:bandit is special case}
\end{align}
\Cref{eq:bandit is special case} states that a standard bandit can be modeled as an SCM associated with the two-node causal diagram in \Cref{fig:foundation_causal_diagram_c}, highlighting that the classical formulation is a special case of a causal model. 
\begin{example}[\textbf{Clinical system; continued}]
    Revisit the clinical-system example. This can be seen as an example of a bandit problem. The environment corresponds to SCM $\mc{M}$ in \Cref{eq:clinical trial scm} and the action space is defined as:
    \begin{align}
        \mc{A} = \{do(X=0), do(X=1)\}.
    \end{align}
    Furthermore, the bandit instance can be written as follows:
    \begin{align}
        \nu = \{
            P_{X=0}(Y) = \Bern(0.4),
            P_{X=1}(Y) = \Bern(0.4 - \eta)\},
    \end{align}
    as shown in \Cref{eq:clinical trial prob do x=0,eq:clinical trial prob do x=1}. 
\end{example}
A careful reader may naturally ask:
\begin{quote}
    \centering
    \bf
    \textit{``Why should we restrict ourselves to this simple setting?''} 
\end{quote}
In particular, there is no need to limit the analysis to a two-node structure with a single manipulable variable $X$. In what follows, we generalize the bandit formulation to arbitrary SCMs, which we refer to as structural causal bandits. 

\paragraph{Structural causal bandits.}
We now formalize the structural causal bandit (SCB) \citep{lee2018structural,lee2019structural} problem in which an agent interacts with an environment modeled by an SCM $\mc{M} = \langle \mf{U}, \mf{V}, \mc{F}, P(\mf{U})\rangle$ that includes a reward variable $Y \in \mf{V}$. In this setting, each action corresponds to an intervention on some set of variables $\mf{X} \subseteq \mf{V} \setminus \{Y\}$, with an assigned value $\mf{x} \in \Omega_{\mf{X}}$. That is:
\begin{align}
    \mc{A} = \{do(\mf{x}) : \mf{x} \in \Omega_{\mf{X}}, \mf{X} \subseteq \mf{V} \setminus \{Y\} \}.
\end{align}
We refer to each \(\mf{X}\) as an \textit{intervention set} (IS). We use the terms \textit{arm}, \textit{action}, and \textit{intervention} interchangeably, depending on the context. The mean reward of an action is denoted by $\mu_\mf{x} = \mb{E}[Y \mid do(\mf{x})]$. Whenever a maximizing action is used, we assume that the maximum is attained and select one maximizer using a fixed tie-breaking rule. This is automatic for finite action spaces. For a fixed intervention set $\mf{X}$, we denote a \textbf{best action} by: 
\begin{align}
    \mf{x}^\ast \in \arg \max_{\mf{x} \in \Omega_{\mf{X}}} \mu_{\mf{x}}.
\end{align}
Similarly, we denote the \textbf{optimal action in $\mc{A}$} by: 
\begin{align}
    \mf{x}^\star \in \operatorname*{arg\,max}_{\substack{\mf{x} \in \Omega_{\mf{X}}\\ \mf{X} \subseteq \mf{V}\setminus \{Y\}}}\mu_{\mf{x}}.
\end{align}
For convenience, we identify $do(\mf{x})$ with the variable-indexed assignment $\mf{x}$ whenever the intervention target is clear. Assignments to different variable sets remain distinct actions, even when their numerical values coincide. The empty assignment denotes $do(\emptyset)$. 

Let $A_t \in \mc{A}$ denote the random intervention selected at round $t$, using the assignment notation above. We define the suboptimality gap by $\Delta_\mf{x} = \mu_{\mf{x}^\star} - \mu_{\mf{x}}$ and the cumulative regret as follows: 
\begin{align}
    R_T^{\mc{A}} = T\mu_{\mf{x}^\star} - \sum_{t=1}^T \mu_{A_t} = \sum_{\mf{x} \in \mc{A}} \Delta_\mf{x} N_T(\mf{x}),
\end{align}
where $N_T(\mf{x})$ denotes the number of times $do(\mf{x})$ is selected up to time $T$. 

In this setting, the agent knows the causal diagram $\mc{G}$ and the reward variable $Y$. The diagram $\mc{G}$ specifies the environment class $\mc{C}$, consisting of SCMs conforming to $\mc{G}$, while the structural functions $\mc{F}$ and the exogenous distribution $P(\mf{U})$ remain unknown. Thus, the agent has access to qualitative structural information but not to the quantitative parameterization that determines the reward distributions. Graphical characterizations range over compatible SCMs, with domains chosen as part of the model; a particular bandit instance fixes these domains. 

\begin{example}[\bf Two manipulable variables] 
Consider an MAB environment induced by the following SCM $\mc{M}$:
\begin{align}
    \mc{M}
= \begin{cases}
        \mf{U} =   \{U_1,U_2\}\\
        \mf{V} = \{X_1,X_2,Y\} \\
        \mc{F} = 
        \begin{cases}
            f_{X_1}(u_1) = u_1\\
            f_{X_2}(x_1,u_2) = x_1 \oplus u_2\\
            f_{Y}(x_2,u_2) = x_2 \oplus u_2\\
        \end{cases}\\
        P(\mf{U}) = \{U \thicksim \Bern(0.5) : \forall U \in \mf{U}\}
\end{cases}\label{eq:iv scm}
\end{align}
where $\oplus$ denotes the exclusive-or operator. Assume that $U_1$ and $U_2$ are mutually independent. The causal diagram $\mc{G}$ associated with $\mc{M}$ is shown in \Cref{fig:scb_standard_bandit_c}. Consider an agent deployed in $\mc{M}$ whose goal is to minimize expected cumulative regret:
\begin{align}
    \mb{E}R_T^{\mc{A}} = \sum_{\mf{x} \in \mc{A}} \Delta_{\mf{x}} \mb{E}N_T(\mf{x})
\end{align}
by intervening on the manipulable variables $\mf{V}\setminus\{Y\} = \{X_1,X_2\}$. The collection of intervention sets is given by: 
\begin{align}
    \{\emptyset, \{X_1\}, \{X_2\}, \{X_1, X_2\}\}.
\end{align}
This collection is precisely the power set of the manipulable variables, and therefore contains $2^{\vert\{X_1, X_2\}\vert} = 4$ possible intervention sets. This setting can be translated into a traditional MAB instance in which each action corresponds to assigning specific values to a subset of $\{X_1, X_2\}$ under intervention. In this example, this yields $9$\footnote{That is, $1+2+2+4$, corresponding to the numbers of actions induced by each of the four intervention sets. } actions: 
\begin{align}
    \mc{A} = \{\mf{x} \in \{0,1\}^{\lvert \mf{X} \rvert}, \mf{X} \in \{\emptyset, \{X_1\}, \{X_2\}, \{X_1, X_2\}\}\}. 
\end{align}
\end{example}
We provide two structural properties that arise in the SCB setting: \textit{equivalence} and \textit{partial orders}. These properties capture dependencies among different actions and can be understood through \textit{do-calculus} \citep{pearl:95} and related tools. 
\begin{theorem}[\textit{do}-calculus \citep{pearl:95}]\label{thm: do-calculus for cd}
    Let $\mc{G}$ be a causal diagram of $\mc{M}$, with endogenous variables $\mf{V}$. For any disjoint sets $\mf{X}, \mf{Y}, \mf{W}, \mf{Z} \subseteq \mf{V}$, the following rules are valid whenever the displayed conditional distributions are defined. 
    \begin{itemize}[leftmargin = 20mm]
        \item[\textbf{Rule~1.}] 
            $P(\mf{y} \mid do(\mf{w}), \mf{x}, \mf{z}) = P(\mf{y} \mid do(\mf{w}), \mf{z})$
                \hfill \textbf{if} $(\mf{X} \Perp \mf{Y} \mid \mf{W} \cup \mf{Z})_{\mc{G}_{\overline{\mf{W}}}}$. 
        \item[\textbf{Rule~2.}]
            $P(\mf{y} \mid do(\mf{w},\mf{x}), \mf{z}) = P(\mf{y} \mid do(\mf{w}), \mf{x}, \mf{z})$ \hfill \textbf{if} $(\mf{X} \Perp \mf{Y} \mid \mf{W} \cup \mf{Z})_{\mc{G}_{\overline{\mf{W}}\underline{\mf{X}}}}$. 
        \item[\textbf{Rule~3.}] 
            $P(\mf{y} \mid do(\mf{w},\mf{x}), \mf{z}) = P(\mf{y} \mid do(\mf{w}), \mf{z})$ 
                \hfill \textbf{if}
                $(\mf{X} \Perp \mf{Y} \mid \mf{W} \cup \mf{Z})_{\mc{G}_{\overline{\mf{W}},\overline{\mf{X(Z)}}}}$\\
            where $\mf{X(Z)} \triangleq \mf{X} \setminus \An(\mf{Z})_{\mc{G}[\mf{V} \setminus \mf{W}]}$. 
    \end{itemize}
\end{theorem}
To provide intuition, let us consider the following example, where the action space can be reduced using graphical information derived from these properties. The first reduction corresponds to removing intervention sets that are \textit{equivalent} to some proper subset, while the second corresponds to removing intervention sets that are \textit{never strictly better} than some other intervention set. 
\begin{example}[\textbf{Continued}]
The main question is whether one can do better by exploiting the causal invariance implied by $\mc{M}$ and represented by $\mc{G}$. 

\bigskip
\noindent\textbf{Property~1.} First, using Rule~3 of do-calculus, we have:
\begin{align}
    P(y \mid do(x_1,x_2)) = P(y \mid do(x_2)),
\end{align}
since $(X_1 \Perp Y)_{\mc{G}_{\overline{X_2}}}$. In words, intervening on $X_1$ has no additional effect on $Y$ once $X_2$ is intervened upon. Therefore, the corresponding expected rewards are equal:
\begin{align}
    \mu_{x_1,x_2} = \mu_{x_2},     
\end{align}
for any $x_1 \in \Omega_{X_1}$ and $x_2 \in \Omega_{X_2}$. Consequently, actions of the form $do(x_1,x_2)$ are redundant, and it suffices to consider only the $5$ actions induced by the remaining three intervention sets:
\begin{align}
    \mc{A} = \{\mf{x} \in \{0,1\}^{\lvert \mf{X} \rvert}, \mf{X} \in \{\emptyset, \{X_1\}, \{X_2\}\}\}. \label{eq: iv scm mis}
\end{align}

\bigskip
\noindent\textbf{Property~2.} Next, the causal structure implies that the observational reward distribution $P(y)=P(y\mid do(\emptyset))$ can be written as a convex combination of $\{P_{x_1}(y)\}_{x_1\in\Omega_{X_1}}$:
\begin{align}
    P(y) & = \sum_{x_1} P(y \mid x_1) P(x_1) \\
        & = \sum_{x_1} P_{x_1}(y) P(x_1),
\end{align}
by Rule~2 of do-calculus with $(X_1 \Perp Y)_{\mc{G}_{\underline{X_1}}}$. Equivalently, $\mu_\emptyset = \sum_{x_1}\mu_{x_1}P(x_1)$. Thus, we can derive: 
\begin{align}
    \mu_\emptyset &= \sum_{x_1}\mu_{x_1}P(x_1)\\
    &\leq \sum_{x_1}\mu_{x_1^\ast}P(x_1)\\
    &= \mu_{x_1^\ast}. 
\end{align}
Hence, $do(\emptyset)$ is never better than $do(x_1^\ast)$ and can therefore be excluded from consideration. Thus, it suffices to consider only the $4$ actions induced by the remaining two intervention sets: 
\begin{align}
    \mc{A} = \{\mf{x} \in \{0,1\}^{\lvert \mf{X} \rvert}, \mf{X} \in \{\{X_1\}, \{X_2\}\}\}. \label{eq: iv scm pomis}
\end{align}
This conclusion holds for any SCM consistent with $\mc{G}$, showing how causal structure can reduce the action space while retaining at least one optimal action. Eliminated actions may tie with retained actions, but cannot achieve a higher reward than the best retained action.
\end{example}

We now formalize these two reductions, beginning with equivalence among actions. 

\subsection{Property 1: Equivalence among Actions}

We begin by recalling Rule~3 of do-calculus, which gives conditions under which an intervention has no effect on the outcome variable. Since actions correspond to interventions (including the null intervention), we consider whether: 
\begin{align}
    P(y \mid do(\mf{x}, \mf{w})) = P(y \mid do(\mf{x}))
\end{align}
holds. By Rule~3 of do-calculus, this equality is valid if $(Y \Perp \mf{W} \mid \mf{X})_{\mc{G}_{\overline{\mf{X} \cup \mf{W}}}}$, which implies that $\mu_{\mf{x} \cup \mf{w}} = \mu_{\mf{x}}$. When this condition holds, the actions $do(\mf{x}, \mf{w})$ and $do(\mf{x})$ are equivalent in terms of their induced reward distributions. Consequently, it suffices to play one action from each equivalence class. 

\paragraph{Minimal intervention sets.}
Ideally, one would like to restrict attention to a smaller subset of actions, provided that it still contains an optimal action. Motivated by this observation, we introduce the notion of a minimal intervention set (MIS). 
\begin{definition} [Minimal intervention set \citep{lee2018structural}] \label{def: mis cd}
    A set $\mf{X} \subseteq \mf{V} \setminus \{Y\}$ is an MIS relative to $\langle \mc{G}, Y \rangle$ if there is no proper subset $\mf{Z} \subsetneq \mf{X}$ such that, for every SCM conforming to $\mc{G}$ and every assignment $\mf{x} \in \Omega_\mf{X}$, $\mu_{\mf{x}} = \mu_{\mf{x} \cap \mf{Z}}$\footnote{We refer to $\mf{X}$ and $\mf{Z}$ as \textit{equivalent} if the equality holds.}. 
\end{definition}
We denote by $\mb{M}_{\mc{G},Y}$ the collection of MISs with respect to $\langle \mc{G},Y \rangle$. In words, an MIS is a set for which no proper subset is equivalent in terms of the induced reward. Whether a set is an MIS can be determined through a simple procedure based on ancestral relationships. 
\begin{proposition}[Proposition 1 in \citet{lee2018structural}]\label{prop: MIS}
    $\mf{X} \subseteq \mf{V} \setminus \{Y\}$ is an MIS with respect to $\langle \mc{G}, Y \rangle$ \textit{if and only if} $\mf{X} \subseteq \An(Y)_{\mc{G}_{\overline{\mf{X}}}}$. 
\end{proposition}
Intuitively, this characterization states that, for $\mf{X}$ to be an MIS, each variable $X \in \mf{X}$ must have a directed path to $Y$ that does not pass through any other variables in $\mf{X} \setminus \{X\}$. Such a path is called \textit{proper} with respect to $\mf{X}$: only its starting vertex belongs to $\mf{X}$. Note that $\emptyset$ and every singleton $\{X\}$ with $X\in\An(Y)_{\mc{G}}\setminus\{Y\}$ are MISs.  
\begin{example}[\textbf{Continued}] Let us consider the causal diagram $\mc{G}$ shown in \Cref{fig:scb_standard_bandit_c}. Since $\emptyset$, $\{X_1\}$, and $\{X_2\}$ are MISs in this graph, it remains only to test $\{X_1,X_2\}$: 
\begin{align}
    \{X_1,X_2\} \nsubseteq \An(Y)_{\mc{G}_{\overline{X_1, X_2}}} = \{X_2, Y\}. 
\end{align}
Therefore, it follows that: 
\begin{align}
    \mb{M}_{\mc{G},Y} = \{\emptyset, \{X_1\}, \{X_2\}\},
\end{align}
and the corresponding action set $\mc{A}$ is given in \Cref{eq: iv scm mis}.
\end{example}

\subsection{Property 2: Partial Orders among MISs} 
\paragraph{Possibly-optimal minimal intervention sets.} 
We now explore partial orders among MISs. Given a causal diagram $\mc{G}$, it is possible that interventions on certain variables always perform at least as well as interventions on others, regardless of the parameterization of the underlying model. Formally, there may exist two sets of variables $\mf{W}$ and $\mf{X}$ such that:
\begin{align}
\mu_{\mf{w}^\ast} \leq \mu_{\mf{x}^\ast} 
\end{align}
for every SCM consistent with $\mc{G}$. In this case, it is unnecessary to consider actions over $\Omega_{\mf{W}}$. Motivated by this observation, we define the notion of \textit{possibly-optimal} MIS, which captures this partial order among MISs. 
\begin{definition} [POMIS \citep{lee2018structural}] \label{def: POMIS for CD}
    Let $\mf{X} \subseteq \mf{V} \setminus \{Y\}$ be an MIS with respect to $\langle \mc{G}, Y \rangle$. If there exists an SCM conforming to $\mc{G}$ such that $\mu_{\mf{x}^\ast} > \mu_{\mf{w}^\ast}$ for all $\mf{W} \in \mb{M}_{\mc{G},Y} \setminus \{\mf{X}\}$ where $\mb{M}_{\mc{G},Y}$ is the collection of MISs with respect to $\langle \mc{G},Y\rangle$, then $\mf{X}$ is a \textit{possibly-optimal minimal intervention set} (POMIS) with respect to $\langle \mc{G}, Y \rangle$.
\end{definition}
We denote by $\mb{P}_{\mc{G},Y}$ the collection of POMISs with respect to $\langle \mc{G},Y\rangle$. Possible optimality means that the graph alone cannot rule out the intervention set; the best action in the actual SCM must still be learned. 

Consider an MIS $\mf{W} \in \mb{M}_{\mc{G},Y}$ and a disjoint set of variables $\mf{X}$. It follows that:
\begin{align}
    \mu_{\mf{w}} = \sum_{\mf{x}} \mb{E}_{\mf{w}}[Y \mid \mf{x}] P_{\mf{w}}(\mf{x}).
\end{align}
Suppose that, by Rule~2 of do-calculus, conditioning on $\mf{x}$ can be exchanged with intervention. Then, we can derive the following for every fixed $\mf{w}$:
\begin{align}
    \mu_{\mf{w}} &= \sum_{\mf{x}} \mb{E}_{\mf{w}}[Y \mid \mf{x}]P_{\mf{w}}(\mf{x})\\
    &= \sum_{\mf{x}} \mu_{\mf{x}\mf{w}} P_{\mf{w}}(\mf{x}) && \textbf{Rule~2}\label{eq:rule2 pomis}\\
    &\leq \sum_{\mf{x}} \mu_{\{\mf{x}\mf{w}\}^\ast} P_{\mf{w}}(\mf{x})\\
    &= \mu_{\{\mf{x}\mf{w}\}^\ast}.
\end{align}
Let $\mf{Z} \subseteq \mf{X} \cup \mf{W}$ be an MIS equivalent to $\mf{X} \cup \mf{W}$. Then, we have: 
\begin{align}
    \mu_{\mf{w}^\ast} \leq \mu_{\mf{z}^\ast}.
\end{align}
To identify such $\mf{X}$, we examine the condition for Rule~2 of do-calculus $(Y \Perp \mf{X} \mid \mf{W})_{\mc{G}_{\overline{\mf{W}}\underline{\mf{X}}}}$, which is equivalent to:
\begin{align}
    (Y \Perp \mf{X})_{\mc{G}'_{\underline{\mf{X}}}},
\end{align}
where $\mc{G}' = \mc{G} \setminus \mf{W}$. Therefore, if every path from $\mf{X}$ to $Y$ in $\mc{G}'_{\underline{\mf{X}}}$ is blocked, then such an MIS $\mf{Z}$ satisfies:
\begin{align}
    \mu_{\mf{w}^\ast} \leq  \mu_{\{\mf{x},\mf{w}\}^\ast} = \mu_{\mf{z}^\ast}. 
\end{align}
Building on this observation, we introduce two graphical concepts for characterizing POMISs. Let $\cc(X)_{\mc{G}}$ be the c-component \citep{tian:02} containing $X$ in $\mc{G}$, namely, the maximal set of vertices connected to $X$ by bidirected paths, and $\cc(\mf{X})_{\mc{G}} = \bigcup_{X \in \mf{X}} \cc(X)_{\mc{G}}$. 

\begin{definition} [Unobserved confounders' territory \citep{lee2018structural}] \label{def: MUCT for cd}
    Given information $\langle \mc{G}, Y\rangle$, let $\mc{H} = \mc{G}[\An(Y)_{\mc{G}}]$. A set of variables $\mf{T} \subseteq \mf{V}(\mc{H})$ containing $Y$ is called a \textit{UC-territory} on $\mc{G}$ with respect to $Y$ if $\De(\mf{T})_\mc{H} = \mf{T}$ and $\cc(\mf{T})_{\mc{H}} = \mf{T}$. A UC-territory $\mf{T}$ is said to be \textit{minimal} if no $\mf{T}' \subsetneq \mf{T}$ is a UC-territory.
\end{definition}
\begin{definition} [Interventional border \citep{lee2018structural}] \label{def: IB for cd}
    Let $\mf{T}$ be a minimal UC-territory on the causal diagram $\mc{G}$ with respect to $Y$. Then $\mf{W} = \Pa(\mf{T})_{\mc{G}} \setminus \mf{T}$ is called an \textit{interventional border} (IB) for $\mc{G}$ with respect to $Y$.
\end{definition}
The minimal unobserved confounders' territory (MUCT) is the smallest set containing $Y$ that is closed under descendants and c-components within $\mc{G}[\An(Y)_{\mc{G}}]$. 

To compute it, start from $\{Y\}$ in $\mc{H} = \mc{G}[\An(Y)_{\mc{G}}]$ and repeatedly add the c-components and descendants of the current set in $\mc{H}$ until the set stops changing. Every UC-territory contains each stage of this construction, so its limit is the unique MUCT. Its closure properties support the use of Rule~2 to exchange conditioning on its border with intervention. 

We denote by $\MUCT(\mc{G}, Y)$ and $\IB(\mc{G}, Y)$ the MUCT and IB with respect to $\langle \mc{G}, Y \rangle$, respectively. The following theorem provides a complete characterization of POMISs. 
\begin{theorem}[Theorem 6 in \citet{lee2018structural}]\label{thm: IB = POMIS for CD}
    $\mf{X} \subseteq \mf{V} \setminus \{Y\}$ is a POMIS with respect to $\langle \mc{G},Y \rangle$ if and only if $\IB(\mc{G}_{\overline{\mf{X}}},Y) = \mf{X}$. 
\end{theorem}
\begin{example}[\textbf{Continued}] Again, consider the causal diagram $\mc{G}$ shown in \Cref{fig:scb_standard_bandit_c}. We now test whether each MIS in $\mb{M}_{\mc{G},Y} = \{\emptyset, \{X_1\}, \{X_2\}\}$ is a POMIS as follows:
\begin{align}
    \text{For $\mf{X} = \emptyset$,~}\IB(\mc{G}_{\overline{\mf{X}}},Y) = \{X_1\} &\neq \mf{X};\\
        \text{For $\mf{X} = \{X_1\}$,~}\IB(\mc{G}_{\overline{\mf{X}}},Y) = \{X_1\} &= \mf{X};\\
        \text{For $\mf{X} = \{X_2\}$,~}\IB(\mc{G}_{\overline{\mf{X}}},Y) = \{X_2\} &= \mf{X}.
\end{align}
Therefore, the collection of POMISs is:  
\begin{align}
    \mb{P}_{\mc{G},Y} = \{\{X_1\}, \{X_2\}\},
\end{align}
and the corresponding action set $\mc{A}$ is given in \Cref{eq: iv scm pomis}. 
\end{example}

\begin{figure}[t]
    \centering
    \tikzset{
        scbmuctgraph/.style={x=0.8cm,y=0.6cm,font=\scriptsize,
            >={Latex[width=1.3mm,length=1.5mm]},line width=0.25mm},
        scbmuctnode/.style={draw=none,circle,fill=white,inner sep=0mm,minimum size=5mm},
        scbmuctterritory/.style={scbmuctnode,draw,fill=none},
        scbmuctborder/.style={scbmuctnode,fill=bettergreen!25,draw},
        scbmuctintervened/.style={draw,line width=0.6mm},
        scbmuctconfounded/.style={<->,dashed},
        scbmuctarea/.style={draw=betterblue!10,fill=betterblue!10,
            line width=6.5mm,line cap=round,line join=round}
    }
    \newcommand{\scbmuctpositions}{
        \coordinate (A) at (0,1.2);
        \coordinate (B) at (0.8,2.4);
        \coordinate (D) at (1.5,1.2);
        \coordinate (C) at (3.2,1.0);
        \coordinate (E) at (2.2,0);
        \coordinate (F) at (0.6,-1.2);
        \coordinate (Y) at (3,-1.2);
        \path[use as bounding box]
            ($(A |- Y)+(-7mm,-5mm)$) rectangle ($(C |- B)+(8mm,5mm)$);
    }
    \newcommand{\scbmuctdirected}{
        \draw[->] (A) -- (B);
        \draw[->] (B) -- (D);
        \draw[->] (D) -- (E);
        \draw[->] (C) -- (E);
        \draw[->] (E) -- (Y);
        \draw[->] (F) -- (Y);
    }
    \newcommand{\scbmuctcy}{
        \draw[scbmuctconfounded] (C) to[bend left=45]
            coordinate[pos=0.5] (scbmuctCYmid) (Y);
    }
    \newcommand{\scbmuctbc}{
        \draw[scbmuctconfounded] (B) to[bend left=40]
            coordinate[pos=0.5] (scbmuctBCmid) (C);
    }
    \newcommand{\scbmuctad}{
        \draw[scbmuctconfounded] (A) to[bend right=25] (D);
    }
    \newcommand{\scbmuctaf}{
        \draw[scbmuctconfounded] (A) to[bend right=45] (F);
    }
    \newcommand{\scbmuctfe}{
        \draw[scbmuctconfounded] (F) -- (E);
    }
    \newcommand{\scbmuctalledges}{
        \scbmuctdirected\scbmuctad\scbmuctbc\scbmuctcy\scbmuctaf\scbmuctfe
    }
    \newcommand{\scbmuctcontext}{
        \begin{scope}[draw=gray!30]
            \scbmuctalledges
        \end{scope}
    }
    \newcommand{\scbmuctregionB}{
        \begin{pgfonlayer}{back}
            \draw[scbmuctarea]
                (F.center) -- (E.center) -- (C.center) to[bend left=45] (Y.center) -- cycle;
        \end{pgfonlayer}
    }
    \newcommand{\scbmuctregionE}{
        \begin{pgfonlayer}{back}
            \fill[scbmuctarea,draw=none]
                (Y.center) circle[radius=0.5\pgflinewidth];
        \end{pgfonlayer}
    }
    \newcommand{\scbmuctregionF}{
        \begin{pgfonlayer}{back}
            \draw[scbmuctarea]
                (D.center) to[bend left=25] (A.center) -- (B.center) to[bend left=40]
                (C.center) to[bend left=45] (Y.center) --
                (E.center) -- cycle;
        \end{pgfonlayer}
    }
    \begin{minipage}[b]{0.47\textwidth}\centering
    \subfloat[$\mc{G}$]{
        \begin{tikzpicture}[scbmuctgraph]
            \scbmuctpositions
            \foreach \v in {A,B,C,D,E,F,Y}
                \node[scbmuctnode] (\v) at (\v) {$\v$};
            \scbmuctalledges
        \end{tikzpicture}\label{fig: scb muct ib a}}
    \end{minipage}
    \hfill
    \begin{minipage}[b]{0.47\textwidth}\centering
    \subfloat[Under intervention on $B$]{
        \begin{tikzpicture}[scbmuctgraph]
            \scbmuctpositions
            \node[scbmuctnode,text=gray!30] (A) at (A) {$A$};
            \foreach \v in {C,E,F,Y}
                \node[scbmuctterritory] (\v) at (\v) {$\v$};
            \node[scbmuctnode,scbmuctintervened] (B) at (B) {$B$};
            \node[scbmuctborder] (D) at (D) {$D$};
            \scbmuctcontext
            \draw[->] (B) -- (D);
            \draw[->] (D) -- (E);
            \draw[->] (C) -- (E);
            \draw[->] (E) -- (Y);
            \draw[->] (F) -- (Y);
            \scbmuctfe\scbmuctcy
            \scbmuctregionB
        \end{tikzpicture}\label{fig: scb muct ib b}}
    \end{minipage}

    \medskip
    \begin{minipage}[b]{0.47\textwidth}\centering
    \subfloat[Under intervention on $E$]{
        \begin{tikzpicture}[scbmuctgraph]
            \scbmuctpositions
            \foreach \v in {A,B,C,D}
                \node[scbmuctnode,text=gray!30] (\v) at (\v) {$\v$};
            \node[scbmuctborder,scbmuctintervened] (E) at (E) {$E$};
            \node[scbmuctborder] (F) at (F) {$F$};
            \node[scbmuctterritory] (Y) at (Y) {$Y$};
            \scbmuctcontext
            \draw[->] (E) -- (Y);
            \draw[->] (F) -- (Y);
            \scbmuctregionE
        \end{tikzpicture}\label{fig: scb muct ib c}}
    \end{minipage}
    \hfill
    \begin{minipage}[b]{0.47\textwidth}\centering
    \subfloat[Under intervention on $F$]{
        \begin{tikzpicture}[scbmuctgraph]
            \scbmuctpositions
            \foreach \v in {A,B,C,D,E,Y}
                \node[scbmuctterritory] (\v) at (\v) {$\v$};
            \node[scbmuctborder,scbmuctintervened] (F) at (F) {$F$};
            \scbmuctcontext
            \scbmuctdirected\scbmuctad\scbmuctbc\scbmuctcy
            \scbmuctregionF
        \end{tikzpicture}\label{fig: scb muct ib d}}
    \end{minipage}
    \caption{\small MUCTs and IBs. Panels (b)--(d) show the ancestral subgraphs of the manipulated graphs. Light blue regions mark MUCTs, green fill marks IB variables, and thick outlines mark intervention targets.}
    \label{fig: scb muct ib}
\end{figure}
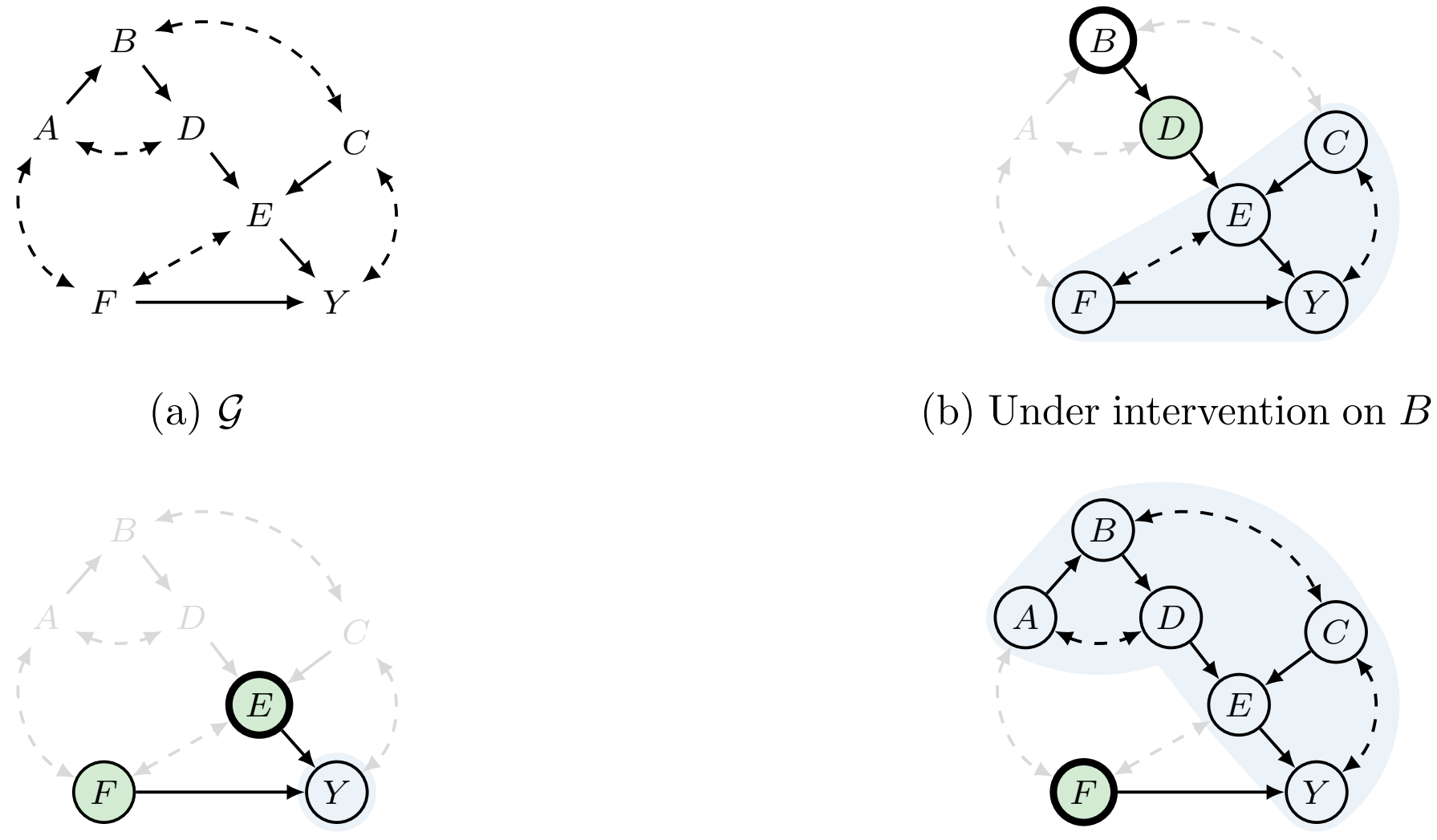

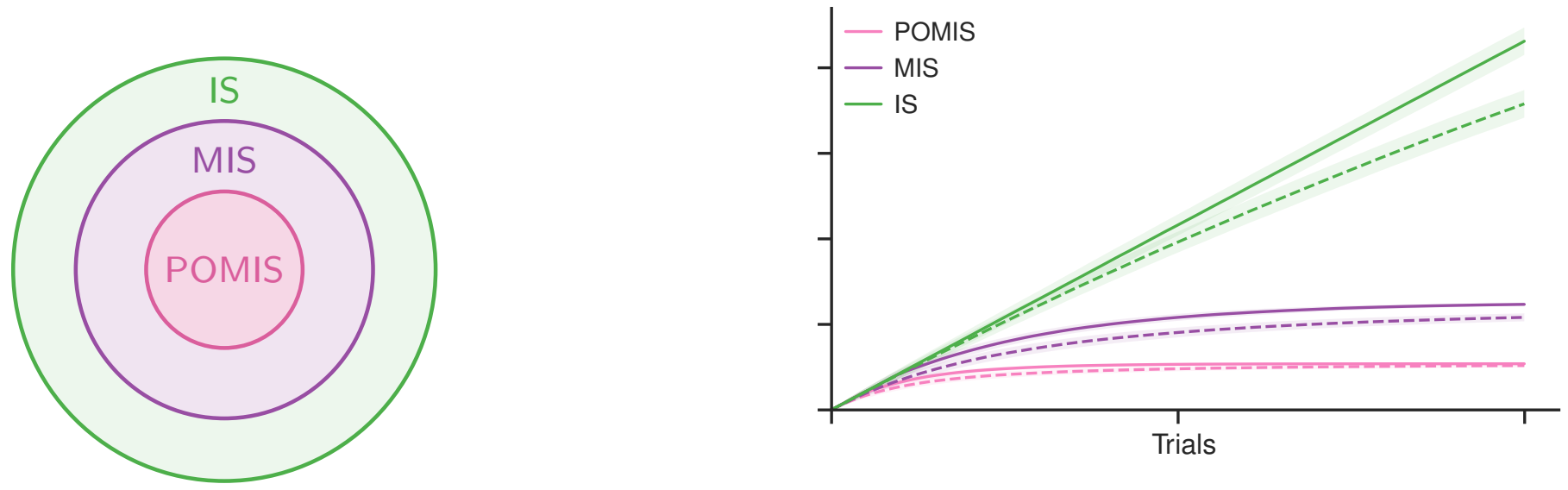
\begin{figure}[t]
    \hfill
    \begin{minipage}[b]{0.4\textwidth}\centering
    \subfloat[Inclusion]{
        \begin{tikzpicture}[
        x=1.2cm,y=1.2cm,
        font=\sffamily\scriptsize,
        line width=0.3mm]
        \def\rIS{1.35}
        \def\rMIS{0.95}
        \def\rPOMIS{0.50}
    
        \fill[bettergreen!10]  (0,0) circle (\rIS);
        \fill[betterpurple!15] (0,0) circle (\rMIS);
        \fill[betterpink!25]   (0,0) circle (\rPOMIS);
    
        \draw[bettergreen]  (0,0) circle (\rIS);
        \draw[betterpurple] (0,0) circle (\rMIS);
        \draw[betterpink]   (0,0) circle (\rPOMIS);
    
        \node[text=bettergreen] at (0,1.15) {IS};
        \node[text=betterpurple] at (0,0.7) {MIS};
        \node[text=betterpink] at (0,0) {POMIS};
        
    \end{tikzpicture}\label{fig:scb_inclusion}}
    \end{minipage}
    \hfill
    \begin{minipage}[b]{0.4\textwidth}
    \subfloat[Regret]{
        \centering
        \includegraphics[width=\textwidth]{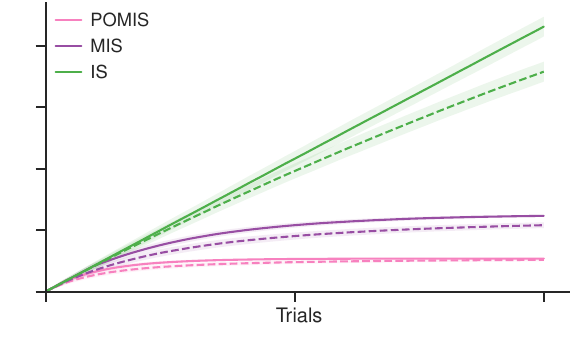}%
        \label{fig:scb_regret}}
    \end{minipage}
    \hfill\null
    \caption{\small (a) Inclusion of POMISs in MISs and of MISs in intervention sets. (b) Schematic comparison of expected cumulative regret for action spaces defined by all intervention sets (IS), MISs, and POMISs. Solid and dashed curves represent distinct policies, e.g., UCB and Thompson sampling. }
    \label{fig:scb_inclusion_and_regret}
\end{figure}

The following example applies the MUCT–IB criterion to several intervention sets in one confounded system. Consider $\mc{G}$ in \Cref{fig: scb muct ib a}. For each intervention set $\mf{X}$, we form $\mc{G}_{\overline{\mf{X}}}$ and compute its MUCT in the subgraph induced by the ancestors of $Y$ in that manipulated graph.  

In \Cref{fig: scb muct ib b}, $A$ is no longer an ancestor of $Y$. Starting from $\{Y\}$, the bidirected edge $C\bedge Y$ adds $C$, the directed edge $C\to E$ adds $E$, and $E\bedge F$ then adds $F$. Consequently:
\begin{align}
    \text{For $\mf{X} = \{B\}$,~} &\MUCT(\mc{G}_{\overline{\mf{X}}},Y) = \{C,E,F,Y\} \\
    &\Rightarrow \IB(\mc{G}_{\overline{\mf{X}}},Y) = \{D\} \neq \mf{X}.
\end{align}
The border gives the comparison $\mu_{b^\ast} \leq \mu_{d^\ast}$, i.e., intervening on $D$ can achieve at least the best reward available by intervening on $B$. Indeed, Rule~2 gives:
\begin{align}
    \mb{E}[Y \mid do(b), d] = \mb{E}[Y \mid do(b,d)],
\end{align}
and Rule~3 removes the intervention on $B$ once $D$ is intervened upon. Therefore, $\mu_b = \sum_{d} \mu_d P(d \mid do(b)) \leq \mu_{d^\ast}$. 

Only $E$, $F$, and $Y$ remain in the ancestral subgraph in \Cref{fig: scb muct ib c}. Thus,
\begin{align}
    \text{For $\mf{X} = \{E\}$,~} &\MUCT(\mc{G}_{\overline{\mf{X}}},Y) = \{Y\} \\
    &\Rightarrow \IB(\mc{G}_{\overline{\mf{X}}},Y) = \{E,F\} \neq \mf{X}.
\end{align}
Here the comparison is $\mu_{e^\ast}\leq\mu_{\{e,f\}^\ast}$, so an optimal joint intervention on $E$ and $F$ performs at least as well as an optimal intervention on $E$ alone. 

The territory in \Cref{fig: scb muct ib d} contains all vertices except $F$:
\begin{align}
    \text{For $\mf{X} = \{F\}$,~} &\MUCT(\mc{G}_{\overline{\mf{X}}},Y) = \{A,B,C,D,E,Y\} \\
    &\Rightarrow \IB(\mc{G}_{\overline{\mf{X}}},Y) = \{F\} = \mf{X}.
\end{align}
By \Cref{thm: IB = POMIS for CD}, $\{F\}$ is a POMIS, whereas $\{B\}$ and $\{E\}$ are not. This illustrates how the MUCT and IB support action space pruning before the agent estimates reward distributions: they identify intervention sets that can be excluded and those that must remain candidates for optimality. 

Equipped with \Cref{thm: IB = POMIS for CD}, one can systematically reduce the action space using the information $\langle \mc{G}, Y \rangle$. Restricting attention to POMISs preserves the optimal achievable reward while reducing the candidate action space. The schematic comparison in \Cref{fig:scb_inclusion_and_regret} illustrates the potential reduction in expected cumulative regret from restricting the action space to MISs or POMISs.

\jmlrsection{Structural Causal Bandits under Markov Equivalence}\label{chap:scb mec}
The SCB formulation in the preceding \unitname{} assumes that the agent knows the causal diagram. Under the causal Markov and faithfulness assumptions, observational conditional independences generally identify only a Markov equivalence class. Causal sufficiency, the absence of unobserved common causes of the observed variables, restricts the graph class but does not, by itself, resolve all orientations \citep{verma1990equivalence,spirtes2000causation,chickering2002optimal,tsamardinos2006max}. Additional assumptions on functional forms, noise distributions, or variation across environments may provide further identification \citep{perry2022causal,peters2016causal,ghassami2017learning,heinze2018invariant,huang2020causal,ghassami2018multi,zeng2021causal}. In the presence of latent variables and without selection bias, we represent the observational Markov equivalence class by a PAG. 

Under these assumptions, observational conditional-independence information identifies a PAG, which may encode a super-exponential number of maximal ancestral graphs (MAGs). Each such MAG, in turn, represents an \textit{infinite} number of causal diagrams over supersets of the observed variables. Therefore, considering all causal diagrams consistent with a PAG is computationally prohibitive. Directly identifying conditions for MIS and POMIS at the level of ancestral graphs would circumvent this difficulty. Our key contributions in this \unitname{} are as follows: 
\begin{itemize}
    \item We generalize MIS and develop graphical criteria for ancestral graphs, enabling an agent to identify and exclude variables that have no effect on the reward. 
    
    \item We extend POMIS to ancestral graphs and establish a necessary and sufficient graphical condition for MAGs. 
    
    \item We present a search algorithm that is sound and complete for POMIS membership in at least one represented MAG. 
\end{itemize}

\paragraph{Relation to the published version.}
This \unitname{} revises \citet{park2025structural}. The counterexample in \Cref{sec:dmis-char} disproves the sufficiency of Theorem~2 of \citet{park2025structural}.
For Proposition~6 of \citet{park2025structural}, only the implication from PAG-level POMIS membership to membership in some represented MAG is retained (\Cref{prop: if POMIS for PAG then POMIS for MAG}). The search associated with Theorem~4 of \citet{park2025structural} instead characterizes the union of the POMIS collections of the represented MAGs (\Cref{sec: pomis for pags}). Definition~6 of \citet{park2025structural} is corrected by removing $\mf{X}$ before computing possible ancestors, excluding paths that pass through intervened variables (\Cref{def: MUCT for PMG}). Our revision of Algorithm~1 of \citet{park2025structural} uses an MIS precheck, omits the preliminary orientation step, and tests proper paths and the corrected IB (\Cref{alg: ispomis,prop: local proper path}). Accordingly, \Cref{sec: scbmec experiment} reports corrected action spaces and new experiments with horizon $50{,}000$ and $100$ repetitions; these replace the published arm counts and regret results.

\section{Generalizing Minimal Intervention Sets}\label{sec: scbmec_mis}

In this section, we generalize the notion of a minimal intervention set (MIS) to ancestral graphs, so that it applies not only to causal diagrams but also to MAGs and PAGs. We first provide complete graphical conditions for MIS in MAGs and PAGs. We then show that MIS membership in a PAG need not have a common witnessing MAG. To address this issue, we introduce definitely minimal intervention sets (DMISs), which require a single compatible SCM to witness the absence of equivalence to every proper subset. Here, \textit{definitely} refers to this common witness, not to minimality in every MAG represented by the PAG. 
\subsection{Minimal Intervention Sets for Ancestral Graphs}

We use $\mc{D}$ to denote either a causal diagram or an ancestral graph (MAG or PAG) over $\mf{V}$. The existence arguments below quantify over the graphical model class, without fixing an observational distribution or a parametric family. 
\begin{definition}[MIS]\label{def: MIS ancestral}
Given $\langle\mc{D},Y\rangle$, a set $\mf{X}\subseteq\mf{V}\setminus\{Y\}$ is an MIS if, for every $\mf{Z}\subsetneq\mf{X}$, there exist an SCM $\mc{S}$ conforming to $\mc{D}$ and an assignment $\mf{x}\in\Omega_{\mf{X}}$ such that $\mu^{\mc{S}}_{\mf{x}}\neq \mu^{\mc{S}}_{\mf{x}\cap\mf{Z}}$\footnote{The superscript identifies the SCM used to evaluate the reward mean.}.  
\end{definition}

For any causal diagram, we can explicitly construct a compatible binary SCM $\mc{S}^\ast$ to provide the witnesses used below (\Cref{lem: binary scm construction}). Each variable equals one when its private uniform noise falls below a threshold that increases with its parents' values, while shared latent variables represent the bidirected edges. Setting a previously unintervened variable to one strictly increases its mean, and positive parental effects propagate this increase along directed paths. Along a proper directed path to $Y$, no other intervention interrupts this propagation, so the reward mean strictly increases. 

\paragraph{Minimal intervention sets for MAGs.} 
We now turn to the graphical characterization of MIS for MAGs. As in the case of causal diagrams, this characterization relies on the explicit ancestral relations among variables in MAGs and on Rule~3 of do-calculus for MAGs \citep{zhang2008a}: 
\begin{itemize}[leftmargin=20mm]
        \item[\textbf{Rule 3.}] 
            $P(\mf{y} \mid do(\mf{w},\mf{x}), \mf{z}) = P(\mf{y} \mid do(\mf{w}), \mf{z})$ 
                \hfill \textbf{if}
                $(\mf{X} \Perp \mf{Y} \mid \mf{W} \cup \mf{Z})_{\mc{M}_{\overline{\mf{W}},\overline{\mf{X(Z)}}}}$\\
            where $\mf{X(Z)} \triangleq \mf{X} \setminus \An(\mf{Z})_{\mc{M}[\mf{V} \setminus \mf{W}]}$. 
\end{itemize}
\begin{restatable}[Characterization of MIS for MAGs]{theorem}{MISforMAGiffAn}\label{thm: MIS for MAG iff An}
    Given information $\langle \mc{M},Y \rangle$, $\mf{X} \subseteq \mf{V} \setminus \{Y\}$ is an MIS relative to $\langle \mc{M}, Y\rangle$ \textit{if and only if} $\mf{X} \subseteq \An(Y)_{\mc{M}_{\overline{\mf{X}}}}$ holds.  
\end{restatable}
\begin{proof}[Proof of \Cref{thm: MIS for MAG iff An}]
    \textbf{(If)} Let $\mc{G}$ have exactly the directed and bidirected edges of $\mc{M}$, interpreted as a causal diagram. Its corresponding MAG is $\mc{M}$, since replacing each bidirected edge by a separate latent common parent preserves its separation model, and maximality and ancestrality recover the same adjacencies and marks. The two graphs have the same directed paths after deleting incoming edges to $\mf{X}$. 

    If $\mf{X} \subseteq \An(Y)_{\mc{M}_{\overline{\mf{X}}}}$, every $X \in \mf{X}$ has a proper directed path to $Y$.   Fix the SCM $\mc{S}^\ast$ provided by \Cref{lem: binary scm construction}, which conforms to $\mc{G}$ and hence to $\mc{M}$. For any proper subset $\mf{Z} \subsetneq \mf{X}$, choose $W \in \mf{X} \setminus \mf{Z}$. Because $W$ has a proper directed path to $Y$ with respect to $\mf{X}$, the lemma gives $\mu_{\mf{X}=\boldsymbol{1}}^{\mc{S}^\ast} > \mu_{\mf{Z}=\boldsymbol{1}}^{\mc{S}^\ast}$ where the subscripts denote interventions that set all variables in the indicated set to one. This proves MIS membership. If $\mf{X} = \emptyset$, there is no proper subset to check. 
    
    \textbf{(Only if)} Suppose $\mf{Z} := \mf{X} \setminus \An(Y)_{\mc{M}_{\overline{\mf{X}}}}$ is nonempty, and let $\mf{W} = \mf{X} \setminus \mf{Z}$. We show: 
    \begin{align}
        (Y \Perp \mf{Z} \mid \mf{W})_{\mc{M}_{\overline{\mf{X}}}}. 
    \end{align}
    Otherwise, an m-connecting path from $Z \in \mf{Z}$ to $Y$ would start with an outgoing edge, since all arrowheads into $\mf{X}$ have been removed. It could contain no collider: every conditioned vertex in $\mf{W}$ has no incoming edge, so no collider can be its ancestor or coincide with it. Such a path must therefore be directed towards $Y$ and avoid $\mf{W}$. This contradicts the choice of $Z$.

    Rule~3 for MAGs now gives $P(y \mid do(\mf{x})) = P(y \mid do(\mf{x} \cap \mf{W}))$ for every assignment $\mf{x}$ and every SCM conforming to $\mc{M}$. Since $\mf{W} \subsetneq \mf{X}$, $\mf{X}$ is not an MIS. 
\end{proof}

\begin{restatable}[]{proposition}{MISforMAGeqMISforCD}\label{prop: MIS for MAG eq MIS for CD}
    $\mf{X} \subseteq \mf{V} \setminus \{Y\}$ is an MIS relative to $\langle \mc{M}, Y\rangle$ \textit{if and only if} there exists a causal diagram $\mc{G} \in [\mc{M}]$ such that $\mf{X}$ is an MIS relative to $\langle \mc{G}, Y \rangle$.
\end{restatable}
\begin{proof}[Proof of \Cref{prop: MIS for MAG eq MIS for CD}]
    \textbf{(If)} Suppose that $\mf{X}$ is an MIS for a causal diagram $\mc{G} \in [\mc{M}]$. For every $\mf{Z} \subsetneq \mf{X}$, there exist an SCM conforming to $\mc{G}$ and an assignment $\mf{x}$ for which $\mu_{\mf{x}} \neq \mu_{\mf{x} \cap \mf{Z}}$. This SCM also conforms to $\mc{M}$; thus, the same witnesses establish that $\mf{X}$ is an MIS for $\mc{M}$. 

    \textbf{(Only if)} By \Cref{thm: MIS for MAG iff An}, $\mf{X} \subseteq \An(Y)_{\mc{M}_{\overline{\mf{X}}}}$. Take the causal diagram $\mc{G}$ with exactly the directed and bidirected edges of $\mc{M}$, as in that theorem's proof. Its corresponding MAG is $\mc{M}$, and the proper directed paths from $\mf{X}$ to $Y$ are unchanged. Thus, \Cref{prop: MIS} implies that $\mf{X}$ is an MIS for $\mc{G}$. 
\end{proof}
This proposition guarantees that if $\mf{X}$ is an MIS relative to $\langle \mc{M}, Y \rangle$ for a given MAG $\mc{M}$, then there exists a causal diagram $\mc{G}$ conforming to $\mc{M}$ such that $\mf{X}$ is an MIS relative to $\langle \mc{G}, Y \rangle$.

\begin{remark}
    Even if $\mf{X}$ is an MIS with respect to $\langle \mc{M}, Y \rangle$, there is no guarantee that it is also an MIS with respect to $\langle \mc{G}, Y \rangle$ for every causal diagram $\mc{G}$ consistent with $\mc{M}$. 
\end{remark}

\paragraph{Minimal intervention sets for PAGs.}
We proceed to the characterization of MIS for PAGs. Rule~3 for PAGs \citep{jaber2022causal} can establish interventional equivalence throughout the represented model class:
\begin{itemize}[leftmargin=20mm]
        \item[\textbf{Rule~3.}] 
            $P(\mf{y} \mid do(\mf{w},\mf{x}), \mf{z}) = P(\mf{y} \mid do(\mf{w}), \mf{z})$ 
                \hfill \textbf{if}
                $(\mf{X} \Perp \mf{Y} \mid \mf{W} \cup \mf{Z})_{\mc{P}_{\overline{\mf{W}},\overline{\mf{X(Z)}}}}$\\
            where $\mf{X(Z)} \triangleq \mf{X} \setminus \Poss\An(\mf{Z})_{\mc{P}[\mf{V} \setminus \mf{W}]}$. 
\end{itemize}
For example, in the PAG $X \cedge Z \cedge Y$, the only path from $X$ to $Y$ has $Z$ as a definite non-collider\footnote{Let $\langle X,Z,Y \rangle$ be any consecutive triple. We say that $Z$ is a \textit{definite collider} if both edges have arrowheads at $Z$. If (1) one of the edges is out of $Z$; or (2) both edges have a circle mark at $Z$ (i.e., $X \acedge Z \caedge Y$) and there is no edge between $X$ and $Y$, then we say that $Z$ is a \textit{definite non-collider}. A path is said to have a \textit{definite status} if every non-endpoint node along it is either a definite collider or a definite non-collider. In a PAG, m-separation is tested over definite-status paths: a path is m-connecting given $\mf{Z}$ if every non-collider is outside $\mf{Z}$ and every collider is an ancestor of $\mf{Z}$. If no such path connects the queried sets, they are m-separated.}. Thus, $(X \Perp Y \mid Z)_{\mc{P}_{\overline{\{X,Z\}}}}$, and Rule~3 gives $P(y \mid do(x,z)) = P(y \mid do(z))$. This excludes $\{X,Z\}$ from MIS membership. To establish a necessary and sufficient characterization, we also need compatible SCMs witnessing the inequalities required for MIS membership. 

For this purpose, we extend the definition of a proper path in \Cref{chap:scb} to possibly directed paths from $X \in \mf{X}$ to $Y$. Each such path is not disturbed by other intervening variables, and thus aligns with the characterizations of MIS for causal diagrams and MAGs. Let $\mc{P}$ be a PAG over the set of variables $\mf{V}$. 

\begin{restatable}[Graphical characterization of MIS for PAGs]{proposition}{MISforPAGeqMISforMAG}\label{prop: MIS for PAG}
    $\mf{X} \subseteq \mf{V} \setminus \{Y\}$ is an MIS relative to $\langle \mc{P}, Y\rangle$ \textit{if and only if} for every variable $X \in \mf{X}$, there exists a proper possibly directed path from $X$ to $Y$ with respect to $\mf{X}$ in $\mc{P}$.    
\end{restatable}   
\begin{proof}[Proof of \Cref{prop: MIS for PAG}]
    \textbf{(If)} Fix an arbitrary $\mf{Z} \subsetneq \mf{X}$ and choose $W \in \mf{X} \setminus \mf{Z}$. By hypothesis and \Cref{lem: zhang2008b lem b.1}, there is an uncovered proper possibly directed path $p$ from $W$ to $Y$ relative to $\mf{X}$. Taking a subsequence preserves properness. 

    Replace every $\cdedge$ edge in $\mc{P}$ by a directed edge and orient its circle components into DAGs without unshielded colliders, orienting every circle edge incident to $W$ out of $W$. Such a choice is possible by the construction in the proof of \Cref{lem: maathuis2015 lem 7.6}; \Cref{lem: zhang2008b thm 2} ensures that the resulting MAG $\mc{M}$ belongs to $[\mc{P}]$. The first edge of $p$ now points out of $W$. Each consecutive triple of $p$ is unshielded and cannot become a collider, so the subsequent edges must also point towards $Y$. Thus, $p$ is a proper directed path in $\mc{M}$. 

    Use the causal diagram with the same directed and bidirected edges as $\mc{M}$. By \Cref{lem: binary scm construction}, it admits an SCM $\mc{S}^\ast$ with $\mu_{\mf{X}=\boldsymbol{1}}^{\mc{S}^\ast} > \mu_{\mf{Z}=\boldsymbol{1}}^{\mc{S}^\ast}$. This SCM conforms to $\mc{P}$. Since $\mf{Z}$ was arbitrary, $\mf{X}$ is an MIS. The witnessing MAG and SCM may depend on $\mf{Z}$, as permitted by \Cref{def: MIS ancestral}. The case $\mf{X} = \emptyset$ is vacuous. 

    \textbf{(Only if)} Suppose that some $Z \in \mf{X}$ has no proper possibly directed path to $Y$. Any proper directed path in a represented MAG would appear as a proper possibly directed path in $\mc{P}$. Hence no $\mc{M} \in [\mc{P}]$ contains such a directed path from $Z$. The separation argument in the only-if part of \Cref{thm: MIS for MAG iff An}, applied to this vertex, gives $P(y \mid do(\mf{x})) = P(y \mid do(\mf{x} \cap (\mf{X} \setminus \{Z\})))$ for every assignment and every SCM conforming to any such MAG. The equality therefore holds throughout the model class of $\mc{P}$, so $\mf{X}$ is not an MIS. 
\end{proof}

\begin{algorithm}[t]
\SetAlgoNoEnd
\DontPrintSemicolon

\textbf{function~}\textsf{MISs}($\mc{D}$, $Y$)\;
\Indp

Set $\mf{A} = \Poss\An(Y)_{\mc{D}}$ and $\mc{H} = \mc{D}[\mf{A}]$.\;

Order $\mf{A} \setminus \{Y\}$ as $\ell$ from sinks to sources as described in the text. \;

\Return{\rm \textsf{subMISs}$(\mc{H}, Y, \emptyset, \ell)$}\;
\Indm

\BlankLine
\textbf{function~}\textsf{subMISs}($\mc{H}$, $Y$, $\mf{X}$, $\ell = [W_1,\cdots,W_r]$)\;
\Indp

Set $\mf{A} = \Poss\An(Y)_{\mc{H} \setminus \mf{X}}$. 

\tcp{Check every selected variable after each addition.}

\lIf{\rm $\exists X\in\mf{X} \text{~such that~} \Poss\Ch(X)_{\mc{H}} \cap \mf{A} = \emptyset$}{\Return{$\emptyset$}}

Set $\mb{M} = \{\mf{X}\}$. \;

\For{$i=1, \cdots , r$}{
    \If{$W_i \in \mf{A}$}{
        Update $\mb{M} =  \mb{M} \cup \textsf{subMISs}(\mc{H},Y,\mf{X}\cup\{W_i\},[W_{i+1},\cdots,W_r])$. \;
    }
}
\Return{$\mb{M}$}\;
\Indm
\caption{\small Enumerating minimal intervention sets by recursive extension.}
\label{alg: miss}
\end{algorithm}

\paragraph{Enumerating minimal intervention sets.} The characterizations in \Cref{thm: MIS for MAG iff An,prop: MIS for PAG} lead to \textsf{MISs} (\Cref{alg: miss}), which returns the complete collection $\mb{M}_{\mc{D},Y}$ for a MAG or PAG $\mc{D}$. The algorithm begins with the empty intervention set and adds one variable at a time. It stops extending a set as soon as one of its variables loses every proper possibly directed path to $Y$. 

First, let $\mf{A} = \Poss\An(Y)_{\mc{D}}$ and retain the induced graph $\mc{H} = \mc{D}[\mf{A}]$. No MIS can contain a variable outside $\mf{A}$. Moreover, every vertex on a possibly directed path to $Y$ belongs to $\mf{A}$, so this restriction preserves all paths needed by the MIS condition. The graph $\mc{H}$ is then kept fixed throughout the recursion. 

The variables in $\mf{A} \setminus \{Y\}$ are arranged in a fixed list $\ell$: 
\begin{itemize}
    \item For a MAG, we use a reverse topological order of its directed edges. That is, whenever $U \to V$, the vertex $V$ precedes $U$.

    \item For a PAG, we order its circle components by reversing the partial topological order returned by \Cref{alg: pto}.
\end{itemize}
For the PAG case, this is the order in which \Cref{alg: pto} extracts the components. Within each component, variables are ordered by increasing shortest-path distance to $Y$ in the underlying undirected graph of $\mc{D}$; ties can be resolved arbitrarily. 

For a PAG, processing circle components in this reverse partial topological order supports early pruning before recursion. To see this, suppose that every possibly directed path from an upstream candidate $W$ to $Y$ passes through already selected downstream variables. Deleting those selected variables excludes $W$ from the remaining possible ancestors of $Y$, so the algorithm discards this extension before making a recursive call. Processing downstream components first therefore allows such upstream candidates to be eliminated as soon as they are considered. 

The same order also preserves paths across different circle components. Each step between distinct components on a possibly directed path leads to a component appearing earlier in the order. A variable selected later from a different component therefore cannot lie on a proper possibly directed path from an earlier selected variable to $Y$. Adding that later variable cannot block the earlier variable's path. Thus, reverse partial topological order prevents this source of failed recursive calls between components. Within a single component, however, a later choice can still block an earlier variable's path. Thus, the algorithm retains the test for all selected variables. 

For a current intervention set $\mf{X}$, \textsf{subMISs} recomputes $\mf{A} = \Poss\An(Y)_{\mc{H} \setminus \mf{X}}$. These are the vertices that can still reach $Y$ without passing through any variable in $\mf{X}$. Write $\Poss\Ch(X)_{\mc{H}}$ for the set of vertices adjacent to $X$ whose incident edge has no arrowhead at $X$. Each selected $X$ must have a first step into $\mf{A}$, giving the test: 
\begin{align}
    \Poss\Ch(X)_{\mc{H}} \cap \mf{A} \neq \emptyset \text{\quad for~every~} X \in \mf{X}. \label{eq: miss test}
\end{align}
Possible ancestors include $Y$ itself, so a single possibly directed edge from $X$ to $Y$ can satisfy this test. For a MAG, the first step and the remaining path must be directed. The test is applied to \textit{all} selected variables after every addition. Although a new variable may have a path to $Y$, selecting it can interrupt a path used by an earlier variable. Keeping $\mc{H}$ fixed allows the algorithm to inspect the first step from every selected variable; only the graph used to compute $\mf{A}$ has those variables removed. 

If the test fails, the call returns an empty collection. Otherwise, it records $\mf{X}$ and tries each remaining variable $W_i \in \mf{A}$. The recursive call receives $\mf{X} \cup \{W_i\}$ and only the variables after $W_i$ in the list. In particular, the initial call records $\emptyset$, because its test has no selected variable to check. 

\begin{proposition}\label{prop: correctness of MISs} $\textsf{MISs}(\mc{D},Y)$ returns exactly $\mb{M}_{\mc{D},Y}$. Every MIS is generated once, and the recursion terminates. 
\end{proposition}
\begin{proof}[Proof of \Cref{prop: correctness of MISs}]
    We first justify the test $\Poss\Ch(X)_{\mc{H}} \cap \mf{A} \neq \emptyset$ for every $X \in \mf{X}$ in \Cref{eq: miss test}. If $X \in \mf{X}$ has a proper possibly directed path to $Y$, its first adjacent vertex $V$ is outside $\mf{X}$. The remaining path avoids all of $\mf{X}$, so $V \in \mf{A}$. Conversely, if $V \in \Poss\Ch(X)_{\mc{H}} \cap \mf{A}$, there is a possibly directed path from $V$ to $Y$ in $\mc{H} \setminus \mf{X}$. Prepending $X$ gives a proper path: $X$ cannot already occur on the remaining path, and no other selected variable occurs there. This also covers $V = Y$, when the remaining path has zero length. Thus, the test is equivalent to requiring a proper path for every selected variable. By \Cref{thm: MIS for MAG iff An,prop: MIS for PAG}, every recorded set is an MIS. 

    Next, suppose an intervention set $\mf{X}$ fails the test at a variable $X \in \mf{X}$. Adding more intervention variables only removes additional vertices from the graph in which the remainder of its path must lie. It cannot create a path for $X$ that avoids the enlarged intervention set. Therefore, no superset of $\mf{X}$ is an MIS, and stopping this branch cannot omit a valid output. Likewise, if an unselected variable $W$ is outside $\mf{A} = \Poss\An(Y)_{\mc{H} \setminus \mf{X}}$, it has no path to $Y$ avoiding $\mf{X}$ and cannot be added in any valid extension. 

    To establish completeness, fix an MIS $\mf{Z}$. All its variables belong to $\Poss\An(Y)_{\mc{D}} \setminus \{Y\}$. List them in the order inherited from $\ell$. Every prefix $\mf{X}$ of this list is an MIS: the proper paths for the variables of $\mf{Z}$ avoid $\mf{Z}$ except at their starting vertices, and hence also avoid the smaller set $\mf{X}$ except at their starting vertices. For the next variable $W \in \mf{Z} \setminus \mf{X}$, its proper path for $\mf{Z}$ avoids $\mf{X}$ entirely. Thus $W \in \mf{A} = \Poss\An(Y)_{\mc{H} \setminus \mf{X}}$, and the algorithm visits the branch that adds it. Inductively, it reaches and records $\mf{Z}$.

    Finally, each recursive call passes only the suffix after its newly selected variable. Selected indices therefore increase strictly. A given set has only one increasing sequence of indices, so it is generated on only one branch. The list becomes shorter at every recursive step, establishing termination. 
\end{proof}

\paragraph{Enumeration cost.} Let $\kappa = \lvert \mb{M}_{\mc{D},Y} \rvert$ be the total number of MISs returned, let $n = \lvert \mf{V} \rvert$ and $m$ be the number of edges, and let $r = \lvert \Poss\An(Y)_{\mc{D}} \setminus \{Y\} \rvert$ be the number of initial candidates. We first express the cost in terms of $\kappa$. Each call that passes \Cref{eq: miss test} records exactly one distinct MIS. Thus, there are exactly $\kappa$ such calls. Each starts at most $r$ recursive calls, while a call that fails the test stops immediately. Thus, the total number of calls is at most $1 + r \kappa$. For a MAG with the reverse topological order above, no call fails the test, and the number of calls is exactly $\kappa$. 

With adjacency-list graph operations, computing the remaining possible ancestors and checking all selected variables takes $\mc{O}(n + m)$ time per call. Copying a suffix or a selected set takes at most $\mc{O}(r)$ time. We also account for the collection unions in the algorithm. Each output belongs to the returned collections of at most $r + 1$ calls along its branch, and each such collection can be copied at most $r$ times while processing the remaining candidates. Repeated copying therefore contributes at most $\mc{O}((r+1)^2 \kappa)$ time. This is absorbed by the bound below because $r+1 \leq n$. A straightforward implementation of the preliminary ordering takes $\mc{O}(n^3)$ time, giving the conservative total bound: 
\begin{align}
    \mc{T}_{\texttt{MISs}} = \mc{O}(n^3 + (r+1) \kappa (n+m)). \label{eq: cost of miss}
\end{align}
For a fixed number $\kappa$ of output MISs, this bound is polynomial in the graph size. It counts both the unsuccessful calls and the cost of collecting the outputs, rather than only the cost of checking a single candidate. 

However, $\kappa$ itself can be exponential. For example, a MAG consisting of $r$ candidate vertices $X_1, \cdots, X_r$, the reward $Y$, and the edges $X_i \to Y$ has $\kappa = 2^r$. Every subset of the candidate vertices is an MIS because each selected variable has a directed path to $Y$. Thus, polynomial dependence on a fixed output count does not imply polynomial time for all input graphs. The recursion reduces unnecessary exploration by rejecting an invalid set together with all of its extensions. 

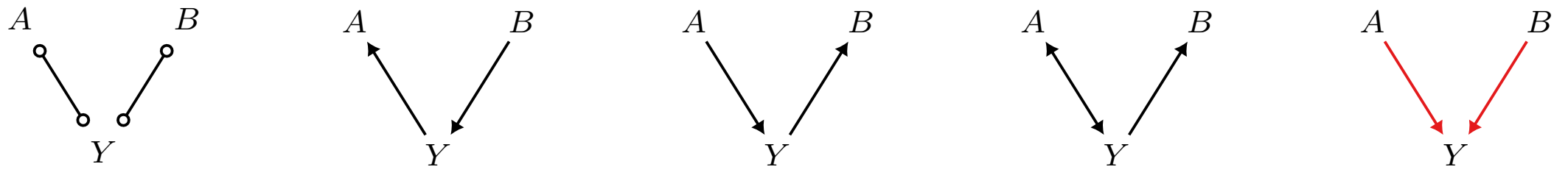
\begin{figure}[t]
    \centering
    \hfill
    \begin{minipage}[b]{0.19\textwidth}
    \subfloat[$\mc{P}$]{
        \begin{tikzpicture}
        [x=1.5cm,y=1.2cm,>={Latex[width=1.4mm,length=1.2mm]},
        line width=0.25mm,
        oo/.tip={Circle[open,length=3.5pt,width=3.5pt]},
        RR/.style={draw=none,circle,inner sep=0mm, minimum size=4mm}
        ]
        
        \node[] (A) at (0,1) {$A$};
        \node[] (Y) at (0.5,0) {$Y$};
        \node[] (B) at (1,1) {$B$};
        
        \draw[{oo}-{oo}] (A) -- (Y);
        \draw[{oo}-{oo}] (Y) -- (B);
        \end{tikzpicture}\label{fig: 3var a}}
    \end{minipage}
    \hfill
    \begin{minipage}[b]{0.19\textwidth} 
    \subfloat[$\mc{M}_1$]{
        \begin{tikzpicture}[x=1.5cm,y=1.2cm,>={Latex[width=1.4mm,length=1.2mm]},
        line width=0.25mm,
        oo/.tip={Circle[open,length=3.5pt,width=3.5pt]},
        RR/.style={draw=none,circle,inner sep=0mm, minimum size=4mm}]
        \node[RR] (A) at (0,1) {$A$};
        \node[RR] (Y) at (0.5,0) {$Y$};
        \node[RR] (B) at (1,1) {$B$};
        
        \draw[->] (B) -- (Y);
        \draw[->] (Y) -- (A);
        
        \end{tikzpicture}\label{fig: 3var b}}
    \end{minipage}
    \hfill
    \begin{minipage}[b]{0.19\textwidth} 
    \subfloat[$\mc{M}_2$]{
        \begin{tikzpicture}[x=1.5cm,y=1.2cm,>={Latex[width=1.4mm,length=1.2mm]},
        line width=0.25mm,
        oo/.tip={Circle[open,length=3.5pt,width=3.5pt]},
        RR/.style={draw=none,circle,inner sep=0mm, minimum size=4mm}]
        \node[RR] (A) at (0,1) {$A$};
        \node[RR] (Y) at (0.5,0) {$Y$};
        \node[RR] (B) at (1,1) {$B$};
        
        \draw[->] (A) -- (Y);
        \draw[->] (Y) -- (B);
        
        \end{tikzpicture}\label{fig: 3var c}}
    \end{minipage}
    \hfill
    \begin{minipage}[b]{0.19\textwidth}
    \subfloat[$\mc{M}_3$]{
        \begin{tikzpicture}[x=1.5cm,y=1.2cm,>={Latex[width=1.4mm,length=1.2mm]},
        line width=0.25mm,
        oo/.tip={Circle[open,length=3.5pt,width=3.5pt]},
        RR/.style={draw=none,circle,inner sep=0mm, minimum size=4mm}]
        \node[RR] (A) at (0,1) {$A$};
        \node[RR] (Y) at (0.5,0) {$Y$};
        \node[RR] (B) at (1,1) {$B$};
        
        \draw[<->] (A) -- (Y);
        \draw[->] (Y) -- (B);
        \end{tikzpicture}\label{fig: 3var d}}
    \end{minipage}
    \hfill
    \begin{minipage}[b]{0.19\textwidth}
    \subfloat[$\mc{M}_4$]{
        \begin{tikzpicture}[x=1.5cm,y=1.2cm,>={Latex[width=1.4mm,length=1.2mm]},
        line width=0.25mm,
        oo/.tip={Circle[open,length=3.5pt,width=3.5pt]},
        RR/.style={draw=none,circle,inner sep=0mm, minimum size=4mm}]
        \node[RR] (A) at (0,1) {$A$};
        \node[RR] (Y) at (0.5,0) {$Y$};
        \node[RR] (B) at (1,1) {$B$};
        
        \draw[->, draw = betterred] (A) -- (Y);
        \draw[<-, draw = betterred] (Y) -- (B);
        \end{tikzpicture}\label{fig: 3var e}}
    \end{minipage}
    \hfill
    \hfill
    \caption{\small Three MAGs $\mc{M}_1$--$\mc{M}_3$ are represented by $\mc{P}$. In contrast, $\mc{M}_4$ is not represented by $\mc{P}$.}
    \label{fig: 3var}
\end{figure}
\paragraph{Possible vacuousness.} One might expect that if $\mf{X}$ is an MIS relative to $\langle \mc{P}, Y \rangle$, then it is also an MIS relative to $\langle \mc{M}, Y \rangle$ for some MAG $\mc{M}$ conforming to the PAG $\mc{P}$. However, this is \textit{not} always the case, i.e., $\mf{X}$ need not be an MIS for any MAG represented by $\mc{P}$. 

For concreteness, consider the PAG $\mc{P}$ in \Cref{fig: 3var} where $\mf{X} = \{A,B\}$ is an MIS with respect to $\langle \mc{P}, Y \rangle$ since both $A$ and $B$ have proper possibly directed paths to $Y$ (i.e., $A \cedge Y$ and $B \cedge Y$, respectively). However, we will demonstrate that at least one of $A$ and $B$ is irrelevant to the reward $Y$ in every conforming MAG. To see this, consider binary SCMs with mutually independent exogenous variables $U_A,U_B,U_Y \sim \Bern(\epsilon)$, where $0 < \epsilon < \frac{1}{2}$. First, for the proper subset $\mf{Z}_1 = \emptyset$, we can construct an SCM $\mc{S}_1$ as follows:
\begin{align}
        \mc{S}_1
    = \begin{cases}
            \mf{U} =   \{U_A,U_B,U_Y\}\\
            \mf{V} = \{A,B,Y\} \\
            \mc{F} = 
            \begin{cases}
                f_A(y,u_A) = y \oplus u_A\\
                f_B(u_B) = u_B\\
                f_Y(b,u_Y) = b \oplus u_Y
            \end{cases}\\
            P(\mf{U}) = \{U_V \thicksim \Bern(\epsilon) : \forall U_V \in \mf{U}\}. 
    \end{cases}
\end{align}
Its corresponding MAG is shown in \Cref{fig: 3var b}. Here, $\mu_{\emptyset} = 2\epsilon(1-\epsilon)$ while $\mu_{\{a,b\}^\ast} = \mu_{b^\ast} = 1-\epsilon$ with $b^\ast = 1$. Thus, $\mu_{a,1} \neq \mu_{\emptyset}$ for every $a \in \{0,1\}$ in $\mc{S}_1$. That is, there exists an SCM $\mc{S}_1$ such that: 
\begin{align}
    \mu_{\mf{x}^\ast} \neq \mu_{\mf{z}_1^\ast}
\end{align}
Next, for the proper subset $\mf{Z}_2 = \{A\}$, since $\mu_a = \mu_{\emptyset} = 2\epsilon(1-\epsilon)$, we again have $\mu_{a,1} \neq \mu_{a}$ for every $a \in \{0,1\}$ in $\mc{S}_1$. 
That is, there also exists an SCM $\mc{S}_1$ such that:
\begin{align}
    \mu_{\mf{x}^\ast} \neq \mu_{\mf{z}_2^\ast}
\end{align}
An analogous construction applies to $\mf{Z}_3 = \{B\}$ using $\mc{S}_2$ (\Cref{fig: 3var c}), defined as follows:
\begin{align}
        \mc{S}_2
    = \begin{cases}
            \mf{U} =  \{U_A,U_B,U_Y\}\\
            \mf{V} = \{A,B,Y\} \\
            \mc{F} = 
            \begin{cases}
                f_A(u_A) = u_A\\
                f_B(y,u_B) = y \oplus u_B\\
                f_Y(a,u_Y) = a \oplus u_Y
            \end{cases}\\
            P(\mf{U}) = \{U_V \thicksim \Bern(\epsilon) : \forall U_V \in \mf{U}\}.
    \end{cases}
\end{align}
Hence, there also exists an SCM $\mc{S}_2$ such that:
\begin{align}
    \mu_{\mf{x}^\ast} \neq \mu_{\mf{z}_3^\ast}
\end{align} 
These witnesses establish that $\{A, B\}$ is an MIS for $\mc{P}$. However, a common witnessing SCM would have a corresponding MAG in which neither intervention variable is redundant. By \Cref{thm: MIS for MAG iff An}, both $A$ and $B$ would then have proper directed paths to $Y$. On this three-vertex skeleton, this requires $A \to Y \gets B$, creating an unshielded collider absent from $\mc{P}$. Hence no such MAG, and therefore no common witnessing SCM, exists. 

\subsection{Definitely Minimal Intervention Sets}\label{sec:dmis-char}

To address this vacuousness, we propose the concept of \textit{definitely} MIS, which ensures that an MIS in a PAG remains an MIS in some consistent MAG. 

For an MIS, the witnessing SCM may depend on the proper subset being compared. A DMIS requires a single SCM to work for all proper subsets; the distinguishing assignment may still depend on the subset.
\begin{definition}[Definitely minimal intervention set]\label{def: DMIS}
Given $\langle\mc{D},Y\rangle$, $\mf{X}\subseteq\mf{V}\setminus\{Y\}$ is a DMIS if there exists an SCM $\mc{S}^\ast$ conforming to $\mc{D}$ such that, for every $\mf{Z}\subsetneq\mf{X}$, there exists an assignment $\mf{x}\in\Omega_{\mf{X}}$ satisfying $\mu^{\mc{S}^\ast}_{\mf{x}}\ne\mu^{\mc{S}^\ast}_{\mf{x}\cap\mf{Z}}$. 
\end{definition}

The collection of DMISs is denoted by $\mb{D}_{\mc{D},Y}$. Since DMIS is stricter than MIS, their inclusion relation is as follows.
\begin{restatable}[]{proposition}{IfDMISthenMIS}\label{prop: if DMIS then MIS}
    If $\mf{X}$ is a DMIS with respect to $\langle \mc{D}, Y \rangle$, then it is also an MIS. 
\end{restatable}
\begin{proof}[Proof of \Cref{prop: if DMIS then MIS}]
    Let $\mc{S}^\ast$ be the common SCM in \Cref{def: DMIS}. For each $\mf{Z} \subsetneq \mf{X}$, that definition supplies an assignment $\mf{x}$ with $\mu^{\mc{S}^\ast}_{\mf{x}} \neq \mu^{\mc{S}^\ast}_{\mf{x}\cap\mf{Z}}$. Choosing the same $\mc{S}^\ast$ for each subset satisfies exactly the quantifiers in \Cref{def: MIS ancestral}. 
\end{proof}
For causal diagrams and MAGs, which have no circle marks, MIS and DMIS coincide. 
\begin{restatable}[]{proposition}{IfMISthenDMISforCDandMAG}\label{prop: if MIS then DMIS for cd and mag}
    Let $\mc{D}$ be either a causal diagram or a MAG. If $\mf{X}$ is an MIS with respect to $\langle \mc{D}, Y \rangle$, then it is also a DMIS.    
\end{restatable}
\begin{proof}[Proof of \Cref{prop: if MIS then DMIS for cd and mag}]
    If $\mc{D}$ is a causal diagram, let $\mc{G} = \mc{D}$. If $\mc{D}$ is a MAG, let $\mc{G}$ have exactly its directed and bidirected edges, so its corresponding MAG is $\mc{D}$. In either case, \Cref{prop: MIS} and \Cref{thm: MIS for MAG iff An} imply that every $X \in \mf{X}$ has a proper directed path to $Y$ in $\mc{G}$. 

    Fix the SCM $\mc{S}^\ast$ supplied by \Cref{lem: binary scm construction}. For any $\mf{Z} \subsetneq \mf{X}$, choose $W \in \mf{X} \setminus \mf{Z}$ and apply that lemma to its proper path. Then $\mu_{\mf{X} = \boldsymbol{1}}^{\mc{S}^\ast} \neq \mu_{\mf{Z} = \boldsymbol{1}}^{\mc{S}^\ast}$. The same SCM works for every proper subset and conforms to $\mc{D}$. This establishes DMIS membership. If $\mf{X} = \emptyset$, any conforming SCM witnesses the vacuous requirement. 
\end{proof}
This equivalence between MIS and DMIS for causal diagrams and MAGs stems from the fact that the relevant ancestral relations are \textbf{determined}, namely, $\mf{X} \subseteq \An(Y)_{\mc{D}_{\overline{\mf{X}}}}$. We now turn to DMIS for PAGs, where ancestral relations are not fully determined, thereby creating a notable gap between MIS and DMIS. Recall the PAG $\mc{P}$ in \Cref{fig: 3var a}, where $\{A,B\}$ is an MIS but not a DMIS. 
\begin{restatable}[]{proposition}{DMISforPAGeqMISforMAG}\label{prop: DMIS for pag eq MIS for MAG}
    $\mf{X} \subseteq \mf{V} \setminus \{Y\}$ is a DMIS relative to $\langle \mc{P},Y \rangle$ \textit{if and only if} there exists a MAG $\mc{M}$ conforming to $\mc{P}$ such that $\mf{X}$ is an MIS relative to $\langle \mc{M},Y \rangle$.    
\end{restatable}    
\begin{proof}[Proof of \Cref{prop: DMIS for pag eq MIS for MAG}]
    \textbf{(If)} Suppose that $\mf{X}$ is an MIS for some $\mc{M} \in [\mc{P}]$. By \Cref{prop: if MIS then DMIS for cd and mag}, there is one SCM $\mc{S}^\ast$ conforming to $\mc{M}$ such that every proper subset of $\mf{X}$ is distinguished by an assignment. Since this SCM also conforms to $\mc{P}$, it witnesses DMIS membership for the PAG. 
    
    \textbf{(Only if)} Let $\mc{S}^\ast$ witness that $\mf{X}$ is a DMIS for $\mc{P}$. Since $\mc{S}^\ast$ conforms to $\mc{P}$, choose a MAG $\mc{M} \in [\mc{P}]$ to which it conforms. The same SCM and assignments witness that $\mf{X}$ is a DMIS for $\mc{M}$. By \Cref{prop: if DMIS then MIS}, it is an MIS for $\mc{M}$. 
\end{proof}
Thus, a DMIS has a common witnessing SCM, whereas an MIS for a PAG may rely on different SCMs for different proper subsets. The following examples illustrate the difficulty of expressing this common-witness requirement as a direct graphical criterion. 

The difficulty for PAGs is that the proper paths required for different intervention variables must occur in the same represented MAG. In \Cref{fig: 3var a}, the vertices $A$ and $B$ cannot both be ancestors of $Y$ in one represented MAG, because $A \to Y \gets B$ would create an unshielded collider on the uncovered path $A \cedge Y \cedge B$. 

\begin{figure}[t]
    \centering
    \begin{minipage}[b]{0.24\textwidth}%
    \centering
    \subfloat[$\mc{P}$]{\begin{tikzpicture}[x=0.8cm,y=1.2cm,>={Latex[width=1.4mm,length=1.7mm]},
        line width=0.25mm,
        oo/.tip={Circle[open,length=3.5pt,width=3.5pt]},
        RR/.style={draw=none,circle,inner sep=0mm, minimum size=4mm}
        ]
        \node[RR] (A) at (0,1.2) {$A$};
        \node[RR] (B) at (2.3,1) {$B$};
        \node[RR] (C) at (0.9,0.5) {$C$};
        \node[RR] (Y) at (1.3,-0.5) {$Y$};
        \node[RR] (D) at (-0.12,0.1) {$D$};

        \draw[{oo}-{oo}] (A) -- (C);
        \draw[{oo}-{oo}] (B) -- (C);
        \draw[{oo}-{oo}] (C) -- (Y);
        \draw[{oo}-{oo}] (B) -- (Y);
        \draw[{oo}-{oo}] (D) -- (Y);
        \end{tikzpicture}\label{fig: 5var a}}\end{minipage}
        \hfill%
        \begin{minipage}[b]{0.24\textwidth}%
        \centering
        \subfloat[$\mc{M}_1$]{\begin{tikzpicture}[x=0.8cm,y=1.2cm,>={Latex[width=1.4mm,length=1.7mm]},
        line width=0.25mm,
        oo/.tip={Circle[open,length=3.5pt,width=3.5pt]},
        RR/.style={draw=none,circle,inner sep=0mm, minimum size=4mm}
        ]
        \node[RR] (A) at (0,1.2) {$A$};
        \node[RR] (B) at (2.3,1) {$B$};
        \node[RR] (C) at (0.9,0.5) {$C$};
        \node[RR] (Y) at (1.3,-0.5) {$Y$};
        \node[RR] (D) at (-0.12,0.1) {$D$};

        \draw[->] (A) -- (C);
        \draw[->] (C) -- (B);
        \draw[->] (C) -- (Y);
        \draw[<-] (B) -- (Y);
        \draw[<-] (D) -- (Y);
        \end{tikzpicture}\label{fig: 5var b}}
        \end{minipage}\hfill%
        \begin{minipage}[b]{0.24\textwidth}%
        \centering
        \subfloat[$\mc{M}_2$]{\begin{tikzpicture}[x=0.8cm,y=1.2cm,>={Latex[width=1.4mm,length=1.7mm]},
        line width=0.25mm,
        oo/.tip={Circle[open,length=3.5pt,width=3.5pt]},
        RR/.style={draw=none,circle,inner sep=0mm, minimum size=4mm}
        ]
        \node[RR] (A) at (0,1.2) {$A$};
        \node[RR] (B) at (2.3,1) {$B$};
        \node[RR] (C) at (0.9,0.5) {$C$};
        \node[RR] (Y) at (1.3,-0.5) {$Y$};
        \node[RR] (D) at (-0.12,0.1) {$D$};

        \draw[->] (A) -- (C);
        \draw[->] (C) -- (B);
        \draw[->] (C) -- (Y);
        \draw[->] (B) -- (Y);
        \draw[<-] (D) -- (Y);

        \end{tikzpicture}\label{fig: 5var c}}\end{minipage}
        \hfill%
        \begin{minipage}[b]{0.24\textwidth}
        \centering
        \subfloat[$\mc{M}_3$]{\begin{tikzpicture}[x=0.8cm,y=1.2cm,>={Latex[width=1.4mm,length=1.7mm]},
        line width=0.25mm,
        oo/.tip={Circle[open,length=3.5pt,width=3.5pt]},
        RR/.style={draw=none,circle,inner sep=0mm, minimum size=4mm}
        ]
        \node[RR] (A) at (0,1.2) {$A$};
        \node[RR] (B) at (2.3,1) {$B$};
        \node[RR] (C) at (0.9,0.5) {$C$};
        \node[RR] (Y) at (1.3,-0.5) {$Y$};
        \node[RR] (D) at (-0.12,0.1) {$D$};

        \draw[<-] (A) -- (C);
        \draw[->] (C) -- (B);
        \draw[<-] (C) -- (Y);
        \draw[<-] (B) -- (Y);
        \draw[->] (D) -- (Y);
        \end{tikzpicture}\label{fig: 5var d}}\end{minipage}
    \caption{\small The three MAGs $\mc{M}_1$--$\mc{M}_3$ are represented by the PAG $\mc{P}$.}
    \label{fig: 5var}
\end{figure}
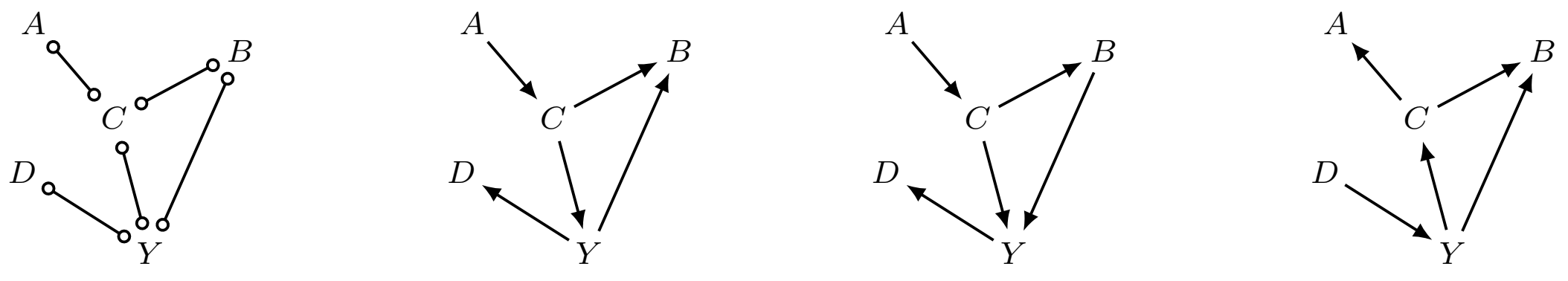
Consider the uncovered circle path $A \cedge C \cedge Y \cedge D$ in the PAG $\mc{P}$ shown in \Cref{fig: 5var a}. Every consecutive triple on this path must remain a non-collider in every represented MAG. This restricts the orientations that could make both $A$ and $D$ ancestors of $Y$, as follows. Consider any MAG represented by $\mc{P}$ (\Cref{fig: 5var b,fig: 5var c,fig: 5var d}) where $A \cedge C$ appears as a directed edge out of $A$ (e.g., $\mc{M}_1$ and $\mc{M}_2$). This results in $C \to Y \to D$, as the path is of definite status. In contrast, if any MAG contains $D \to Y$, this leads to $Y \to C \to A$, as in $\mc{M}_3$. The important observation is that $A \to C$ ensures that $D \notin \An(Y)_{\mc{M}}$, whereas $D \to Y$ ensures that $A \notin \An(Y)_{\mc{M}}$ for any MAG $\mc{M} \in [\mc{P}]$. This shows that $A$ and $D$ cannot simultaneously be ancestors of $Y$ in $\mc{M}$. Hence, $\{A, D\}$ is not a DMIS relative to $\langle \mc{P}, Y \rangle$.

\begin{figure}[t]
    \centering
    \begin{minipage}[b]{0.24\textwidth}%
        \centering
        \subfloat[$\mc{P}$]{%
        \begin{tikzpicture}[x=0.8cm,y=1.2cm,>={Latex[width=1.4mm,length=1.7mm]},
            line width=0.25mm,
            oo/.tip={Circle[open,length=3.5pt,width=3.5pt]},
            RR/.style={draw=none,circle,inner sep=0mm,minimum size=4mm}]
            \node[RR] (X) at (1.25,1.4) {$X$};
            \node[RR] (Z) at (1.25,0.48) {$Z$};
            \node[RR] (A) at (0,0) {$A$};
            \node[RR] (B) at (2.5,0) {$B$};
            \node[RR] (Y) at (1.25,-0.8) {$Y$};

            \draw[{oo}-{oo}] (A) -- (B);
            \draw[{oo}-{oo}] (X) -- (A);
            \draw[{oo}-{oo}] (X) -- (B);
            \draw[{oo}-{oo}] (Z) -- (A);
            \draw[{oo}-{oo}] (Z) -- (B);
            \draw[{oo}-{oo}] (A) -- (Y);
            \draw[{oo}-{oo}] (B) -- (Y);
        \end{tikzpicture}\label{fig: scbmec dmis 5var a}}%
    \end{minipage}
    \hfill%
    \begin{minipage}[b]{0.24\textwidth}%
        \centering
        \subfloat[$\mc{M}_1$]{%
        \begin{tikzpicture}[x=0.8cm,y=1.2cm,>={Latex[width=1.4mm,length=1.7mm]},
            line width=0.25mm,
            oo/.tip={Circle[open,length=3.5pt,width=3.5pt]},
            RR/.style={draw=none,circle,inner sep=0mm,minimum size=4mm}]
            \node[RR] (X) at (1.25,1.4) {$X$};
            \node[RR] (Z) at (1.25,0.48) {$Z$};
            \node[RR] (A) at (0,0) {$A$};
            \node[RR] (B) at (2.5,0) {$B$};
            \node[RR] (Y) at (1.25,-0.8) {$Y$};

            \draw[->] (A) -- (B);
            \draw[->] (X) -- (A);
            \draw[->] (X) -- (B);
            \draw[->] (A) -- (Z);
            \draw[->] (B) -- (Z);
            \draw[->] (A) -- (Y);
            \draw[->] (B) -- (Y);
        \end{tikzpicture}\label{fig: scbmec dmis 5var b}}%
    \end{minipage}
    \hfill%
    \begin{minipage}[b]{0.24\textwidth}%
        \centering
        \subfloat[$\mc{M}_2$]{%
        \begin{tikzpicture}[x=0.8cm,y=1.2cm,>={Latex[width=1.4mm,length=1.7mm]},
            line width=0.25mm,
            oo/.tip={Circle[open,length=3.5pt,width=3.5pt]},
            RR/.style={draw=none,circle,inner sep=0mm,minimum size=4mm}]
            \node[RR] (X) at (1.25,1.4) {$X$};
            \node[RR] (Z) at (1.25,0.48) {$Z$};
            \node[RR] (A) at (0,0) {$A$};
            \node[RR] (B) at (2.5,0) {$B$};
            \node[RR] (Y) at (1.25,-0.8) {$Y$};

            \draw[->] (B) -- (A);
            \draw[->] (A) -- (X);
            \draw[->] (B) -- (X);
            \draw[->] (Z) -- (A);
            \draw[->] (Z) -- (B);
            \draw[->] (A) -- (Y);
            \draw[->] (B) -- (Y);
        \end{tikzpicture}\label{fig: scbmec dmis 5var c}}%
    \end{minipage}
    \hfill%
    \begin{minipage}[b]{0.24\textwidth}%
        \centering
        \subfloat[Invalid]{%
        \begin{tikzpicture}[x=0.8cm,y=1.2cm,>={Latex[width=1.4mm,length=1.7mm]},
            line width=0.25mm,
            oo/.tip={Circle[open,length=3.5pt,width=3.5pt]},
            RR/.style={draw=none,circle,inner sep=0mm,minimum size=4mm}]
            \node[RR] (X) at (1.25,1.4) {$X$};
            \node[RR] (Z) at (1.25,0.48) {$Z$};
            \node[RR] (A) at (0,0) {$A$};
            \node[RR] (B) at (2.5,0) {$B$};
            \node[RR] (Y) at (1.25,-0.8) {$Y$};

            \draw[{oo}-{oo}] (A) -- (B);
            \draw[->, betterred] (X) -- (A);
            \draw[->, betterred] (B) -- (X);
            \draw[->, betterred] (A) -- (Z);
            \draw[->, betterred] (Z) -- (B);
            \draw[->] (A) -- (Y);
            \draw[->] (B) -- (Y);
        \end{tikzpicture}\label{fig: scbmec dmis 5var d}}%
    \end{minipage}
    \caption{\small The intervention set $\{X,Z\}$ is an MIS but not a DMIS for $\mc{P}$. The MAGs $\mc{M}_1$ and $\mc{M}_2$ realize the two paths separately; the proper paths for $\{X,Z\}$ cannot be jointly realized in a MAG represented by $\mc{P}$.}
    \label{fig: scbmec dmis 5var}
\end{figure}
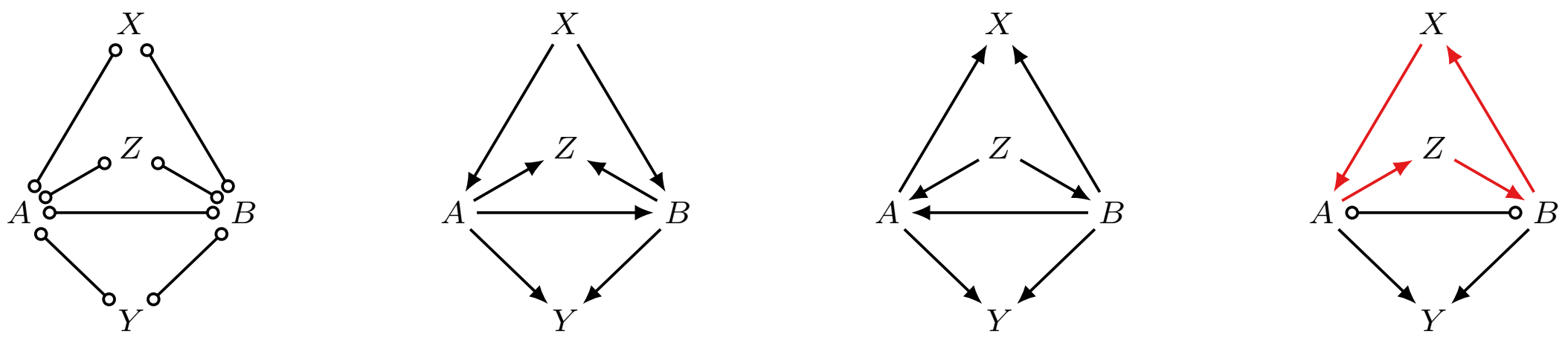
We consider a more involved example in \Cref{fig: scbmec dmis 5var}. In $\mc{P}$, $\mf{X} = \{X,Z\}$ is an MIS because $X$ and $Z$ each have a proper possibly directed path to $Y$ with respect to $\mf{X}$, namely $X \cedge A \cedge Y$ and $Z \cedge B \cedge Y$, respectively. These paths are realized as directed paths in $\mc{M}_1$ and $\mc{M}_2$, separately. 

However, realizing both paths in a single MAG requires $X \to A$ and $Z \to B$, which forces $A \to Z$ and $B \to X$ to avoid new v-structures. These orientations produce the directed cycle $X \to A \to Z \to B \to X$. The alternative choice of distinct first neighbors, $X \to B$ and $Z \to A$, is symmetric. A shared first neighbor instead creates a v-structure, either $X \to A \gets Z$ or $X \to B \gets Z$. Since these cases exhaust the possible first neighbors, no MAG represented by $\mc{P}$ admits proper directed paths from both $X$ and $Z$ to $Y$ with respect to $\mf{X}$. Hence $\mf{X}$ is not a DMIS.

\begin{remark}[On the graphical characterization of DMISs]
The example in \Cref{fig: scbmec dmis 5var} provides a counterexample to the graphical characterization of DMISs for PAGs in \citet{park2025structural}. The paths $X \cedge A \cedge Y$ and $Z \cedge B \cedge Y$ meet the condition in Theorem~2 of \citet{park2025structural}: no uncovered path can have both starting edges as its end edges, because any connecting path contains a shielded triple involving the adjacent vertices $A$ and $B$. Hence, $\{X,Z\}$ satisfies the stated graphical condition but is not a DMIS, disproving its sufficiency. 

We leave the development of a direct graphical criterion for DMIS membership for PAGs, together with a polynomial-time decision algorithm, to future work. 
\end{remark}

\section{Generalizing Possibly-Optimal Minimal Intervention Sets}\label{sec: scbmec pomis}

We now refine the notion of a \textit{possibly-optimal minimal intervention set} (POMIS) by defining it over DMISs rather than MISs. This refined notion is consistent with the existing literature and serves as a natural extension, as supported by \Cref{prop: if MIS then DMIS for cd and mag,prop: if DMIS then MIS}, which imply that MIS and DMIS coincide for causal diagrams and MAGs.
\begin{definition}[Possibly-optimal minimal intervention set (POMIS)] \label{def: POMIS for PAG}
    Let $\mf{X} \subseteq \mf{V} \setminus \{Y\}$ be a DMIS relative to $\langle \mc{D}, Y\rangle $. If there exists an SCM conforming to $\mc{D}$ such that $\mu_{\mf{x}^\ast} >  \mu_{\mf{w}^\ast}$ for \textit{all} $\mf{W} \in \mb{D}_{\mc{D},Y} \setminus \{\mf{X}\}$, we say that $\mf{X}$ is a \textit{possibly-optimal minimal intervention set} (POMIS) relative to $\langle \mc{D}, Y\rangle$. 
\end{definition}
For a PAG, this definition compares $\mf{X}$ against the DMISs of the entire represented model class within one SCM. All reward means and maximizing assignments in this comparison are evaluated in that same SCM. The following proposition relates this comparison to POMIS membership in a represented MAG. 
\begin{restatable}[]{proposition}{POMISforPAGeqPOMISforMAG}\label{prop: if POMIS for PAG then POMIS for MAG}
    If $\mf{X} \subseteq \mf{V} \setminus \{Y\}$ is a POMIS relative to $\langle \mc{P}, Y\rangle$, then there exists a MAG $\mc{M}$ conforming to $\mc{P}$ such that $\mf{X}$ is a POMIS relative to $\langle \mc{M}, Y \rangle$. 
\end{restatable}
\begin{proof}[Proof of
\Cref{prop: if POMIS for PAG then POMIS for MAG}] Let $\mc{S}^\ast$ witness that $\mf{X}$ is a POMIS relative to $\langle\mc{P},Y\rangle$. Since $\mc{S}^\ast$ conforms to $\mc{P}$, choose a MAG $\mc{M}\in[\mc{P}]$ to which it conforms.

First, we will show that $\mf{X}$ is an MIS relative to $\langle\mc{M},Y\rangle$. For the sake of contradiction, suppose $\mf{X}$ is not an MIS relative to $\langle \mc{M}, Y \rangle$. Then, repeatedly removing
a graphically redundant subset yields an MIS $\mf{Z}\subsetneq\mf{X}$ such that $\mu_{\mf{x}}=\mu_{\mf{x}\cap\mf{Z}}$ for every assignment $\mf{x}$ and every SCM conforming to $\mc{M}$.
In particular, under $\mc{S}^\ast$, maximizing over assignments gives $\mu_{\mf{x}^\ast} = \mu_{\mf{z}^\ast}$. By \Cref{prop: if MIS then DMIS for cd and mag}, $\mf{Z}$ is a DMIS for $\mc{M}$ and hence for $\mc{P}$. This contradicts the strict optimality of $\mf{X}$ against every other DMIS in $\mb{D}_{\mc{P},Y}$ under $\mc{S}^\ast$.

Thus, $\mf{X}$ is an MIS, and therefore a DMIS, for $\mc{M}$. Moreover, $\mb{D}_{\mc{M},Y}\subseteq\mb{D}_{\mc{P},Y}$, since an SCM witnessing DMIS membership for $\mc{M}$ also conforms
to $\mc{P}$. The strict inequalities witnessed by $\mc{S}^\ast$ therefore hold against every member of $\mb{D}_{\mc{M},Y}\setminus\{\mf{X}\}$. Consequently, $\mf{X}$ is a POMIS relative to $\langle\mc{M},Y\rangle$. 
\end{proof}

\paragraph{Failure of the converse.}
The converse of \Cref{prop: if POMIS for PAG then POMIS for MAG} does \textbf{not} hold. To see this, consider the PAG shown in \Cref{fig: 3var a}. The empty set is a POMIS for both MAGs shown in \Cref{fig: 3var b,fig: 3var c}. In each MAG, the edge entering $Y$ is invisible, so its endpoints may be confounded in a represented causal diagram. POMIS membership here asserts the existence of a compatible SCM in which the empty intervention is strictly optimal among MISs; it need not be optimal in the SCM displayed earlier. However, in every compatible SCM, at least one of $A$ and $B$ is not an ancestor of $Y$, so intervening on that variable leaves the distribution of $Y$ unchanged; that is, either $P(y \mid do(a)) = P(y)$ or $P(y \mid do(b)) = P(y)$. Since both singleton intervention sets are DMISs for the PAG, the empty set cannot be strictly better than every other DMIS.

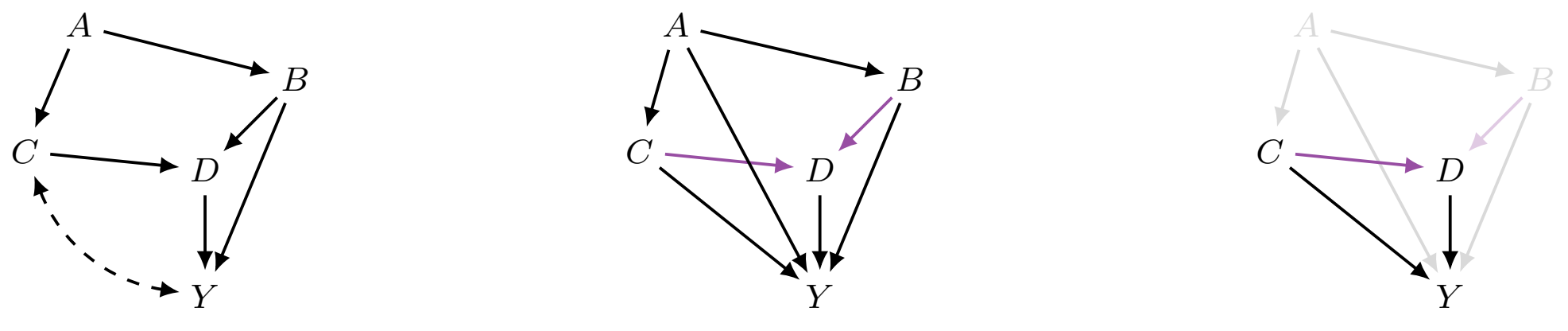
\begin{figure}[t]
    \centering  
    \begin{minipage}[b]{0.32\textwidth}\centering  
    \subfloat[$\mc{G}$]{
        \begin{tikzpicture}[x=1.5cm,y=1.5cm,>={Latex[width=1.4mm,length=1.6mm]},
        line width=0.25mm,
        oo/.tip={Circle[open,length=3.5pt,width=3.5pt]},
        RR/.style={draw=none,circle,inner sep=0mm, minimum size=4mm}
        ]
        \node[RR] (C) at (0,0.8) {$C$};
        \node[RR] (D) at (1,0.7) {$D$};
        \node[RR] (Y) at (1,0) {$Y$};
        \node[RR] (B) at (1.5,1.2) {$B$};
        \node[RR] (A) at (0.3,1.5) {$A$};

        \draw[->] (A) -- (B);
        \draw[->] (A) -- (C);
        \draw[->] (C) -- (D);
        \draw[->] (D) -- (Y);
        \draw[->] (B) -- (D);
        \draw[->] (B) -- (Y);
        
        \draw[dashed, <->] (C) to[bend right = 30] (Y);
        \end{tikzpicture}\label{fig: 5var2 a}}
    \end{minipage}\hfill
    \begin{minipage}[b]{0.32\textwidth}\centering  
    \subfloat[$\mc{M}$]{
        \begin{tikzpicture}[x=1.5cm,y=1.5cm,>={Latex[width=1.4mm,length=1.6mm]},
        line width=0.25mm,
        oo/.tip={Circle[open,length=3.5pt,width=3.5pt]},
        RR/.style={draw=none,circle,inner sep=0mm, minimum size=4mm}
        ]
        \node[RR] (C) at (0,0.8) {$C$};
        \node[RR] (D) at (1,0.7) {$D$};
        \node[RR] (Y) at (1,0) {$Y$};
        
        \node[RR] (B) at (1.5,1.2) {$B$};
        \node[RR] (A) at (0.2,1.5) {$A$};

        \draw[->] (A) -- (B);
        \draw[->] (A) -- (C);
        \draw[<-,draw = betterpurple] (D) -- (C);
        \draw[->] (D) -- (Y);
        \draw[->,draw = betterpurple] (B) -- (D);
        \draw[->] (B) -- (Y);
        
        \draw[->] (C) -- (Y);
        \draw[->] (A) -- (Y);
        
        \end{tikzpicture}\label{fig: 5var2 b}}
    \end{minipage} \hfill
    \begin{minipage}[b]{0.32\textwidth}\centering  
    \subfloat[$\mc{H}$]{
        \begin{tikzpicture}[x=1.5cm,y=1.5cm,>={Latex[width=1.4mm,length=1.6mm]},
        line width=0.25mm,
        oo/.tip={Circle[open,length=3.5pt,width=3.5pt]},
        RR/.style={draw=none,circle,inner sep=0mm, minimum size=4mm}
        ]
        \node[RR] (C) at (0,0.8) {$C$};
        \node[RR] (D) at (1,0.7) {$D$};
        \node[RR] (Y) at (1,0) {$Y$};
        \node[RR] (B) at (1.5,1.2) {\color{gray!30}$B$};
        \node[RR] (A) at (0.2,1.5) {\color{gray!30}$A$};

        \draw[->, draw = gray!30] (A) -- (B);
        \draw[->, draw = gray!30] (A) -- (C);
        \draw[->, draw = gray!30] (A) -- (Y);
        \draw[->, draw = betterpurple!30] (B) -- (D);
        \draw[->, draw = gray!30] (B) -- (Y);
        
        \draw[<-, draw = betterpurple] (D) -- (C);
        \draw[->] (D) -- (Y);
        \draw[->] (C) -- (Y);
        
        \end{tikzpicture}\label{fig: 5var2 c}}
    \end{minipage}
    \caption{\small (a, b) Causal diagram and corresponding MAG. (c) Induced graph $\mc{M} \setminus \{A,B\}$.}
    \label{fig: 5var2}
\end{figure}
\paragraph{MUCT and IB for ancestral graphs.}
We now investigate the graphical characterization of POMISs for MAGs and PAGs. To this end, we extend the notions of MUCT and IB to these graph classes. The main challenge is that ancestral graphs do not display the underlying latent structure or the inducing paths that give rise to their adjacencies. We therefore use edge visibility to identify directed edges that cannot be confounded in any represented causal diagram. 
To generalize the UC-territory, we introduce a \textit{possible c-component} \citep{jaber2018causal}, which provides a necessary condition for nodes to belong to the same c-component in an underlying causal diagram.
\begin{definition} [pc-component \citep{jaber2018causal}]\label{def: pc}
    Two nodes are in the same \textit{possible c-component} (pc-component) if there is a path between them such that: 
    \begin{itemize}
        \item[1.] all non-endpoint nodes along the path are colliders; and
        \item[2.] none of the edges are visible.
    \end{itemize}
\end{definition}
The pc-component relation need not be transitive. We use $\pc(X)_{\mc{Q}}$ for the set of vertices related to $X$ by the path condition in the definition, including $X$ itself, and define $\pc(\mf{X})_{\mc{Q}} \triangleq \bigcup_{X \in \mf{X}}\pc(X)_{\mc{Q}}$. Thus, these sets need not form a partition. 

For example, $A$ and $B$ are in the same pc-component in the PAG $\mc{P}$ in \Cref{fig:foundation_ancestral_pag} because they are connected by the collider path $A \cdedge C \dcedge B$, whose edges are all invisible, and indeed $\pc(A)_\mc{P} = \{A,B,C\}$. Furthermore, because $A \notin \pc(Y)_\mc{P} = \{B,Y\}$, $A$ and $Y$ cannot belong to the same c-component in any causal diagram conforming to $\mc{P}$. We now generalize MUCT and IB for PMGs. 

\begin{definition} [Unobserved confounders' territory for PMGs] \label{def: MUCT for PMG}
    Given information $\langle \mc{Q}, Y\rangle$ and intervention set $\mf{X} \subseteq \mf{V} \setminus \{Y\}$, let $\mc{H} = \mc{Q}[\Poss\An(Y)_{\mc{Q} \setminus \mf{X}}]$. A set of variables $\mf{T} \subseteq \Poss\An(Y)_{\mc{Q} \setminus \mf{X}}$ containing $Y$ is called a \textit{UC-territory} with respect to $\langle \mc{Q}, Y, \mf{X} \rangle$ if $\Poss\De(\mf{T})_\mc{H} = \mf{T}$ and $\pc(\mf{T})_{\mc{H}} = \mf{T}$. A UC-territory $\mf{T}$ is called a \textit{minimal} UC-territory (MUCT) if no $\mf{T}' \subsetneq \mf{T}$ is a UC-territory.
\end{definition}
\begin{definition} [Interventional border for PMGs] \label{def: IB for PMG}
    Let $\mf{T}$ be a minimal UC-territory with respect to $\langle \mc{Q}, Y, \mf{X} \rangle$. Then $\Pa(\mf{T})_{\mc{Q}} \setminus \mf{T}$ is called an \textit{interventional border} (IB) with respect to $\langle \mc{Q}, Y, \mf{X} \rangle$.
\end{definition}
We denote MUCT and IB with respect to $\langle \mc{Q}, Y , \mf{X}\rangle$ as $\MUCT(\mc{Q}, Y, \mf{X})$ and $\IB(\mc{Q}, Y, \mf{X})$, respectively. Note that the MUCT is unique. Starting from $\{Y\}$ and repeatedly adding possible descendants and pc-related vertices in $\mc{H}$ yields a closed set contained in every UC-territory. Since the graph is finite, this process terminates at the unique minimal UC-territory. 

For concreteness, consider $\mc{M}$ and $\mf{X} = \{A,B\}$ in \Cref{fig: 5var2 b}. Here, we omit $\Poss$ because we are working with a MAG. Let $\mc{H}$ be the induced graph $\mc{M}[\An(Y)_{\mc{M}\setminus \mf{X}}]$ (shown in \Cref{fig: 5var2 c}). In $\mc{H}$, the nodes $C$ and $Y$ are in the same pc-component, and $D$ is a descendant of $C$. This implies that $\mf{T} = \{C,D,Y\}$ is the minimal closed set for $\De_\mc{H}$ and $\pc_\mc{H}$. Therefore, we have:
\begin{align}
    \MUCT(\mc{M},Y, \mf{X}) &= \mf{T} = \{C,D,Y\}\\
    \IB(\mc{M},Y, \mf{X}) &= \Pa(\mf{T})_\mc{M} \setminus \mf{T} \\
    &= \{A,B\}.
\end{align}

\subsection{Characterization of POMIS for MAGs}\label{sec: pomis for mags}

The MUCT and IB construction gives a necessary and sufficient condition for POMIS membership in a MAG. 

\begin{restatable}[Graphical characterization of POMIS for MAGs]{theorem}{IBeqPOMISforMAG}\label{thm: IB = POMIS for MAG}
    $\mf{X} \subseteq \mf{V} \setminus \{Y\}$  is a POMIS relative to $\langle \mc{M}, Y \rangle$ \textit{if and only if} $\mf{X} = \mathsf{IB}(\mc{M}, Y, \mf{X})$.
\end{restatable}
The proof is given in \Cref{sec: proof of thm: IB = POMIS for MAG}. For the empty intervention in each MAG in \Cref{fig: 3var b,fig: 3var c}, the MUCT consists of $Y$ and its parent, and the IB is empty. Thus, \Cref{thm: IB = POMIS for MAG} confirms the POMIS membership stated above. 

For any intervention set, we seek a causal diagram whose corresponding MAG is $\mc{M}$ and whose MUCT and IB after intervention agree with those computed from $\mc{M}$. \Cref{lem: causal diagram construction} establishes that such a diagram always exists. We also retain exactly the directed edges of $\mc{M}$, so that the two graphs have the same proper directed paths and hence the same MISs. 

Start by copying all directed and bidirected edges of $\mc{M}$ into a causal diagram, and compute its MUCT after intervention. If this set is smaller than the MUCT computed from the MAG, choose a variable inside the current set that has invisible parents outside it but within $\An(Y)_{\mc{M} \setminus \mf{X}}$. For every such parent, add a bidirected edge to the chosen variable, while keeping the original directed edge. All these edges to the chosen variable are added together. Recompute the MUCT and repeat until it agrees with the MAG's MUCT. The proof explains both why a variable can always be chosen when the sets differ and why these additions preserve the corresponding MAG. 

\begin{figure}[t]
    \centering
    \tikzset{
        scbconstructiongraph/.style={x=1.5cm,y=1.5cm,font=\scriptsize,
            >={Latex[width=1.4mm,length=1.6mm]},line width=0.25mm,
            draw=black,text=black},
        scbconstructionnode/.style={draw=none,circle,fill=white,
            inner sep=0mm,minimum size=5mm,text=black},
        scbconstructionterritory/.style={scbconstructionnode,draw=black,fill=none},
        scbconstructionborder/.style={scbconstructionnode,draw=black,fill=bettergreen!25},
        scbconstructionintervened/.style={draw=black,line width=0.6mm},
        scbconstructionarea/.style={draw=betterblue!10,fill=betterblue!10,
            line width=5mm,line cap=round,line join=round}
    }
    \newcommand{\scbconstructionpositions}{
        \coordinate (A) at (0.2,1.5);
        \coordinate (B) at (1.5,1.2);
        \coordinate (C) at (0,0.8);
        \coordinate (D) at (1,0.7);
        \coordinate (Y) at (1,0);
        \path[use as bounding box] (-0.32,-0.30) rectangle (1.78,1.72);
    }
    \newcommand{\scbconstructionretained}[1][]{
        \draw[->] (A) -- (C);
        \draw[->] (A) -- (Y);
        \draw[->,#1] (B) -- (D);
        \draw[->] (B) -- (Y);
        \draw[->,#1] (C) -- (D);
        \draw[->] (C) -- (Y);
        \draw[->] (D) -- (Y);
    }
    \begin{minipage}[b]{0.32\textwidth}\centering
    \subfloat[$\mc{M}$]{
        \begin{tikzpicture}[scbconstructiongraph]
            \scbconstructionpositions
            \foreach \v in {A,B,C,D,Y}
                \node[scbconstructionnode] (\v) at (\v) {$\v$};
            \draw[->] (A) -- (B);
            \scbconstructionretained[draw=betterpurple,line width=0.25mm]
        \end{tikzpicture}\label{fig:scbmec_mag_construction_a}}
    \end{minipage}\hfill
    \begin{minipage}[b]{0.32\textwidth}\centering
    \subfloat[Initial diagram]{
        \begin{tikzpicture}[scbconstructiongraph]
            \scbconstructionpositions
            \foreach \v in {A,B}
                \node[scbconstructionborder,scbconstructionintervened] (\v) at (\v) {$\v$};
            \foreach \v in {C,D}
                \node[scbconstructionborder] (\v) at (\v) {$\v$};
            \node[scbconstructionterritory] (Y) at (Y) {$Y$};
            \draw[->,draw=gray!30] (A) -- (B);
            \scbconstructionretained
            \begin{pgfonlayer}{back}
                \fill[betterblue!10] (Y.center) circle[radius=3.5mm];
            \end{pgfonlayer}
        \end{tikzpicture}\label{fig:scbmec_mag_construction_b}}
    \end{minipage}\hfill
    \begin{minipage}[b]{0.32\textwidth}\centering
    \subfloat[After addition]{
        \begin{tikzpicture}[scbconstructiongraph]
            \scbconstructionpositions
            \foreach \v in {A,B}
                \node[scbconstructionborder,scbconstructionintervened] (\v) at (\v) {$\v$};
            \foreach \v in {C,D,Y}
                \node[scbconstructionterritory] (\v) at (\v) {$\v$};
            \draw[->,draw=gray!30] (A) -- (B);
            \scbconstructionretained
            \draw[<->,dashed] (C) to[bend right=35] (Y);
            \draw[<->,dashed] (D) to[bend left=60] (Y);
            \begin{pgfonlayer}{back}
                \draw[scbconstructionarea]
                    (C.center) to[bend right=35] (Y.center)
                    to[bend right=60] (D.center) -- cycle;
            \end{pgfonlayer}
        \end{tikzpicture}\label{fig:scbmec_mag_construction_c}}
    \end{minipage}
    \caption{\small Constructing a causal diagram for the MAG in \Cref{fig: 5var2 b}, with $\mf{X}=\{A,B\}$. Copying the MAG's edges initially gives IB $\{A,B,C,D\} \neq \mf{X}$ in (b). Adding $C\bedge Y$ and $D\bedge Y$ together gives IB $\{A,B\} = \mf{X}$ in (c). Light blue regions mark MUCTs, green fill marks IB variables, and thick outlines mark intervention targets.}
    \label{fig:scbmec_mag_construction}
\end{figure}
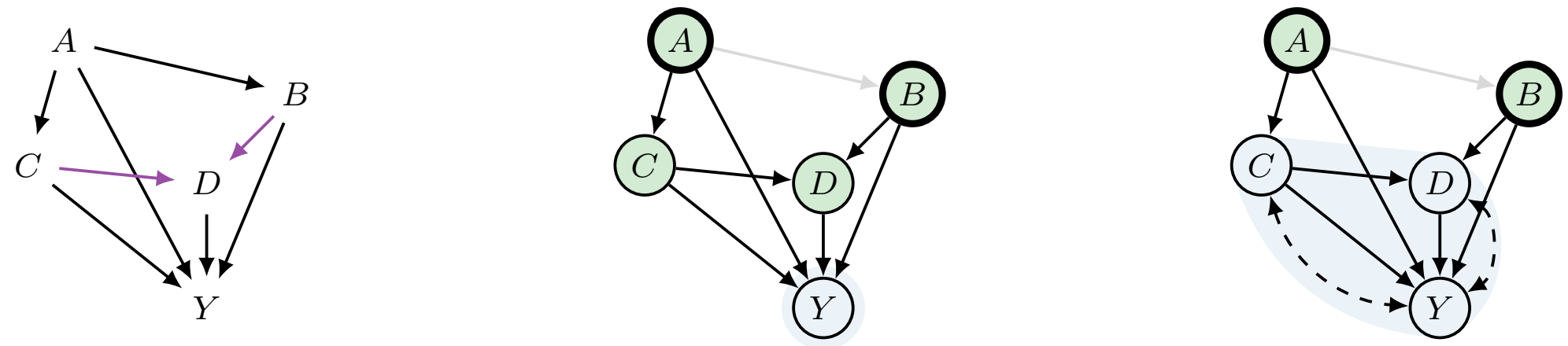

\Cref{fig:scbmec_mag_construction} applies the construction to the running example in \Cref{fig: 5var2 b}. With $\mf{X} = \{A,B\}$, the relevant non-intervened ancestors are $\An(Y)_{\mc{M} \setminus \mf{X}} = \{C,D,Y\}$. The copied causal diagram has no bidirected edges, so its MUCT after intervention is initially just $\{Y\}$. In the original MAG, however, both $C \to Y$ and $D \to Y$ are invisible: every other vertex is adjacent to $Y$, so neither edge can have a visibility witness. The construction therefore adds $C \bedge Y$ and $D \bedge Y$ together. The three vertices are now connected by bidirected edges, giving IB $\{A,B\} = \mf{X}$, as required. Therefore, we conclude that $\mf{X} = \{A,B\}$ is a POMIS with respect to $\langle \mc{M}, Y \rangle$. 

The constructed diagram retains exactly the directed edges of the MAG and need not coincide with $\mc{G}$ in \Cref{fig: 5var2 a}. That original diagram is also represented by $\mc{M}$ and has $\{A,B\}$ as a POMIS: after intervention, $C \bedge Y$ places $C$ in the territory, descendant closure includes $D$, and the border is again $\{A,B\}$. 

\subsection{Searching for POMISs in Represented MAGs}\label{sec: pomis for pags}
The remainder of this section targets $\widehat{\mb{P}}_{\mc{P},Y} \triangleq \bigcup_{\mc{M} \in [\mc{P}]} \mb{P}_{\mc{M},Y}$, where $\mb{P}_{\mc{M},Y}$ is the collection of POMISs of a represented MAG. By \Cref{prop: if POMIS for PAG then POMIS for MAG}, the PAG-wide collection of POMISs in \Cref{def: POMIS for PAG} satisfies $\mb{P}_{\mc{P},Y} \subseteq \widehat{\mb{P}}_{\mc{P},Y}$, and this inclusion can be strict. The search below is sound and complete for the union. 

This union retains an optimal action for every compatible SCM. Fix such an SCM and its corresponding MAG $\mc{M}$. For any intervention set $\mf{X}$, let $\mf{T}=\MUCT(\mc{M},Y,\mf{X})$ and $\mf{W}=\IB(\mc{M},Y,\mf{X})$. The necessity proof of \Cref{thm: IB = POMIS for MAG} gives $\mu_{\mf{x}^\ast} \leq \mu_{\mf{w}^\ast}$ in this SCM. Moreover, $\An(Y)_{\mc{M}\setminus\mf{W}}=\mf{T}$: descendant closure keeps a directed path from every member of $\mf{T}$ to $Y$ inside $\mf{T}$, while any directed path from outside $\mf{T}$ to $Y$ enters $\mf{T}$ through a parent in $\mf{W}$. Every path used to generate $\mf{T}$ by descendant and pc closure also stays inside $\mf{T}$. With visibility inherited from the same full MAG, restricting the closure construction to $\mc{M}[\mf{T}]$ therefore leaves its result unchanged. Thus $\MUCT(\mc{M},Y,\mf{W})=\mf{T}$ and $\IB(\mc{M},Y,\mf{W})=\mf{W}$. By \Cref{thm: IB = POMIS for MAG}, $\mf{W}$ is a POMIS of $\mc{M}$. Applying this argument to an optimal intervention set proves that the union retains an optimal action. 

To search for $\widehat{\mb{P}}_{\mc{P},Y}$, we partition $[\mc{P}]$ according to the possible orientations of circle marks at vertices in $\mf{X} \cup \{Y\}$. Each part of this partition is represented by a locally oriented PMG, whose graphical structure is used in the terminal IB test. To this end, we first review local transformations of \citet{wang2023sound}, which orient the circle marks incident to a given variable.

\paragraph{Local transformation.} 
We define a \textit{local transformation} \citep{wang2023sound} at $A \in \mf{V}$ as a set:
\begin{align}
    \mf{C}^{\mc{Q}}_A \subseteq \{V \in \Adj(A)_{\mc{Q}} \mid A \caedge V\}.
\end{align}
The vertices in $\mf{C}^{\mc{Q}}_A$ are those whose edges with a circle mark at $A$ will be oriented with arrowheads at $A$. That is, for $V \in \mf{C}_A^{\mc{Q}}$ and $V' \in \{U \in \Adj(A)_{\mc{Q}} : A \caedge U\} \setminus \mf{C}_A^{\mc{Q}}$, the local transformation places an arrowhead at $A$ on the first edge and a tail at $A$ on the second edge, with $\mc{R}_{\texttt{SB}}$ completing the latter as a directed edge. 
\begin{itemize}
    \item[$\mc{R}_{\texttt{SB}}$]  If $X \ucedge Y$, then orient $X \ucedge Y$ as $X \to Y$.
\end{itemize}

The rule $\mc{R}_{\texttt{SB}}$ naturally follows from the absence of selection bias\footnote{Additional rules for more general local transformations are studied by \citet{venkateswaran2024towards,wang2024newrulescausalidentification} and \citet{wangpolynomial}.}. The resulting local orientation is: 
\begin{align}
    V \acedge A \caedge V' \implies V \adedge A \to V'.
\end{align}
We say that a local transformation is \textit{valid} if at least one MAG in $[\mc{Q}]$ is consistent with all of its chosen orientations. A local transformation can be viewed as a candidate statement: it resolves the circle marks at one vertex. In the search below, we consider alternative statements of this kind and retain those compatible with at least one represented MAG. Notably, \citet{wang2022sound,wang2023sound} and \citet{wangefficient} demonstrated that the rules $\mc{R}_1$--$\mc{R}_3$, $\mc{R}_4'$, $\mc{R}_8$--$\mc{R}_{10}$, and $\mc{R}_{\texttt{SB}}$ are complete for incorporating local transformations. 
\begin{itemize}
    \item[$\mc{R}'_4$] If $\langle X, \cdots, W,Z,Y \rangle$ is a discriminating path between $X$ and $Y$ for $Z$, and $Z \caedge Y$, then orient $Z \caedge Y$ as $Z \to Y$.
\end{itemize}
In other words, after applying a valid local transformation and then applying the rules until no further orientation is possible, the orientations in the resulting PMG are complete. 
The following theorem shows that the validity of a given local transformation can be verified in $\mc{O}(\lvert \mf{V} \rvert^3)$ time. 
\begin{theorem}[Theorem 3 and Remark 3 in \citet{wang2023sound}]\label{thm: wang2023 thm 3}
    Let $\mc{Q}$ be a PMG obtained from a PAG $\mc{P}$ by applying valid local transformations and the orientation rules $\mc{R}_1$--$\mc{R}_3$, $\mc{R}_4'$, $\mc{R}_8$--$\mc{R}_{10}$, and $\mc{R}_{\texttt{SB}}$ to closure after each transformation. Given a local transformation $\mf{C}_A^{\mc{Q}}$, there exists a MAG $\mc{M} \in [\mc{Q}]$ with $A \daedge V$ for all $V \in \mf{C}_A^{\mc{Q}}$, and $A \to V'$ for all $V' \in \{V \in \Adj(A)_{\mc{Q}} \mid A \caedge V\} \setminus \mf{C}_A^{\mc{Q}}$ \textit{if and only if} $\mf{C}_A^{\mc{Q}}$ satisfies the following conditions:
    {
\begin{itemize}[leftmargin=10mm]
    \item[1.] $\Poss\De(A)_{\mc{Q}\setminus \mf{C}_A^{\mc{Q}}} \cap \Pa(\mf{C}_A^{\mc{Q}})_\mc{Q} = \emptyset$;
    
    \item[2.] $\mc{Q}[\mf{C}_A^{\mc{Q}}]$ is a complete graph;
    \item[3.] Orient the subgraph $\mc{Q}[\Poss\De(A)_{\mc{Q}\setminus \mf{C}_A^{\mc{Q}}}\setminus\{A\}]$ as follows until no further orientations are required: 
    
    For any circle edge $V_l \cedge V_j$, orient the edge as $V_l \to V_j$ if:
    \begin{itemize}
        \item[1.] $\mf{F}_{V_l} \setminus \mf{F}_{V_j} \neq \emptyset$; or
        
        \item[2.] $\mf{F}_{V_l} = \mf{F}_{V_j}$ and there exists $V_m \in \Poss\De(A)_{\mc{Q}\setminus \mf{C}_A^{\mc{Q}}}\setminus\{A\}$ not adjacent to $V_j$ such that $V_m \to V_l \cedge V_j$.
    \end{itemize}
    where $\mf{F}_{V_l} = \{V \in \mf{C}_A^{\mc{Q}} \cup \{A\} \mid V \acedge V_l \in \mc{Q}\}$ is computed before these orientations. These orientations must introduce no new v-structures and must not require both $V_l \to V_j$ and $V_j \to V_l$ for the same original circle edge. 
\end{itemize}
}
\end{theorem}
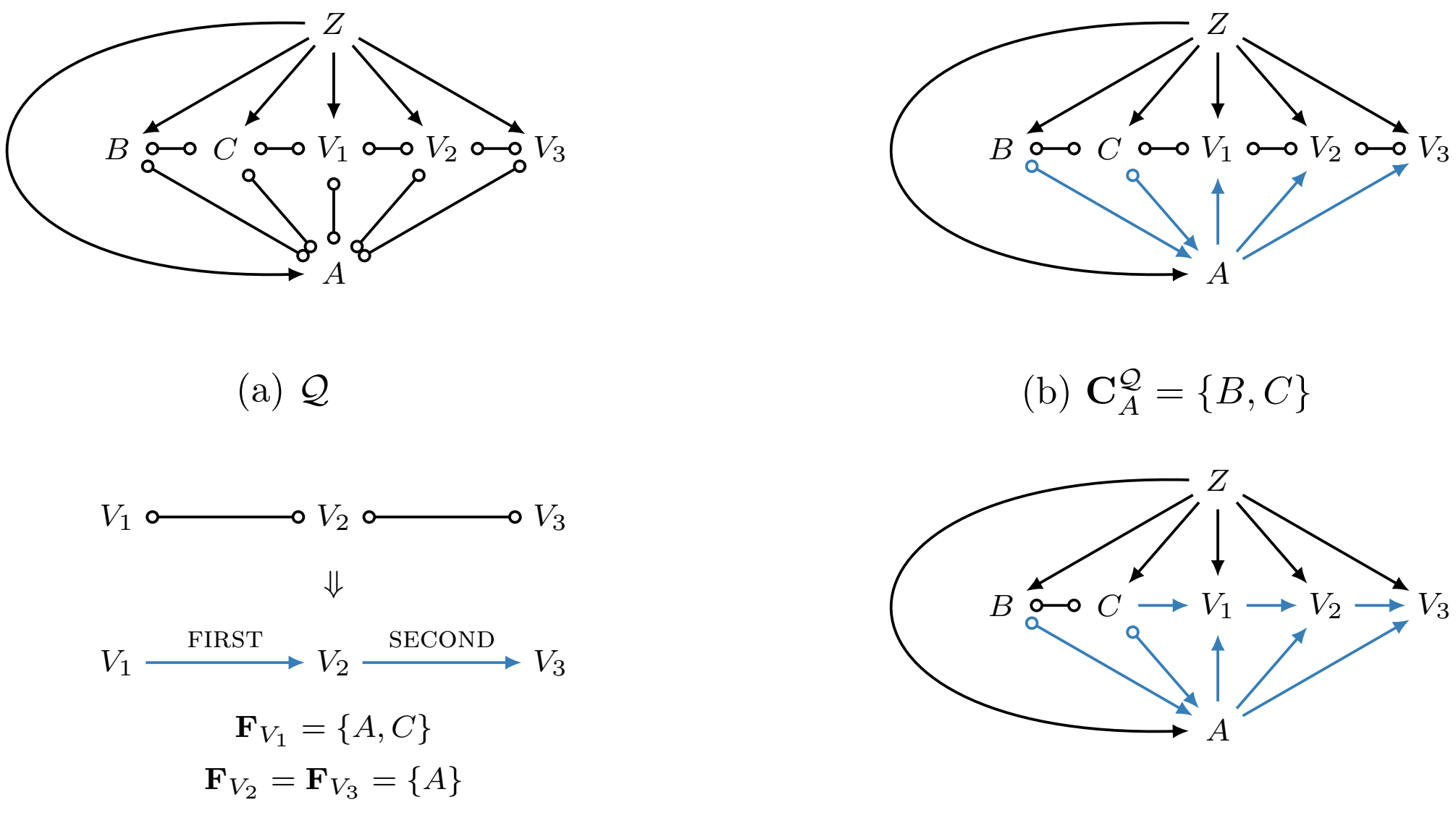
\begin{figure}[t]
    \centering
    \tikzset{
        ltgraph/.style={x=0.9cm,y=0.95cm,font=\scriptsize,
            >={Latex[width=1.3mm,length=1.5mm]},line width=0.25mm},
        ltcircle/.tip={Circle[open,length=3.5pt,width=3.5pt]},
        ltvertex/.style={draw=none,circle,inner sep=0mm,minimum size=5mm}
    }
    \newcommand{\ltvertices}{
        \path[use as bounding box] (-1.25,-1.9) rectangle (4.8,1.55);
        \node[ltvertex] (Z) at (2.2,1.2) {$Z$};
        \node[ltvertex] (B) at (0,0) {$B$};
        \node[ltvertex] (C) at (1.1,0) {$C$};
        \node[ltvertex] (V1) at (2.2,0) {$V_1$};
        \node[ltvertex] (V2) at (3.3,0) {$V_2$};
        \node[ltvertex] (V3) at (4.4,0) {$V_3$};
        \node[ltvertex] (A) at (2.2,-1.2) {$A$};
    }
    \newcommand{\ltparents}{
        \foreach \v in {B,C,V1,V2,V3}
            \draw[->,black ] (Z) -- (\v);
        \draw[->,black ] (Z) .. controls (-2.1,1.3) and (-2.1,-1.3) .. (A);
    }
    \newcommand{\ltcirclepath}{
        \draw[{ltcircle}-{ltcircle}] (B) -- (C);
        \draw[{ltcircle}-{ltcircle}] (C) -- (V1);
        \draw[{ltcircle}-{ltcircle}] (V1) -- (V2);
        \draw[{ltcircle}-{ltcircle}] (V2) -- (V3);
    }
    \newcommand{\ltlocalmarks}{
        \foreach \v in {B,C}
            \draw[{ltcircle}->,betterblue] (\v) -- (A);
        \foreach \v in {V1,V2,V3}
            \draw[->,betterblue] (A) -- (\v);
    }
    \begin{minipage}[b]{0.47\textwidth}\centering
    \subfloat[$\mc{Q}$]{
        \begin{tikzpicture}[ltgraph]
            \ltvertices\ltparents\ltcirclepath
            \foreach \v in {B,C,V1,V2,V3}
                \draw[{ltcircle}-{ltcircle}] (A) -- (\v);
        \end{tikzpicture}\label{fig: local transformation a}}
    \end{minipage}
    \hfill
    \begin{minipage}[b]{0.47\textwidth}\centering
    \subfloat[$\mf{C}_A^{\mc{Q}}=\{B,C\}$]{
        \begin{tikzpicture}[ltgraph]
            \ltvertices\ltparents\ltcirclepath\ltlocalmarks
        \end{tikzpicture}\label{fig: local transformation b}}
    \end{minipage}
    
    \medskip
    \begin{minipage}[b]{0.47\textwidth}\centering
    \subfloat[Condition~3]{
        \begin{tikzpicture}[ltgraph]
            \path[use as bounding box] (-1.25,-1.9) rectangle (4.8,1.55);
            \node[ltvertex] (P1) at (0,0.85) {$V_1$};
            \node[ltvertex] (P2) at (2.2,0.85) {$V_2$};
            \node[ltvertex] (P3) at (4.4,0.85) {$V_3$};
            \draw[{ltcircle}-{ltcircle}] (P1) -- (P2);
            \draw[{ltcircle}-{ltcircle}] (P2) -- (P3);
            \node at (2.2,0.2) {$\Downarrow$};
            \node[ltvertex] (V1) at (0,-0.55) {$V_1$};
            \node[ltvertex] (V2) at (2.2,-0.55) {$V_2$};
            \node[ltvertex] (V3) at (4.4,-0.55) {$V_3$};
            \draw[->,draw = betterblue] (V1) -- node[above,font=\scriptsize]{\sc first} (V2);
            \draw[->,draw = betterblue] (V2) -- node[above,font=\scriptsize]{\sc second} (V3);
            \node at (2.2,-1.2) {$\mf{F}_{V_1}=\{A,C\}$};
            \node at (2.2,-1.7) {$\mf{F}_{V_2}=\mf{F}_{V_3}=\{A\}$};
        \end{tikzpicture}\label{fig: local transformation c}}
    \end{minipage}
    \hfill
    \begin{minipage}[b]{0.47\textwidth}\centering
    \subfloat[The resulting PMG]{
        \begin{tikzpicture}[ltgraph]
            \ltvertices\ltparents\ltlocalmarks
            \draw[{ltcircle}-{ltcircle}] (B) -- (C);
            \draw[->,betterblue] (C) -- (V1);
            \draw[->,betterblue] (V1) -- (V2);
            \draw[->,betterblue] (V2) -- (V3);
        \end{tikzpicture}\label{fig: local transformation d}}
    \end{minipage}
    \caption{\small Local transformation and the conditions in \Cref{thm: wang2023 thm 3}. Panels (a)--(b) specify the local transformation, (c) illustrates the third condition, and (d) shows the PMG after orientation closure. Blue edges show local orientations and their consequences. In (c), \textsc{first} and \textsc{second} refer to the two subconditions of condition~3.}
    \label{fig: local transformation}
\end{figure}

For concreteness, consider the PMG $\mc{Q}$ shown in \Cref{fig: local transformation a}, in which $Z$ is a parent of every other vertex. The choice $\mf{C}_A^{\mc{Q}} = \{B,C\}$ supplies the local transformation shown in \Cref{fig: local transformation b}:
\begin{align}
    B \cdedge A \dcedge C \quad\text{~and~}\quad A \to V_i \quad\text{~for~} i = 1,2,3.  
\end{align}
The marks at $B$ and $C$ remain circles. Validity requires a MAG consistent with both $\mc{Q}$ and these local orientations. 

The first condition of \Cref{thm: wang2023 thm 3} checks whether the local transformation would create a directed or almost directed cycle. 
\begin{align}
    \text{For~} \mf{C}_A^{\mc{Q}} &= \{B,C\},~\Poss\De(A)_{\mc{Q} \setminus \mf{C}_A^{\mc{Q}}} \cap \Pa(\mf{C}_A^{\mc{Q}})_{\mc{Q}} \\
    & = \{A, V_1,V_2,V_3 \} \cap \{Z\}\\
    & = \emptyset. 
\end{align}
All arrows incident to $Z$ point away from $Z$, preventing a possibly directed path from $A$ back to this parent of $B$ and $C$. Thus, the local choice cannot close a directed or almost directed cycle through $Z$. 

Next, the second condition checks whether the local transformation introduces a new v-structure at $A$. Since $B$ and $C$ are adjacent, $\mc{Q}[\{B,C\}]$ is complete. Their adjacency shields the triple $B\cdedge A\dcedge C$, so the two new arrowheads do not create a v-structure at $A$. 

Finally, the third condition checks whether orientations propagated to possible descendants of $A$ create new v-structures or require conflicting directions. The induced subgraph in \Cref{fig: local transformation c} is $V_1\cedge V_2\cedge V_3$, with: 
\begin{align}
    \mf{F}_{V_1} = \{A,C\}, \qquad \mf{F}_{V_2} = \mf{F}_{V_3} = \{A\}. 
\end{align}
Here $A$ is excluded from the induced subgraph but retained in the $\mf{F}$ sets. The first subcondition requires $V_1 \to V_2$ because $\mf{F}_{V_1} \setminus \mf{F}_{V_2} = \{C\} \neq \emptyset$. Since $\mf{F}_{V_2} = \mf{F}_{V_3}$ and $V_1 \to V_2 \cedge V_3$ is unshielded, the second subcondition then requires $V_2 \to V_3$. Both orientations avoid new v-structures and conflicting directions. 

All three conditions hold, so the local transformation is valid. Orientation closure yields the PMG shown in \Cref{fig: local transformation d}, with $C \to V_1 \to V_2 \to V_3$ fixed. The edges $B \cedge C$, $B \cdedge A$, and $C \cdedge A$ retain four circle marks. Therefore, the resulting PMG represents multiple MAGs consistent with the local transformation. 

The PMG incorporating local transformations satisfies the properties below. 
\begin{theorem}[Theorem 1 in \citet{wang2023sound}]\label{thm: wang2023 thm 1}
    Let $\mc{Q}$ be a PMG obtained from a PAG $\mc{P}$ by applying valid local transformations and applying the orientation rules $\mc{R}_1$--$\mc{R}_3$, $\mc{R}_4'$, $\mc{R}_8$--$\mc{R}_{10}$, and $\mc{R}_{\texttt{SB}}$ to closure after each transformation. Then $\mc{Q}$ satisfies the following properties:
    \begin{itemize}
        \item\textbf{Closed.~} $\mc{Q}$ is closed under the orientation rules.
        
        \item\textbf{Invariant.~} The arrowheads ($\mathord{>}$) and tails ($-$) in $\mc{Q}$ are invariant in all the MAGs consistent with $\mc{Q}$.

        \item\textbf{Chordal.~}\footnote{A graph is \textit{chordal} if any cycle of length four or more has a chord, which refers to an edge joining two vertices that are not adjacent in the cycle.} Circle components in $\mc{Q}$ are chordal.
        
        \item\textbf{Balanced.~} For any three nodes $A, B, C$ in $\mc{Q}$, if $A \adedge B \caedge C$, then there is an edge between $A$ and $C$ with an arrowhead at $C$, namely, $A \adedge C$. Furthermore, if the edge between $A$ and $B$ is $A \to B$, then the edge between $A$ and $C$ is either $A \to C$ or $A \cdedge C$ (i.e., it is not $A \bedge C$).
        
        \item\textbf{Complete.~} For each circle at $A$ on any edge $A \caedge B$ in $\mc{Q}$, there exist MAGs $\mc{M}_1$ and $\mc{M}_2$ consistent with $\mc{Q}$ such that $A \daedge B$ in $\mc{M}_1$ and $A \to B$ in $\mc{M}_2$. 
    \end{itemize}    
\end{theorem}
Thus, PMGs obtained by valid local transformations retain the listed graphical properties of PAGs, while representing a class of MAGs restricted by the incorporated local transformations.  

\paragraph{Properties: inherited from PAGs.} 
Let $\mc{P}$ be a PAG, and let $\mc{Q}$ be obtained from $\mc{P}$ through a sequence of valid local transformations, applying the orientation rules $\mc{R}_1$--$\mc{R}_3$, $\mc{R}_4'$, $\mc{R}_8$--$\mc{R}_{10}$, and $\mc{R}_{\texttt{SB}}$ to closure after each transformation. For any $\mf{A} \subseteq \mf{V}$, let $\mc{Q}[\mf{A}]$ be the induced graph on $\mf{A}$. 

\begin{lemma}\label{lem: PMG balanced}
  For any distinct vertices $A,B,C \in \mf{A}$, if $A \adedge B \caedge C$ in $\mc{Q}[\mf{A}]$, then $A \adedge C$ is also present. Furthermore, if $A \to B$, then the edge between $A$ and $C$ is either $A \to C$ or $A \cdedge C$.
\end{lemma}
\begin{proof}[Proof of \Cref{lem: PMG balanced}]
  By \Cref{thm: wang2023 thm 1}, $\mc{Q}$ satisfies the balanced property. Since $A,B,C \in \mf{A}$, every edge among these vertices, together with its endpoint marks, is preserved in the induced graph $\mc{Q}[\mf{A}]$. Both conclusions therefore hold in $\mc{Q}[\mf{A}]$.
\end{proof}

\begin{lemma}\label{lem: PMG circle balanced}
    Suppose that distinct vertices $A,B \in \mf{A}$ are connected by a circle path in $\mc{Q}[\mf{A}]$. Then the following properties hold:
    \begin{itemize}
        \item [1.] If $A$ and $B$ are adjacent, their edge is $A \cedge B$.
        
        \item [2.] For every $C \in \mf{A} \setminus \{A,B\}$, $C \adedge A$ if and only if $C \adedge B$. Furthermore, $C \bedge A$ if and only if $C \bedge B$.
    \end{itemize}
\end{lemma}
\begin{proof}[Proof of \Cref{lem: PMG circle balanced}]
    Let $p: V_0 (= A) \cedge V_1 \cedge \cdots \cedge V_k (=B)$
    be a circle path in $\mc{Q}[\mf{A}]$.
    
    For the first property, the case $k=1$ is immediate. Suppose that $k \geq 2$ and the edge between $A$ and $B$ has an arrowhead at $B$. Applying \Cref{lem: PMG balanced} to $A \adedge V_k \cedge V_{k-1}$ gives $A \adedge V_{k-1}$. Repeating this argument along $p$ yields $A \adedge V_1$, contradicting $A \cedge V_1$. Reversing the roles of $A$ and $B$ also excludes an arrowhead at $A$. Since selection bias is absent and $\mc{Q}$ is closed under $\mc{R}_{\texttt{SB}}$, the edge must therefore be $A \cedge B$.
    
    For the second property, if $C$ belongs to the same
    circle component as $A$ and $B$, the first property
    excludes all four relations under consideration.
    Otherwise, repeated application of \Cref{lem: PMG balanced} along $p$ shows that $C \adedge A$ implies $C \adedge B$.
    Traversing $p$ in reverse gives the converse.
    
    It remains to establish the equivalence for bidirected edges.
    Consider consecutive vertices $U,V$ on $p$ and suppose that $C \bedge U$.
    The balanced property gives $C \adedge V$.
    If this edge were $C \to V$, applying the stronger
    part of \Cref{lem: PMG balanced} to
    $C \to V \cedge U$ would exclude $C \bedge U$.
    If it were $C \cdedge V$, applying the balanced
    property to $U \bedge C \cdedge V$ would give
    $U \adedge V$, contradicting $U \cedge V$.
    Thus, $C \bedge V$.
    Iterating along $p$, and then applying the same
    argument in reverse, proves the desired equivalence.
  \end{proof}

\begin{lemma}\label{lem: PMG ancestral}
   Let $\mc{G}$ be any causal diagram consistent with $\mc{Q}$, and let $X,Y \in \mf{A}$. If $X \in \An(Y)_{\mc{G}[\mf{A}]}$, then $X \in \Poss\An(Y)_{\mc{Q}[\mf{A}]}$.
\end{lemma}
\begin{proof}[Proof of \Cref{lem: PMG ancestral}]
    The case $X=Y$ is immediate. Otherwise, let $p$ be a directed path from $X$ to $Y$ in $\mc{G}[\mf{A}]$, and let $\mc{M}$ be the MAG corresponding to $\mc{G}$. Each directed edge on $p$ is also present with the same orientation in $\mc{M}$. Since $\mc{M} \in [\mc{Q}]$, no endpoint mark in $\mc{Q}$ can contradict these orientations. Hence, the same vertex sequence forms a possibly directed path from $X$ to $Y$ in $\mc{Q}$. All its vertices belong to $\mf{A}$, so this path is also present in $\mc{Q}[\mf{A}]$. Therefore, $X \in \Poss\An(Y)_{\mc{Q}[\mf{A}]}$. 
\end{proof}

\begin{lemma}\label{lem: PMG visible}
   For every directed edge $X \to Y$ in $\mc{Q}[\mf{A}]$, if $X \to Y$ is visible in $\mc{Q}$, then it remains visible in $\mc{Q}[\mf{A}]$.
\end{lemma}
\begin{proof}[Proof of \Cref{lem: PMG visible}]
    Let $\mc{G}$ be any DAG with explicit latent vertices, whose corresponding MAG belongs to $[\mc{Q}]$, and let $\mf{U}$ be its latent vertex set. Since $X \to Y$ is visible in $\mc{Q}$, $\mc{G}$ contains no inducing path between $X$ and $Y$, relative to $\mf{U}$, that is into $X$.

    Suppose that such a path existed in $\mc{G}[\mf{A} \cup \mf{U}]$. Its edges would also be present in $\mc{G}$, as would the directed paths from each collider to at least one endpoint. It would therefore be an inducing path in $\mc{G}$ relative to the same latent set $\mf{U}$, contradicting visibility in $\mc{Q}$. As $\mc{G}$ was arbitrary, $X \to Y$ is visible in $\mc{Q}[\mf{A}]$. 
\end{proof}

\begin{lemma}\label{lem: PMG pc-component}
   Let $\mc{G}$ be any causal diagram consistent with $\mc{Q}$, and let $X,Y \in \mf{A}$. If $X \in \cc(Y)_{\mc{G}[\mf{A}]}$, then $X \in \pc(Y)_{\mc{Q}[\mf{A}]}$. 
\end{lemma}
 \begin{proof}[Proof of \Cref{lem: PMG pc-component}]
      The case $X=Y$ is immediate. Otherwise, let $p$ be a bidirected path between $X$ and $Y$ in $\mc{G}[\mf{A}]$, and let $\mc{M} \in [\mc{Q}]$ be the MAG corresponding to $\mc{G}$. By \Cref{lem: lem 6 in jaber2018}, some subsequence of the vertices of $p$ forms a collider path $p'$ in $\mc{M}$ whose directed edges, if any, are invisible in $\mc{M}$.

      Among the paths between $X$ and $Y$ in $\mc{M}$ whose vertex sequences are subsequences of $p'$, choose a shortest path $q$ subject to the requirements that every internal vertex is a collider and every directed edge is invisible in $\mc{M}$. Such a path exists because $p'$ satisfies these requirements.
      The proof of \Cref{lem: lem 7 in jaber2018} in \citet{jaber2018causal} shows that the internal vertices of this chosen path are definite colliders in the corresponding PAG $\mc{P}$. Their arrowheads are preserved when $\mc{P}$ is refined to $\mc{Q}$. Moreover, every vertex of $q$ belongs to $\mf{A}$. Thus, the corresponding path in $\mc{Q}[\mf{A}]$ has only definite colliders as internal vertices.

      We next show that no edge on this path is visible in $\mc{Q}[\mf{A}]$. Consider any directed edge $V \to W$ on the
      path in $\mc{Q}[\mf{A}]$. Because $\mc{M} \in [\mc{Q}]$, the corresponding edge in $\mc{M}$ is also $V \to W$, and it is invisible by the choice of $q$. By \Cref{lem: zhang2008a lem 10}, a DAG represented by $\mc{M}$ can be chosen to contain $V \gets U \to W$ for a latent vertex $U$. Since $V,W \in \mf{A}$, this common-parent path survives restriction to $\mf{A}$ and the latent vertices. As $\mc{M} \in [\mc{Q}]$, this DAG belongs to the class used to interpret visibility in $\mc{Q}[\mf{A}]$. Hence, $V \to W$ is invisible in $\mc{Q}[\mf{A}]$. The witnessing DAG may be chosen separately for each directed edge, since visibility is defined edge by edge. 

      Every remaining edge on the path is non-directed and is therefore invisible by definition. Consequently, $q$ is a path in $\mc{Q}[\mf{A}]$ whose internal vertices are colliders and whose edges are all invisible. By \Cref{def: pc}, this proves
      $X \in \pc(Y)_{\mc{Q}[\mf{A}]}$.
\end{proof}

\begin{lemma}\label{lem: PMG pto}
   \textsc{pto} (\Cref{alg: pto}) is sound over $\mc{Q}[\mf{A}]$. 
\end{lemma}
\begin{proof}[Proof of \Cref{lem: PMG pto}]
      Construct an auxiliary directed graph $\mc{T}$ whose vertices are the circle components of $\mc{Q}[\mf{A}]$. For distinct components $\mf{B}$ and $\mf{C}$, include the directed edge $\mf{B} \to \mf{C}$ if $\mc{Q}[\mf{A}]$ contains $U \to V$ or
      $U \cdedge V$ for some $U \in \mf{B}$ and $V \in \mf{C}$.

      We first show that $\mc{T}$ is acyclic. Suppose instead that it contains a directed cycle $\mf{B}_1 \to \cdots \to \mf{B}_k \to \mf{B}_1$. For each directed edge on this cycle, choose an edge $U_i \qdedge V_i$ in $\mc{Q}[\mf{A}]$ with $U_i \in \mf{B}_i$ and $V_i \in \mf{B}_{i+1}$, where indices are interpreted cyclically. Since $V_i$ and $U_{i+1}$ lie in the same circle component, the arrowhead and bidirected-edge equivalences in \Cref{lem: PMG circle balanced} imply $U_i \qdedge U_{i+1}$. Indeed, the arrowhead at $V_i$ propagates to $U_{i+1}$, while a bidirected edge to $U_{i+1}$ would imply a bidirected edge to $V_i$, contrary to the choice of $U_i \qdedge V_i$.

      By the chordal property in \Cref{thm: wang2023 thm 1} and \Cref{lem: wang2023 lem 16.1}, there exists a MAG $\mc{M} \in [\mc{Q}]$ obtained by replacing every $\cdedge$ edge with a directed edge and orienting the circle components into DAGs without v-structures. The edges identified above would then form $U_1 \to U_2 \to \cdots \to U_k \to U_1$ in $\mc{M}$, contradicting its ancestral property. Therefore, $\mc{T}$ is acyclic. 

      Every nonempty induced subgraph of $\mc{T}$ consequently has a vertex with no outgoing directed edge. Such a vertex corresponds to a remaining circle component $\mf{B}$ for which every edge to another remaining component has an arrowhead at its endpoint in $\mf{B}$. Thus, $\mf{B}$ satisfies the condition for removal in \textsc{pto}. Removing a whole circle component leaves the other circle components unchanged. Repeatedly choosing and removing a component with this property therefore removes every component and terminates. 
      
      Finally, let $\mc{G}$ be any causal diagram consistent with $\mc{Q}$. Every directed edge $U \to V$ in $\mc{G}[\mf{A}]$ is also directed from $U$ to $V$ in its corresponding MAG. If $U$ and $V$ belong to different circle components of $\mc{Q}[\mf{A}]$, their edge in $\mc{Q}$ must be $U \to V$ or $U \cdedge V$. There is therefore a directed edge in $\mc{T}$ from the component of $U$ to the component of $V$. The component of $V$ must be removed before that of $U$, because the latter still has an outgoing directed edge while both remain. Reversing the removal order thus places the component of $U$ before that of $V$, as required by every directed edge between components in $\mc{G}[\mf{A}]$. The resulting partial topological order is therefore valid for $\mc{G}[\mf{A}]$. 
\end{proof}
Building on these properties, we can treat PMGs obtained through local transformations in much the same way as PAGs.
\paragraph{Orienting according to local transformations.}

Fix an ordering of the vertices in $\mf{X} \cup \{Y\}$. Starting from $\mc{P}$, process these vertices in this order, considering every valid local transformation at each step. After each transformation, apply the orientation rules $\mc{R}_1$--$\mc{R}_3$, $\mc{R}_4'$, $\mc{R}_8$--$\mc{R}_{10}$, and $\mc{R}_{\texttt{SB}}$ until no further orientation is possible. Let $\mc{Q}_{\mf{X}}^i$ denote the resulting PMGs, where $i$ indexes the sequences of local transformation choices. The constructed PMGs $\mc{Q}_\mf{X}^i$ inherit these graphical properties. 
\begin{corollary}[Inherited properties]\label{cor: constructed PMGs inherit properties}
\Cref{lem: PMG balanced,lem: PMG circle balanced,lem: PMG ancestral,lem: PMG visible,lem: PMG pc-component,lem: PMG pto} apply with $\mc{Q}=\mc{Q}_{\mf{X}}^i$. 
\end{corollary}
\begin{proof}[Proof of \Cref{cor: constructed PMGs inherit properties}]
    By \Cref{thm: wang2023 thm 1}, the resulting PMG is orientation-complete. It also satisfies the common hypotheses of the lemmas. Their conclusions therefore hold for $\mc{Q}_{\mf{X}}^i[\mf{A}]$. 
\end{proof}
\begin{proposition}\label{prop: complete PMG} 
    Let $\mc{M} \in [\mc{P}]$, and suppose that $\mf{X}$ is an MIS relative to $\langle \mc{M},Y \rangle$. Then there exists a PMG $\mc{Q}_\mf{X}^i$ representing $\mc{M}$ such that:
    \begin{itemize}[leftmargin=10mm]
        \item[1.] Every circle mark at a vertex in $\mf{X} \cup \{Y\}$ in $\mc{P}$ is replaced by a tail or an arrowhead in $\mc{Q}_\mf{X}^i$.
        
        \item[2.] $\mc{Q}_\mf{X}^i$ is closed under $\mc{R}_1$--$\mc{R}_3$, $\mc{R}_4'$, $\mc{R}_8$--$\mc{R}_{10}$, and $\mc{R}_{\texttt{SB}}$.
        
        \item[3.] For every $X \in \mf{X}$, $\mc{Q}_{\mf{X}}^i$ contains a proper directed path from $X$ to $Y$ with respect to $\mf{X}$.

    \end{itemize}    
\end{proposition}
 \begin{proof}[Proof of \Cref{prop: complete PMG}]
    We first construct a sequence of local transformations that preserves $\mc{M}$. Initially, $\mc{M} \in [\mc{P}]$. At each step, let $\mc{Q}$ be the current PMG and let $A \in \mf{X} \cup \{Y\}$ be the vertex being processed. Choose the local transformation at $A$ so that every circle mark at $A$ is oriented as in $\mc{M}$. This choice is valid because $\mc{M} \in [\mc{Q}]$ provides a compatible MAG realizing all the selected marks simultaneously. The subsequent orientation rules are sound and
    therefore preserve compatibility with $\mc{M}$. Inductively, the resulting PMG $\mc{Q}_{\mf{X}}^i$ satisfies $\mc{M} \in [\mc{Q}_{\mf{X}}^i]$. Every vertex in $\mf{X} \cup \{Y\}$ is processed, and orientation closure is applied after each step. Hence, the first two conditions hold.
    
    To prove the third condition, fix $X \in \mf{X}$. By \Cref{thm: MIS for MAG iff An}, choose a shortest proper directed path $p: V_0 (=X) \to V_1 \to \cdots \to V_k (= Y)$ with respect to $\mf{X}$ in $\mc{M}$. This path is uncovered; otherwise, a chord between nonconsecutive vertices must point forward, since the earlier vertex is an ancestor of the later one and $\mc{M}$ is ancestral. Such a chord would give a shorter proper directed path, a contradiction. 
    
    Since $\mc{M}$ and $\mc{Q}_{\mf{X}}^i$ have the same adjacencies, the corresponding path is also uncovered in $\mc{Q}_{\mf{X}}^i$. The mark at $X$ on the edge between $X$ and $V_1$ is fixed in $\mc{Q}_{\mf{X}}^i$ and agrees with $\mc{M}$. It is therefore a tail. Thus, closure under $\mc{R}_{\texttt{SB}}$ gives $X \to V_1$ in $\mc{Q}_{\mf{X}}^i$. 
    
    Now suppose that $V_{j-1} \to V_j$ has been established for some $1 \leq j < k$. Because $\mc{M} \in [\mc{Q}_{\mf{X}}^i]$ and $V_j \to V_{j+1}$ in $\mc{M}$, the mark at $V_j$ on the edge between $V_j$ and $V_{j+1}$ cannot be an arrowhead. If it is a circle, the uncovered configuration $V_{j-1} \to V_j \caedge V_{j+1}$ is oriented as $V_j \to V_{j+1}$ by $\mc{R}_1$, since $V_{j-1}$ and $V_{j+1}$ are not adjacent. If the mark is already a tail, the edge is $V_j \to V_{j+1}$, using
    $\mc{R}_{\texttt{SB}}$ if necessary. Thus, closure implies $V_j \to V_{j+1}$
    in either case.

    Thus, the path is a proper directed path from $X$ to $Y$ with respect to $\mf{X}$ in $\mc{Q}_{\mf{X}}^i$. The same PMG $\mc{Q}_{\mf{X}}^i$ was constructed by preserving $\mc{M}$ throughout all local transformations. Hence, this conclusion holds simultaneously for every $X \in \mf{X}$, establishing the third condition.
\end{proof}

\begin{algorithm}[t]
\SetAlgoNoEnd
\DontPrintSemicolon

\textbf{function~}\textsf{IsPOMIS}($\mc{P}$, $Y$, $\mf{X}$)\;

\Indp{
    \tcp{Check whether the given set $\mf{X}$ is an MIS.} 

    \lIf{\rm $\mf{X}$ does not satisfy \Cref{prop: MIS for PAG} }{\Return{\rm \textsf{False}}}

    \Return{\rm \textsf{subIsPOMIS}$(\mc{P}, \mf{X} \cup \{Y\}, Y, \mf{X})$}\;

}

\Indm

\BlankLine 

\textbf{function~}{\textsf{subIsPOMIS}$(\mc{Q}, \mf{A}, Y, \mf{X})$}\;

\Indp{
\tcp{Apply the terminal IB test justified by \Cref{thm: isPOMIS sound and complete}.}        
    \lIf {\rm $\mf{A}$ is empty}{\Return{\rm $\IB(\mc{Q}, Y, \mf{X}) = \mf{X}$}}
    {
    
     Pick the first vertex $A\in\mf{A}$ in a fixed order.\;

    \For{\rm each local transformation $\mf{C}^{\mc{Q}}_A \subseteq \{V \in \Adj(A)_{\mc{Q}} \mid A \caedge V\}$}{
        
        \tcp{Check whether $\mf{C}^{\mc{Q}}_A$ is valid.}        
        \If {\rm $\mf{C}^{\mc{Q}}_A$ satisfies \Cref{thm: wang2023 thm 3}}{

          \tcp{Ensure a proper directed path after closure.}

            \If{$A \in \Poss\An(Y)_{\mc{Q}\setminus
                    \left(
                              \mf{C}_A^{\mc{Q}}
                              \cup
                              (\mf{X}\setminus\{A\})
                          \right)
                      }
$}{
                    Let $\mc{Q}'$ be the PMG obtained by orienting the circle marks around $A$ according to $\mf{C}^{\mc{Q}}_A$ and then applying the orientation rules $\mc{R}_1$--$\mc{R}_3$, $\mc{R}_4'$, $\mc{R}_8$--$\mc{R}_{10}$, and $\mc{R}_{\texttt{SB}}$ until none applies.\;
                            
                    \If{\rm \textsf{subIsPOMIS}$(\mc{Q}',\mf{A}\setminus \{A\}, Y, \mf{X})$}{\Return{\rm \textsf{True}}}
                }
            }
        }
    \Return{\rm \textsf{False}}    
    }
}
\caption{\small Recursive search over valid local transformations.}
\label{alg: ispomis}
\end{algorithm}


\paragraph{Algorithmic characterization.}
Building on the theoretical results above, we present \textsf{IsPOMIS}~(\Cref{alg: ispomis}), which searches locally oriented PMGs and applies a terminal IB test. The following theorem establishes that the algorithm decides whether $\mf{X}$ is a POMIS in at least one MAG represented by $\mc{P}$. 

\begin{theorem}
    \label{thm: isPOMIS sound and complete}
    \textsf{IsPOMIS} (\Cref{alg: ispomis}) returns {\rm \textsf{True}} \textit{if and only if} $\mf{X} \in \widehat{\mb{P}}_{\mc{P},Y}$. 
\end{theorem}

The proof in \Cref{sec: proof of thm: isPOMIS sound and complete} combines \Cref{thm: IB = POMIS for MAG} with the terminal PMG--MAG correspondence in \Cref{lem: terminal pmg-mag correspondence}. The latter constructs one compatible MAG with the required IB. The algorithm can therefore stop while circle marks remain at vertices outside $\mf{X} \cup \{Y\}$. 

The algorithm first checks whether $\mf{X}$ is an MIS using \Cref{prop: MIS for PAG}. If this necessary condition fails, it returns \textsf{False}. Otherwise, it calls \textsf{subIsPOMIS} directly on $\mc{P}$, with $\mf{A}=\mf{X}\cup\{Y\}$. When enumerating the full union, the candidates for these individual calls are the sets returned by \textsf{MISs} in \Cref{alg: miss}. 

At each recursive call, the algorithm selects the first vertex $A\in\mf{A}$ in a fixed order and considers local transformations $\mf{C}_A^{\mc{Q}}$ satisfying \Cref{thm: wang2023 thm 3}. 

For each valid transformation, it additionally checks:
\begin{align}
    A \in \Poss\An(Y)_{\mc{Q}\setminus
    \left(\mf{C}_A^{\mc{Q}} \cup (\mf{X}\setminus\{A\})
    \right)}.
\end{align}
Here, we set $\Poss\An(Y)_{\mc{H}}=\emptyset$ whenever $Y$ is not a vertex of $\mc{H}$. When $A=Y$, the condition holds automatically because possible ancestors include the vertex itself and $Y\notin\mf{C}_Y^{\mc{Q}}\cup\mf{X}$. When $A\in\mf{X}$, the condition checks for a possibly directed path to $Y$ that avoids both $\mf{C}_A^{\mc{Q}}$ and the other intervention variables.

If the condition holds, the algorithm applies the selected local transformation and then the orientation rules $\mc{R}_1$--$\mc{R}_3$, $\mc{R}_4'$, $\mc{R}_8$--$\mc{R}_{10}$, and $\mc{R}_{\texttt{SB}}$ until none applies, obtaining $\mc{Q}'$. The following proposition shows that, for $A \in \mf{X}$, this update establishes a proper directed path from $A$ to $Y$.
\begin{proposition} \label{prop: local proper path} 
    Let $\mc{Q}$ be obtained from a PAG by valid local transformations and orientation closure, let $\mf{X} \subseteq \mf{V} \setminus \{Y\}$ and let $A \in \mf{X} \cup \{Y\}$. Consider a MAG $\mc{M} \in [\mc{Q}]$ consistent with the valid local transformation $\mf{C}_A^{\mc{Q}}$. Then the following equivalence holds:
    \begin{align}
      A \in \Poss\An(Y)_{\mc{Q}\setminus
          \left(
              \mf{C}_A^{\mc{Q}}
              \cup
              (\mf{X}\setminus\{A\})
          \right)
      } \iff
      A \in \An(Y)_{ \mc{M}\setminus(\mf{X}\setminus\{A\})}.
    \end{align}
    If either graph does not contain $Y$, the corresponding ancestor set of $Y$ is interpreted as empty. 
\end{proposition}
\begin{proof}[Proof of \Cref{prop: local proper path}]
The case $A = Y$ is immediate because ancestors include the vertex itself. Hence assume $A \in \mf{X}$. 

\textbf{(Only if)} Suppose that the left-hand condition holds. There is a possibly directed path from $A$ to $Y$ in
$\mc{Q}$ avoiding $\mf{C}_A^{\mc{Q}}\cup(\mf{X}\setminus\{A\})$. By \Cref{lem: wang2023 lem 2}, the path has an uncovered possibly directed subsequence with the same endpoints. Since taking a subsequence introduces no vertices, the resulting path still avoids the deleted vertices. 

Its first neighbor of $A$ lies outside $\mf{C}_A^{\mc{Q}}$. If the first edge already has a tail at $A$, it points out of $A$ by closure under $\mc{R}_{\texttt{SB}}$. Otherwise, its mark at $A$ is a circle, which the selected local transformation replaces by a tail; $\mc{R}_{\texttt{SB}}$ then completes the directed edge. The rules $\mc{R}_1$ and $\mc{R}_{\texttt{SB}}$ then propagate this orientation along the uncovered path. Consequently, the resulting PMG $\mc{Q}'$ contains a proper directed path from $A$ to $Y$ with respect to $\mf{X}$. The fixed MAG $\mc{M}$ agrees with the selected local transformations, so sound orientation closure preserves $\mc{M} \in [\mc{Q}']$. The directed path in $\mc{Q}'$ is therefore also present in this $\mc{M}$, establishing the right-hand condition. 

\textbf{(If)} Conversely, suppose that the right-hand condition holds. Then $\mc{M}$ contains a directed path from $A$ to $Y$ avoiding $\mf{X}\setminus\{A\}$. Since $\mc{M}$ is consistent with the selected local transformation, the edge between $A$ and every $C\in\mf{C}_A^{\mc{Q}}$ has an arrowhead at $A$. By ancestrality, no such $C$ can be a descendant of $A$. Hence, the directed path also avoids $\mf{C}_A^{\mc{Q}}$. Its corresponding path in $\mc{Q}$ is possibly directed, establishing the left-hand condition. 
\end{proof}

The recursion continues with $\mf{A}\setminus\{A\}$. Once established, the proper directed paths are preserved by all subsequent valid transformations and orientation updates. Thus, when $\mf{A} = \emptyset$, every intervention variable has a proper directed path to $Y$ in the resulting orientation-complete PMG $\mc{Q}_{\mf{X}}^i$. The algorithm then checks $\IB(\mc{Q}_{\mf{X}}^i,Y,\mf{X})=\mf{X}$. It returns \textsf{True} if any recursive branch satisfies this condition and \textsf{False} otherwise. 

\begin{corollary}\label{cor: dmis search sound and complete}
    Define \textsf{IsDMIS} by running \Cref{alg: ispomis} with the terminal return value $\IB(\mc{Q},Y,\mf{X})=\mf{X}$ replaced by \textsf{True}, while retaining all other tests and recursive steps. For any $\mf{X} \subseteq \mf{V} \setminus \{Y\}$, this procedure terminates and returns \textsf{True} if and only if $\mf{X} \in \mb{D}_{\mc{P},Y}$.
\end{corollary}
\begin{proof}[Proof of \Cref{cor: dmis search sound and complete}]
    The search has finite branching, and each recursive descent removes one vertex from $\mf{A}$, so it terminates.
    
    If it accepts, the valid transformations and sound orientation closure yield a terminal PMG $\mc{Q}$ with $\emptyset \neq [\mc{Q}] \subseteq[\mc{P}]$. By \Cref{prop: local proper path}, every $X \in \mf{X}$ has a proper directed path to $Y$ with respect to $\mf{X}$ in $\mc{Q}$, preserved by subsequent orientations. Any $\mc{M} \in [\mc{Q}]$ therefore makes $\mf{X}$ an MIS by \Cref{thm: MIS for MAG iff An}, and \Cref{prop: DMIS for pag eq MIS for MAG} gives DMIS membership. 

    Conversely, a DMIS has a witnessing MAG $\mc{M} \in [\mc{P}]$ in which it is an MIS, and it passes the initial MIS screen. Follow the local choices agreeing with this same $\mc{M}$, as in \Cref{prop: complete PMG}. These choices are valid and preserve $\mc{M}$; \Cref{prop: local proper path} ensures that every proper path test passes. The test at $Y$ holds by reflexivity. Thus an accepting terminal branch is reached. The argument includes $\mf{X}=\emptyset$, for which the proper path requirements are vacuous. 
\end{proof}

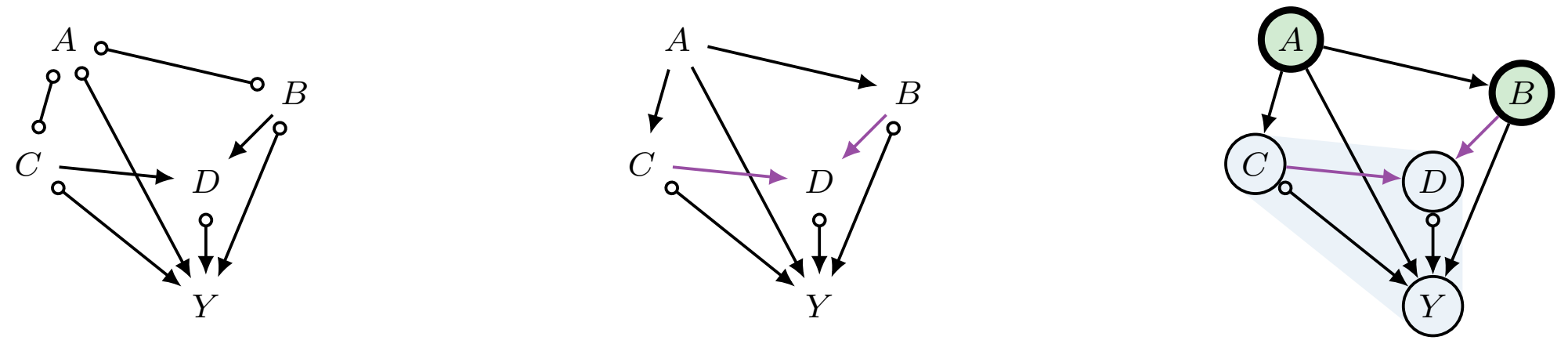
\begin{figure}[t]
    \centering
    \tikzset{
        scbpmggraph/.style={
            x=1.5cm,y=1.5cm,font=\scriptsize,
            >={Latex[width=1.4mm,length=1.6mm]},
            line width=0.25mm,draw=black,text=black
        },
        scbpmgcircle/.tip={Circle[open,length=3.5pt,width=3.5pt]},
        scbpmgnode/.style={
            draw=none,circle,fill=white,inner sep=0mm,
            minimum size=5mm,text=black
        },
        scbpmgterritory/.style={scbpmgnode,draw=black,fill=none},
        scbpmgborder/.style={
            scbpmgnode,draw=black,fill=bettergreen!25,
            line width=0.6mm
        },
        scbpmgarea/.style={
            draw=betterblue!10,fill=betterblue!10,
            line width=5mm,line cap=round,line join=round
        }
    }

    \begin{subfigure}[b]{0.32\textwidth}
        \centering
        \begin{tikzpicture}[scbpmggraph]
            \path[use as bounding box]
                (-0.32,-0.30) rectangle (1.78,1.72);
            \node[scbpmgnode] (A) at (0.2,1.5) {$A$};
            \node[scbpmgnode] (B) at (1.5,1.2) {$B$};
            \node[scbpmgnode] (C) at (0,0.8) {$C$};
            \node[scbpmgnode] (D) at (1,0.7) {$D$};
            \node[scbpmgnode] (Y) at (1,0) {$Y$};

            \draw[{scbpmgcircle}-{scbpmgcircle}] (A) -- (B);
            \draw[{scbpmgcircle}-{scbpmgcircle}] (A) -- (C);
            \draw[{scbpmgcircle}->] (A) -- (Y);
            \draw[->] (B) -- (D);
            \draw[->] (C) -- (D);
            \draw[{scbpmgcircle}->] (B) -- (Y);
            \draw[{scbpmgcircle}->] (C) -- (Y);
            \draw[{scbpmgcircle}->] (D) -- (Y);
        \end{tikzpicture}
        \caption{$\mathcal{P}$}
        \label{fig: ispomis_example_a}
    \end{subfigure}\hfill
    \begin{subfigure}[b]{0.32\textwidth}
        \centering
        \begin{tikzpicture}[scbpmggraph]
            \path[use as bounding box]
                (-0.32,-0.30) rectangle (1.78,1.72);
            \node[scbpmgnode] (A) at (0.2,1.5) {$A$};
            \node[scbpmgnode] (B) at (1.5,1.2) {$B$};
            \node[scbpmgnode] (C) at (0,0.8) {$C$};
            \node[scbpmgnode] (D) at (1,0.7) {$D$};
            \node[scbpmgnode] (Y) at (1,0) {$Y$};

            \draw[->] (A) -- (B);
            \draw[->] (A) -- (C);
            \draw[->] (A) -- (Y);
            \draw[->,draw=betterpurple,line width=0.25mm] (B) -- (D);
            \draw[<-,draw=betterpurple,line width=0.25mm] (D) -- (C);
            \draw[{scbpmgcircle}->] (B) -- (Y);
            \draw[{scbpmgcircle}->] (C) -- (Y);
            \draw[{scbpmgcircle}->] (D) -- (Y);
        \end{tikzpicture}
        \caption{After $A$}
        \label{fig: ispomis_example_b}
    \end{subfigure}\hfill
    \begin{subfigure}[b]{0.32\textwidth}
        \centering
        \begin{tikzpicture}[scbpmggraph]
            \path[use as bounding box]
                (-0.32,-0.30) rectangle (1.78,1.72);
            \coordinate (A) at (0.2,1.5);
            \coordinate (B) at (1.5,1.2);
            \coordinate (C) at (0,0.8);
            \coordinate (D) at (1,0.7);
            \coordinate (Y) at (1,0);

            \draw[scbpmgarea] (C) -- (D) -- (Y) -- cycle;
            \foreach \v in {A,B}
                \node[scbpmgborder] (\v) at (\v) {$\v$};
            \foreach \v in {C,D,Y}
                \node[scbpmgterritory] (\v) at (\v) {$\v$};

            \draw[->] (A) -- (B);
            \draw[->] (A) -- (C);
            \draw[->] (A) -- (Y);
            \draw[->,draw=betterpurple,line width=0.25mm] (B) -- (D);
            \draw[<-,draw=betterpurple,line width=0.25mm] (D) -- (C);
            \draw[->] (B) -- (Y);
            \draw[{scbpmgcircle}->] (C) -- (Y);
            \draw[{scbpmgcircle}->] (D) -- (Y);
        \end{tikzpicture}
        \caption{Terminal PMG $\mathcal{Q}$}
        \label{fig: ispomis_example_c}
    \end{subfigure}

    \caption{\small An \textsf{IsPOMIS} example for $\mathbf{X}=\{A,B\}$. (a) The input PAG $\mathcal{P}$. (b) The PMG after the valid choice $\mathbf{C}^{\mathcal{P}}_A = \emptyset$. (c) The terminal PMG after choosing $\mathbf{C}_B = \emptyset$ and then $\mathbf{C}_Y = \emptyset$, with two circle marks remaining. The light blue region marks the MUCT $\{C,D,Y\}$, green fill marks the IB $\{A,B\}$, and thick outlines mark the intervention targets. Purple directed edges are visible in their full PMG.}
    \label{fig: ispomis_example}
\end{figure}
To illustrate the POMIS search, we return to the MAG in \Cref{fig:scbmec_mag_construction_a} and consider its PAG in \Cref{fig: ispomis_example_a}. The candidate is $\mf{X} = \{A,B\}$, and the processing order is $\langle A,B,Y \rangle$. Both $A \cdedge Y$ and $B \cdedge Y$ are proper possibly directed paths, so this candidate belongs to the collection returned by \textsf{MISs} and passes the initial MIS check. The subsequent search determines whether this candidate is a POMIS in at least one represented MAG.

At $A$, choose $\mf{C}_A^{\mc{P}} = \emptyset$. This is valid, and the additional path test deletes only $B$, leaving the path $A \cdedge Y$. The transformation orients $A \to B$, $A \to C$, and $A \to Y$. Orientation closure gives the PMG $\mc{Q}_A$ in \Cref{fig: ispomis_example_b}. To see the role of the additional test, compare the valid choice $\mf{C}_A^{\mc{P}} = \{Y\}$. This choice instead places an arrowhead at $A$ on its edge to $Y$, but the path test deletes $Y$ and therefore rejects the branch. Validity alone does not ensure a proper path to $Y$.  

At $B$, the only remaining circle mark is on $B \cdedge Y$. Choose $\mf{C}_B^{\mc{Q}_A} = \emptyset$. The path test deletes $A$ and retains $B \cdedge Y$, so the branch continues and fixes $B \to Y$. 

At $Y$, all incident marks are already arrowheads. Its only local choice is empty, and the reflexive possible ancestor condition holds after deleting $\{A,B\}$. No further orientations are needed. The resulting PMG $\mc{Q}$ is shown in \Cref{fig: ispomis_example_c}. After deleting $\mf{X}$, the possible ancestors of $Y$ are $\{C,D,Y\}$. The invisible edges $C \cdedge Y$ and $D \cdedge Y$ place both $C$ and $D$ in the territory reached from $Y$. Hence:
\begin{align}
    \text{For $\mf{X} = \{A,B\}$,~} &\MUCT(\mc{Q},Y, \mf{X}) = \{C,D,Y\} \\
    &\Rightarrow \IB(\mc{Q},Y, \mf{X}) = \{A,B\} = \mf{X}.
\end{align}
The algorithm therefore returns \textsf{True}. Two circle marks remain, one at $C$ and one at $D$. Thus, the terminal graph is still a PMG. Replacing these two circles by tails gives the MAG in \Cref{fig:scbmec_mag_construction_a}. Its construction example in \Cref{fig:scbmec_mag_construction_c} realizes the same territory and border in a compatible causal diagram. Thus, the local search and the earlier MAG construction establish the same candidate's POMIS membership from their respective graph representations.

\paragraph{Runtime analysis.} Let $n=\lvert\mf{V}\rvert$, let $m$ be the number of edges in $\mc{P}$, and let $k=\lvert\mf{X}\rvert+1$. The initial MIS check can be implemented in $\mc{O}(n + m)$ time using a reverse reachability search from $Y$ along possibly directed adjacencies. Vertices in $\mf{X}$ are marked when reached but are not expanded, ensuring that the corresponding paths avoid other intervention variables. 

Let $p\leq n-1$ bound the number of circle marks at any vertex being processed. Each recursive call considers at most $2^p$ local transformations. Checking the validity of one transformation takes $\mc{O}(n^3)$ time \citep{wang2023sound}, while the condition in \Cref{prop: local proper path} can be evaluated by a reachability search in $\mc{O}(n+m)$ time. 

The recursion has depth $k$ and examines at most $k \cdot 2^{kp}$ candidate transformations and $2^{kp}$ terminal states.
Let $\mc{T}_{\mathtt{closure}}$ denote the worst-case cost of one orientation closure and $\mc{T}_{\mathtt{IB}}$ the worst-case cost of evaluating the terminal IB condition. A conservative upper bound for one call to \textsf{IsPOMIS} is therefore: 
  \begin{align}
      \mc{O}
      \left(n + m + 
      2^{kp}\left[
              k\bigl(n^3+\mc{T}_{\mathtt{closure}}\bigr)
              +\mc{T}_{\mathtt{IB}}
          \right]
      \right). \label{eq: cost of ispomis}
\end{align}
To enumerate all members of $\widehat{\mb{P}}_{\mc{P},Y}$, first obtain $\mb{M}_{\mc{P},Y}$ using \textsf{MISs}, and then retain the sets accepted by \textsf{IsPOMIS}. Every POMIS in a represented MAG is an MIS for the PAG, so this restriction preserves every member of the union. There are $\kappa$ candidates. The total cost is the MIS enumeration cost in \Cref{eq: cost of miss}, plus the sum of the individual \textsf{IsPOMIS} costs in \Cref{eq: cost of ispomis} over these $\kappa$ sets. 

An alternative is to enumerate the MAGs represented by $\mc{P}$ and evaluate their IB conditions. For example, \textsc{MagList} \citep{wangefficient} enumerates these MAGs without duplication by recursively determining local structures throughout $\mf{V}$. For a given candidate $\mf{X}$, \textsf{IsPOMIS} branches only on local transformations at $\mf{X}\cup\{Y\}$, while applying orientation closure to the entire graph. A terminal PMG may therefore represent multiple MAGs. Its IB test is justified by \Cref{thm: isPOMIS sound and complete}. 

For the PMGs visited by the algorithm, \Cref{cor: graphical characterization of visibility in a PMG} shows that visibility is exactly characterized by the graphical criterion in \Cref{lem: visible}, evaluated in the full PMG. For each directed edge $R \to B$, this criterion can be checked by finding the bidirected connections among parents of $B$ and inspecting the edges entering those vertices. Starting from $R$, retain only parents of $B$ along these bidirected connections, and check whether an edge entering a reached vertex starts at a vertex nonadjacent to $B$. Each search takes $\mc{O}(n + m)$ time, so all directed edges can be checked in $\mc{O}(m(n + m))$ time. With their visibility fixed, the territory is reached from $Y$ by traversing invisible edges in either direction and visible edges only from tail to arrowhead, within the possible-ancestor graph. Ancestor restriction, this traversal, and the border calculation each take $\mc{O}(n + m)$ time. Thus $\mc{T}_{\mathtt{IB}} = \mc{O}((m + 1)(n + m))$ is a valid bound.

\section{Experiments}\label{sec: scbmec experiment}
\input{figures/fig_scbmec_experiment_pags_tasks}
\begin{table}[t]
    \centering
    \caption{\small Candidate action spaces. Each strategy column gives the number of intervention sets / arms. }
    \label{tab:scbmec-experiment-design}
    \small
    \setlength{\tabcolsep}{10pt}
    \begin{tabular}{crrrrr}
        \toprule
        \bf Task & $\vert \mf{V} \rvert$ & $T$ & \bf POMIS union & \bf DMIS & \bf BF \\
        \midrule
        1 & 5 & 50k  & 5 / 11 & 7 / 17 & 16 / 81 \\
        2 & 6 & 50k  & 18 / 89 & 30 / 195 & 32 / 243 \\
        3 & 10 & 50k  & 8 / 54 & 120 / 1,593 & 512 / 19,683 \\
        4 & 6 & 50k  & 5 / 11 & 12 / 41 & 32 / 243 \\
        5 & 8 & 50k   & 16 / 81 & 32 / 189 & 128 / 2,187 \\
        6 & 8 & 50k   & 37 / 211 & 70 / 627 & 128 / 2,187 \\
        \bottomrule
    \end{tabular}
\end{table}

We evaluate six settings, each with a fixed binary SCM whose structural functions combine AND ($\land$), OR ($\lor$), exclusive OR ($\oplus$), and negation ($\neg$), with independent Bernoulli exogenous variables. The complete structural functions and exogenous distributions are specified in \Cref{sec: scbmec experimental SCMs}. The agent receives the oracle PAG of the model's causal diagram, and observes the reward of its selected intervention at each round. This PAG is obtained from graphical separation relations in the causal diagram, rather than from conditional independence tests on samples of the fixed SCM. Task~1 uses the PAG already shown in \Cref{fig: 5var a}. \Cref{fig:scbmec-experiment-pags} places the causal diagrams of the SCMs for Tasks~2--6 beside their oracle PAGs. 

\begin{figure}[t]
    \centering
    \captionsetup[subfigure]{skip=2pt}
    {\color{black}\includegraphics[width=\linewidth]{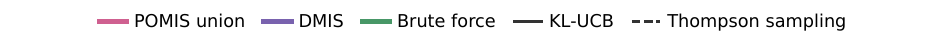}}
    \begin{subfigure}[b]{0.3498\linewidth}
        \centering
        {\color{black}\includegraphics[width=\linewidth]{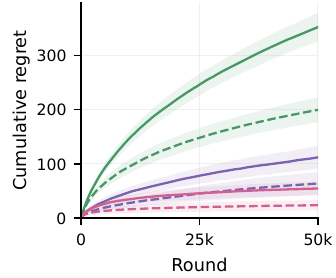}}
        \caption{Task 1}
        \label{fig:scbmec-experiment-task1}
    \end{subfigure}%
    \begin{subfigure}[b]{0.325\linewidth}
        \centering
        {\color{black}\includegraphics[width=\linewidth]{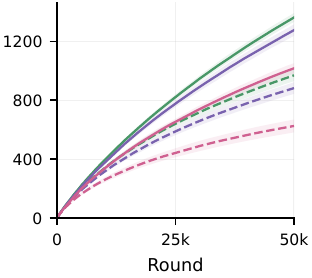}}
        \caption{Task 2}
        \label{fig:scbmec-experiment-task2}
    \end{subfigure}%
    \begin{subfigure}[b]{0.325\linewidth}
        \centering
        {\color{black}\includegraphics[width=\linewidth]{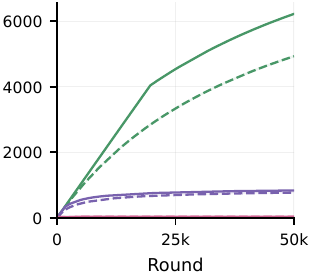}}
        \caption{Task 3}
        \label{fig:scbmec-experiment-task3}
    \end{subfigure}
    \par\vspace{2pt}
    \begin{subfigure}[b]{0.3498\linewidth}
        \centering
        {\color{black}\includegraphics[width=\linewidth]{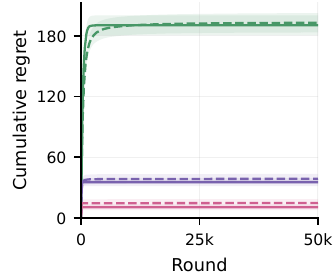}}
        \caption{Task 4}
        \label{fig:scbmec-experiment-task4}
    \end{subfigure}%
    \begin{subfigure}[b]{0.325\linewidth}
        \centering
        {\color{black}\includegraphics[width=\linewidth]{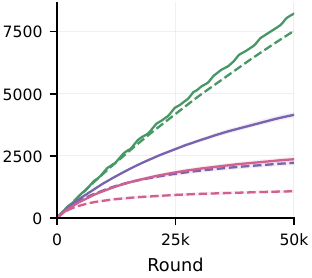}}
        \caption{Task 5}
        \label{fig:scbmec-experiment-task5}
    \end{subfigure}%
    \begin{subfigure}[b]{0.325\linewidth}
        \centering
        {\color{black}\includegraphics[width=\linewidth]{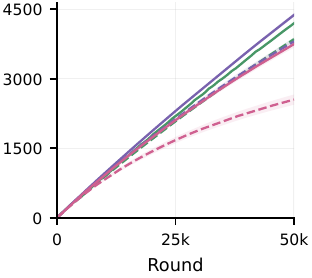}}
        \caption{Task 6}
        \label{fig:scbmec-experiment-task6}
    \end{subfigure}
    \caption{\small Experiments on Tasks 1--6. Curves show mean cumulative regret over 50,000 rounds; shaded regions show one standard deviation across 100 independent runs. Pink, purple, and green denote the POMIS union, DMIS, and brute force, respectively. Solid and dashed lines denote KL-UCB and Thompson sampling.}
    \label{fig:scbmec-experiment-main}
\end{figure}

We compare three candidate action spaces. Brute force (BF) includes every assignment to every subset of $\mf{V} \setminus \{Y\}$, including the empty intervention. With binary variables, its size is $\lvert \mc{A}_{\texttt{BF}} \rvert = \sum_{k = 0}^{\lvert \mf{V} \rvert - 1} \binom{\lvert \mf{V} \rvert - 1}{k} 2^k = 3^{\vert \mf{V} \rvert - 1}$. The \texttt{DMIS} strategy retains all assignments to DMISs. We obtain these sets by applying \textsf{IsDMIS} in \Cref{cor: dmis search sound and complete} to the MIS candidates. The \texttt{POMIS~union} strategy retains all assignments to the intervention sets in the union $\widehat{\mb{P}}_{\mc{P},Y}$ over represented MAGs, as returned by the POMIS membership procedure. Thus, the label \texttt{POMIS~union} in the figures and tables denotes this union, rather than $\mb{P}_{\mc{P},Y}$. \Cref{tab:scbmec-experiment-design} reports the number of intervention sets and arms in each action space. 

\begin{table}[t]
    \centering
    \caption{\small Final cumulative regret at $T = 50{,}000$ (mean $\pm$ standard deviation across 100 runs).}
    \label{tab:scbmec-experiment-regret}
    \small
    \setlength{\tabcolsep}{10pt}
    \begin{tabular}{clrrr}
        \toprule
        \bf Task & \bf Solver & \bf POMIS union & \bf DMIS & \bf BF \\
        \midrule
        \multirow{2}{*}{1}
            & TS     & \regret{\boldsymbol{23.3}}{9.4} & \regret{63.4}{20.9}  & \regret{199.6}{22.6} \\
            & KL-UCB & \regret{\boldsymbol{54.2}}{12.6}             & \regret{111.8}{21.4} & \regret{352.5}{25.9} \\
        \addlinespace[3pt]

        \multirow{2}{*}{2}
            & TS     & \regret{\boldsymbol{625.3}}{43.9} & \regret{882.9}{39.4}  & \regret{970.2}{40.0} \\
            & KL-UCB & \regret{\boldsymbol{1017.8}}{34.7}             & \regret{1277.4}{39.1} & \regret{1363.9}{30.1} \\
        \addlinespace[3pt]

        \multirow{2}{*}{3}
            & TS     & \regret{\boldsymbol{31.0}}{4.5}                    & \regret{770.7}{19.2} & \regret{4929.9}{40.1} \\
            & KL-UCB & \regret{\boldsymbol{30.5}}{4.1}       & \regret{835.2}{21.6} & \regret{6222.6}{26.8} \\
        \addlinespace[3pt]

        \multirow{2}{*}{4}
            & TS     & \regret{\boldsymbol{14.6}}{3.8}                    & \regret{38.6}{4.8} & \regret{192.9}{9.6} \\
            & KL-UCB & \regret{\boldsymbol{10.5}}{2.8}       & \regret{35.3}{4.0} & \regret{190.4}{9.8} \\
        \addlinespace[3pt]

        \multirow{2}{*}{5}
            & TS     & \regret{\boldsymbol{1071.3}}{69.7} & \regret{2216.2}{99.6}  & \regret{7531.8}{45.8} \\
            & KL-UCB & \regret{\boldsymbol{2359.8}}{99.3}              & \regret{4147.6}{114.7} & \regret{8224.0}{25.4} \\
        \addlinespace[3pt]

        \multirow{2}{*}{6}
            & TS     & \regret{\boldsymbol{2548.8}}{110.3} & \regret{3804.5}{68.4} & \regret{3852.4}{41.6} \\
            & KL-UCB & \regret{\boldsymbol{3745.7}}{74.5}               & \regret{4374.8}{42.0} & \regret{4194.8}{30.4} \\
        \bottomrule
    \end{tabular}
\end{table}

Each action space is paired with Thompson sampling (TS) \citep{thompson1933likelihood} and Bernoulli KL-UCB \citep{garivier2011kl,cappe2013kullback}. TS uses a $\mathtt{Beta}(1,1)$ prior for each arm. KL-UCB initially samples arms once in a random order and then uses Bernoulli KL upper confidence indices with exploration function $f(s) = \log s + 3 \log \log s$ for $s \geq 3$ and $f(s) = 1$ otherwise, where $s$ is the number of completed rounds. Thus, after initialization, the decision at round $t$ uses $f(t-1)$. All tasks use $T = 50{,}000$ rounds, so each candidate action space completes this initial exploration. The reward mean of each intervention is computed from the fixed SCM to simulate Bernoulli rewards and evaluate regret. Arm selection uses only the selected arms and observed rewards from previous rounds. 

The performance measure is the cumulative (pseudo-)regret defined in \Cref{chap:scb}: 
\begin{align}
    R_T = \sum_{t=1}^T (\mu^\star - \mu_{A_t}), \qquad \mu^\star = \max_{a \in \mc{A}_{\texttt{BF}}} \mu_{a}, 
\end{align}
using the same full action space benchmark for all strategies. For each of the six fixed SCMs, we verify that all three candidate action spaces retain an arm attaining this optimal mean. Each of the 36 configurations---six tasks, three strategies, and two solvers---is run $100$ times. In \Cref{fig:scbmec-experiment-main}, curves show the mean cumulative pseudo-regret and shaded regions show one standard deviation across these runs. 

At $T = 50{,}000$, \texttt{POMIS~union} has the lowest mean cumulative regret in all twelve task--solver comparisons in \Cref{tab:scbmec-experiment-regret}. Task~3 illustrates the benefit of substantial action reduction. \texttt{POMIS~union} retains $54$ arms, compared with $19{,}683$ for BF, and both solvers incur much less regret. However, fewer arms do not guarantee lower finite-horizon regret. In Task~6, KL-UCB incurs mean regret $4{,}374.8$ with DMIS, compared with $4{,}194.8$ with BF. These results illustrate the benefit of the \texttt{POMIS~union} action space on the six fixed SCMs with oracle PAGs. 

\jmlrsection{On Transportability for Structural Causal Bandits}\label{chap:scb tr}
\Cref{chap:scb,chap:scb mec} used causal structure to reduce the action space while preserving an optimal action. Substantial exploration may nevertheless remain, and each intervention can be costly. A complementary strategy is to use data already collected from related source environments. Such data can narrow the possible rewards before interaction begins and reduce the \textit{cold-start} problem. 

However, an intervention that performs well in one environment may perform poorly in another. Transferring its reward distribution therefore requires accounting for the mechanisms that may change.

This \unitname{} considers a known causal diagram and addresses these environmental discrepancies through \textit{transportability} \citep{pearl2011transportability,bareinboim2012transportability,bareinboim2016causal}. \citet{zhang2017transfer} studied the transfer of prior observations for particular structures, including bow and instrumental-variable graphs. \citet{bellot2023transportability} and \citet{deng2025transfer} considered more general causal diagrams with single-variable interventions. Here, the available source experiments may involve several variables, and some variables may be non-manipulable in the target environment. 

We first explain how non-manipulable variables change the POMIS action space. Then, we introduce the following contributions of this \unitname{}. 

\begin{itemize}
    \item We derive \textit{causal bounds} on individual action rewards by combining source distributions. 

    \item We next provide \textit{dominance bounds}, which compare the best rewards under different intervention constraints.

    \item Finally, we use these bounds to guide exploration in the target environment.  
\end{itemize}

\paragraph{Relation to the published version.}
This \unitname{} revises \citet{park2025transportability}. The assumptions of Proposition~1 of \citet{park2025transportability} are corrected to allow $\mf{C} \subseteq \mf{C}' \subseteq \mf{D}$ with $\mf{C}$ ancestral in $\mc{G}[\mf{C}']$ (\Cref{tr:prop: Q trans bound}). For the online policy, the variance interval uses the action-specific lower bound $\ell_{\mf{x}}$, and a Bernstein index and its corresponding regret guarantee replace the published index and bound (\Cref{tr:eq:variance-bound,tr:eq:ucb,tr:thm: regret}). The experiments in \Cref{tr:sec:experiments} use this revised policy. Task~2 also uses a revised reward mechanism and controlled comparisons to isolate the effect of the upper dominance bound: the improvement reported for the original Task~2 did not establish that effect. The resulting regret measurements replace the published numerical results. 

\section{Structural Causal Bandits with Non-manipulable Variables} 

\Cref{chap:scb} allowed interventions on every observed variable except the reward. In many applications, an observed variable can be measured but cannot be set directly. For example, a treatment can be assigned while a patient's physiological response must follow its natural mechanism. Let $\mf{N} \subseteq \mf{V} \setminus \{Y\}$ denote these \textit{non-manipulable} variables. The agent may now intervene only on sets $\mf{X} \subseteq \mf{V} \setminus \{Y\} \setminus \mf{N}$ \citep{lee2019structural}. The variables in $\mf{N}$ continue to respond to their causes, including interventions on other variables. Thus, this restriction changes which actions are available while preserving the SCM and reward notation introduced earlier. As before, the agent chooses an intervention based on the past history, before observing the variables generated in the current round. 

\paragraph{MISs and POMISs under constraint.} 
The definitions of MIS and POMIS in \Cref{chap:scb} extend by restricting their candidate intervention sets to the manipulable variables. We denote the resulting collections by $\mb{M}_{\mc{G},Y}^{\mf{N}}$ and $\mb{P}_{\mc{G},Y}^{\mf{N}}$, respectively. Minimality is unchanged because every subset of a feasible intervention set is also feasible. Consequently:
\begin{align}
    \mb{M}_{\mc{G},Y}^{\mf{N}} = \{ \mf{X} \in \mb{M}_{\mc{G},Y} : \mf{X} \cap \mf{N} = \emptyset \}.
\end{align}
In particular, the ancestral criterion in \Cref{chap:scb} applies directly after checking that $\mf{X}$ is feasible. 

Possible optimality also depends on which \textit{competing} intervention sets remain available. A feasible MIS $\mf{X}$ belongs to $\mb{P}_{\mc{G},Y}^{\mf{N}}$ if some SCM conforming to $\mc{G}$ satisfies:
\begin{align}
    \mu_{\mf{x}^\ast} > \mu_{\mf{w}^\ast} \quad \text{for~every~} \mf{W} \in \mb{M}_{\mc{G},Y}^{\mf{N}} \setminus \{\mf{X}\}. 
\end{align}
An unconstrained POMIS that remains feasible still satisfies this requirement, because its witnessing SCM has fewer competitors to beat. Therefore:
\begin{align}
    \{ \mf{X} \in \mb{P}_{\mc{G},Y} : \mf{X} \cap \mf{N} = \emptyset\} 
    \subseteq
    \mb{P}_{\mc{G},Y}^{\mf{N}}
    \subseteq
    \mb{M}_{\mc{G},Y}^{\mf{N}}.
\end{align}
The first inclusion can be strict. An MIS previously dominated by an intervention on a non-manipulable variable may become a necessary candidate once that intervention is unavailable. Hence, simply removing infeasible sets from $\mb{P}_{\mc{G},Y}$ can discard an optimal feasible action.

\paragraph{Graphical characterization for POMISs under constraint.} To account for this change, we use the latent projection from \Cref{def: latent projection} of $\mc{G}$ onto $\mf{V} \setminus \mf{N}$, denoted by $\mc{H} = \mc{G} \langle \mf{V} \setminus \mf{N} \rangle$. 

The reason for projection is that an unavailable intervention cannot interrupt a causal pathway through $\mf{N}$. Algebraically, each non-manipulable variable's structural function can be substituted into the functions of its children. Repeating this substitution gives a model on $\mc{H}$ with the same reward distribution for every feasible intervention. 

The graphical characterization ranges over SCMs compatible with the diagram,
allowing the domains of non-manipulable variables to vary so their mechanisms
can encode the inputs needed by their children. 

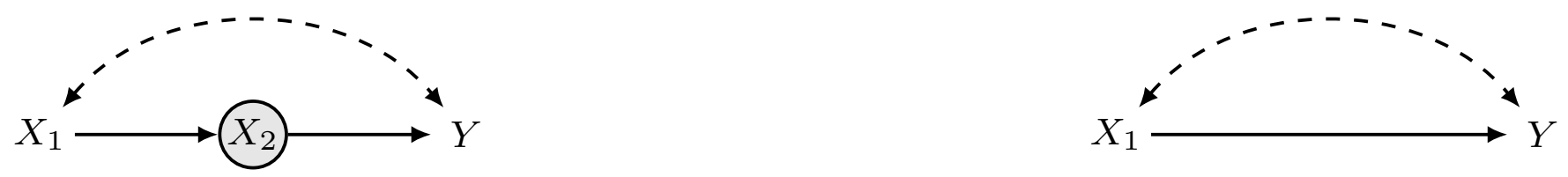
\begin{figure}[t]
    \centering
    \tikzset{
        trnonmanipgraph/.style={x=1.6cm,y=1cm,font=\scriptsize,
            >={Latex[width=1.3mm,length=1.5mm]},line width=0.25mm},
        trnonmanipnode/.style={draw=none,circle,inner sep=0mm,minimum size=5mm}
    }
    \begin{minipage}[b]{0.47\textwidth}\centering
    \subfloat[$\mc{G}$]{
        \begin{tikzpicture}[trnonmanipgraph]
            \path[use as bounding box] (-0.2,-0.3) rectangle (2.2,1.0);
            \node[trnonmanipnode] (X1) at (0,0) {$X_1$};
            \node[trnonmanipnode,draw,fill=gray!20] (X2) at (1,0) {$X_2$};
            \node[trnonmanipnode] (Y) at (2,0) {$Y$};
            \draw[->] (X1) -- (X2);
            \draw[->] (X2) -- (Y);
            \draw[<->,dashed] (X1) to[bend left=50] (Y);
        \end{tikzpicture}\label{fig: scbtr_nonmanip-projection-full}}
    \end{minipage}
    \hfill
    \begin{minipage}[b]{0.47\textwidth}\centering
    \subfloat[$\mc{G}\langle\{X_1,Y\}\rangle$]{
        \begin{tikzpicture}[trnonmanipgraph]
            \path[use as bounding box] (-0.2,-0.3) rectangle (2.2,1.0);
            \node[trnonmanipnode] (X1) at (0,0) {$X_1$};
            \node[trnonmanipnode] (Y) at (2,0) {$Y$};
            \draw[->] (X1) -- (Y);
            \draw[<->,dashed] (X1) to[bend left=50] (Y);
        \end{tikzpicture}\label{fig: scbtr_nonmanip-projection-latent}}
    \end{minipage}
    \caption{\small A non-manipulable mediator changes which intervention sets are POMISs. The shaded variable $X_2$ is observed but cannot be intervened on. The unconstrained POMISs are $\{\emptyset,\{X_2\}\}$, whereas the constraint $\mf{N}=\{X_2\}$ yields $\{\emptyset,\{X_1\}\}$, as obtained from the latent projection in (b).}
    \label{fig: scbtr_nonmanip-projection}
\end{figure}

\begin{theorem}[Constrained POMISs \citep{lee2019structural}]\label{thm: constrained pomis}
    With $\mc{H} = \mc{G} \langle \mf{V} \setminus \mf{N} \rangle$, $\mb{P}_{\mc{G},Y}^{\mf{N}} = \mb{P}_{\mc{H},Y}$, and a feasible set $\mf{X}$ is a constrained POMIS if and only if $\IB(\mc{H}_{\overline{\mf{X}}},Y) = \mf{X}$. 
\end{theorem}
\begin{proof}[Proof of \Cref{thm: constrained pomis}]
    Latent projection preserves directed paths between retained vertices that avoid an intervention set. The ancestral MIS criterion therefore gives $\mb{M}_{\mc{G},Y}^{\mf{N}} = \mb{M}_{\mc{H},Y}$. Substituting the mechanisms in $\mf{N}$ preserves every feasible reward distribution, so each constrained POMIS witness on $\mc{G}$ is also a witness on $\mc{H}$.

    Conversely, the constructive proof of \Cref{thm: IB = POMIS for CD} provides a POMIS witness on $\mc{H}$ whose independent exogenous variables each affect at most two observed variables \citep{lee2018structural}. The reverse construction of Proposition~3 in \citet{lee2019structural} expands this witness to $\mc{G}$ by letting restored variables encode the inputs needed by their children. The allowed choice of their domains permits this construction. It preserves the feasible reward distributions and hence the strict comparisons against the common MIS collection. This proves equality; the IB criterion follows from \Cref{thm: IB = POMIS for CD}. 
\end{proof}
For instance, consider the causal diagram in \Cref{fig: scbtr_nonmanip-projection-full}. Its MISs are:
\begin{align}
    \mb{M}_{\mc{G},Y} = \{\emptyset, \{X_1\}, \{X_2\}\}. 
\end{align}
The joint set $\{X_1, X_2\}$ is redundant because fixing $X_2$ removes the effect of $X_1$ on $Y$. 

When both $X_1$ and $X_2$ are manipulable, the minimal UC-territory of $\mc{G}$ contains all three variables, and thus its border is empty. Hence $\emptyset$ is a POMIS. After intervening on $X_2$, the border is $\{X_2\}$. By contrast, intervening on $X_1$ removes its bidirected edge and leaves the border $\{X_2\}$. Hence:
\begin{align}
    \mb{P}_{\mc{G},Y} = \{\emptyset, \{X_2\}\}. 
\end{align}
Now suppose that $X_2$ is non-manipulable, so $\mf{N} = \{X_2\}$. Projecting out $X_2$ gives $X_1 \to Y$ together with $X_1 \bedge Y$, as shown in \Cref{fig: scbtr_nonmanip-projection-latent}. In this projected graph, the border is empty before intervention and becomes $\{X_1\}$ after intervening on
$X_1$. Therefore:
\begin{align}
    \text{For $\mf{N} = \{X_2\}$,} \quad \mb{P}_{\mc{G},Y}^{\mf{N}} = \{\emptyset, \{X_1\}\}. 
\end{align}
Filtering the unconstrained POMISs would retain only $\emptyset$ and miss $\{X_1\}$. 

To see this, consider the following binary SCM conforming to the causal diagram in \Cref{fig: scbtr_nonmanip-projection-full}:
\begin{align}
    \mc{M}
= \begin{cases}
        \mf{U} =   \{U,U_1,U_2,U_Y\}\\
        \mf{V} = \{X_1,X_2,Y\} \\
        \mc{F} = 
        \begin{cases}
            f_{X_1}(u,u_1) = u \oplus u_1\\
            f_{X_2}(x_1,u_2) = x_1 \oplus u_2\\
            f_{Y}(x_2,u, u_Y) = x_2 \oplus u \oplus u_Y\\
        \end{cases}\\
        P(\mf{U}) = \{U \thicksim \Bern(\frac{2}{5}), U_1,U_2,U_Y \sim \Bern(\frac{1}{4})\}.
\end{cases}
\end{align}
The exogenous variables are mutually independent. The shared variable $U$ induces $X_1 \bedge Y$, while $U_1$, $U_2$, and $U_Y$ supply separate random variation in the three mechanisms. In particular, $P(X_2 = 1 \mid X_1 = 0) = \frac{1}{4}$ and $P(X_2 = 1 \mid X_1 = 1) = \frac{3}{4}$. Every assignment to $(X_1, X_2, Y)$ has positive observational probability. The observational and interventional means are:
\begin{align}
    \mu_{\emptyset} &= \frac{7}{16}, &
    \mu_{X_1=0} &= \frac{19}{40}, &
    \mu_{X_1=1} &= \frac{21}{40},\\
    \mu_{X_2=0} &= \frac{9}{20}, &
    \mu_{X_2=1} &= \frac{11}{20}. &&
\end{align}
Thus, when both variables can be manipulated, $do(X_2 = 1)$ attains the optimal mean $\frac{11}{20}$. When $\mf{N} = \{X_2\}$, the optimal feasible action is $do(X_1=1)$, with mean $\frac{21}{40}$. 

For the remainder of this \unitname{}, we write the actions induced by constrained POMISs as:
\begin{align}
    \mc{A}_{\mc{G},Y}^{\mf{N}} = \{\mf{x} : \mf{X} \in \mb{P}_{\mc{G},Y}^{\mf{N}}, \mf{x} \in \Omega_{\mf{X}} \}.
\end{align}
This reduction preserves the best achievable expected reward under the constraint. It does not determine which feasible actions are most informative during learning: observations from an excluded action may still help estimate rewards of retained actions. In the transportability setting considered next, the same distinction applies to prior experiments, including experiments on variables that cannot be manipulated in the target environment. 

\section{Structural Causal Bandits under Transportability}

Let $\pi^\ast$ denote the target environment and $\mc{M}^\ast$ its SCM, with reward $Y \in [0,1]$ and non-manipulable variables $\mf{N}^\ast$. The agent also has information from source environments $\Pi = \{\pi^1, \cdots, \pi^n\}$, associated with SCMs $\mc{M}^1,\cdots,\mc{M}^n$ over the same observed variables and domains, all conforming to the known causal diagram $\mc{G}$. We use $P^\ast_{\mf{x}}$ and $P^i_{\mf{x}}$ for the distributions under $do(\mf{x})$ in the target and source $\pi^i$, respectively. Thus, $\mu^\ast_{\mf{x}} = \mb{E}_{P^\ast_{\mf{x}}}Y$ is the target mean reward. A superscript $\ast$ on a distribution or mean specifies the target environment; the superscripts on assignments $\mf{x}^\ast$ and $\mf{x}^\star$ retain their meanings from \Cref{chap:scb}.

The target action space is $\mc{A}_{\mc{G},Y}^{\mf{N}^\ast}$. At round $t$, the agent selects $A_t$ from this space and observes the outcome of its intervention. The cumulative pseudo-regret from \Cref{chap:scb}, evaluated in the target, is:
\begin{align}
    R_T
    =T\mu^\ast_{\mf{x}^\star}-\sum_{t=1}^T\mu^\ast_{A_t}
    =\sum_{\mf{x}\in\mc{A}_{\mc{G},Y}^{\mf{N}^\ast}}
        \Delta_{\mf{x}}N_T(\mf{x}),
\end{align}
where $\Delta_{\mf{x}} = \mu^\ast_{\mf{x}^\star}-\mu^\ast_{\mf{x}}$ and $\mf{x}^\star$ is optimal under the target constraint. The objective is to minimize $\mb{E}R_T$. Source data do not contribute rewards to this objective; they provide information for choosing target interventions.

\paragraph{Environmental discrepancies.} For source $\pi^i$, let $\mb{Z}^i$ contain the intervention sets for which experimental distributions are available. The empty set denotes observational data. We write:
\begin{align}
    \mb{Z}=\{\mb{Z}^i\}_{i=1}^n,
    \qquad
    \mc{P}^{\Pi}_{\mb{Z}}
    =\{P^i_{\mf{z}}:
        \pi^i\in\Pi,\ \mf{Z}\in\mb{Z}^i,\
        \mf{z}\in\Omega_{\mf{Z}}\}.
\end{align}
A source may permit interventions that are unavailable in the target. For example, a laboratory experiment may directly control a biological mediator that cannot be assigned in a deployment population. Such experiments may still inform the rewards of feasible target actions.

For the probability bound algorithms below, observed variables have finite domains, with $\Omega_Y \subseteq [0,1]$. Source probabilities are treated as exact population quantities. 

\begin{definition}[Environment discrepancy and selection diagram]
\label{tr:def:selection}
Let $\Delta^i=\Delta^{\ast,i}$\footnote{The superscript in $\Delta^i$ denotes an environmental discrepancy; the subscript in $\Delta_{\mf{x}}$ denotes a reward gap.} contain the variables whose mechanisms may differ between $\pi^\ast$ and $\pi^i$. For each $V\in\Delta^i$, add a root node $S_V$ and an edge $S_V\to V$ to $\mc{G}$. The resulting graph is the \textit{selection diagram} $\mc{G}^{\Delta^i}$.
\end{definition}

A selection node marks a possible discrepancy between environments. Its absence asserts invariance of the corresponding mechanism, including the relevant exogenous variation. When a shared unobserved cause changes, its effects on all affected mechanisms must be represented. Equality of observed marginal distributions alone does not establish this invariance.

Let $\boldsymbol{\Delta}=\{\Delta^i\}_{i=1}^n$, $\mf{S}^i=\{S_V:V\in\Delta^i\}$, and $\mf{S}=\bigcup_i\mf{S}^i$. Adding all these selection edges gives the \textit{collective selection diagram} $\mc{G}^{\boldsymbol{\Delta}}$. We assume that the environments admit a unified SCM with this diagram, in which selection nodes are independent roots controlling the environment-specific mechanisms. With $\mf{S}^{-i}=\mf{S}\setminus\mf{S}^i$, its distributions satisfy: 
\begin{align}
    P^{\boldsymbol{\Delta}}_{\mf{x}}
        (\mf{y}\mid\mf{S}=\boldsymbol{0})
        &=P^\ast_{\mf{x}}(\mf{y}), \\
    P^{\boldsymbol{\Delta}}_{\mf{x}}
        (\mf{y}\mid\mf{S}^i=\mf{i},\mf{S}^{-i}=\boldsymbol{0})
        &=P^i_{\mf{x}}(\mf{y}).
    \label{tr:eq:selection-switch}
\end{align}
The assignment $\mf{S}^i=\mf{i}$ sets every selection node in $\mf{S}^i$ to $i$. In this representation, do-calculus can compare distributions across environments by determining whether the relevant selection nodes affect the queried outcome. 

\begin{figure}[t]
    \centering
    \tikzset{
        trselectiongraph/.style={x=1.1cm,y=0.95cm,font=\scriptsize,
            >={Latex[width=1.4mm,length=1.7mm]},line width=0.25mm},
        trselectionnode/.style={draw=none,circle,inner sep=0mm,minimum size=4mm},
        trselectionindicator/.style={rectangle,fill=betterred,draw=none,
            inner sep=0mm,minimum size=1.5mm},
        trselectionedge/.style={->,draw=betterred,line width=0.35mm}
    }
    \begin{minipage}[b]{0.31\textwidth}\centering
    \subfloat[Population shift]{
        \begin{tikzpicture}[trselectiongraph,rotate=-10]
            \path[use as bounding box] (-0.25,-0.3) rectangle (2.3,2.2);
            \node[trselectionnode] (X1) at (0,0) {$X_1$};
            \node[trselectionnode] (X2) at (1,1.5) {$X_2$};
            \node[trselectionnode] (W) at (1,0) {$W$};
            \node[trselectionnode] (Y) at (2,0) {$Y$};
            \draw[->] (X1) -- (W);
            \draw[->] (X2) -- (Y);
            \draw[->] (W) -- (Y);
            \node[trselectionnode,text=betterred] (U) at (0.4,0.8) {$U$};
            \draw[->,dashed] (U) to[bend right=20] (W);
            \draw[->,dashed] (U) to[bend left=20] (X2);
        \end{tikzpicture}\label{fig_scbtr: cardiovascular}}
    \end{minipage}
    \hfill
    \begin{minipage}[b]{0.31\textwidth}\centering
    \subfloat[$\mc{G}^{\boldsymbol{\Delta}}$]{
        \begin{tikzpicture}[trselectiongraph,rotate=-10]
            \path[use as bounding box] (-0.25,-0.3) rectangle (2.3,2.2);
            \node[trselectionnode] (X1) at (0,0) {$X_1$};
            \node[trselectionnode] (X2) at (1,1.5) {$X_2$};
            \node[trselectionnode] (W) at (1,0) {$W$};
            \node[trselectionnode] (Y) at (2,0) {$Y$};
            \draw[->] (X1) -- (W);
            \draw[->] (X2) -- (Y);
            \draw[->] (W) -- (Y);
            \draw[<->,dashed] (W) to[bend left=50] (X2);
            \node[trselectionindicator] (WT) at (0.2,0.4) {};
            \draw[trselectionedge] (WT) -- (W);
            \node[trselectionindicator] (X2T) at (0.2,1.2) {};
            \draw[trselectionedge] (X2T) -- (X2);
        \end{tikzpicture}\label{fig_scbtr: cardiovascular selection}}
    \end{minipage}
    \hfill
    \begin{minipage}[b]{0.31\textwidth}\centering
    \subfloat[Multiple sources]{
        \begin{tikzpicture}[trselectiongraph,x=1cm,rotate=-5]
            \path[use as bounding box] (-0.95,-0.3) rectangle (2.45,2.2);
            \node[trselectionnode] (A) at (0,0) {$A$};
            \node[trselectionnode] (B) at (0,1.8) {$B$};
            \node[trselectionnode] (C) at (1,0.9) {$C$};
            \node[trselectionnode] (Y) at (2,0) {$Y$};
            \draw[->] (A) -- (Y);
            \draw[->] (A) -- (C);
            \draw[->] (B) -- (C);
            \draw[->] (C) -- (Y);
            \node[trselectionindicator] (AT) at (-0.8,0.7) {};
            \draw[trselectionedge] (AT) -- (A);
            \node[trselectionindicator,fill=betterblue] (CT) at (0.2,1) {};
            \draw[trselectionedge,draw=betterblue] (CT) -- (C);
            \draw[<->,dashed] (A) to[bend left=30] (B);
            \draw[<->,dashed] (B) to[bend left=50] (Y);
        \end{tikzpicture}\label{fig_scbtr: 4var selection}}
    \end{minipage}
    \caption{\small (a) The distribution of the unobserved variable $U$ differs between source and target populations. (b) The corresponding collective selection diagram marks the affected mechanisms at $W$ and $X_2$. (c) A collective selection diagram with $\Delta^1 = \{A\}$ (red) and $\Delta^2=\{C\}$ (blue). Squares mark selection variables.}
    \label{fig_scbtr: selections}
\end{figure}
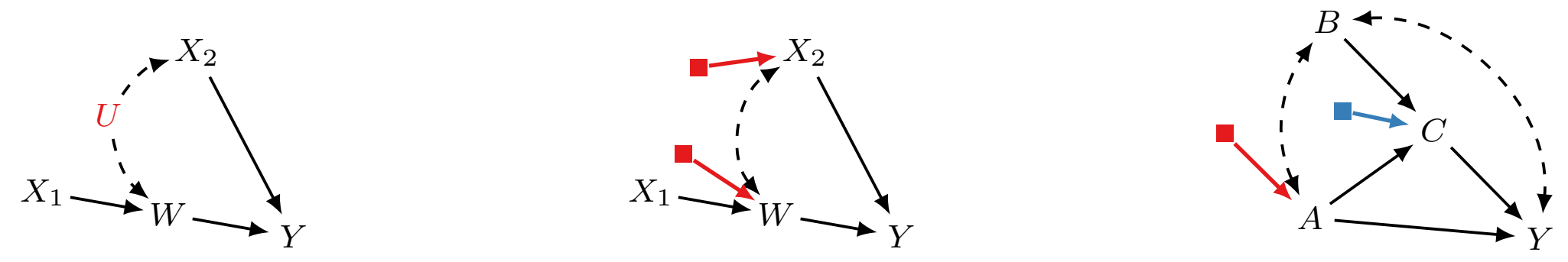

\begin{example}[\bf Treatment selection]\label{tr:ex:cardiovascular}
Consider the schematic treatment system in \Cref{fig_scbtr: cardiovascular}. Let $X_1$ indicate an antihypertensive treatment, $X_2$ an anti-diabetic treatment, $W$ blood pressure, and $Y$ a favorable health outcome. The unobserved variable $U$ influences both $W$ and $X_2$.

Suppose that the agent can assign the two treatments but cannot directly set blood pressure, so $\mf{N}^\ast=\{W\}$. Prior data from another population can help evaluate treatment choices. However, if the distribution of $U$ differs across populations, both mechanisms involving $W$ and $X_2$ may change. The collective selection diagram in \Cref{fig_scbtr: cardiovascular selection} represents this possibility by $\Delta^1=\{W,X_2\}$. The mechanism for $Y$, conditional on its causes and exogenous inputs, remains shared. We will return to this distinction when comparing feasible treatments with hypothetical interventions on $W$. 
\end{example}

\section{Partial Transportability and Causal Bounds}

A target mean $\mu_{\mf{x}}^\ast$ is \textit{transportable} from $\mc{P}_{\mb{Z}}^{\Pi}$ if it has the same value in every collection of SCMs compatible with both these source distributions and the collective selection diagram \citep{lee2020general}. Equivalently, the available information uniquely determines the target mean. Different factors in a formula may come from different environments. 

For the collective selection diagram shown in \Cref{fig_scbtr: 4var selection}, suppose that source $\pi^1$ supplies observational data and source $\pi^2$ supplies experiments on $\{A,C\}$, i.e., $\mb{Z}^1 = \{\emptyset\}$ and $\mb{Z}^2 = \{ \{A,C\}\}$. Let $\mf{N}^\ast = \{C\}$. Although $C$ cannot be manipulated in the target, experiments on $C$ in $\pi^2$ can still identify the target reward of $do(a)$. Specifically:
\begin{align}
    P^\ast_a(y)
    &=\sum_{b,c}P^\ast_a(b)P^\ast_a(c\mid b)P^\ast_a(y\mid b,c) \\
    &\overset{\mathrm{(a)}}{=}
      \sum_{b,c}P^\ast_{a,c}(b)P^\ast_\emptyset(c\mid a,b)
                     P^\ast_{a,c}(y\mid b) && \text{Rules~2 and~3}\\
    &=\sum_{b,c}P^\ast_\emptyset(c\mid a,b)P^\ast_{a,c}(b,y) \\
    &\overset{\mathrm{(b)}}{=}
      \sum_{b,c}P^1_\emptyset(c\mid a,b)P^2_{a,c}(b,y).
    \label{tr:eq: transport example}
\end{align}
Intermediate conditional probabilities are evaluated on conditioning events with positive probability; the final expression requires only the displayed source conditionals. Step~(a) uses Rules~2 and~3 of do-calculus: the intervention on $C$ does not affect $B$, conditioning on $C$ can be exchanged with intervention in the reward factor, and the conditional mechanism for $C$ is recovered observationally given $A,B$. In step~(b), source $\pi^1$ supplies this conditional mechanism because its discrepancy is confined to $A$. Source $\pi^2$ supplies the joint distribution of $B,Y$ under $do(a,c)$ because intervening on $C$ removes its changed mechanism. Neither source needs to reproduce the entire target system. Multiplying these two contributions and summing over $B, C$ gives the target reward distribution. The example illustrates why heterogeneous experiments can be useful even when they cannot be repeated in the target. 

\subsection{Partial Transportability}

Exact transportability need not hold for every action. We call a mean \textit{partially transportable} when the source information restricts it to an informative interval $[\ell_{\mf{x}}, u _{\mf{x}}] \subsetneq [0,1]$ that is valid for every compatible target SCM. To construct such intervals, we decompose the query into causal factors and bound the factors that cannot be determined exactly. 

\paragraph{Q-decomposition.} 
Recall the c-components from \Cref{chap:scb}, and write $\cc_{\mc{G}}$ for their collection. For $\mf{C}\subseteq\mf{V}$, define the causal factor: 
\begin{align}
    Q[\mf{C}](\mf{v})=P_{\mf{v} \setminus \mf{C}}(\mf{c}).
    \label{tr:eq:q-factor}
\end{align}
It is the distribution of $\mf{C}$ when every observed variable outside $\mf{C}$ is fixed by intervention. Thus, $Q[\mf{C}]$ is shorthand for an interventional distribution; its displayed arguments also record those fixed values. We usually suppress its assignment argument $\mf{v}$. Superscripts distinguish target and source factors, as in $Q^\ast[\mf{C}]$ and $Q^i[\mf{C}]$. 

Let $\mf{Y}^{+}=\An(Y)_{\mc{G}\setminus\mf{X}}$. Only these remaining ancestors can affect $Y$ after $do(\mf{x})$. The Q-decomposition \citep{tian:02} gives: 
\begin{align}
    P^\ast_{\mf{x}}(y)
    &=\sum_{\mf{y}^{+}\setminus\{Y\}}
        \prod_{\mf{C}\in\cc_{\mc{G}[\mf{Y}^{+}]}}Q^\ast[\mf{C}], \\
    \mu^\ast_{\mf{x}}
    &=\sum_{\mf{y}^{+}}y
        \prod_{\mf{C}\in\cc_{\mc{G}[\mf{Y}^{+}]}}Q^\ast[\mf{C}].
    \label{tr:eq: decompose reward}
\end{align}
The fixed intervention arguments in each factor agree with $\mf{x}$. Other fixed variables outside $\mf{Y}^{+}\cup\mf{X}$ do not affect these remaining mechanisms.

If $\mf{C}\cap\Delta^i=\emptyset$, then all mechanisms retained in $Q[\mf{C}]$ are shared, and $Q^\ast[\mf{C}]=Q^i[\mf{C}]$. When the latter factor is identifiable from an available source distribution and its graph, this equality transports the factor. Transporting every factor in \Cref{tr:eq: decompose reward} is sufficient to determine the reward distribution and hence its mean.

\begin{figure}[t]
    \centering
    \newcommand{\trjigsawside}[1]{
        (0,0) .. controls (0,0) and (0.4,-0.04*#1) ..
        (0.4,0.04*#1) .. controls (0.4,0.11*#1) and (0.2,0.26*#1) ..
        (0.5,0.26*#1) .. controls (0.8,0.26*#1) and (0.6,0.11*#1) ..
        (0.6,0.04*#1) .. controls (0.6,-0.04*#1) and (1,0) .. (1,0)
    }
    \tikzset{
        trjigsawgraph/.style={x=1.25cm,y=1.25cm,font=\scriptsize,
            line width=0.25mm,
            execute at begin picture={\path[use as bounding box]
                (-0.5,0) rectangle (2.5,2);}},
        trjigsawlabel/.style={draw=none,xshift=0.625cm,yshift=0.625cm},
        pics/trjigsawpiece/.style n args={4}{code={
            \path[pic actions]
                \trjigsawside{#1}
                [rotate around={90:(0.5,0.5)}] -- \trjigsawside{#2}
                [rotate around={90:(0.5,0.5)}] -- \trjigsawside{#3}
                [rotate around={90:(0.5,0.5)}] -- \trjigsawside{#4}
                -- cycle;
        }}
    }
    \begin{minipage}[b]{0.31\textwidth}\centering
    \subfloat[Pieces]{
        \begin{tikzpicture}[trjigsawgraph]
            \path (0,1) pic[fill=white,draw=black,dashed]
                {trjigsawpiece={1}{1}{0}{0}} node[trjigsawlabel] {$Q_1$};
            \path (1,1) pic[fill=white,draw=black,dashed]
                {trjigsawpiece={1}{0}{0}{-1}} node[trjigsawlabel] {$Q_2$};
            \path (0,0) pic[fill=white,draw=black,dashed]
                {trjigsawpiece={0}{-1}{-1}{0}} node[trjigsawlabel] {$Q_3$};
            \path (1,0) pic[fill=white,draw=black,dashed]
                {trjigsawpiece={0}{0}{-1}{1}} node[trjigsawlabel] {$Q_4$};
        \end{tikzpicture}\label{fig_scbtr: jigsaw pieces}}
    \end{minipage}
    \hfill
    \begin{minipage}[b]{0.31\textwidth}\centering
    \subfloat[Transportable]{
        \begin{tikzpicture}[trjigsawgraph]
            \path (0,1) pic[fill=betterorange!70, draw = black]
                {trjigsawpiece={1}{1}{0}{0}} node[trjigsawlabel] {$Q_1$};
            \path (1,1) pic[fill=betterblue!70, draw = black]
                {trjigsawpiece={1}{0}{0}{-1}} node[trjigsawlabel] {$Q_2$};
            \path (0,0) pic[fill=bettergreen!70, draw = black]
                {trjigsawpiece={0}{-1}{-1}{0}} node[trjigsawlabel] {$Q_3$};
            \path (1,0) pic[fill=betterorange!70, draw = black]
                {trjigsawpiece={0}{0}{-1}{1}} node[trjigsawlabel] {$Q_4$};
        \end{tikzpicture}\label{fig_scbtr: jigsaw transportable}}
    \end{minipage}
    \hfill
    \begin{minipage}[b]{0.31\textwidth}\centering
    \subfloat[Non-transportable]{
        \begin{tikzpicture}[trjigsawgraph]
            \path (1,1) pic[fill=betterblue!70, draw = black]
                {trjigsawpiece={1}{0}{0}{-1}} node[trjigsawlabel] {$Q_2$};
            \path (0,0) pic[fill=bettergreen!70, draw = black]
                {trjigsawpiece={0}{-1}{-1}{0}} node[trjigsawlabel] {$Q_3$};
            \path (1,0) pic[fill=betterorange!70, draw = black]
                {trjigsawpiece={0}{0}{-1}{1}} node[trjigsawlabel] {$Q_4$};
            \path (0,1) pic[fill=white,draw=black,dashed]
                {trjigsawpiece={1}{1}{0}{0}} node[trjigsawlabel] {$Q_1$};
        \end{tikzpicture}\label{fig_scbtr: jigsaw nontransportable}}
    \end{minipage}
    \caption{\small A jigsaw abstraction of the decomposition in \Cref{tr:eq: decompose reward}. Each piece represents a factor, and each color denotes a source environment supplying that factor. All pieces are available in (b). In (c), $Q_1$ remains unavailable, motivating partial transportability bounds.}
    \label{fig_scbtr: jigsaw_partial_tr}
\end{figure}
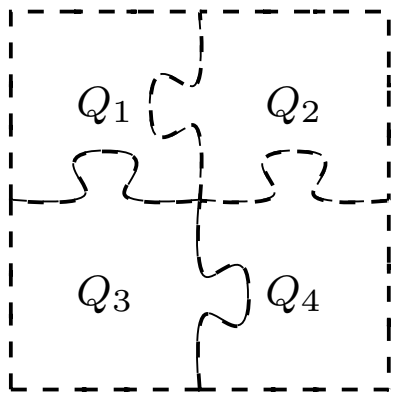
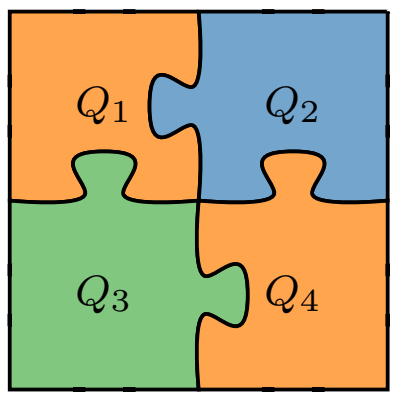
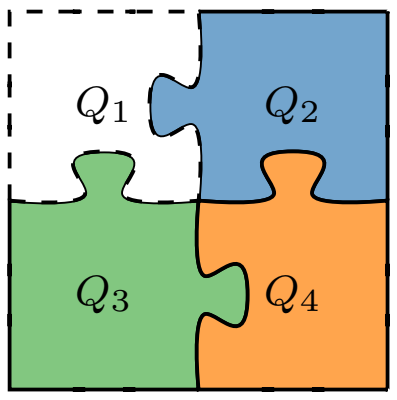

\Cref{fig_scbtr: jigsaw_partial_tr} depicts the decomposition as a jigsaw puzzle. The query determines the pieces needed, and different source experiments may provide different pieces. A missing exact piece does not require discarding the entire query: a lower or upper bound for that piece may still constrain the final reward. 

\paragraph{Bounds for unavailable factors.} 
Return to the collective selection diagram in \Cref{fig_scbtr: 4var selection}, but now suppose that $\mb{Z}^1 = \{\emptyset, \{C\}\}$ and $\mb{Z}^2 = \{\{B\}\}$. The target mean decomposes into:
\begin{align}
    \mu^\ast_a=\sum_{b,c,y} y Q^\ast[C] Q^\ast[B,Y].
\end{align}
Source $\pi^1$ still identifies $Q^\ast[C] = P^1_\emptyset(c\mid a,b)$. It does not identify $Q^1[B,Y]$: within the available factor on $\{A,B,Y\}$, latent confounding prevents removal of the mechanism for $A$. Source $\pi^2$ intervenes on $B$ and therefore does not provide the required joint factor on $B,Y$. Nevertheless, $Q^\ast[B,Y] = Q^1[B,Y]$, since neither $B$ nor $Y$ is in $\Delta^1$. 

The natural bounds \citep{manski1990nonparametric,robins1989analysis},
\begin{align}
    P(\mf{x},\mf{y})
    \leq P_{\mf{x}}(\mf{y})
    \leq P(\mf{x},\mf{y})+1-P(\mf{x}),
    \label{tr:eq:natural-bound}
\end{align}
give the starting point. For units whose natural value is $\mf{x}$, replacing the corresponding mechanisms by $\mf{x}$ leaves the values received by all other mechanisms unchanged. Their outcome is therefore unchanged, so their observed joint probability supplies the lower bound. The remaining units can contribute at most their total probability, $1 - P(\mf{x})$, to the upper bound. The same argument applies within an available causal factor. 

For the next proposition, take $\mf{C}\subseteq\mf{C}'\subseteq\mf{D}$, where $\mf{C}$ is ancestral in $\mc{G}[\mf{C}']$, $\mf{C}\cap\Delta^i=\emptyset$, and $Q^i[\mf{D}]$ is available or identifiable from source information. Ancestral marginalization gives $Q^i[\mf{C}]=\sum_{\mf{c}'\setminus\mf{C}}Q^i[\mf{C}']$.

\begin{proposition}[Bounds for a shared causal factor] \label{tr:prop: Q trans bound}
The target factor satisfies:
\begin{align}
    \sum_{\mf{c}'\setminus\mf{C}}Q^i[\mf{D}]
    \leq Q^\ast[\mf{C}] \le\sum_{\mf{c}'\setminus\mf{C}}
        \left\{Q^i[\mf{D}]+1-\sum_{\mf{c}'}Q^i[\mf{D}]\right\}.
    \label{tr:eq:q-bounds}
\end{align}
If $Q^i[\mf{C}']$ is identifiable, then $Q^\ast[\mf{C}]=\sum_{\mf{c}'\setminus\mf{C}}Q^i[\mf{C}']$.
\end{proposition}
\begin{proof}[Proof of \Cref{tr:prop: Q trans bound}]
    Fix the arguments outside $\mf{D}$ and apply \Cref{tr:eq:natural-bound} within the resulting SCM, with the intervention variables $\mf{D} \setminus \mf{C}'$. This gives, at every assignment:
    \begin{align}
        Q^i[\mf{D}] \leq 
        Q^i[\mf{C}'] \leq
        Q^i[\mf{D}] + 1 - \sum_{\mf{c}'} Q^i[\mf{D}]. 
    \end{align}
    The inner sum varies only $\mf{C}'$, holding all other arguments fixed. Because $\mf{C}$ is ancestral in $\mc{G}[\mf{C}']$, ancestral marginalization gives $Q^i[\mf{C}] = \sum_{\mf{c}' \setminus \mf{C}} Q^i[\mf{C}']$ by Lemma~3 in \citet{tian:02}. Also $Q^\ast[\mf{C}] = Q^i[\mf{C}]$ by the shared mechanisms. Summing the pointwise bounds proves the claim; the same marginalization of an identified $Q^i[\mf{C}']$ gives the exact case. 
\end{proof}

In the example, use $\mf{C} = \mf{C}' = \{B,Y\}$ and $\mf{D}=\{A,B,Y\}$. The enclosing factor is observed under the source experiment on $C$:
\begin{align}
    Q^1[A,B,Y]=P^1_c(a,b,y).
\end{align}
Combining its bounds with the exact factor for $C$ yields:
\begin{align}
    \mu^\ast_a
    &\geq \sum_{b,c,y}yP^1_\emptyset(c\mid a,b)P^1_c(a,b,y) \\
    \mu^\ast_a
    &\le\sum_{b,c,y}yP^1_\emptyset(c\mid a,b) \left\{P^1_c(a,b,y)+1-P^1_c(a)\right\}.
\end{align}
Thus, the experiment on $C$ contributes information about a target intervention on $A$, even though it does not identify its reward exactly. 

\begin{algorithm}[t]
\small
\SetAlgoNoEnd
\DontPrintSemicolon
\textbf{function~}\textsc{paTR}$(\mf{y},\mf{x},\mc{G},\boldsymbol{\Delta},\mathsf{type})$\;
\Indp{
    \tcp{Use an available experiment when its discrepancy is irrelevant.}
    \If{\rm some $\mf{Z}\in\mb{Z}^i$ satisfies $\mf{X}=\mf{Z}\cap\mf{V}$ and $(\mf{S}^i\Perp\mf{Y})_{\mc{G}^{\boldsymbol{\Delta}}\setminus\mf{X}}$}{
        \textbf{yield} $P^i_{\mf{z}\setminus\mf{V},\mf{x}\cap\mf{Z}}(\mf{y})$ for every such source experiment; \Return\;
    }
    
    \BlankLine
    \tcp{Remove variables outside the ancestral query.}
    \If{\rm $\mf{V}'=\mf{V}\setminus\An(\mf{Y})_{\mc{G}}\ne\emptyset$}{
        \textbf{yield} \textsc{paTR}$(\mf{y},\mf{x}\setminus\mf{V}',\mc{G}\setminus\mf{V}',\{\Delta^i\setminus\mf{V}'\}_i,\mathsf{type})$; \Return\;
    }
    
    \BlankLine
    \tcp{Intervene on irrelevant remaining variables by Rule~3.}
    \If{\rm $\mf{V}'=(\mf{V}\setminus\mf{X})\setminus\An(\mf{Y})_{\mc{G}\setminus\mf{X}}\ne\emptyset$}{
        \For{\rm each $\mf{v}' \in \Omega_{\mf{V}'}$}{
            \textbf{yield} \textsc{paTR}$(\mf{y},\mf{x} \cup \mf{v}',\mc{G},\boldsymbol{\Delta},\mathsf{type})$\;
        }
        \Return\;
    }

    \BlankLine
    \tcp{Combine expressions for the c-components.}
    \If{\rm $|\cc_{\mc{G}\setminus\mf{X}}|>1$}{
        \textbf{yield} $\displaystyle\sum_{\mf{v}\setminus(\mf{X}\cup\mf{Y})}\prod_{\mf{C}\in\cc_{\mc{G}\setminus\mf{X}}}$ \textsc{paTR}$(\mf{c},\mf{v}\setminus\mf{C},\mc{G},\boldsymbol{\Delta},\mathsf{type})$; \Return\;
    }

    \BlankLine
    \tcp{Try each source that preserves the remaining mechanisms.}
    \For{\rm $\pi^i\in\Pi$ with $\Delta^i\cap(\mf{V}\setminus\mf{X})=\emptyset$, and $\mf{Z}\in\mb{Z}^i$ with $\mf{Z}\cap\mf{V}\subseteq\mf{X}$}{
        \textbf{yield} \textsc{paID}$(\mf{y},\mf{x}\setminus\mf{Z},P^i_{\mf{z}\setminus\mf{V},\mf{x}\cap\mf{Z}},\mc{G}\setminus(\mf{Z}\cap\mf{X}),\mathsf{type})$\;
    }
}
\Indm
\caption{\small Partial transportability.}
\label{tr:alg: paTR}
\end{algorithm}

\paragraph{Algorithmic partial transportability.}
\textsc{paTR} in \Cref{tr:alg: paTR} constructs lower or upper expressions for a target query, following the recursive structure of general transportability \citep{lee2020general}. 

Its arguments are the outcome assignment $\mf{y}$, intervention assignment $\mf{x}$, current graph $\mc{G}$, current discrepancy sets $\boldsymbol{\Delta}$, and bound direction $\mathsf{type}\in\{\ell,u\}$. In every call, $\mf{V}=\mf{V}(\mc{G})$ denotes the current vertex set, and $\mf{X},\mf{Y}\subseteq\mf{V}$ are the disjoint variable sets of $\mf{x},\mf{y}$. The sources $\Pi$, experimental sets $\mb{Z}$, and available distributions $\mc{P}_{\mb{Z}}^\Pi$ are fixed global information. Each yielded expression is a candidate lower bound when $\mathsf{type}=\ell$ or upper bound when $\mathsf{type}=u$ for the current target probability; an exact expression serves both directions.

A \textbf{yield} emits an expression, and a recursive \textbf{yield} forwards every expression obtained by that call. A following \textbf{return} ends the current call. If no source supplies a required expression, that call yields nothing. Whenever a source experiment is used, all available assignments $\mf{z}$ agreeing with the current query on $\mf{Z}\cap\mf{V}$ are considered. Values $\mf{z}\setminus\mf{V}$ remain fixed external arguments. A source distribution supplied to a recursive call is first marginalized to the vertices of that call's graph, with its intervention arguments held fixed.

The lines of \Cref{tr:alg: paTR} have the following roles:
\begin{itemize}
    \item \textbf{Use a direct source expression (lines~2--3).}
    The experiment must set exactly $\mf{X}$ among the current vertices. The separation condition checks that its source discrepancy does not affect $\mf{Y}$ after those vertices are removed. Every qualifying source expression is then exact for the current target query, and the call ends.

    \item \textbf{Remove non-ancestors (lines~4--5).}
    Variables outside $\An(\mf{Y})_{\mc{G}}$ cannot affect the query. The recursive call removes them from the graph, intervention assignment, and discrepancy sets, then forwards its expressions.

    \item \textbf{Add irrelevant interventions (lines~6--9).}
    Among the variables remaining after $\mf{X}$ is removed, those outside the ancestors of $\mf{Y}$ may be fixed to any assignment $\mf{v}'$ by Rule~3. For every assignment $\mf{v}'\in\Omega_{\mf{V}'}$, the target query satisfies $P^\ast_{\mf{x}}(\mf{y})=P^\ast_{\mf{x}\mf{v}'}(\mf{y})$. The algorithm forwards the candidate expressions from all these assignments before ending the call. Although the target probability is invariant to this choice, the available source factors used to bound it may differ. \textsc{CausalBound} intersects the resulting bounds, which can tighten the interval without claiming sharpness. 

    \item \textbf{Decompose the query (lines~10--11).}
    If several c-components remain, the algorithm queries the causal factor of each component $\mf{C}$ by fixing the other current variables. It forms every combination of one yielded expression per component, multiplies them, and sums over variables outside $\mf{X}\cup\mf{Y}$. Components may use different sources; a combination contributes no expression if any required component yields none.

    \item \textbf{Choose a source for the remaining component (lines~12--13).}
    A source is eligible when its discrepancies avoid all remaining mechanisms, $\mf{V}\setminus\mf{X}$, and its experiment intervenes only within $\mf{X}$ among the current vertices. The variables already fixed by that experiment are removed from the graph and from the intervention still to be performed. The resulting source distribution and query are passed to \textsc{paID}, which supplies exact expressions or natural bounds.
\end{itemize}

\begin{algorithm}[t]
\small
\SetAlgoNoEnd
\DontPrintSemicolon
\textbf{function~}\textsc{paID}$(\mf{y},\mf{x},P,\mc{G},\mathsf{type})$\;
\Indp{
    \tcp{Marginalize when no intervention remains.}
    \If{\rm $\mf{X}=\emptyset$}{\textbf{yield} $\sum_{\mf{v}\setminus\mf{Y}}P(\mf{v})$; \Return\;}
    \If{\rm $\mf{V}\ne\An(\mf{Y})_{\mc{G}}$}{
        Set $\mf{D}=\An(\mf{Y})_{\mc{G}}$.\;
        \textbf{yield} \textsc{paID}$(\mf{y},\mf{x}\cap\mf{D},\sum_{\mf{v}\setminus\mf{D}}P(\mf{v}),\mc{G}[\mf{D}],\mathsf{type})$; \Return\;
    }
    
    Let $\{\mf{C}\}=\cc_{\mc{G}\setminus\mf{X}}$.\;
    
    \BlankLine
    \tcp{Use natural bounds when exact identification stops.}
    \If{\rm $\cc_{\mc{G}}=\{\mf{V}\}$}{
        \eIf{\rm $\mathsf{type}=\ell$}{
            \textbf{yield} $\sum_{\mf{v}\setminus(\mf{X}\cup\mf{Y})}P(\mf{v})$\;
        }{
            \textbf{yield} $\sum_{\mf{v}\setminus(\mf{X}\cup\mf{Y})}\{P(\mf{v})+1-\sum_{\mf{v}\setminus\mf{X}}P(\mf{v})\}$\;
        }
        \Return\;
    }
    
    \BlankLine
    \tcp{Extract an identifiable factor or recurse inside its c-component.}
    \If{\rm $\mf{C}\in\cc_{\mc{G}}$}{
        \textbf{yield} $\sum_{\mf{c}\setminus\mf{Y}}\prod_{V_j\in\mf{C}}P(v_j\mid\mf{v}_{\prec j})$; \Return\;
    }
    Let $\mf{C}'\in\cc_{\mc{G}}$ contain $\mf{C}$.\;
    Set $Q[\mf{C}']=\prod_{V_j\in\mf{C}'}P(v_j\mid\mf{v}_{\prec j})$.\;
    \textbf{yield} \textsc{paID}$(\mf{y},\mf{x}\cap\mf{C}',Q[\mf{C}'],\mc{G}[\mf{C}'],\mathsf{type})$\;
}
\Indm
\caption{\small Partial identification subroutine for \textsc{paTR}.}
\label{tr:alg: paID}
\end{algorithm}

The subroutine \textsc{paID} in \Cref{tr:alg: paID} works within the selected source. Its arguments are $\mf{y},\mf{x},P,\mc{G},\mathsf{type}$, with the same local graph, assignment, and bound-direction conventions. Here $P$ is the source distribution or a causal factor over the current vertices, with outside assignments held fixed. The reductions in \textsc{paTR} ensure that each call satisfies:
\begin{align}
    \mf{C}=\mf{V}\setminus\mf{X}
    =\An(\mf{Y})_{\mc{G}\setminus\mf{X}},
    \qquad \cc_{\mc{G}\setminus\mf{X}}=\{\mf{C}\}.
\end{align}
The recursive calls of \textsc{paID} preserve this condition. The subroutine yields lower or upper expressions for the intervention query within $P$; the source-invariance check in \textsc{paTR} makes these expressions valid for the corresponding target query. In the factor products in \Cref{tr:alg: paID}, $\mf{v}_{\prec j}$ denotes values preceding $V_j$ in a topological ordering of the current graph, and the conditionals are formed from the current $P$ while holding its outside assignments fixed.

\begin{itemize}
    \item \textbf{Finish an observational query (lines~2--3).}
    If no intervention remains, marginalizing $P$ to $\mf{Y}$ gives an exact expression.

    \item \textbf{Restrict to ancestors (lines~4--6).}
    Set $\mf{D}=\An(\mf{Y})_{\mc{G}}$, sum out variables outside $\mf{D}$ from $P$, and retain only the graph and intervention on $\mf{D}$.

    \item \textbf{Bound an unavailable factor (lines~7--13).}
    Line~7 names the unique remaining c-component $\mf{C}$. If the whole current graph is one c-component, exact extraction stops. Lines~9--12 apply the lower or upper natural bound from \Cref{tr:prop: Q trans bound} and marginalize to $\mf{Y}$, after which the call ends. In the preceding example, this is the step that bounds $Q^1[B,Y]$ using $Q^1[A,B,Y]$.

    \item \textbf{Extract an exact factor (lines~14--15).}
    If $\mf{C}$ is already a c-component of the current graph, its Q-factor is the product of the displayed conditionals for $V_j\in\mf{C}$. Summing over $\mf{C}\setminus\mf{Y}$ gives the exact query.

    \item \textbf{Reduce to an enclosing component (lines~16--18).}
    Otherwise, locate the c-component $\mf{C}'$ containing $\mf{C}$ and extract its identifiable factor $Q[\mf{C}']$ from $P$. Recurse using this factor, the induced graph on $\mf{C}'$, and the interventions still inside $\mf{C}'$. Assignments outside $\mf{C}'$ remain fixed arguments of the extracted factor.
\end{itemize}

\begin{theorem}\label{tr:thm: sound patr}
    Every expression yielded by \textsc{paTR} is a valid lower or upper bound, according to $\mathsf{type}$, for the queried target probability. 
\end{theorem}

The proof is given in \Cref{sec: proof of tr:thm: sound patr}. Several expressions can bound the same probability. The theorem guarantees that these bounds are valid, although the source information may permit tighter bounds. 

\subsection{Causal Bounds}
\begin{algorithm}[t]
\small
\SetAlgoNoEnd
\DontPrintSemicolon
\tcp{Retain the trivial bounds when no source gives an expression.}
Initialize $[\ell_{\mf{x}},u_{\mf{x}}]=[0,1]$.\;
\BlankLine
\tcp{Intersect all lower bounds after evaluating the source quantities.}
\For{\rm each expression $P^\ell_{\mf{x}}(y)$ yielded by \textsc{paTR}$(y,\mf{x},\mc{G},\boldsymbol{\Delta},\ell)$}{
    
    \lIf{\rm after the stated chain-rule simplifications, a required source conditional is undefined at an evaluated assignment}{\textbf{continue}}
    
    Set $\ell_{\mf{x}} = \max\{\ell_{\mf{x}},\sum_y yP^\ell_{\mf{x}}(y)\}$.\;
}

\BlankLine

\tcp{A probability upper bound can be truncated at one.}

\For{\rm each expression $P^u_{\mf{x}}(y)$ yielded by \textsc{paTR}$(y,\mf{x},\mc{G},\boldsymbol{\Delta},u)$}{
    \lIf{\rm after the stated chain-rule simplifications, a required source conditional is undefined at an evaluated assignment}{\textbf{continue}}
    
    Set $u_{\mf{x}} = \min\{u_{\mf{x}},\sum_y y\min\{1,P^u_{\mf{x}}(y)\}\}$.\;
}
\Return{$[\ell_{\mf{x}},u_{\mf{x}}]$}\;
\caption{\small \textsc{CausalBound}: partially transportable reward bounds.}
\label{tr:alg: causal bound}
\end{algorithm}

\textsc{CausalBound} in \Cref{tr:alg: causal bound} takes the reward $Y\in[0,1]$, an intervention assignment $\mf{x}$, the diagram $\mc{G}$, and discrepancies $\boldsymbol{\Delta}$, together with the fixed source experiments $\mb{Z}$ and distributions $\mc{P}_{\mb{Z}}^\Pi$. It returns an interval $[\ell_{\mf{x}},u_{\mf{x}}]$ containing the target mean $\mu_{\mf{x}}^\ast$. Its calls to \textsc{paTR} use the fixed source information specified above. 

Before checking support, we combine products of source probabilities by the chain rule, preserving their source and fixed experimental arguments. For example, for disjoint assignments $a,b,c,d$, the product $P(a,b \mid c,d)P(c\mid d)$ is evaluated through the merged expression $P(a,b,c \mid d)$. The merged conditional requires $P(d)>0$ and is defined even when $P(c,d)=0$. This rule uses the joint-probability expression supplied by the chain rule; it does not permit arbitrary cancellation of zero-denominator ratios or numerical evaluation of undefined conditionals.

After these simplifications, we evaluate a formula only when every remaining source conditional it uses has a conditioning event with positive probability. Otherwise, that formula is omitted; another formula may still provide a bound. If none is usable, we retain $0 \leq \mu_{\mf{x}}^\ast \leq 1$. Thus, the graph determines which source mechanisms can be transferred, while the source probabilities determine whether a particular formula can be evaluated.

Line~1 starts from the valid interval $[0,1]$. Lines~2--4 evaluate each available lower expression, sum it against the nonnegative reward values, and retain the largest resulting lower bound. Lines~5--7 apply the same operation to upper expressions, truncating each probability bound at one before taking the smallest reward upper bound. Lines~3 and~6 apply the support test after the stated chain-rule simplifications and discard a formula if any remaining source conditional required for its evaluation is undefined. Thus a missing formula does not invalidate the other bounds, and a direction with no usable formula retains its trivial endpoint. Line~8 returns the intersection of the usable bounds. These operations are sound because multiplying pointwise probability bounds by $y \geq 0$ and summing preserves their direction. 

\begin{corollary}\label{tr:cor: causal bound}
\textsc{CausalBound} returns $\ell_{\mf{x}}\le\mu^\ast_{\mf{x}}\leq u_{\mf{x}}$. If this interval is a  
proper subinterval of $[0,1]$, the target mean is partially transportable.
\end{corollary}
\begin{proof}
By \Cref{tr:thm: sound patr}, every usable lower and upper expression satisfies $0\leq P^\ell_{\mf{x}}(y)\leq P^\ast_{\mf{x}}(y) \le\min\{1,P^u_{\mf{x}}(y)\}$. Multiplying by $y \geq 0$ and summing over $y$ gives valid bounds on $\mu^\ast_{\mf{x}}$. Taking the maximum of these lower bounds and the minimum of these upper bounds, starting from $[0,1]$, preserves validity. A proper resulting subinterval is
informative by the definition of partial transportability.
\end{proof}

\paragraph{Computational cost.}
The cost includes graph operations, expression generation, and evaluation of source probabilities. Enumeration can be large: if each of $d$ components has $z$ candidate expressions, their product can yield $z^d$ combinations. Each Rule~3 branch also visits $\lvert \Omega_{\mf{V}'} \rvert$ assignments; these finite-domain branching factors compound across recursive calls. The total cost must therefore account for the number and size of returned expressions, as well as the assignments summed over in their evaluation. Source quantities and repeated subqueries can be reused. These computations occur before target interaction; once the bounds are available, the online policy uses their numerical endpoints. 

\section{Dominance Bounds}
\label{tr:sec: dominance bounds}

Causal bounds constrain individual rewards. The graph also relates the best rewards attainable from different action spaces. We say that one action space \textit{dominates} another when its maximum expected reward is at least as large. This comparison concerns the maxima over the two spaces, rather than every pair of their actions. 

\subsection{Lower Dominance Bound}

Under the target constraint $\mf{N}^\ast$, the POMIS action space attains at least the reward of every feasible MIS action. Therefore, any lower bound for a feasible MIS action also lower-bounds the target optimum. Define:
\begin{align}
    \ell^\star
    =\max_{\substack{\mf{W}\in\mb{M}_{\mc{G},Y}^{\mf{N}^\ast}\\
                    \mf{w}\in\Omega_{\mf{W}}}}\ell_{\mf{w}}.
    \label{tr:eq: ldb}
\end{align}
If $\mf{w}^\dagger$ attains this maximum, then:
\begin{align}
    \ell^\star
    =\ell_{\mf{w}^\dagger}
    \le\mu^\ast_{\mf{w}^\dagger}
    \le\mu^\ast_{\mf{x}^\star}.
\end{align}
The maximizing lower bound may come from an MIS that is not a POMIS. Its reward is still achievable, so it can certify that another action is suboptimal. In particular, $u_{\mf{x}}<\ell^\star$ implies $\mu^\ast_{\mf{x}}<\mu^\ast_{\mf{x}^\star}$ and permits safe removal of $\mf{x}$.

\subsection{Upper Dominance Bound}

A non-manipulable variable restricts the interventions available to the agent. Allowing an additional intervention cannot reduce the best attainable reward. The constrained POMIS construction makes this comparison explicit.

\begin{proposition}\label{tr:prop: POMIS to MIS}
    A POMIS under $\langle\mc{G},Y,\mf{N}\rangle$ is an MIS under $\langle\mc{G},Y\rangle$.
\end{proposition}
\begin{proof}[Proof of \Cref{tr:prop: POMIS to MIS}]
    A constrained POMIS $\mf{X}$ is a minimal feasible intervention set. If it were not an unconstrained MIS, a proper subset of $\mf{X}$ would be interventionally equivalent for $Y$ in every compatible SCM. That subset is also feasible because $\mf{X}\cap\mf{N}=\emptyset$, contradicting constrained minimality. 
\end{proof}
A constrained POMIS can nevertheless fail to be an unconstrained POMIS: an intervention that dominates it may become available when the constraint is removed. 

\begin{example}[\textbf{Continued}]
Return to \Cref{tr:ex:cardiovascular}, with $\mf{N}^\ast=\{W\}$. Its constrained POMISs are:
\begin{align}
    \mb{P}_{\mc{G},Y}^{\{W\}}
    =\{\{X_1\},\{X_1,X_2\}\}.
\end{align}
Thus, an optimal feasible treatment can be selected among interventions on $X_1$ alone or on both treatments. Now consider the unconstrained intervention set $\mf{R}=\{W,X_2\}$. Fixing both direct causes of $Y$ can match or improve the mean attained by any distribution over their natural values. For the best intervention on $X_1$:
\begin{align}
    \mu^\ast_{x_1^\ast}
    &=\sum_{\mf{r}}\mb{E}_{P^\ast_{x_1^\ast}}[Y\mid\mf{r}]
                        P^\ast_{x_1^\ast}(\mf{r}) \\
    &\overset{\mathrm{(a)}}{=}
      \sum_{\mf{r}}\mb{E}_{P^\ast_{x_1^\ast,\mf{r}}}[Y]
                        P^\ast_{x_1^\ast}(\mf{r}) && \text{Rule~2}\\
    &\overset{\mathrm{(b)}}{=}
      \sum_{\mf{r}}\mu^\ast_{\mf{r}}P^\ast_{x_1^\ast}(\mf{r}) && \text{Rule~3} \\
    &\leq\mu^\ast_{\mf{r}^\ast}.
\end{align}
Step~(a) uses Rule~2, since the relevant paths from $W,X_2$ to $Y$ are unconfounded after fixing $X_1$. Step~(b) uses Rule~3: once $W$ is fixed, $X_1$ no longer affects $Y$. The last inequality bounds an average by its largest term.

The same argument applies to an intervention on $\{X_1,X_2\}$, averaging only over $W$. Consequently, the best reward under the target constraint is no greater than the best reward under $do(w,x_2)$. The latter interventions remain unavailable to the target agent, but an upper bound on their rewards can still constrain every feasible treatment.
\end{example}

\begin{theorem}\label{tr:thm: dominance relations}
    Let $\mf{N}\subsetneq\mf{N}^\ast$, and let $\mf{r}^\star$ be optimal in $\mc{A}_{\mc{G},Y}^{\mf{N}}$. Then $\mu^\ast_{\mf{x}^\star}\le\mu^\ast_{\mf{r}^\star}$.
\end{theorem}
\begin{proof}[Proof of \Cref{tr:thm: dominance relations}]
Every intervention feasible under $\mf{N}^\ast$ is feasible under $\mf{N}\subsetneq\mf{N}^\ast$. The maximum target mean over all feasible interventions therefore cannot decrease when the constraint is relaxed. POMIS reduction preserves each maximum, giving $\mu^\ast_{\mf{x}^\star}\le\mu^\ast_{\mf{r}^\star}$.
\end{proof}

The optimal action $\mf{r}^\star$ under the weaker constraint is unknown. To obtain a computable upper bound, maximize the available upper bounds over \textit{all} POMIS actions under the weaker constraint: 
\begin{align}
    \mu^\ast_{\mf{x}^\star}
    \leq\max_{\mf{r}\in\mc{A}_{\mc{G},Y}^{\mf{N}}}\mu^\ast_{\mf{r}}
    \leq\max_{\mf{r}\in\mc{A}_{\mc{G},Y}^{\mf{N}}}u_{\mf{r}}.
    \label{tr:eq:weaker-upper}
\end{align}
Combining these bounds, including the original constraint, gives:
\begin{align}
    u^\star
    =\min_{\mf{N}\subseteq\mf{N}^\ast}
        \max_{\mf{r}\in\mc{A}_{\mc{G},Y}^{\mf{N}}}u_{\mf{r}}.
    \label{tr:eq:udb}
\end{align}
Every reward in this calculation is a \textit{target} reward. Source experiments are used to bound those rewards; the optimum of a source environment is not itself a bound on the target optimum.

\begin{algorithm}[t]
\small
\SetAlgoNoEnd
\DontPrintSemicolon
\textbf{function~}\textsc{udb}$(\mc{G},Y,\mf{N},\mb{L},\mb{U})$\;
\Indp{
    \tcp{Share stored results across recursive calls.}
    Return a stored result for $\mf{N}$ if available; otherwise store each return value below.\;
    
    \tcp{Construct the POMIS action space for the current constraint.}
    
    Set $\mc{H}=\mc{G}\langle\mf{V}\setminus\mf{N}\rangle$.\;
    
    Set $\mc{A}_{\mc{H},Y}=\{\mf{r}:\mf{R}\in\mb{P}_{\mc{H},Y},\ \mf{r}\in\Omega_{\mf{R}}\}$.\;
    Choose $\mf{r}^{\dagger}\in\arg\max_{\mf{r}\in\mc{A}_{\mc{H},Y}}u_{\mf{r}}$ preferring larger lower bounds in a tie.\;
    
    Set $u^\star=u_{\mf{r}^{\dagger}}$.\;
    
    \BlankLine
    \tcp{An exact maximizing bound determines this constrained optimum.}
    
    \lIf{\rm $u_{\mf{r}^{\dagger}}=\ell_{\mf{r}^{\dagger}}$}{\Return{$u^\star$}}
    
    \lIf{\rm $\mf{N}=\emptyset$\nllabel{tr:line:udb:base}}{\Return{$u^\star$}}
    \BlankLine
    
    \tcp{Every weaker constraint supplies another valid upper bound.}
    
    \For{\rm each $\mf{N}'\subsetneq\mf{N}$ with $|\mf{N}'|=|\mf{N}|-1$}{
        Set $u^\star\gets\min\{u^\star,\textsc{udb}(\mc{G},Y,\mf{N}',\mb{L},\mb{U})\}$.\;
    }
    \Return{$u^\star$}\;
}
\Indm
\caption{\small Upper dominance bound by traversing weaker constraints.}
\label{tr:alg: UDB}
\end{algorithm}

\subsection{Traversing the Constraint Hierarchy}

Removing one variable from $\mf{N}$ gives an immediately weaker constraint. Repeating this step reaches every subset of $\mf{N}$. Exact optimal rewards increase weakly along these paths, but their computed upper bounds need not: source experiments can constrain a hypothetical intervention more tightly than the original feasible actions.

\textsc{udb} in \Cref{tr:alg: UDB} takes $\mc{G}$, $Y$, the current non-manipulable set $\mf{N}$, and valid target mean bounds $\mb{L}=\{\ell_{\mf{r}}\}$ and $\mb{U}=\{u_{\mf{r}}\}$ for the actions queried during recursion. It returns an upper bound on the best target reward under $\mf{N}$. The diagram, reward, and causal bounds remain fixed, and an initially empty collection of stored results is shared across the recursive calls.

The lines of \Cref{tr:alg: UDB} have the following roles:
\begin{itemize}
    \item \textbf{Construct the current candidates (lines~2--6).} Reuse a stored result if this constraint was visited before. Otherwise, project out $\mf{N}$, enumerate its POMIS actions, and select an action $\mf{r}^\dagger$ with the largest upper bound. Ties are resolved by preferring the larger lower bound, with a fixed rule for any remaining tie. Its upper endpoint supplies the local candidate.
    
    \item \textbf{Stop when no improvement is possible (lines~7--8).} If the maximizing action has $\ell_{\mf{r}^\dagger}=u_{\mf{r}^\dagger}$, the exact constrained optimum is known. An exact interval for another action does not suffice. At $\mf{N}=\emptyset$, no constraint remains to relax.
    
    \item \textbf{Visit weaker constraints (lines~9--11).} Remove one non-manipulable variable at a time and take the minimum of the resulting bounds and the local candidate. Store the returned value for later visits.
\end{itemize}

\begin{corollary}\label{tr:cor: udb}
\textsc{udb} returns the bound in \Cref{tr:eq:udb}; in particular, $\mu^\ast_{\mf{x}^\star}\leq u^\star$.
\end{corollary}
\begin{proof}[Proof of \Cref{tr:cor: udb}]
At a constraint $\mf{N}$, the local value $\max_{\mf{r}\in\mc{A}_{\mc{G},Y}^{\mf{N}}}u_{\mf{r}}$ bounds its constrained optimum. Every recursive value bounds an optimum under a weaker constraint, which is at least the current optimum by \Cref{tr:thm: dominance relations}. Taking their minimum is therefore valid. Removing one variable recursively visits every proper subset, so induction on $\lvert \mf{N} \rvert$ gives the subset-minimum in \Cref{tr:eq:udb}. For an early return, the maximizing action $\mf{r}^\dagger$ satisfies $\ell_{\mf{r}^\dagger}=u_{\mf{r}^\dagger}$, so its known mean equals the local maximum upper bound and hence the exact constrained optimum. Every bound at a weaker constraint is at least this value; omitting those calls therefore leaves the subset-minimum unchanged.
\end{proof}

The same constraint can be reached by removing variables in different orders. Reusing its result limits the traversal to at most $2^{\lvert \mf{N}^\ast \rvert}$ distinct constraints and $\lvert \mf{N}^\ast \rvert 2^{\lvert \mf{N}^\ast \rvert - 1}$ one-variable relaxations when $\mf{N}^\ast\neq\emptyset$. The cost at each constraint additionally includes latent projection, POMIS enumeration, and any causal bounds not already computed. Early termination can reduce this cost. 

\paragraph{Combining the bounds.}
The lower dominance bound certifies an achievable target reward; the upper dominance bound certifies a ceiling on the target optimum. Together with an action's causal bounds, they provide two ways to reduce exploration. If $u_{\mf{x}}<\ell^\star$, the action can be removed. Otherwise, its reward still satisfies:
\begin{align}
    \ell_{\mf{x}}
    \leq 
    \mu^\ast_{\mf{x}} 
    \leq
    \min\{u_{\mf{x}},u^\star\}.
\end{align}
The upper dominance bound can therefore improve the interval even when no additional action can be pruned. 

\section{Worked Examples of Bound Construction}

Every endogenous and exogenous variable is binary, and the exogenous variables within each SCM are mutually independent. For these worked examples, the source environments use the displayed structural functions and exogenous distributions, and their distributions are known at the population level. The discrepancy sets specify mechanisms that are allowed to differ, without requiring an actual change. The agent uses the collective selection diagram and available source distributions, while the displayed target means serve only to check the bounds. Numerical results are rounded to four decimal places; all comparisons use unrounded values. 

\begin{figure}[t]
    \centering
    \tikzset{
        trworkedgraph/.style={x=0.9cm,y=0.85cm,font=\scriptsize,
            >={Latex[width=1.3mm,length=1.6mm]},line width=0.25mm},
        trworkednode/.style={draw=none,circle,inner sep=0mm,minimum size=4.5mm},
        trworkedfixed/.style={trworkednode,draw,fill=gray!20},
        trworkedselection/.style={rectangle,fill=betterred,draw=none,
            inner sep=0mm,minimum size=1.5mm},
        trworkedselectionedge/.style={->,draw=betterred,line width=0.35mm}
    }
    \begin{minipage}[b]{0.25\textwidth}\centering
    \subfloat[]{
        \begin{tikzpicture}[trworkedgraph]
            \path[use as bounding box] (-0.6,-0.6) rectangle (2.2,2.3);
            \node[trworkednode] (A) at (0,0) {$A$};
            \node[trworkednode] (B) at (1,1.6) {$B$};
            \node[trworkednode] (Y) at (2,0) {$Y$};
            \draw[->] (B) -- (A);
            \draw[->] (B) -- (Y);
            \draw[->] (A) -- (Y);
            \draw[<->,dashed] (A) to[bend right=35] (Y);
            \node[trworkedselection] (SA) at (-0.45,0.9) {};
            \draw[trworkedselectionedge] (SA) -- (A);
        \end{tikzpicture}\label{tr:fig: worked-one}}
    \end{minipage}
    \hfill
    \begin{minipage}[b]{0.29\textwidth}\centering
    \subfloat[]{
        \begin{tikzpicture}[trworkedgraph]
            \path[use as bounding box] (-0.75,-0.6) rectangle (2.5,2.3);
            \node[trworkedfixed] (A) at (0,0) {$A$};
            \node[trworkednode] (B) at (0,2) {$B$};
            \node[trworkedfixed] (C) at (1,1) {$C$};
            \node[trworkednode] (Y) at (2,0) {$Y$};
            \draw[->] (A) -- (Y);
            \draw[->] (A) -- (C);
            \draw[->] (B) -- (C);
            \draw[->] (C) -- (Y);
            \draw[<->,dashed] (C) to[bend right=32] (B);
            \draw[<->,dashed] (B) to[bend left=55] (Y);
            \node[trworkedselection] (SA) at (-0.6,0.65) {};
            \draw[trworkedselectionedge] (SA) -- (A);
            \node[trworkedselection,fill=betterblue] (SB) at (-0.6,1.6) {};
            \draw[trworkedselectionedge,draw=betterblue] (SB) -- (B);
        \end{tikzpicture}\label{tr:fig: worked-two}}
    \end{minipage}
    \hfill
    \begin{minipage}[b]{0.40\textwidth}\centering
    \subfloat[]{
        \begin{tikzpicture}[trworkedgraph,x=0.95cm]
            \path[use as bounding box] (-0.6,-0.6) rectangle (3.4,2.3);
            \node[trworkedfixed] (T) at (0,1.8) {$T$};
            \node[trworkednode] (R) at (1.2,1.8) {$R$};
            \node[trworkedfixed] (W) at (0,0) {$W$};
            \node[trworkednode] (X) at (1.1,0) {$X$};
            \node[trworkednode] (Z) at (2.2,0) {$Z$};
            \node[trworkednode] (Y) at (3.3,0) {$Y$};
            \draw[->] (T) -- (R);
            \draw[->] (R) -- (Z);
            \draw[->] (W) -- (X);
            \draw[->] (X) -- (Z);
            \draw[->] (Z) -- (Y);
            \draw[<->,dashed] (W) to[bend right=32] (X);
            \draw[<->,dashed] (W) to[bend left=25] (R);
            \draw[<->,dashed] (W) to[bend right=48] (Z);
            \draw[<->,dashed] (X) to[bend left=32] (Y);
            \draw[<->,dashed] (R) to[bend left=35] (Y);
            \node[trworkedselection] (ST) at (-0.45,1.0) {};
            \draw[trworkedselectionedge] (ST) -- (T);
            \node[trworkedselection,fill=betterblue] (SR) at (2.05,2.1) {};
            \draw[trworkedselectionedge,draw=betterblue] (SR) -- (R);
        \end{tikzpicture}\label{tr:fig: worked-three}}
    \end{minipage}
    \caption{\small Collective selection diagrams for the worked examples. Gray circles mark variables that are non-manipulable in the target. Red and blue squares denote source~1 and source~2 selections, respectively: (a) $\Delta^1=\{A\}$; (b) $\Delta^1=\{A\}$ and $\Delta^2=\{B\}$; (c) $\Delta^1=\{T\}$ and $\Delta^2=\{R\}$.}
    \label{tr:fig: worked-examples}
\end{figure}
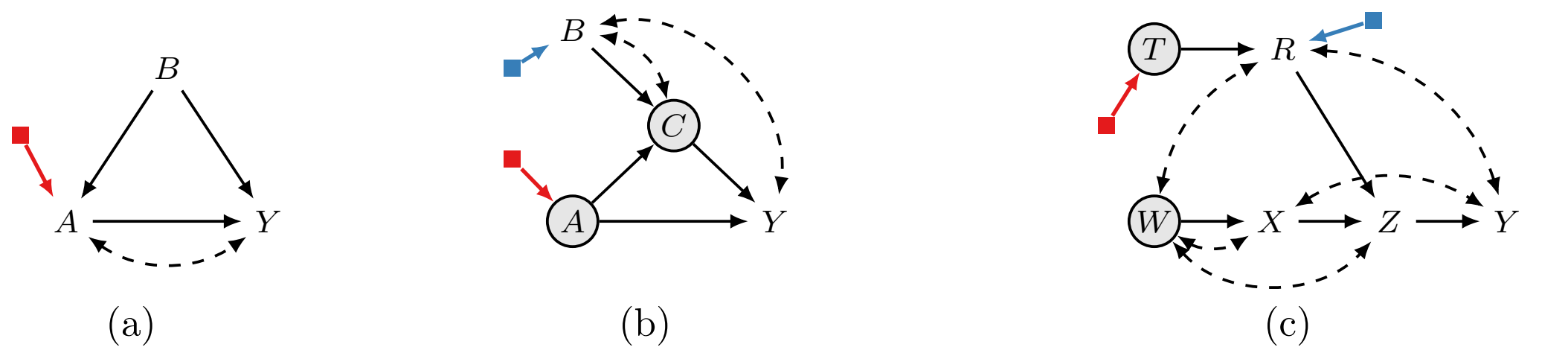
\begin{table}[t]
    \centering
    \caption{\small Bounds for all six initial POMIS actions in \Cref{tr:fig: worked-one}. The lower dominance bound is $\ell^\star=0.8325$.}
    \label{tr:tab: worked-one}
    \small
    \setlength{\tabcolsep}{9pt}
    \begin{tabular}{lrrrl}
        \toprule
        \bf Action & $\ell_{\mf{x}}$ & $\mu^\ast_{\mf{x}}$ & $u_{\mf{x}}$ & \bf Decision \\
        \midrule
        $do(B=0)$       & 0.0000 & 1.0000 & 1.0000 & Retain \\
        $do(B=1)$       & 0.0000 & 0.5900 & 1.0000 & Retain \\
        $do(A=0,B=0)$   & 0.1675 & 1.0000 & 1.0000 & Retain \\
        $do(A=0,B=1)$   & 0.2965 & 0.5450 & 0.7940 & Remove \\
        $do(A=1,B=0)$   & 0.8325 & 1.0000 & 1.0000 & Retain \\
        $do(A=1,B=1)$   & 0.2935 & 0.5900 & 0.7960 & Remove \\
        \bottomrule
    \end{tabular}
\end{table}

\paragraph{Eliminating actions before interaction.} Consider the diagram in \Cref{tr:fig: worked-one} and the following SCM. Let $\mf{V}=\{A,B,Y\}$. The exogenous variables are: 
\begin{align}
  \mf{U} = \{U_A,U_B,U_Y,U_{AY}\}.  
\end{align}
The structural functions are: 
\begingroup
\begin{align}
\begin{aligned}
 f_A &= u_A\lor(b\oplus u_{AY})\\
 f_B &= u_B\\
 f_Y &= (1-b)\lor((u_{AY}\lor a)\oplus u_Y).
\end{aligned}
\end{align}
\endgroup
The exogenous distributions are:
\begingroup
\begin{align}
\begin{aligned}
 U_A&\sim\Bern(0.33) & U_B&\sim\Bern(0.38)\\
 U_Y&\sim\Bern(0.41) & U_{AY}&\sim\Bern(0.75).
\end{aligned}
\end{align}
\endgroup
Here $\mf{N}^\ast=\emptyset$, $\Delta^1=\{A\}$, and $\mb{Z}^1=\{\{B\}\}$. The constrained POMIS collection is $\mb{P}_{\mc{G},Y}^{\emptyset}=\{\{B\},\{A,B\}\}$, giving six initial actions.

For $do(B=b)$, the remaining c-component $\{A,Y\}$ intersects the discrepancy set at $A$, so \textsc{paTR} supplies no informative bound and retains $[\ell_b,u_b]=[0,1]$. For $do(A=a,B=b)$, the mechanism at $A$ is replaced by intervention. Applying the natural bounds within the source experiment on $B$ gives:
\begin{align}
    \ell_{a,b}&=P^1_b(a,Y=1), &
    u_{a,b}&=P^1_b(a,Y=1)+1-P^1_b(a).
\end{align}
\Cref{tr:tab: worked-one} gives all six intervals. The largest feasible lower bound is $\ell^\star=\ell_{A=1,B=0}=0.8325$, while $u^\star=1$. The two actions $do(A=0,B=1)$ and $do(A=1,B=1)$ satisfy $u_{\mf{x}}<\ell^\star$ and are removed. The four retained actions are $do(B=0)$, $do(B=1)$, $do(A=0,B=0)$, and $do(A=1,B=0)$. For each retained action, $\min\{u_{\mf{x}},u^\star\}=1$, so the upper dominance bound does not further change its interval. 

\paragraph{Clipping bounds through weaker constraints.} Consider the diagram in \Cref{tr:fig: worked-two} and the following SCM. Let $\mf{V}=\{A,B,C,Y\}$. The exogenous variables are: 
\begin{align}
    \mf{U}=\{U_A,U_B,U_C,U_Y,U_{BC},U_{BY}\}.    
\end{align}
The structural functions are: 
\begingroup
\begin{align}
\begin{aligned}
 f_A &= u_A\\
 f_B &= (u_{BY}\land u_{BC})\oplus u_B\\
 f_C &= b\oplus\bigl(u_{BC}\land((1-u_C)\oplus a)\bigr)\\
 f_Y &= c\oplus\bigl(u_{BY}\land((1-u_Y)\oplus a)\bigr).
\end{aligned}
\end{align}
\endgroup
The exogenous distributions are:
\begingroup
\begin{align}
\begin{aligned}
 U_A&\sim\Bern(0.28) & U_B&\sim\Bern(0.42) & U_C&\sim\Bern(0.52)\\
 U_Y&\sim\Bern(0.47) & U_{BC}&\sim\Bern(0.62) & U_{BY}&\sim\Bern(0.49).
\end{aligned}
\end{align}
\endgroup
The target constraint is $\mf{N}^\ast=\{A,C\}$. Source~1 supplies observational data, with $\Delta^1=\{A\}$ and $\mb{Z}^1=\{\emptyset\}$; source~2 supplies experiments on $C$, with $\Delta^2=\{B\}$ and $\mb{Z}^2=\{\{C\}\}$. The initial POMIS collection is $\{\emptyset,\{B\}\}$, giving three target actions. 

For any fixed experimental assignment $c'$, the observational action is exactly transportable:
\begin{align}
    \mu^\ast_{\emptyset}
    &=\sum_{a,b,c,y}yP^2_{c'}(a)P^1_{\emptyset}(b,c,y\mid a)
      \approx0.4844.
\end{align}
For $do(B=b)$, the factor $Q^\ast[C]$ is not identified from the available sources. Its natural bounds combine with the exact factors for $A$ and $Y$:
\begin{align}
    \ell_b&=\sum_{a,c}P^2_{c'}(a)P^1_{\emptyset}(b,c\mid a)P^2_c(Y=1\mid a),\\
    u_b&=\min\biggl\{1,\sum_{a,c}P^2_{c'}(a)P^2_c(Y=1\mid a)\bigl[P^1_{\emptyset}(b,c\mid a)+1-P^1_{\emptyset}(b\mid a)\bigr]\biggr\}.
\end{align}
Thus $[\ell_0,u_0]\approx[0.2097,0.6783]$ and $[\ell_1,u_1]\approx[0.2752,0.8066]$. The lower dominance bound is the exact observational mean, $\ell^\star\approx0.4844$, and every initial action satisfies $u_{\mf{x}}\ge\ell^\star$.

The upper dominance calculation visits the four constraints in \Cref{tr:tab: worked-two-udb}; their relations are shown in \Cref{tr:fig: hierarchy}. At $\mf{N}=\{A\}$, the exact action $do(C=1)$ has mean approximately $0.7485$. However, $do(B=1)$ has the larger upper bound approximately $0.8066$, and its interval does not collapse. Hence the presence of an exact action does not justify early termination. At $\mf{N}=\{C\}$, the largest upper bound is approximately $0.8070$, attained by $do(A=0,B=1)$, whose interval also does not collapse. Both branches reach $\mf{N}=\emptyset$, where:
\begin{align}
    \mu^\ast_{A=1,C=1}
    =P^2_{C=1}(Y=1\mid A=1)=0.7697.
\end{align}
This action has the largest upper bound among the unconstrained POMIS actions, and its lower and upper bounds agree. Therefore the call returns $0.7697$; the other branch reuses this stored result. The final upper dominance bound is:
\begin{align}
    u^\star=\mu^\ast_{A=1,C=1}=0.7697.
\end{align}
All three target actions remain, but the upper limit for $do(B=1)$ decreases from approximately $0.8066$ to $0.7697$, as shown in \Cref{tr:tab: worked-two}. The true best target mean is $\mu^\ast_{B=1}\approx0.5969$; the intervention $do(A=1,C=1)$ is used only to bound the target optimum because $A$ and $C$ cannot be manipulated in the target.

\begin{figure}[t]
    \centering
    \begin{tikzpicture}[x=6.2cm,y=2cm,font=\scriptsize,
        >={Latex[width=2mm,length=2.2mm]},line width=0.4mm,
        trhierarchynode/.style={align=center,inner sep=3mm}]
        \node[trhierarchynode] (AC) at (0,1) {
            (a) $\mf{N}^\ast=\{A,C\}$\\
            $\mb{P}_{\mc{G},Y}^{\mf{N}^\ast}=\{\emptyset,\{B\}\}$};
        \node[trhierarchynode] (A) at (1,1) {
            (b) $\mf{N}=\{A\}$\\
            $\mb{P}_{\mc{G},Y}^{\mf{N}}=\{\emptyset,\{B\},\{C\}\}$};
        \node[trhierarchynode] (C) at (0,0) {
            (c) $\mf{N}=\{C\}$\\
            $\mb{P}_{\mc{G},Y}^{\mf{N}}=\{\{A\},\{A,B\}\}$};
        \node[trhierarchynode] (none) at (1,0) {
            (d) $\mf{N}=\emptyset$\\
            $\mb{P}_{\mc{G},Y}^{\mf{N}}=\{\{A\},\{A,C\}\}$};
        \draw[->] (AC) -- (A);
        \draw[->] (AC) -- (C);
        \draw[->] (A) -- (none);
        \draw[->] (C) -- (none);
    \end{tikzpicture}
    \caption{\small Dominance among POMIS action spaces for the graph in \Cref{tr:fig: worked-two}. Each arrow points toward a weaker manipulability constraint, whose action space has an optimal expected reward at least as large as that of its predecessor. The labels (a)--(d) identify the four constraint settings.}
    \label{tr:fig: hierarchy}
\end{figure}
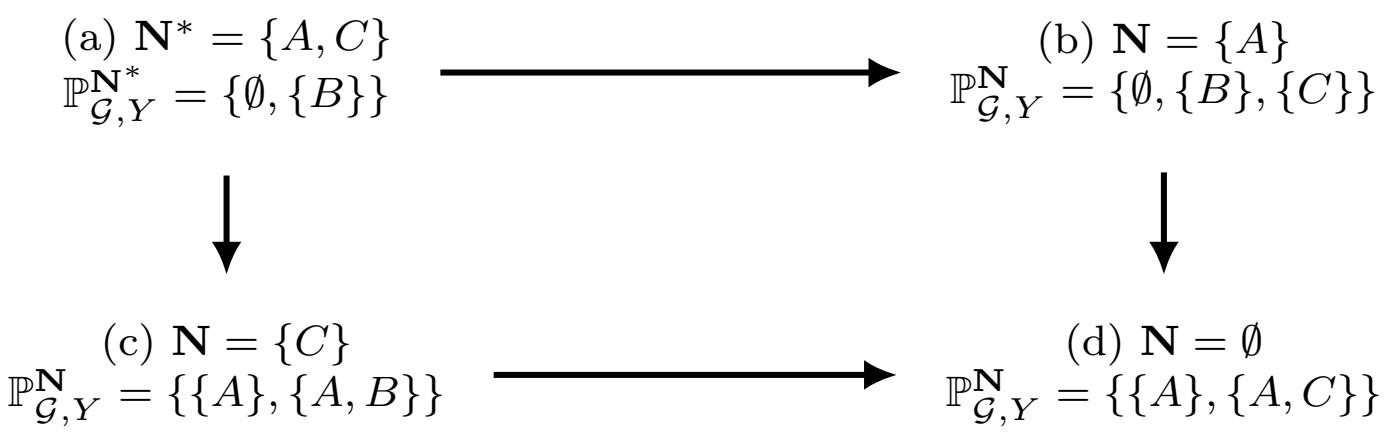
\begin{table}[t]
    \centering
    \caption{\small The three target actions in \Cref{tr:fig: worked-two}. All are retained. The dominance bound reduces the upper limit for $do(B=1)$ to $u^\star=0.7697$.}
    \label{tr:tab: worked-two}
    \small
    \setlength{\tabcolsep}{10pt}
    \begin{tabular}{lrrrr}
        \toprule
        \bf Action & $\ell_{\mf{x}}$ & $\mu^\ast_{\mf{x}}$ & $u_{\mf{x}}$ & $b_{\mf{x}}=\min\{u_{\mf{x}},u^\star\}$ \\
        \midrule
        $do(\emptyset)$ & 0.4844 & 0.4844 & 0.4844 & 0.4844 \\
        $do(B=0)$       & 0.2097 & 0.4031 & 0.6783 & 0.6783 \\
        $do(B=1)$       & 0.2752 & 0.5969 & 0.8066 & 0.7697 \\
        \bottomrule
    \end{tabular}
\end{table}
\begin{table}[t]
    \centering
    \caption{\small The four distinct constraints visited by \textsc{udb} in \Cref{tr:fig: worked-two}. Both one-variable constraints share the stored result for $\mf{N}=\emptyset$. The local upper bound is maximized over the displayed POMIS actions.}
    \label{tr:tab: worked-two-udb}
    \small
    \setlength{\tabcolsep}{7pt}
    \begin{tabular}{clrr}
        \toprule
        $\mf{N}$ & $\mb{P}_{\mc{G},Y}^{\mf{N}}$ & \bf Local upper bound & \bf Returned bound \\
        \midrule
        $\{A,C\}$ & $\{\emptyset,\{B\}\}$ & 0.8066 & 0.7697 \\
        $\{A\}$   & $\{\emptyset,\{B\},\{C\}\}$ & 0.8066 & 0.7697 \\
        $\{C\}$   & $\{\{A\},\{A,B\}\}$ & 0.8070 & 0.7697 \\
        $\emptyset$ & $\{\{A\},\{A,C\}\}$ & 0.7697 & 0.7697 \\
        \bottomrule
    \end{tabular}
\end{table}

\paragraph{An exact optimal value from source experiments.} Consider the diagram in \Cref{tr:fig: worked-three} and the following SCM. Let $\mf{V}=\{R,T,W,X,Z,Y\}$. The exogenous variables are: 
\begin{align}
    \mf{U}=\{U_V:V\in\mf{V}\}\cup\{U_{RW},U_{RY},U_{XY},U_{WX},U_{WZ}\}.
\end{align}
The structural functions are:
\begingroup
\begin{align}
\begin{aligned}
 f_R &= u_R\lor\bigl((1-u_{RW})\land(t\land u_{RY})\bigr)\\
 f_T &= u_T\\
 f_W &= u_W\lor\bigl((1-u_{WX})\land(u_{RW}\land u_{WZ})\bigr)\\
 f_X &= w\lor\bigl((1-u_{WX})\land(u_{XY}\land u_X)\bigr)\\
 f_Z &= x\lor\bigl((1-u_{WZ})\land(r\land u_Z)\bigr)\\
 f_Y &= z\lor\bigl((1-u_{XY})\land(u_{RY}\land u_Y)\bigr).
\end{aligned}
\end{align}
\endgroup
The exogenous distributions are:
\begingroup
\begin{align}
\begin{aligned}
 U_R&\sim\Bern(0.53) & U_T&\sim\Bern(0.63) & U_W&\sim\Bern(0.38)\\
 U_X&\sim\Bern(0.27) & U_Z&\sim\Bern(0.40) & U_Y&\sim\Bern(0.26)\\
 U_{RW}&\sim\Bern(0.52) & U_{RY}&\sim\Bern(0.63) & U_{XY}&\sim\Bern(0.79)\\
 U_{WX}&\sim\Bern(0.74) & U_{WZ}&\sim\Bern(0.31).
\end{aligned}
\end{align}
\endgroup
Let $\mf{N}^\ast=\{T,W\}$, $\Delta^1=\{T\}$, $\Delta^2=\{R\}$, $\mb{Z}^1=\{\emptyset,\{Z\}\}$, and $\mb{Z}^2=\{\{Z\}\}$. The initial POMIS collection is $\{\emptyset,\{R\},\{X\},\{Z\}\}$, giving seven actions.

The observational mean and both interventions on $Z$ are exactly transportable:
\begin{align}
    \mu^\ast_{\emptyset}
    &=\sum_{r,t,w,x,z,y}yP^2_{z'}(t)P^1_{\emptyset}(r,w,x,z,y\mid t)
      \approx0.5514,\\
    \mu^\ast_z&=P^1_z(Y=1)=P^2_z(Y=1).
\end{align}
Here $z'$ is any fixed source assignment. The joint source factor also has the form
$P^1_{\emptyset}(r,w,x,z,y\mid t)=P^1_{\emptyset}(w)P^1_{\emptyset}(r,x,z,y\mid t,w)$, since $T$ and $W$ are independent in the diagram.

\begin{table}[t]
    \centering
    \caption{\small Bounds for the seven initial POMIS actions in \Cref{tr:fig: worked-three}. The exact source value for $do(Z=1)$ gives $\ell^\star=u^\star=1$, so three actions are removed.}
    \label{tr:tab: worked-three}
    \small
    \setlength{\tabcolsep}{10pt}
    \begin{tabular}{lrrrl}
        \toprule
        \bf Action & $\ell_{\mf{x}}$ & $\mu^\ast_{\mf{x}}$ & $u_{\mf{x}}$ & \bf Decision \\
        \midrule
        $do(\emptyset)$ & 0.5514 & 0.5514 & 0.5514 & Remove \\
        $do(R=0)$       & 0.2140 & 0.4553 & 1.0000 & Retain \\
        $do(R=1)$       & 0.4092 & 0.6110 & 1.0000 & Retain \\
        $do(X=0)$       & 0.0713 & 0.1990 & 0.7901 & Remove \\
        $do(X=1)$       & 0.2176 & 1.0000 & 1.0000 & Retain \\
        $do(Z=0)$       & 0.0344 & 0.0344 & 0.0344 & Remove \\
        $do(Z=1)$       & 1.0000 & 1.0000 & 1.0000 & Retain \\
        \bottomrule
    \end{tabular}
\end{table}

For $do(X=x)$, two required factors are identified:
\begin{align}
    Q^\ast[T]&=P^2_{z'}(t), &
    Q^\ast[R,Y]&=P^1_z(r,y\mid t).
\end{align}
The remaining factor $Q^\ast[Z]$ is bounded, for each fixed $w'$, by
\begin{align}
    P^1_{\emptyset}(r,w',x,z\mid t) 
    \leq Q^\ast[Z]
    \leq P^1_{\emptyset}(r,w',x,z\mid t)+1-P^1_{\emptyset}(r,w',x\mid t).
\end{align}
Multiply these bounds by the two exact factors, sum over $(R,T,Z)$ with $Y=1$, and take the largest lower bound and smallest upper bound over $w'\in\{0,1\}$, as enumerated by the Rule~3 loop of \textsc{paTR}. The target reward under $do(X=x)$ is unchanged by additionally fixing $W$; the natural bounds may differ because they use different slices of the source distribution. The resulting intervals appear in \Cref{tr:tab: worked-three}. This calculation illustrates the component-product step of \textsc{paTR}: source~2 supplies $T$, while source~1 supplies the remaining exact factor and a bound on $Z$. 

For $do(R=r)$, the natural-bound step is applied to: 
\begin{align}
    Q^1[R,W,X,Z,Y] = P^1_{\emptyset}(r,w,x,z,y\mid t).
\end{align}
For this query, the Rule~3 loop enumerates $t'\in\{0,1\}$, yielding:
\begin{align}
    \ell_r&=\max_{t'}P^1_{\emptyset}(r,Y=1\mid t'),\\
    u_r&=\min\left\{1,\min_{t'}\left[
        P^1_{\emptyset}(r,Y=1\mid t')
        +8\{1-P^1_{\emptyset}(r\mid t')\}\right]\right\}=1.
\end{align}
The coefficient $8$ comes from the eight assignments to $(W,X,Z)$ in the outer sum. The inner marginal in the pointwise upper bound sums over all of $(W,X,Z,Y)$.

\Cref{tr:tab: worked-three} lists the resulting intervals. The source experiment $do(Z=1)$ identifies a mean of one, so $\ell^\star=u^\star=1$. Among actions tied at the largest upper bound, the tie rule prefers the larger lower bound and selects $do(Z=1)$. Its interval collapses, so \textsc{udb} returns at the initial constraint. The actions $do(\emptyset)$, $do(X=0)$, and $do(Z=0)$ have upper bounds below one and are removed. The four retained actions are $do(R=0)$, $do(R=1)$, $do(X=1)$, and $do(Z=1)$, all with clipped upper limit one. Thus source information identifies the optimal target value even though several other action means remain only partially transportable.

\section{Online Learning with Transferred Bounds}
\label{tr:sec: lucb}

We now use the bounds in a UCB policy. The clipping principle \citep{zhang2017transfer,Gong2023Efficient} limits an optimistic empirical estimate by a valid prior upper bound. \textsc{trUCB} in \Cref{tr:alg: trUCB} first computes the causal and dominance bounds, removes actions with $u_{\mf{x}}<\ell^\star$, and then interacts only through the retained action space $\mc{A}^\ast$.

Its inputs are $\mc{G}$, $Y$, $\mf{N}^\ast$, the source information $(\boldsymbol{\Delta},\mb{Z},\mc{P}_{\mb{Z}}^\Pi)$, and an integer horizon $T\geq1$. It produces the target actions $A_1,\cdots,A_T$ and records their rewards. The causal queries include feasible MIS actions used for $\ell^\star$ and POMIS actions under the weaker constraints visited by \textsc{udb}; their bounds can be computed when first needed and reused. 

\paragraph{Clipped exploration.}
Let $\kappa = \lvert \mc{A}^\ast \rvert$ and suppose that the horizon $T$ is known. The analysis uses the stationary, independent reward draws from \Cref{chap:scb}. For each retained action, set: 
\begin{align}
    b_{\mf{x}}&=\min\{u_{\mf{x}},u^\star\}, &
    \sigma_{\mf{x}}^2
    &=\max_{\mu\in[\ell_{\mf{x}},b_{\mf{x}}]}\mu(1-\mu).
    \label{tr:eq:variance-bound}
\end{align}
Because $Y\in[0,1]$, its variance is no greater than $\mu^\ast_{\mf{x}}(1-\mu^\ast_{\mf{x}})$, and hence no greater than $\sigma_{\mf{x}}^2$. For example, an interval lying near zero or one gives a smaller variance bound than an interval containing $\frac{1}{2}$.

Using the counts $N_t(\mf{x})$ from \Cref{chap:scb}, the empirical target mean after at least one pull is:
\begin{align}
    \widehat{\mu}^\ast_{\mf{x},t}
    =\frac{1}{N_t(\mf{x})}
        \sum_{s=1}^{t}Y_s\,\indicate\{A_s=\mf{x}\}.
\end{align}
Each retained action is initially played once. With $\zeta_T = \log (2\kappa T^2)$, a Bernstein confidence radius gives the index:
\begin{align}
    U_{\mf{x}}(t)
    &=\widehat{\mu}^\ast_{\mf{x},t-1}
      +\sqrt{\frac{2\sigma_{\mf{x}}^2\zeta_T}{N_{t-1}(\mf{x})}}
      +\frac{2\zeta_T}{3N_{t-1}(\mf{x})}, \\
    \bar U_{\mf{x}}(t)
    &=\min\{U_{\mf{x}}(t),b_{\mf{x}}\}.
    \label{tr:eq:ucb}
\end{align}
The square-root term uses the variance information supplied by the bounds. The linear term ensures the confidence guarantee for general bounded rewards. Both the count and the empirical mean use observations through round $t-1$, as required when selecting $A_t$.

\begin{algorithm}[t]
\small
\SetAlgoNoEnd
\DontPrintSemicolon
\tcp{Compute source and structural bounds before interaction.}

Initialize $\mc{A}^\ast=\mc{A}_{\mc{G},Y}^{\mf{N}^\ast}$.\nllabel{tr:line:trucb:actions}\;

Compute $[\ell_{\mf{x}},u_{\mf{x}}]$ for the required actions using \textsc{CausalBound}.\nllabel{tr:line:trucb:causal}\;

Compute $u^\star$ using  \textsc{udb}$(\mc{G},Y,\mf{N}^\ast,\mb{L},\mb{U})$.\nllabel{tr:line:trucb:upper}\;

Set $\ell^\star=\max_{\mf{W}\in\mb{M}_{\mc{G},Y}^{\mf{N}^\ast},\,\mf{w}\in\Omega_{\mf{W}}}\ell_{\mf{w}}$.\nllabel{tr:line:trucb:lower}\;

\BlankLine
\tcp{Remove actions that cannot attain the target optimum.}

Remove every $\mf{x}\in\mc{A}^\ast$ with $u_{\mf{x}}<\ell^\star$.\nllabel{tr:line:trucb:prune}\;

Set $\kappa = \vert \mc{A}^\ast \rvert$ and $\zeta_T=\log(2\kappa T^2)$.\nllabel{tr:line:trucb:confidence}\;

Set $b_{\mf{x}}=\min\{u_{\mf{x}},u^\star\}$ and
$\sigma_{\mf{x}}^2=\max_{\mu\in[\ell_{\mf{x}},b_{\mf{x}}]}\mu(1-\mu)$ for each retained action.\nllabel{tr:line:trucb:variance}\;
\BlankLine
\tcp{Initialize the empirical means using one pull per retained action.}
Play each retained action once in a fixed order, recording its reward and stopping after $T$ total pulls.\nllabel{tr:line:trucb:initialize}\;

\For{\rm each round $t=\kappa+1,\cdots,T$\nllabel{tr:line:trucb:round}}{
    \tcp{Use only observations available before the current decision.}
    
    Set $U_{\mf{x}}(t)=\hat\mu_{\mf{x},t-1}^\ast+\sqrt{\frac{2\sigma_{\mf{x}}^2\zeta_T}{N_{t-1}(\mf{x})}}+\frac{2\zeta_T}{3N_{t-1}(\mf{x})}$ for each $\mf{x}\in\mc{A}^\ast$.\nllabel{tr:line:trucb:index}\;
    
    Set $\bar U_{\mf{x}}(t)=\min\{U_{\mf{x}}(t),b_{\mf{x}}\}$.\nllabel{tr:line:trucb:clip}\;
    Choose $A_t\in\arg\max_{\mf{x}\in\mc{A}^\ast}\bar U_{\mf{x}}(t)$ using a fixed tie-breaking rule.\nllabel{tr:line:trucb:choose}\;
    
    Intervene with $do(A_t)$ and observe $Y_t\sim P^\ast_{A_t}(Y)$.\nllabel{tr:line:trucb:observe}\;
    
    Update the count and empirical reward mean of $A_t$.\nllabel{tr:line:trucb:update}\;
}
\caption{\small \textsc{trUCB}: online learning with transferred bounds.}
\label{tr:alg: trUCB}
\end{algorithm}

Lines~1--5 perform the action reduction and bound construction before interaction. Lines~6--8 set the confidence parameter and variance bounds, then obtain one reward per retained action. At each subsequent round, lines~10--14 form the indices, clip them, select an action maximizing $\bar U_{\mf{x}}(t)$, observe its target reward, and update only that action's empirical mean and count. Directly evaluating the $\kappa$ indices costs $\mc{O}(\kappa)$ per round; updating the selected mean and count costs $\mc{O}(1)$. Since $b_{\mf{x}} \geq \mu^\ast_{\mf{x}}$, clipping preserves optimism whenever the empirical confidence bounds are valid. On the event that all confidence bounds hold, an action whose prior upper bound lies below the optimal mean need not be explored after initialization. 

\paragraph{Regret guarantee.} Assume valid population bounds and $T \geq \kappa$. If the horizon is shorter, initialization stops after $T$ pulls. 

\begin{theorem}\label{tr:thm: regret}
The expected regret satisfies:
\begin{align}
    \mb{E}R_T
    \le \sum_{\mf{x}\in\mc{A}^\ast}\Delta_{\mf{x}}
    +\zeta_T\sum_{\substack{\mf{x}\in\mc{A}^\ast:\\
                 b_{\mf{x}}\ge\mu^\ast_{\mf{x}^\star},\
                 \Delta_{\mf{x}}>0}}
       \left(\frac{32\sigma_{\mf{x}}^2}{\Delta_{\mf{x}}}+\frac83\right)+1.
    \label{tr:eq:regret-bound}
\end{align}
\end{theorem}
The proof is given in \Cref{tr:sec: proof regret}. The three terms have direct interpretations. The first accounts for playing each retained action once. The logarithmic term accounts for suboptimal actions whose upper bounds still permit optimality. The last term accounts for failure of the simultaneous confidence event, whose probability is at most $\frac{1}{T}$. In particular, initialization can incur regret even when no suboptimal action remains in the logarithmic sum.

More explicitly, actions removed before interaction are never selected. For a retained action with $b_{\mf{x}}<\mu^\ast_{\mf{x}^\star}$, the expected number of pulls is at most two. For the other retained suboptimal actions, it is at most $2+\frac{32\sigma_{\mf{x}}^2 \zeta_T}{\Delta_{\mf{x}}^2} +\frac{8\zeta_T}{3\Delta_{\mf{x}}}$.
Thus, transferred bounds can remove actions entirely, prevent repeated exploration, or reduce the variance term for actions that remain competitive. Since $u^\star\ge\mu^\ast_{\mf{x}^\star}$, the condition $b_{\mf{x}}\ge\mu^\ast_{\mf{x}^\star}$ is equivalent to $u_{\mf{x}}\ge\mu^\ast_{\mf{x}^\star}$. The dominance bound nevertheless affects the indices and $\sigma_{\mf{x}}^2$. 

\section{Experiments}
\label{tr:sec:experiments}

We evaluate the bounds and online policy on three binary SCMs. Tasks~1 and~3 use the models in the previous examples with \Cref{tr:fig: worked-one,tr:fig: worked-three}, respectively. Task~2 uses the diagram, source experiments, discrepancy sets, and target constraint of \Cref{tr:fig: worked-two}, with the reward mechanism and exogenous probabilities specified below. 

\paragraph{Task~2: upper dominance and interval variance.}
The agent cannot manipulate $A$ or $C$ in the target environment. Source~1 supplies observational data with $\Delta^1=\{A\}$, and source~2 supplies experiments on $C$ with $\Delta^2=\{B\}$. All exogenous variables are mutually independent. The SCM is specified as follows: 
Let $\mf{V}=\{A,B,C,Y\}$. The exogenous variables are:
\begin{align}
    \mf{U}=\{U_A,U_B,U_C,U_Y,U_{BC},U_{BY},U_G\}.
\end{align}
The structural functions are:
\begingroup
\begin{align}
\begin{aligned}
 f_A &= u_A\\
 f_B &= (u_{BY}\land u_{BC})\oplus u_B\\
 f_C &= b\oplus\bigl(u_{BC}\land((1-u_C)\oplus a)\bigr)\\
 f_Y &= u_G\land\bigl[(c\land(1-u_Y))\oplus(a\land u_{BY})\bigr].
\end{aligned}
\end{align}
\endgroup
The exogenous distributions are:
\begingroup
\begin{align}
\begin{aligned}
 U_A&\sim\Bern(0.8) & U_B&\sim\Bern(0.9) & U_C&\sim\Bern(0.6)\\
 U_Y&\sim\Bern(0.6) & U_{BC}&\sim\Bern(0.1) & U_{BY}&\sim\Bern(0.4)\\
 U_G&\sim\Bern(0.2).
\end{aligned}
\end{align}
\endgroup
The additional private noise $U_G$ enters only $f_Y$, so the causal diagram is unchanged. The three feasible actions are observation, $do(B=0)$, and $do(B=1)$, with $do(B=1)$ attaining the largest mean. Their source-derived intervals appear in \Cref{tab:scbtr-task2-bounds}.

\begin{table}[t]
    \centering
    \caption{\small Source bounds and target means for experimental Task~2. All three actions are retained, and $u^\star=0.096$. Values are rounded to four decimal places.}
    \label{tab:scbtr-task2-bounds}
    \small
    \setlength{\tabcolsep}{10pt}
    \begin{tabular}{lrrrr}
        \toprule
        \bf Action & $\ell_{\mf{x}}$ & $\mu^\ast_{\mf{x}}$ & $u_{\mf{x}}$ & $b_{\mf{x}}$ \\
        \midrule
        $do(\emptyset)$ & 0.0883 & 0.0883 & 0.0883 & 0.0883 \\
        $do(B=0)$ & 0.0090 & 0.0654 & 0.1451 & 0.0960 \\
        $do(B=1)$ & 0.0797 & 0.0914 & 0.1004 & 0.0960 \\
        \bottomrule
    \end{tabular}
\end{table}

For the hypothetical intervention $do(A=a,C=c)$, source~2 identifies:
\begin{align}
    \mu^\ast_{A=a,C=c}
    =P^2_{C=c}(Y=1\mid A=a)
    =0.2\bigl(0.4c+0.4a-0.32ac\bigr).
\end{align}
These means are $0$, $0.08$, $0.08$, and $0.096$ for $(a,c)=(0,0),(0,1),(1,0),(1,1)$, respectively. The other unconstrained POMIS, $\{A\}$, gives means approximately $0.0689$ and $0.0931$. The largest local upper bounds under the other constraints all exceed $0.096$. Thus \textsc{udb} returns $u^\star=0.096$, reducing the upper limits of both interventions on $B$ to this value. All three actions remain after pruning. Since their intervals lie below $\frac{1}{2}$, the reduced upper limits also tighten the Bernoulli variance bounds in \Cref{tr:eq:variance-bound}: from approximately $0.1241$ to $0.0868$ for $do(B=0)$ and from $0.0903$ to $0.0868$ for $do(B=1)$. The larger reduction applies to the suboptimal action.

\begin{table}[t]
    \centering
    \caption{\small Candidate action spaces. Each strategy column gives the number of intervention sets / arms. The retained action space is used by \textsc{trUCB}.}
    \label{tab:scbtr-experiment-design}
    \small
    \setlength{\tabcolsep}{9pt}
    \begin{tabular}{crrrrr}
        \toprule
        \bf Task & $\lvert\mf{V}\rvert$ & $T$ & \bf UCB & \bf \textsc{poUCB} & \bf \textsc{trUCB} \\
        \midrule
        1 & 3 & 10k & 4 / 9 & 2 / 6 & 2 / 4 \\
        2 & 4 & 10k & 2 / 3 & 2 / 3 & 2 / 3 \\
        3 & 6 & 10k & 8 / 27 & 4 / 7 & 3 / 4 \\
        \bottomrule
    \end{tabular}
\end{table}

\paragraph{Source information and target interaction.}
The agent receives the causal diagram, the target constraint, and the permitted population source distributions. The source and target SCMs share the displayed numerical parameters in these tasks; the discrepancy sets specify which mechanisms may differ without requiring an actual change. Thus, these experiments study the use of source information under the asserted invariances. They do not assess uncertainty from finite source datasets. \Cref{tab:scbtr-experiment-design} reports the numbers of intervention sets and arms considered by the three main strategies in each task.

At every target round, the agent observes a fresh reward from its chosen intervention. We compute each target mean by summing over the SCM's exogenous assignments and generate an independent Bernoulli reward with that mean. This gives the same reward distribution as simulating the full binary SCM. Exact target means are used only for reward generation and evaluation; the policy uses its permitted prior bounds and previously observed rewards. All methods are evaluated against the same optimum over target-feasible interventions: $1$, approximately $0.0914$, and $1$ in Tasks~1--3. In particular, hypothetical interventions used by \textsc{udb} do not enlarge the agent's feasible actions or change its regret benchmark.

\paragraph{Compared policies.}
We first compare \texttt{UCB}, \texttt{poUCB}, and \texttt{trUCB}. The first uses all assignments to all manipulable subsets, including the empty intervention. The second restricts these assignments to constrained POMISs. After playing each candidate action once, both use the standard UCB1 index \citep{auer2002finite}, $\widehat\mu^\ast_{\mf{x},t-1} + \sqrt{\frac{2 \log t}{N_{t-1}(\mf{x})}}$. The third uses \Cref{tr:alg: trUCB}, including the Bernstein index in \Cref{tr:eq:ucb}. These main comparisons therefore combine differences in action selection and confidence rules.

To distinguish the effects of the transferred bounds, we also compare four policies using the same Bernstein rule as \texttt{trUCB}. \texttt{Bernstein-POMIS} uses all POMIS actions with upper limit one and variance bound $\frac{1}{4}$. \texttt{Pruning only} removes actions with $u_{\mf{x}}<\ell^\star$ but otherwise keeps these trivial bounds. \texttt{Pruning + clipping} additionally uses $b_{\mf{x}}=\min\{u_{\mf{x}},u^\star\}$, while retaining variance bound $\frac{1}{4}$. Finally, \texttt{trUCB without UDB} uses the causal upper bound $u_{\mf{x}}$ in both clipping and the variance interval, omitting only the upper dominance bound. Each policy sets $\kappa$ to its number of retained actions when computing $\zeta_T$.

\begin{figure}[t]
    \centering
    \captionsetup[subfigure]{skip=2pt}
    {\color{black}\includegraphics[width=\linewidth]{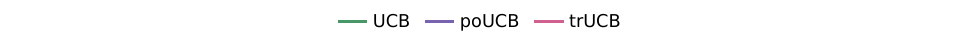}}
    \begin{subfigure}[b]{0.350\linewidth}
        \centering
        {\color{black}\includegraphics[width=\linewidth]{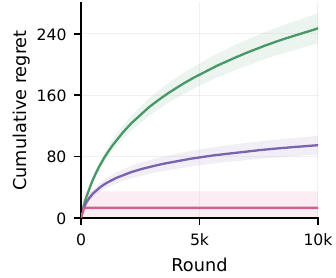}}
        \caption{Task~1}
        \label{fig:scbtr-experiment-main-task1}
    \end{subfigure}%
    \begin{subfigure}[b]{0.325\linewidth}
        \centering
        {\color{black}\includegraphics[width=\linewidth]{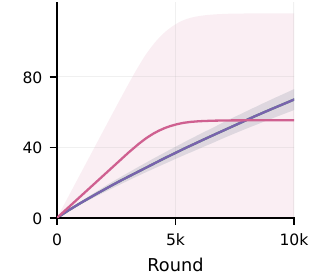}}
        \caption{Task~2}
        \label{fig:scbtr-experiment-main-task2}
    \end{subfigure}%
    \begin{subfigure}[b]{0.325\linewidth}
        \centering
        {\color{black}\includegraphics[width=\linewidth]{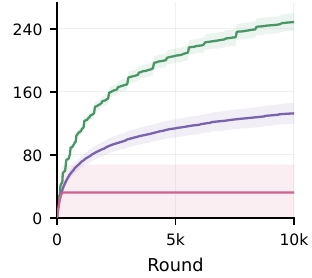}}
        \caption{Task~3}
        \label{fig:scbtr-experiment-main-task3}
    \end{subfigure}
    \par\vspace{5pt}
    {\color{black}\includegraphics[width=\linewidth]{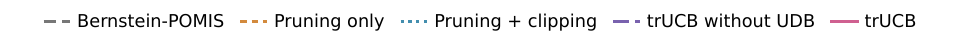}}
    \begin{subfigure}[b]{0.350\linewidth}
        \centering
        {\color{black}\includegraphics[width=\linewidth]{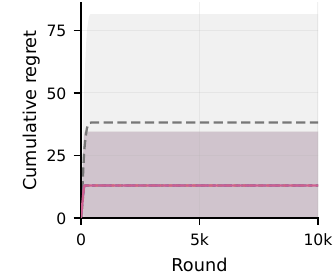}}
        \caption{Task~1}
        \label{fig:scbtr-experiment-ablation-task1}
    \end{subfigure}%
    \begin{subfigure}[b]{0.325\linewidth}
        \centering
        {\color{black}\includegraphics[width=\linewidth]{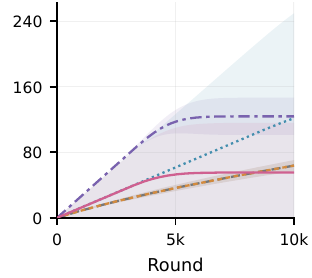}}
        \caption{Task~2}
        \label{fig:scbtr-experiment-ablation-task2}
    \end{subfigure}%
    \begin{subfigure}[b]{0.325\linewidth}
        \centering
        {\color{black}\includegraphics[width=\linewidth]{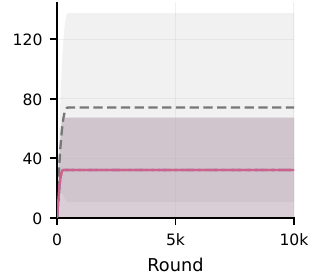}}
        \caption{Task~3}
        \label{fig:scbtr-experiment-ablation-task3}
    \end{subfigure}
        \caption{\small Mean cumulative pseudo-regret over $10{,}000$ rounds and 1,000 independent runs; shaded regions show one standard deviation. The upper row compares UCB and \textsc{poUCB} (standard UCB indices) with \textsc{trUCB} (Bernstein indices); the lower row compares controlled Bernstein-index ablations. UCB and \textsc{poUCB} overlap in Task~2. The curves for \textsc{trUCB} and \textsc{trUCB} without UDB overlap in Tasks~1 and~3.}
    \label{fig:scbtr-experiment-main}
\end{figure}

\paragraph{Evaluation.}
Every task--policy configuration is run for $T=10{,}000$ rounds and repeated $1{,}000$ times. Before interaction, each repetition draws an action order independently of the reward draws and keeps it fixed for initialization and index ties. All policies in that repetition use the same order restricted to their candidate actions and share the same sequence of independent uniform draws for generating rewards. The random sequences used for this evaluation were not used to choose the parameters of Task~2. We record cumulative pseudo-regret. Curves in \Cref{fig:scbtr-experiment-main} show its mean, and shaded regions show one population standard deviation across repetitions, matching the presentation in \Cref{sec: scbmec experiment}. \Cref{tab:scbtr-experiment-regret} reports the final mean and standard deviation.

\begin{table}[t]
    \centering
        \caption{\small Final cumulative pseudo-regret at $T=10{,}000$ (mean $\pm$ standard deviation across 1,000 independent runs). The lower block gives Bernstein-index ablations. Bold entries mark the lowest mean for each task.}
    \label{tab:scbtr-experiment-regret}
    \small
    \setlength{\tabcolsep}{10pt}
    \begin{tabular}{lrrr}
        \toprule
        \bf Policy & \bf Task 1 & \bf Task 2 & \bf Task 3 \\
        \midrule
        UCB & \regret{247.1}{19.6} & \regret{67.1}{5.9} & \regret{248.6}{11.3} \\
        \textsc{poUCB} & \regret{94.8}{12.1} & \regret{67.1}{5.9} & \regret{132.7}{13.2} \\
        \textsc{trUCB} & \regret{\boldsymbol{13.0}}{21.6} & \regret{\boldsymbol{55.4}}{60.7} & \regret{\boldsymbol{32.2}}{35.2} \\
        \midrule
        Bernstein-POMIS & \regret{38.1}{43.5} & \regret{63.8}{6.7} & \regret{74.1}{63.4} \\
        Pruning only & \regret{\boldsymbol{13.0}}{21.6} & \regret{63.8}{6.7} & \regret{\boldsymbol{32.2}}{35.2} \\
        Pruning + clipping & \regret{\boldsymbol{13.0}}{21.6} & \regret{121.7}{128.1} & \regret{\boldsymbol{32.2}}{35.2} \\
        \textsc{trUCB} without UDB & \regret{\boldsymbol{13.0}}{21.6} & \regret{123.9}{22.7} & \regret{\boldsymbol{32.2}}{35.2} \\
        \bottomrule
    \end{tabular}
\end{table}

\paragraph{Performance.}
In Task~1, POMIS reduction decreases the number of actions from nine to six, and the source bounds remove two more. At $T=10{,}000$, the mean regret decreases from $247.1$ for \texttt{UCB} to $94.8$ for \texttt{poUCB} and $13.0$ for \texttt{trUCB}. The controlled comparison isolates the additional effect of pruning: \texttt{Bernstein-POMIS} incurs mean regret $38.1$, whereas \texttt{Pruning only} incurs $13.0$. Since every retained upper limit is one, clipping does not further tighten these limits. 

In Task~2, all methods retain the same three actions. Thus \texttt{UCB} and \texttt{poUCB} coincide, with mean regret $67.1$, and pruning alone does not change the Bernstein policy, whose mean regret is $63.8$. Full \texttt{trUCB} gives $55.4$, compared with $123.9$ without upper dominance. However, \texttt{Pruning + clipping}, which fixes the variance bound at $\frac{1}{4}$, gives $121.7$. A common clipping level can produce index ties, so tightening a cap alone need not improve finite-horizon regret. The full policy also uses the smaller variance bounds supplied by the intervals. The additional comparison below isolates how upper dominance contributes through these two uses.

In Task~3, the action counts decrease from $27$ to $7$ after POMIS reduction and to $4$ after pruning. The corresponding main-policy regrets are $248.6$, $132.7$, and $32.2$. With the Bernstein rule held fixed, pruning reduces mean regret from $74.1$ to $32.2$. The source experiment identifies an optimal mean of one, but several retained actions still have uncertain means. Consequently, initialization and subsequent exploration can incur regret even though the optimal value is known.

In Tasks~1 and~3, \texttt{trUCB without UDB} and \texttt{Pruning + clipping} select exactly the same actions as full \texttt{trUCB} under the shared random draws. Task~2 instead exhibits an additional gain from upper dominance and interval variance information. Together, the tasks distinguish action reduction from changes to the exploration of retained actions.

\begin{figure}[t]
    \centering
    {\color{black}\includegraphics[width=0.4\linewidth]{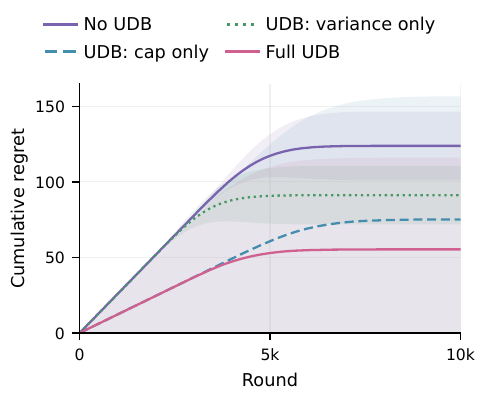}}
    \caption{\small Separating the two uses of upper dominance in Task~2. All policies use the same three actions and Bernstein indices. \texttt{UDB cap only} adjusts the clipping level, \texttt{UDB variance only} adjusts the interval variance bound, and \texttt{Full UDB} applies both adjustments. Curves show mean cumulative pseudo-regret over 1,000 repetitions; shaded regions show one standard deviation.}
    \label{fig:scbtr-task2-udb-isolation}
\end{figure}

\paragraph{Separating the two uses of upper dominance.}
For Task~2, \Cref{fig:scbtr-task2-udb-isolation} further separates the effect of $u^\star$ on clipping from its effect on the variance interval. All four policies retain the same three actions and use the same Bernstein rule, initialization, reward draws, and action order. Starting from \texttt{trUCB without UDB}, \texttt{UDB cap only} replaces $u_{\mf{x}}$ by $b_{\mf{x}}$ only in the clipping step, while retaining the variance bound computed from $[\ell_{\mf{x}},u_{\mf{x}}]$. Conversely, \texttt{UDB variance only} uses the variance bound computed from $[\ell_{\mf{x}},b_{\mf{x}}]$ but clips at $u_{\mf{x}}$. Full \texttt{trUCB} applies both changes. The cap-only comparison therefore differs from \texttt{Pruning + clipping}, which fixes the variance bound at $\frac{1}{4}$.

The mean regrets are $123.9$ without upper dominance, $75.2$ with the cap adjustment alone, $91.2$ with the variance adjustment alone, and $55.4$ with both. Full \texttt{trUCB} reduces mean regret by $55.3\%$ relative to \texttt{trUCB without UDB}. The mean paired reduction is $68.5$, with a $95\%$ confidence interval of $[64.8,72.3]$ across repetitions. Both uses of upper dominance therefore contribute to the improvement. The common cap removes the larger upper limit initially assigned to the suboptimal action, and the tighter variance bound further reduces its exploration. These effects do not require removing an action.

The improvement also persists when the horizon is changed: the mean regrets without and with upper dominance are $51.9$ and $24.7$ at $T=2{,}000$, and $145.4$ and $64.6$ at $T=50{,}000$. Each horizon uses its own $\zeta_T$ and $1{,}000$ repetitions. In a further sensitivity check, changing each of the seven exogenous success probabilities by $\pm0.05$, one at a time, gives $14$ settings. Using $256$ new repetitions per setting at $T=10{,}000$, the mean reductions range from $40.0\%$ to $58.2\%$. These results illustrate a setting in which upper dominance improves learning; they do not imply an improvement for every SCM or every reward sequence.

\jmlrsection{Counterfactual Structural Causal Bandits}\label{chap:ctf scb}
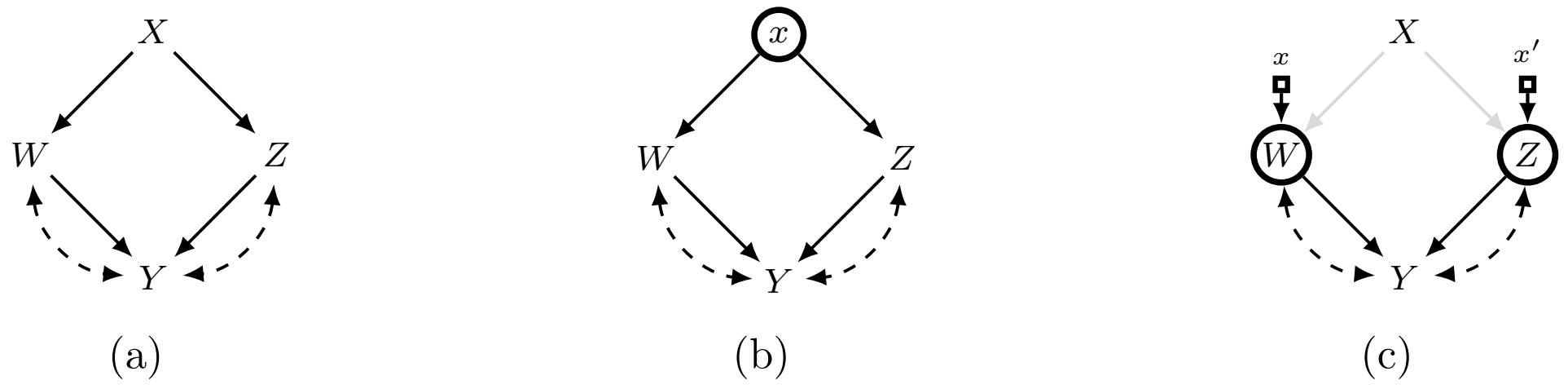
\begin{figure}[t]
    \begin{minipage}[b]{0.3\textwidth}\centering
    \subfloat[]{
        \begin{tikzpicture}[x=1cm,y=1cm,>={Latex[width=1.4mm,length=1.7mm]},
        line width=0.25mm,
        R/.style={draw = none,circle,inner sep=0mm, minimum size=4.5mm}]
        \node[R] (A) at (1,2) {$X$};
        \node[R] (B) at (0,1) {$W$};
        \node[R] (C) at (2,1) {$Z$};
        \node[R] (Y) at (1,0) {$Y$};
        \draw[->] (A) -- (B);
        \draw[->] (A) -- (C);
        \draw[->] (B) -- (Y);
        \draw[->] (C) -- (Y);

        \draw[dashed, <->] (C) to[bend left = 40] (Y);
        \draw[dashed, <->] (B) to[bend right = 40] (Y);
        \end{tikzpicture}\label{fig:ctfscb_introduction_example_a}}
    \end{minipage}
    \hfill
    \begin{minipage}[b]{0.3\textwidth}\centering
    \subfloat[]{
        \begin{tikzpicture}[x=1cm,y=1cm,>={Latex[width=1.4mm,length=1.7mm]},
        line width=0.25mm,
        R/.style={draw = none,circle,inner sep=0mm, minimum size=4mm}]

        \node[R,draw,line width=0.5mm] (A) at (1,2) {$x$};
        \node[R] (B) at (0,1) {$W$};
        \node[R] (C) at (2,1) {$Z$};
        \node[R] (Y) at (1,0) {$Y$};
        \draw[->] (A) -- (B);
        \draw[->] (A) -- (C);
        \draw[->] (B) -- (Y);
        \draw[->] (C) -- (Y);

        \draw[dashed, <->] (C) to[bend left = 40] (Y);
        \draw[dashed, <->] (B) to[bend right = 40] (Y);
        \end{tikzpicture}\label{fig:ctfscb_introduction_example_b}}
    \end{minipage}
    \hfill
    \begin{minipage}[b]{0.3\textwidth}\centering
    \subfloat[]{
        \begin{tikzpicture}[x=1cm,y=1cm,>={Latex[width=1.4mm,length=1.7mm]},
        line width=0.25mm,
        transp/.style={regime, minimum size=1mm, draw = black,inner sep=0mm},
        regime line/.style={draw=black, thick},
        R/.style={draw = none,circle,inner sep=0mm, minimum size=4.5mm}]
        \node[R] (A) at (1,2) {$X$};
        \node[R, draw, line width=0.5mm] (B) at (0,1) {$W$};
        \node[R, draw, line width=0.5mm] (C) at (2,1) {$Z$};
        \node[R] (Y) at (1,0) {$Y$};
        \draw[->, gray!30] (A) -- (B);
        \draw[->, gray!30] (A) -- (C);
        \draw[->] (B) -- (Y);
        \draw[->] (C) -- (Y);

        \draw[dashed, <->] (C) to[bend left = 40] (Y);
        \draw[dashed, <->] (B) to[bend right = 40] (Y);

        \node[transp, above = 2.5mm of B] (BT){};
        \node[above = -0.5mm of BT,font=\sffamily\tiny] (a1){$x$};
        \draw[->,regime line] (BT) -- (B);

        \node[transp, above = 2.5mm of C] (CT){};
        \node[above = -0.5mm of CT, yshift=0.2mm,font=\sffamily\tiny] (a2){$x'$};
        \draw[->,regime line] (CT) -- (C);
        \end{tikzpicture}\label{fig:ctfscb_introduction_example_c}}
    \end{minipage}
    \hfill\null
    \caption{\small (a) A structural causal bandit environment. (b) The submodel induced by $do(x)$. (c) The submodel induced by the counterfactual action $do(W = W_x, Z= Z_{x'})$. }
    \label{fig:ctfscb_introduction_example}
\end{figure}

The preceding \unitnames{} use causal structure to reduce the search space of a decision-making agent. This \unitname{} extends the available interventions themselves. 

\begin{example}[\bf AI-assisted clinical decision support system]
    Consider a hospital aiming to improve the quality of patient care. Let $Y$ denote the treatment outcome, $X$ a patient's report of their health status, $W$ an expert physician's assessment, and $Z$ the score from an AI-assisted clinical tool. \Cref{fig:ctfscb_introduction_example_a} illustrates the setting, where bidirected edges represent unobserved confounders such as historical bias \citep{peiffer2020machine}. 
    
    For fixed report values $x$ and $x'$, a standardized symptom-reporting interface can supply $x$ to the expert assessment and $x'$ to the AI tool. Supplying $x$ to both assessments gives the same expected outcome as $do(X=x)$, namely $\mathbb{E}Y_x$ (\Cref{fig:ctfscb_introduction_example_b}). Alternatively, $x$ can describe a current health condition, while $x'$ describes a prospective condition, such as the anticipated health benefits of beginning regular exercise (\Cref{fig:ctfscb_introduction_example_c}). These report values are selected for the experiment before the patient's natural report is observed. Supplying the two values separately gives $\mathbb{E}Y_{W_x, Z_{x'}}$, allowing one to ask counterfactual-level questions: 
    \begin{center}
       \bf 
       \textit{``What if the AI assistant had perceived the patient's prospective health status $x'$ rather than the current one $x$?''}
    \end{center}
    and to design decision strategies that improve patient outcomes. Here $W_x$ is the physician's assessment under report $x$, and $Z_{x'}$ is the AI score under report $x'$. Both concern the same patient, so their evaluation retains the same unobserved characteristics. The reporting interface changes the information received by each assessment; it does not itself change the patient's physical health. 
\end{example}

The example raises two questions.
\begin{itemize}
    \item First, which counterfactuals can actually be produced through interaction with the system? 
    
    \item Second, among the available actions, which can be removed without losing an optimal decision? 
    
\end{itemize}
The first question requires a notion of realizability. The second extends the equivalence and partial-order arguments developed for ordinary interventions in \Cref{chap:scb}.

We extend the SCB framework to accommodate a broader class of decision-making scenarios. Our key contributions in this \unitname{} are as follows:   
\begin{itemize}
    \item We first formulate the counterfactual structural causal bandit problem, enabling an agent to consider actions beyond $\mc{L}_2$-level interventions. 
    
    \item We investigate equivalence and partial-order relations among these actions and develop graphical characterizations for reducing the action space. 
    
    \item Building on these results, we present an algorithm that removes redundant or weakly dominated choices while retaining an optimal action. 
\end{itemize}

\paragraph{Relation to the published version.}
This \unitname{} revises \citet{park2026counterfactual}. The counterfactual regime graph in Section~3.2 of \citet{park2026counterfactual} is corrected by deleting directed edges whose inputs are fixed to constants, retaining bidirected edges, and then restricting to reward ancestors (\Cref{thm: ctf-pomis}). The sufficiency proof of Theorem~2 of \citet{park2026counterfactual} is replaced by a construction using reserved values and propagation of invalid inputs (\Cref{sec: proof of thm: ctf-pomis}). The illustrative SCM in Section~3.2 of \citet{park2026counterfactual} is also corrected: $f_X$ is set to $u_X \lor z$, and $f_Z$ is specified as $u_Z \lor (w \oplus u_{ZY})$. Finally, \Cref{sec: ctfscb experiment} uses distinct realizable parent-input assignments, corrected \ctf-POMIS arm counts, and a \ctf-MIS baseline that includes nested actions. Its action counts and regret results replace those in Tables~1 and~2 of \citet{park2026counterfactual}. 

\section{Background: Counterfactuals and Realizability}

\subsection{Counterfactual Distributions}

We use the SCM, submodel, and PCH notation of \Cref{chap:logical foundation}. The counterfactual $X_{\mf{w}}(\mf{u})$ evaluates $X$ in the submodel $\mc{M}_{\mf{w}}$. Its random counterpart is $X_{\mf{w}} = X_{\mf{w}}(\mf{U})$. Layer~3 of the PCH permits joint events involving counterfactuals from several such submodels. All counterfactuals are evaluated at the \textbf{same} exogenous realization \citep{bareinboim2022pearl}. 
\begin{definition}[Counterfactual distribution \citep{bareinboim2022pearl}]
    An SCM $\mc{M}$ induces joint distributions over counterfactual events by:
    \begin{align}
        P(\mf{y}_\mf{x}, \cdots, \mf{z}_{\mf{w}}) = \sum_{\mf{u}} \indicate\{\mf{Y}_{\mf{x}}(\mf{u}) = \mf{y}, \cdots, \mf{Z}_{\mf{w}}(\mf{u}) = \mf{z}\}P(\mf{u}).
    \end{align}
\end{definition}
Shared exogenous inputs preserve the dependence between counterfactuals, including dependence induced by unobserved confounders. In the clinical example, $W_x$ and $Z_{x'}$ describe two assessments of one patient under different reports. Separately sampling patients for the two assessments generally produces a different joint distribution. A fully specified SCM determines the joint counterfactual distribution, but its semantic definition alone supplies neither an identification formula from available data nor an experiment that measures the counterfactuals. 

We write a collection of counterfactuals as $\mf{X}_\ast = \{X_{i[\mf{w}_i]}\}_{i=1}^k$ and its targets as $\mf{X} = \mf{V}(\mf{X}_\ast) = \{X_i\}_{i=1}^k$. The brackets separate the variable index $i$ from its subscript assignment, and the symbol $\ast$ indexes the collection of counterfactual subscripts. Targets are distinct and exclude $Y$; every specified counterfactual has a nonempty subscript set. For example, $\mf{X}_\ast = \{W_x,Z_{x'}\}$ has target set $\mf{X} = \{W,Z\}$. Different terms may use different values, including $x \neq x'$. Within a single counterfactual $Z_w$, however, the same assignment $w$ is used throughout its defining submodel. 

\paragraph{Nested counterfactuals.} Counterfactuals can themselves supply the values of an intervention. For instance, $Y_{\mf{Z}_\mf{x}} (\mf{u})$, equivalently $Y_{\mf{Z} = \mf{Z}_\mf{x}(\mf{u})}(\mf{u})$, first computes $\mf{Z}_{\mf{x}}(\mf{u})$ and then substitutes those values for $\mf{Z}$ in the mechanism for the outer counterfactual \citep{correa2021nested}. The inner and outer counterfactuals use the same $\mf{u}$ even though they may be evaluated under different interventions. More generally, $Y_{\mf{X}_\ast}(\mf{u})$ evaluates the specified counterfactuals and substitutes their values for the corresponding targets.

We write $\mf{X}_{\ast\ast}$ when the assigned counterfactuals may themselves have counterfactuals in their subscripts. The double asterisk allows arbitrary nesting and
includes $\mf{X}_\ast$ as a special case. This distinction concerns the assigned counterfactuals; the reward $Y_{\mf{X}_\ast}$ already involves a counterfactual substitution.

\paragraph{Counterfactual calculus.} The \textit{\ctf-calculus} provides rules for manipulating counterfactual probability quantities \citep{correa2025counterfactual}. We recall the consistency and exclusion rules used below. 

\textbf{CTF-Rule~1~(Consistency).~} Once a counterfactual already takes the value $\mf{x}$, explicitly setting it to that same value does not change the remaining counterfactuals:
\begin{align}
    P(\mf{y}_{\mf{T}_{\ast\ast},\mf{x}}, \mf{x}_{\mf{T}_{\ast\ast}}, \mf{w}_\ast) = P(\mf{y}_{\mf{T}_{\ast\ast}}, \mf{x}_{\mf{T}_{\ast\ast}}, \mf{w}_\ast)
\end{align}
This rule allows a computed counterfactual to replace an explicit assignment when their values agree.

The extension from $\mf{T}_{\ast}$ in the original rule to $\mf{T}_{\ast\ast}$ follows from unit-level composition: at a fixed exogenous realization $\mf{u}$, evaluate the nested assignments recursively, apply consistency to the resulting values, and then average over $\mf{u}$. All nested counterfactuals use that same exogenous realization.

\textbf{CTF-Rule~3~(Exclusion).~} An assignment can be removed when its variables cannot affect the counterfactual after the other interventions:
\begin{align}
    P(\mf{y}_{\mf{x}\mf{z}}, \mf{w}_\ast) = P(\mf{y}_{\mf{z}}, \mf{w}_\ast), \qquad \text{~if~} \mf{X} \cap \An(\mf{Y})_{\mc{G}_{\overline{\mf{Z}}}} = \emptyset.
\end{align}
The ancestor condition is checked after cutting the incoming edges of $\mf{Z}$. In the clinical example, fixing both $W$ and $Z$ blocks every directed path from $X$ to $Y$, so an additional assignment to $X$ cannot change $Y_{wz}$. Consistency and exclusion will be used to identify counterfactual assignments that do not change the reward. 

\paragraph{Counterfactual unnesting theorem.} Nested distributions can also be expanded into joint events involving counterfactuals with less nesting. This is useful because each term then states explicitly which values must occur together on the same unit. 
\begin{lemma}[Counterfactual Unnesting Theorem (CUT); Theorem~4 in \citet{correa2021nested}]\label{cut}
   For a collection of counterfactuals $\mf{T}_{\ast\ast}$: 
    \begin{align}
        P(Y_{\mf{T}_{\ast\ast}, X_{\mf{z}}} = y) = \sum_{x} P(Y_{\mf{T}_{\ast\ast},{x}} = y, X_{\mf{z}} = {x}).
    \end{align}  
\end{lemma}

For example, applying CUT to the two assessments gives:
\begin{align}
    P(Y_{W_x,Z_{x'}} = y) = \sum_{wz} P(Y_{wz} = y, W_x = w,Z_{x'} = z). 
\end{align}
For each possible pair $w,z$, the event requires both assessments to produce that pair and the outcome under those values to equal $y$. Summing over the pairs recovers the nested distribution. These counterfactuals cannot generally be sampled independently because they retain the same exogenous realization. CUT is an identity between distributions; the expansion alone does not supply an experiment for the joint events or a single unnested action equivalent to the nested regime. 

\subsection{Realizability}
Throughout this \unitname{}, endogenous variables have finite discrete domains, as in \citet{raghavan2025counterfactual}. These domains are fixed within each bandit instance but may vary across the compatible SCMs used in a graphical characterization, as specified in \Cref{chap:scb}. The exogenous variables may have general distributions. 

\paragraph{Experimental capabilities.} 
Layer~1 distributions can be sampled through passive observation of the system's natural behavior. Layer~2 distributions require interventions, such as Fisherian randomization \citep{fisher1935design}. Counterfactual randomization further permits changing the value transmitted to selected child mechanisms while allowing other children to receive the original value \citep{raghavan2025counterfactual}. 

\begin{definition}[Feasible operations]\label{def: feasible operation}
    An operation collection $\mb{A}$ specifies which of the following physical operations are available:
    {
    \begin{itemize}
        \item[1.] \textsc{Select}: select a unit from the target population. 
        
        \item[2.] \textsc{Read}$(V)$: observe the output of the mechanism generating $V$ on that unit. 
        
        \item[3.] \textsc{Rand}$(X)$: replace the natural mechanism of $X$ by an externally randomized value whose support is $\Omega_{X}$. 
        
        \item[4.] \textsc{ctf-Rand}$(X \to \mf{Z})$: randomize the value of $X$ received by selected directed children $\mf{Z} \subseteq \Ch(X)_{\mc{G}}$, using suitable counterfactual mediators.
    \end{itemize}     
    }   
\end{definition}
A \textit{counterfactual mediator} for $X \to Z$ is a manipulable interface through which $f_Z$ receives the information about $X$. Its ordinary states must encode $X$ so that the input received by $f_Z$ agrees with the original mechanism; changing the interface reproduces the intended alternative input. The reporting interface plays this role in the introductory example. Availability of such an interface is an experimental capability assumption: observing the edge $X \to Z$ alone does not establish it. The formal mediator conditions are given by \citet{raghavan2025counterfactual}.

\paragraph{Realizability.}
Counterfactual semantics specify a counterfactual even when no available experiment can measure it. Realizability asks whether physical operations can produce samples from the required distribution, including the dependence among counterfactuals on the same unit. 
\begin{definition}[Realizability]
    A counterfactual distribution is realizable from $\mb{A}$ if operations in $\mb{A}$ produce samples from it for every SCM conforming to $\mc{G}$ and the specified experimental capabilities. 
\end{definition}
Throughout this \unitname{}, we use the maximal collection $\mb{A}^\dagger$, which permits the operations above at every relevant variable and parent-child interface. Each external randomizer assigns positive probability to every value in the corresponding finite domain. The randomizers for separately selected inputs are independent of one another and of the unit's exogenous variables. Assigned inputs can therefore be obtained by retaining units whose randomized inputs match the selected assignment. Retention depends only on these randomized inputs. As in \Cref{chap:scb}, each bandit round uses a fresh independent retained unit. Discarded randomization attempts are outside the round count. Even with these capabilities, a mechanism cannot be evaluated twice under conflicting inputs on the same unit. Different reports sent to \textit{different} mechanisms, as in $W_x$ and $Z_{x'}$ in the clinical example, are allowed. The difficulty arises when the requested counterfactuals require two different versions of the \textit{same} mechanism. Counterfactual ancestors make these requirements explicit. 

\paragraph{Counterfactual ancestors.} 
Before listing required mechanisms, we simplify the counterfactuals. \textit{Effectiveness} gives $X_{x\mf{w}} = x$, so a directly assigned constant requires no evaluation of $f_X$. For other counterfactuals, \textit{interventional minimization} removes assignments that cannot affect the counterfactual \citep{correa2021nested}:
\begin{align}
    \lVert X_{\mf{w}} \rVert = X_{\mf{w} \cap \An(X)_{\mc{G}_{\overline{\mf{W}}}}}. 
\end{align}
For a collection, $\lVert \mf{X}_\ast \rVert$ applies this operation to each term. We apply this simplification before checking realizability. 

The counterfactual ancestors of $X_{\mf{w}}$, denoted $\An(X_{\mf{w}})_{\mc{G}}$, record the counterfactuals needed to compute it. A counterfactual $T_{\mf{z}}$ belongs to this set when:  
\begin{itemize}
    \item[1.] $T \in \An(X)_{\mc{G}_{\overline{\mf{W}}}} \setminus \mf{W}$, and

    \item[2.] $\mf{z}  = \mf{w} \cap \An(T)_{\mc{G}_{\overline{\mf{W}}}}$.
\end{itemize}
The first condition traces the mechanisms that remain upstream of $X$ after fixing $\mf{W}$. The second retains only the assignments that can affect each such mechanism. For a collection, $\An(\mf{X}_\ast)_{\mc{G}}$ is  the union of the counterfactual ancestor sets of its terms. Natural variables may occur as intermediate ancestors; they need not be among the assigned counterfactuals. Two required versions conflict when their subscript assignments differ after interventional minimization, including the variables to which those assignments apply. 

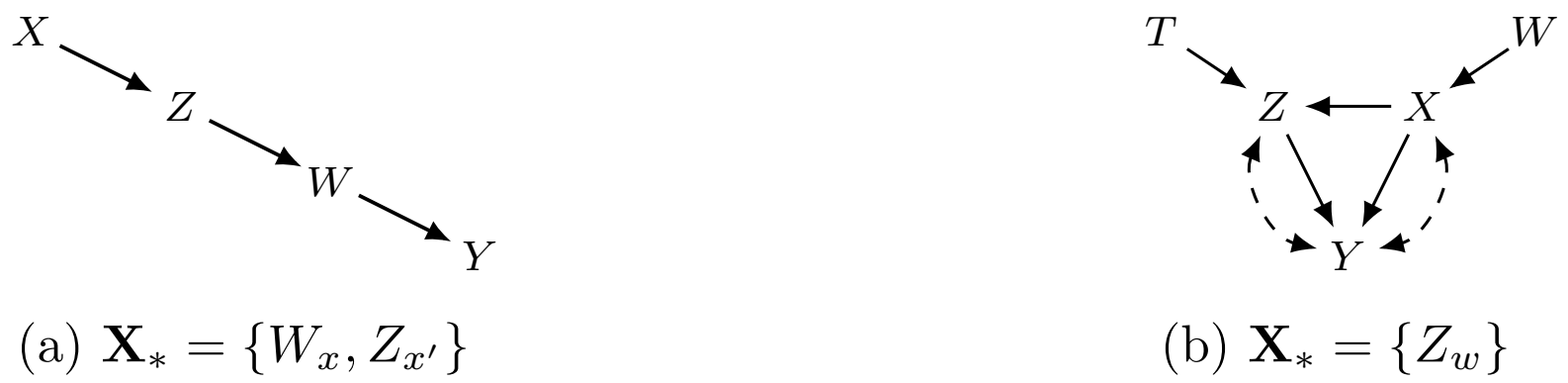
\begin{figure}[t]
    \hfill
    \begin{minipage}[t]{0.43\textwidth}\centering
    \subfloat[$\mf{X}_\ast = \{W_x,Z_{x'}\}$]{
    \begin{tikzpicture}[x=1cm,y=1cm,>={Latex[width=1.4mm,length=1.7mm]},
        line width=0.25mm,
        RR/.style={draw = none,circle,inner sep=0mm, minimum size=4mm}]
        \centering
        \node[RR] (X) at (0,1.5) {$X$};
        \node[RR] (Z) at (1,1) {$Z$};
        \node[RR] (W) at (2,0.5) {$W$};
        \node[RR] (Y) at (3,0) {$Y$};

        \draw[->] (X) -- (Z);
        \draw[->] (Z) -- (W);
        \draw[->] (W) -- (Y);
    \end{tikzpicture}\label{fig: realizable_a}}
    \end{minipage}
    \hfill
    \begin{minipage}[b]{0.43\textwidth}\centering
    \subfloat[$\mf{X}_\ast = \{Z_w\}$]{
    \begin{tikzpicture}[x=1cm,y=1cm,>={Latex[width=1.4mm,length=1.7mm]},
        line width=0.2mm,
        RR/.style={draw = none,circle,inner sep=0mm, minimum size=4mm}]\centering
        \node[RR] (W) at (1.75,1.5) {$W$};
        \node[RR] (T) at (-0.75,1.5) {$T$};
        \node[RR] (X) at (1,1) {$X$};
        \node[RR] (Z) at (0,1) {$Z$};
        \node[RR] (Y) at (0.5,0) {$Y$};
        \draw[->] (X) -- (Z);
        \draw[->] (Z) -- (Y);
        \draw[->] (W) -- (X);
        \draw[->] (T) -- (Z);
        \draw[->] (X) -- (Y);
        \draw[<->, dashed] (X) to[bend left = 50] (Y);
        \draw[<->, dashed] (Z) to[bend right = 50] (Y);
    \end{tikzpicture}\label{fig: realizable_b}}
    \end{minipage}
    \hfill\null
    \caption{\small Conflicting mechanism evaluations. (a) For $x \neq x'$, realizing $\{W_x,Z_{x'}\}$ requires both $Z_x$ and $Z_{x'}$. (b) The counterfactual $Z_w$ requires $X_w$, while the outer reward also uses the natural $X$. } 
    \label{fig: realizable}
\end{figure}

For instance, the causal diagram in \Cref{fig: realizable_a} with $\mf{X}_\ast = \{W_x, Z_{x'}\}$ gives: 
\begin{align}
    \An(\mf{X}_\ast)_{\mc{G}}  &= \An(W_x)_{\mc{G}} \cup \An(Z_{x'})_{\mc{G}}\\
    &= \{W_x,Z_x\} \cup \{Z_{x'}\} = \{W_x,Z_x, Z_{x'}\}. 
\end{align}
Computing $W_x$ first requires $Z_x$, because $Z$ lies between $X$ and $W$. The separate counterfactual $Z_{x'}$ then asks for a second version of $Z$. If $x' = x$, these requirements coincide and one evaluation of $f_Z$ supplies both. If $x' \neq x$, however, they require conflicting inputs to that mechanism. This leads to the graphical criterion below. 

\begin{lemma}[Realizability; Corollary~3.7 of \citet{raghavan2025counterfactual}] \label{lem: cor 3.7 in raghavan2025}
    Under $\mb{A}^\dagger$, a counterfactual distribution is realizable if and only if its ancestor set contains no conflicting required versions of the same mechanism. 
\end{lemma}

Thus $P(W_x, Z_{x'})$ in \Cref{fig: realizable_a} is not realizable when $x \neq x'$. The obstacle concerns producing both counterfactuals on one unit, even though either counterfactual can be measured separately. 

For an action, realizability concerns the mechanisms actually required to evaluate its reward on the same unit, including the inner counterfactuals that the reward uses. In \Cref{fig: realizable_b}, $Z_w$ requires $X_w$ as an input. However, the outer counterfactual $Y_{Z_w}$ also receives the natural $X$ through the direct edge $X \to Y$. Producing the required counterfactuals would therefore require both versions of $X$. If the action instead supplies $X_w$ to the outer reward as well, giving $Y_{X_w,Z_w}$, the conflicting requirement disappears. This example explains why checking the assigned counterfactuals alone is insufficient: the complete nested evaluation determines which mechanisms are needed. 

\section{Counterfactual Structural Causal Bandits}

We now formalize the \textit{counterfactual structural causal bandit} (\ctf-SCB) problem. As in \Cref{chap:scb}, an agent interacts with a target system modeled by an SCM with a reward variable $Y$. The agent knows $\mc{G}$ and $Y$, while the structural functions and the exogenous distribution remain unknown. Traditional SCB \citep{lee2018structural} optimizes the mean reward of an ordinary intervention: 
\begin{align}
    \mu_{\mf{x}} = \mb{E}Y_{\mf{x}} = \mb{E}[Y\mid do(\mf{X}=\mf{x})].
\end{align}
A counterfactual action $\mf{X}_{\ast\ast}$ can instead supply computed counterfactuals to the system. We write its mean reward as:
\begin{align}
    \mu_{\mf{X}_{\ast\ast}} = \mb{E}Y_{\mf{X}_{\ast\ast}}.
\end{align}
When the action is given by counterfactual assignments, each $X_{i[\mf{w}_i]}$ is computed under the indicated submodel and substituted for its target $X_i$. The required mechanisms and the reward are evaluated on the same unit. 

For concreteness, return to \Cref{fig:ctfscb_introduction_example}. The mean reward $\mu_x$ is obtained by setting $X = x$, whereas $\mu_{W_x,Z_{x'}}$ uses the physician's counterfactual under $x$ and the AI counterfactual under $x'$. Since $W$ and $Z$ are the only children of $X$ in this graph, $\mu_x = \mu_{W_x,Z_x}$. Indeed, setting $X=x$ sends $x$ to both assessments, and assigning $x$ separately to the two inputs produces the same reward. Allowing $x'$ to differ from $x$ makes additional decisions available. The ordinary intervention is recovered when both input values agree.

\paragraph{Parent inputs of counterfactual actions.} In the preceding example, counterfactuals describe the action. For computation, we describe the same experiment through the inputs received by each mechanism. At $f_X$, the map $\pa[X]$ records the parents whose inputs are assigned constants. Each selected input has a separately selectable value, and every other input receives the output of its parent's mechanism on the same unit. Inputs to $f_Y$ may also be fixed, while the reward mechanism itself is retained. Each required mechanism is evaluated once, in a topological order of $\mc{G}$. 

We use $\mf{X}_{\ast\ast}$ for this counterfactual action specified by these inputs and $Y_{\mf{X}_{\ast\ast}}$ for its reward. The parent-input specification preserves the constant inputs and counterfactual substitutions of the original action. Values within each counterfactual subscript remain unchanged. Inputs that cannot affect $Y$ may be discarded. 

\begin{algorithm}[t]
\small
\SetAlgoNoEnd
\DontPrintSemicolon

Initialize $\mc{F}_{\mf{X}_{\ast\ast}}$ as the collection of original mechanisms $\mc{F}$.\;

\For{\rm each $X\in\mf{V}$ in topological order}{
    \For{\rm each $W\in\pa[X]$}{
        Substitute the selected value for the argument $W$ in $f_X\in\mc{F}_{\mf{X}_{\ast\ast}}$.\;
    }
    Retain every other parent argument and every exogenous argument of $f_X$.\;
}
\Return{$\mc{M}_{\mf{X}_{\ast\ast}}=\langle\mf{U},\mf{V},\mc{F}_{\mf{X}_{\ast\ast}},P(\mf{U})\rangle$}\;
\caption{\small Constructing the SCM for an assigned counterfactual action.}
\label{alg: ctf-realizable-scm}
\end{algorithm}

Let $\mc{M}_{\mf{X}_{\ast\ast}}$ be the SCM obtained by fixing these selected parent arguments, retaining every other argument and the original exogenous inputs. Furthermore, let $\mc{G}_{\mf{X}_{\ast\ast}}$ be a graph obtained from $\mc{G}$ by deleting $W \to V$ whenever a required evaluation of $f_V$ receives a constant in place of its $W$ argument. All vertices, other directed edges, and bidirected edges are retained. 

A directed path in $\mc{G}_{\mf{X}_{\ast\ast}}$ therefore follows inputs that still receive their parent's computed counterfactual. In particular, a vertex loses a parent precisely when that incoming edge is replaced by a constant input. 
 
\begin{definition}[Action space]\label{def: ctfscb action space}
    The action space $\mc{A}_{\ctf}$ consists of all counterfactual actions realizable under $\mb{A}^\dagger$, with their selected input values specified. 
\end{definition}
 
This space includes realizable nested counterfactual actions. Their constant inputs can be recorded at the receiving mechanisms, as established below. 

\begin{proposition}\label{prop: cover realizable nested counterfactuals}
    Every realizable nested counterfactual action can be implemented by the parent-input choices described above. 
\end{proposition}
\begin{proof}[Proof of \Cref{prop: cover realizable nested counterfactuals}]
    Expand $Y_{\mf{X}_{\ast\ast}}$ into its required structural equations, applying effectiveness and identifying equal counterfactuals. Realizability gives one evaluation of each required mechanism. Record in $\pa[X]$ every parent input fixed to a constant at $f_X$, retaining its value. All other arguments are outputs of parent mechanisms on the same unit. These choices define an action in $\mc{A}_{\ctf}$.  

    To compare the two evaluations, proceed in topological order. Each first required mechanism receives the same fixed inputs and exogenous realization in both evaluations, and hence produces the same output. At a subsequent mechanism, the parent outputs already agree, and the fixed inputs and exogenous realization also agree. Its output is therefore unchanged. Continuing to $Y$ proves equality of the reward on every unit. 
\end{proof}

\paragraph{Constructing the SCM for a counterfactual action.} \Cref{alg: ctf-realizable-scm} starts from the original mechanisms and replaces only the selected parent arguments. Each remaining argument receives its parent's output on the same unit, and every exogenous input is retained. Evaluating these mechanisms in topological order gives the intended reward $Y_{\mf{X}_{\ast\ast}}$. Selecting an input of $f_Y$ changes that argument while leaving its other inputs available; the reward mechanism itself is retained. 

The construction specifies how an action is evaluated in an SCM. As in the previous \unitnames{}, the bandit agent need not know these mechanisms. Instead, it performs the corresponding experiment and observes the reward. 

\paragraph{Counterfactual structural causal bandits objective.} As in \Cref{chap:scb}, at round $t$ the agent chooses $A_t \in \mc{A}_{\ctf}$, performs the corresponding experiment on a fresh unit, and observes its reward $Y_t$. Repeated selections provide samples for estimating the mean of that action, just as ordinary interventions do in SCB. 

We use the best-action convention of \Cref{chap:scb}: a superscript $\ast$ optimizes the subscript values of a fixed set of counterfactuals, and $\star$ selects an optimal action over the entire space. The cumulative pseudo-regret is:
\begin{align}
R_T^{\mc{A}_{\ctf}}
 =T\mu_{\mf{X}_{\ast\ast}^\star}-\sum_{t=1}^T\mu_{A_t} 
 =\sum_{\mf{X}_{\ast\ast} \in \mc{A}_{\ctf}} \Delta_{\mf{X}_{\ast\ast}}N_T(\mf{X}_{\ast\ast}),
\end{align}
where $\Delta_{\mf{X}_{\ast\ast}} = \mu_{\mf{X}_{\ast\ast}^\star} - \mu_{\mf{X}_{\ast\ast}}$ and $N_T(\mf{X}_{\ast\ast})$ counts the rounds $t\leq T$ with $A_t=\mf{X}_{\ast\ast}$. The learning objective has the same form as before; the available experiments and the benchmark reward have been extended. 

\begin{corollary}\label{cor: l2 dominance}
    The optimal ordinary intervention cannot outperform the optimal counterfactual action, i.e., $\mu_{\mf{x}^\star} \leq \mu_{\mf{X}_{\ast\ast}^\star}$.
\end{corollary}
\begin{proof}[Proof of \Cref{cor: l2 dominance}]
    An ordinary intervention is obtained by assigning the same value to every outgoing input of each intervened variable. Hence its reward is represented in $\mc{A}_{\ctf}$. 
\end{proof}

A larger space can offer a higher reward, but it also gives the agent more actions to explore. The following sections use causal structure to reduce this exploration burden. We first ask when removing a counterfactual assignment leaves the reward unchanged, and then compare the best rewards attainable by non-equivalent counterfactual actions. 

\section{Equivalence among Actions}

In \Cref{chap:scb}, an ordinary intervention set is reduced by removing variables whose assigned values cannot affect the reward once the other interventions are in place. For example, if every directed path from $Z$ to $Y$ passes through an intervened variable $W$, the additional assignment to $Z$ does not change $Y$. An MIS retains the intervention variables needed to distinguish that intervention set from its proper subsets. 

\paragraph{Counterfactual minimal intervention sets.}
The same question arises for a set of counterfactuals $\mf{X}_{\ast\ast}$ with outer targets $\mf{X} = \mf{V}(\mf{X}_{\ast\ast})$. Each target now receives a counterfactual, which may itself contain nested counterfactuals. Such an assignment can be redundant for two reasons:
\begin{itemize}
    \item[1.] The target may have no remaining directed path to the reward.

    \item[2.] The counterfactuals assigned to the other targets may already induce the same counterfactual at that target. 
\end{itemize}
We examine both possibilities using the consistency and exclusion rules of \ctf-calculus.

We first apply effectiveness and interventional minimization within the nested counterfactuals. For the special case $\mf{X}_\ast = \{X_{i[\mf{w}_i]}\}_i$, this gives $\mf{X}_\ast  = \lVert \mf{X}_\ast \rVert$. We then ask whether the assignment to a target variable can be removed altogether. Retained counterfactuals keep all their inner substitutions and constant assignments. As in the MIS definition of \Cref{chap:scb}, this comparison concerns an expression with variable assignments: retained terms keep their assignments, and equivalence must hold for all admissible values and all compatible SCMs. 
\begin{definition}[Counterfactual minimal intervention set]\label{def: ctf mis}
    A set of counterfactuals $\mf{X}_{\ast\ast}$ is a \textit{counterfactual minimal intervention set} (\ctf-MIS) if no proper subset $\mf{Z}_{\ast\ast} \subsetneq \mf{X}_{\ast\ast}$ satisfies $\mu_{\mf{X}_{\ast\ast}} = \mu_{\mf{Z}_{\ast\ast}}$ for every compatible SCM and every admissible subscript assignment. 
\end{definition}
Thus, deleting a term requires equality for every admissible assignment and every compatible SCM. The following examples show how the causal diagram can justify removing an assignment for all such values and SCMs. 

For example, consider the causal diagram shown in \Cref{fig: realizable_a}, and take the realizable counterfactual $\mf{X}_\ast = \{W_x, Z_x\}$. The two assigned counterfactuals use the same assignment $x$. To generate the reward, the outer system receives $W_x$ at $W$ and $Z_x$ at $Z$. However, once $W$ receives its assigned counterfactual, the value supplied separately at $Z$ has no directed path to $Y$. CUT makes this explicit:
\begin{align}
    \mu_{W_x,Z_x} &= \sum_{ywz} y P(y_{wz},w_x,z_x) \\
    &= \sum_{ywz} y P(y_{w},w_x,z_x) = \sum_{yw} y P(y_{w},w_x) = \mu_{W_x}. 
\end{align}
The second line uses \ctf-Rule~3 and then sums over $z$. Removing the outer assignment $Z_x$ still leaves the evaluation of $Z_x$ needed inside the submodel that computes $W_x$. It is the separate assignment at the outer target $Z$ that is redundant. Therefore $\{W_x,Z_x\}$ is not a \ctf-MIS. 

In \Cref{fig: realizable_b}, consider $\mf{X}_\ast = \{X_w, Z_{w'}\}$. Both targets have directed edges to $Y$, so neither is excluded by the ordinary MIS condition. Nevertheless, computing $Z_{w'}$ requires $X_{w'}$, while the first assignment requires $X_w$. Realizability therefore requires $w' = w$. 

With this shared value, assigning $X_w$ already makes $Z$ behave as $Z_w$ since the only directed path from $W$ to $Z$ passes through $X$. \ctf-Rule~1 permits substituting $X_w$ into the submodel for $Z_w$, and \ctf-Rule~3 then removes the assignment to $W$: $Z_w = Z_{w,X_w} = Z_{X_w}$ and thus $Y_{X_w,Z_w} = Y_{X_w}$. 

The other parent $T$ of $Z$ retains its natural value on the same unit in both evaluations. Thus every admissible assignment satisfies $\mu_{X_w,Z_w} = \mu_{X_w}$, and the assignment $Z_w$ can be removed even though $Z$ has a directed edge to $Y$. 

\paragraph{Graphical characterization for CTF-MISs.}
Throughout the remainder of this section, we assume that $\mf{X}_{\ast\ast}$ is realizable and every assigned counterfactual is nonconstant, i.e., after effectiveness, it still requires its target mechanism. Constant values are free apart from equalities required by realizability. 

The graph $\mc{G}_{\mf{X}_{\ast\ast}}$ deletes the directed edges replaced by constant inputs. Thus a vertex $V$ receives a fixed input exactly when its parent set in this graph is strictly smaller than in $\mc{G}$. In the second condition below, deleting the other outer targets from this graph ensures that the entire path, including its starting vertex, avoids those targets. 

\begin{figure}[t]
    \hfill
    \begin{minipage}[b]{0.43\textwidth}\centering
    \subfloat[$\mc{G}$]{
    \begin{tikzpicture}[x=1cm,y=0.9cm,>={Latex[width=1.4mm,length=1.7mm]},
        font=\scriptsize,
        line width=0.25mm,
        RR/.style={draw=none,circle,inner sep=0mm, minimum size=4mm, font=\scriptsize},
        transp/.style={regime,fill=betterred,minimum size=1mm,draw=none,inner sep=1mm},
        regime/.style={rectangle,draw=betterred,very thick,draw=betterred,minimum size=1mm},
        regime line/.style={draw=betterred, very thick},
        rotate = 0]\centering
        \node[RR] (X) at (0,3) {$X$};
        \node[RR] (W) at (-1,2) {$W$};
        \node[RR] (Z) at (1,2) {$Z$};
        \node[RR] (T) at (-1,1) {$T$};
        \node[RR] (R) at (1,1) {$R$};
        \node[RR] (Y) at (0,0) {$Y$};
        \draw[->] (X) -- (W);
        \draw[->] (X) -- (Z);
        \draw[->] (X) -- (Y);
        \draw[->] (W) -- (T);
        \draw[->] (Z) -- (R);
        \draw[->] (T) -- (Y);
        \draw[->] (R) -- (Y);
        \draw[<->, dashed] (R) to[bend left = 50] (Y);
        \draw[<->, dashed] (R) to[bend right = 30] (T);
        \draw[<->, dashed] (R) to[bend right = 50] (Z);
        \draw[<->, dashed] (R) to[bend right = 50] (W);
        \draw[<->, dashed] (X) to[bend left = 50] (Z);
    \end{tikzpicture}\label{fig:ctfscb 6var_a}}
    \end{minipage}
    \hfill
    \begin{minipage}[b]{0.43\textwidth}\centering
    \subfloat[]{
    \begin{tikzpicture}[x=1cm,y=.9cm,>={Latex[width=1.4mm,length=1.7mm]},
        font=\scriptsize,
        line width=0.25mm,
        RR/.style={draw=none,circle,inner sep=0mm, minimum size=4mm,font=\scriptsize},
        transp/.style={rectangle,fill=white,minimum size=1mm,draw=black,inner sep=0mm},
        regime/.style={rectangle, draw=betterred, thick,minimum size=1mm},
        regime line/.style={draw=black, thick, line width=0.25mm},
        rotate = 0]\centering
        \node[RR] (X) at (0,3) {$X$};
        \node[RR] (W) at (-1,2) {$W$};
        \node[RR] (Z) at (1,2) {$Z$};
        \node[RR] (T) at (-1,1) {$T$};
        \node[RR] (R) at (1,1) {$R$};
        \node[RR] (Y) at (0,0) {$Y$};
        \draw[->] (X) -- (Y);
        \draw[->, draw = gray!30] (X) -- (W);
        \draw[->, draw = gray!30] (X) -- (Z);
        \draw[->] (W) -- (T);
        \draw[->] (Z) -- (R);
        \draw[->] (T) -- (Y);
        \draw[->] (R) -- (Y);
        \draw[<->, dashed] (R) to[bend left = 50] (Y);
        \draw[<->, dashed] (R) to[bend right = 30] (T);
        \draw[<->, dashed] (R) to[bend right = 50] (Z);
        \draw[<->, dashed] (R) to[bend right = 50] (W);
        \draw[<->, dashed] (X) to[bend left = 50] (Z);
        \node[transp, right = 2.5mm of Z] (ZT){};
        \node[above = -0.5mm of ZT, font=\scriptsize] (wd){{\color{black} $x'$}};
        \draw[->,regime line] (ZT) -- (Z);
        \node[transp, left = 2.5mm of W] (WT){};
        \node[above = -0.5mm of WT, font=\scriptsize] (wo){{\color{black} $x$}};
        \draw[->,regime line] (WT) -- (W);
    \end{tikzpicture}\label{fig:ctfscb 6var_b}}
    \end{minipage}
    \hfill\null
    \caption{\small (a) A causal diagram. (b) Equivalent counterfactual actions induce the same propagation of $x$ and $x'$ to $Y$. }
    \label{fig:ctfscb 6var}
\end{figure}
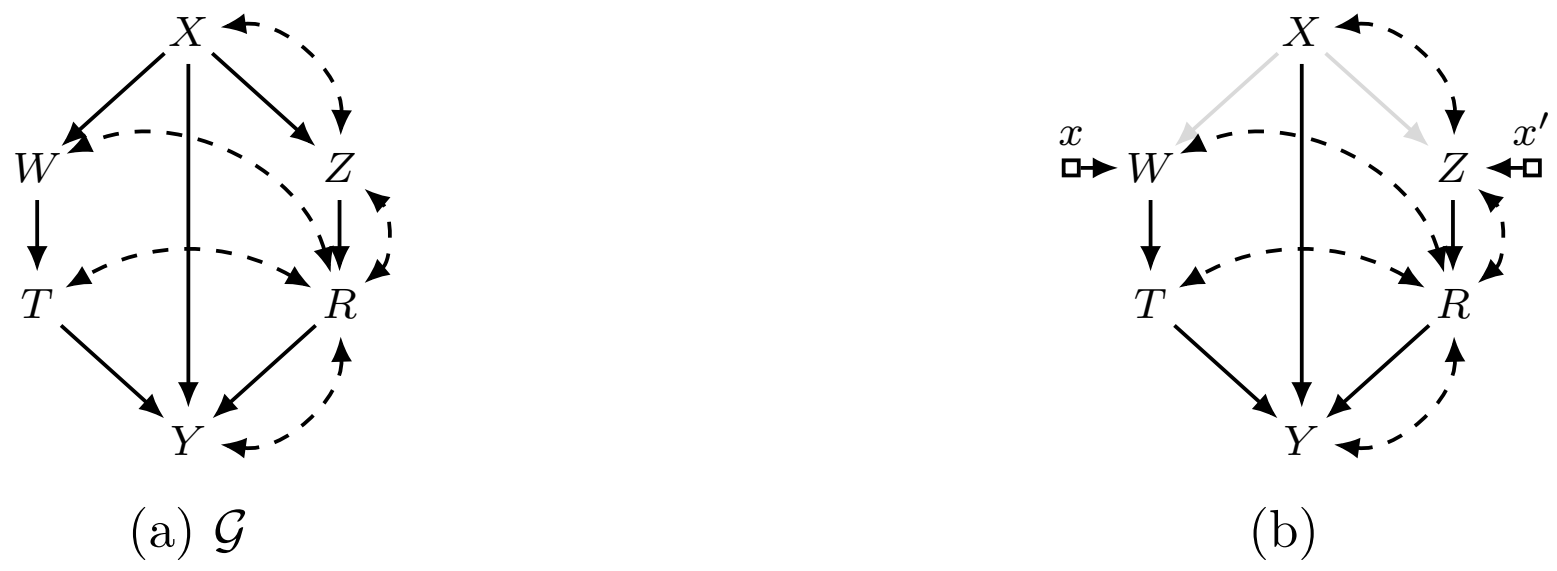

\begin{theorem}[Graphical characterization of \ctf-MIS]\label{thm: ctf-mis}
    A counterfactual action $\mf{X}_{\ast\ast}$ is a \ctf-MIS if and only if the following two conditions hold:
    \begin{itemize}
        \item[1.] $\mf{X} \subseteq \An(Y)_{\mc{G}_{\overline{\mf{X}}}}$.
        \item[2.] For every $X_i \in \mf{X}$, there exists $V \in \An(X_i)_{\mc{G}_{\mf{X}_{\ast\ast}} \setminus (\mf{X} \setminus \{X_i\})}$ such that \\
        $\Pa(V)_{\mc{G}_{\mf{X}_{\ast\ast}}} \subsetneq \Pa(V)_{\mc{G}}$. 
    \end{itemize}
\end{theorem}
The proof is given in \Cref{sec: proof of thm: ctf-mis}. The first condition is the ordinary MIS condition applied to the outer target set $\mf{X}$. Each target must have a directed path to $Y$ that avoids the other targets. In the first example, $Z$ fails this test because its only path to $Y$ passes through $W$. In contrast, $\{W_x, Z_{x'}\}$ in \Cref{fig:ctfscb 6var_a} has the two paths $W \to T \to Y$ and $Z \to R \to Y$. Therefore, both targets satisfy the first condition. 

The second condition traces the counterfactual's computation in the other direction. A directed edge deleted from $\mc{G}$ must enter a vertex with a retained directed path to $X_i$ avoiding every other outer target. The path may have length zero: a fixed input received by $X_i$ itself suffices. A fixed input received only at another outer target does not suffice, because that target's counterfactual is still supplied when the assignment to $X_i$ is removed. 

In the second example, the fixed input $W \to X$ enters $X$. The counterfactual $X_w$ therefore satisfies the second condition. For $Z_w$, however, this input reaches $Z$ only through the other target $X$. Deleting $X$ from the graph removes that path, and $Z_w$ fails the condition. In contrast, for $\{W_x, Z_{x'}\}$ in \Cref{fig:ctfscb 6var_a}, both $W$ and $Z$ receive a fixed input directly, so both conditions hold. 

\begin{corollary}\label{cor: ctfmis for unnested counterfactual}
    For $\mf{X}_\ast = \{X_{i[\mf{w}_i]}\}_i$ satisfying the first condition, the second condition is equivalent to $\mf{W}_i \cap \An(X_i)_{\mc{G}_{\overline{\mf{X} \setminus \{X_i\}}}} \neq \emptyset$ for every $X_i$. 
\end{corollary}
\begin{proof}[Proof of \Cref{cor: ctfmis for unnested counterfactual}]
    The first condition in \Cref{thm: ctf-mis} ensures that every assigned target computation is required by the reward. For the second condition, suppose a vertex of $\mf{W}_i$ has a directed path to $X_i$ in $\mc{G}_{\overline{\mf{X} \setminus \{X_i\}}}$. Take the last vertex of $\mf{W}_i$ on this path. Its input to the next mechanism is fixed, whereas the remaining path to $X_i$ contains no further subscript intervention or other outer target. Realizability preserves these inputs in $\mc{G}_{\mf{X}_{\ast}}$, giving the mechanism and path required by the general condition. 
    
    Conversely, a fixed input received by a mechanism on such a path to $X_i$ must be an assignment from $\mf{W}_i$: these are exactly the constants used in the submodel computing $X_{i[\mf{w}_i]}$. Prepending that input edge to the path gives a directed path from $\mf{W}_i$ to $X_i$ whose internal vertices avoid the other outer targets. Consequently, the general condition is equivalent to $\mf{W}_i \cap \An(X_i)_{\mc{G}_{\overline{\mf{X} \setminus \{X_i\}}}} \neq \emptyset$, proving the specialization. 
\end{proof}

\begin{figure}[t]
    \hfill
    \begin{minipage}[t]{0.3\textwidth}\centering
    \subfloat[$\mc{G}$]{
    \begin{tikzpicture}[x=1cm,y=1.1cm,>={Latex[width=1.4mm,length=1.7mm]},
        font=\scriptsize,
        line width=0.25mm,
        RR/.style={draw=none,circle,inner sep=0mm, minimum size=4mm, font=\scriptsize}]\centering
        \node[RR] (W) at (0,2) {$W$};
        \node[RR] (X) at (-1,1) {$X$};
        \node[RR] (Z) at (1,1) {$Z$};
        \node[RR] (Y) at (0,0) {$Y$};
        \draw[->] (W) -- (X);
        \draw[->] (W) -- (Z);
        \draw[->] (X) -- (Z);
        \draw[->] (X) -- (Y);
        \draw[->] (Z) -- (Y);

        \draw[<->, dashed] (Z) to[bend left = 50] (Y);
        \draw[<->, dashed] (W) to[bend left = 50] (Z);
        \draw[<->, dashed] (X) to[bend left = 50] (W);
    \end{tikzpicture}\label{fig:ctfscb nested ctfmis_a}}
    \end{minipage}
    \hfill
    \begin{minipage}[b]{0.3\textwidth}\centering
    \subfloat[$\mc{G}_{\mf{X}_{\ast\ast}}$]{
    \begin{tikzpicture}[x=1cm,y=1.1cm,>={Latex[width=1.4mm,length=1.7mm]},
        font=\scriptsize,
        line width=0.25mm,
        RR/.style={draw=none,circle,inner sep=0mm, minimum size=4mm, font=\scriptsize}]
        \centering
        \node[RR] (W) at (0,2) {$W$};
        \node[RR] (X) at (-1,1) {$X$};
        \node[RR] (Z) at (1,1) {$Z$};
        \node[RR] (Y) at (0,0) {$Y$};
        \draw[->,gray!30] (W) -- (X);
        \draw[->,gray!30] (W) -- (Z);
        \draw[->] (X) -- (Z);
        \draw[->] (X) -- (Y);
        \draw[->] (Z) -- (Y);

        \draw[<->, dashed] (Z) to[bend left = 50] (Y);
        \draw[<->, dashed] (W) to[bend left = 50] (Z);
        \draw[<->, dashed] (X) to[bend left = 50] (W);
    \end{tikzpicture}\label{fig:ctfscb nested ctfmis_b}}
    \end{minipage}
    \hfill \null
    \caption{\small (a) A causal diagram and (b) the graph for $\mf{X}_{\ast\ast} = \{X_w,Z_{w',X_w}\}$. } 
    \label{fig:ctfscb nested ctfmis}
\end{figure}
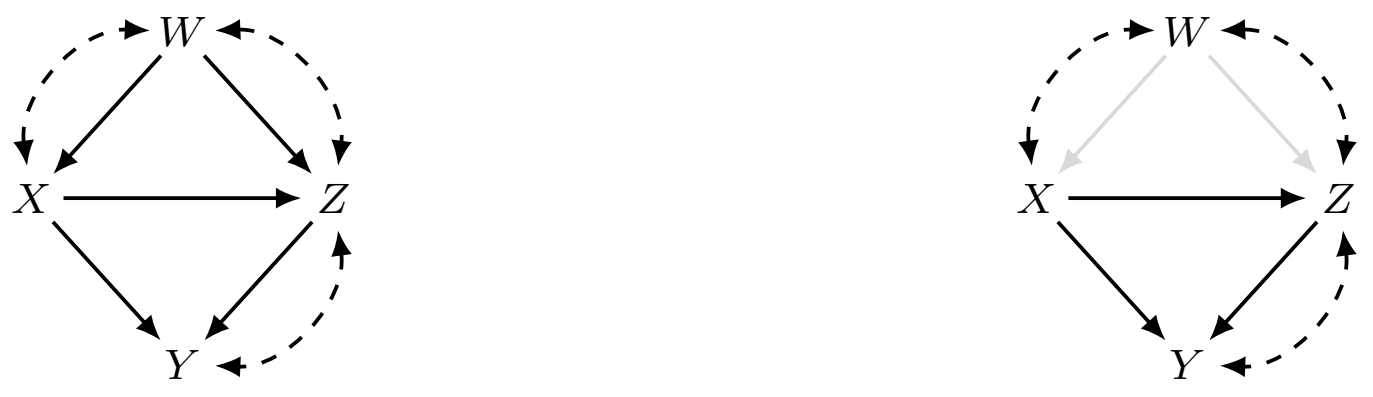

Consider the causal diagram in \Cref{fig:ctfscb nested ctfmis_a} with $\mf{X}_{\ast\ast} = \{X_w,Z_{w',X_{w}}\}$. The first counterfactual gives $f_X$ the input $W = w$. The second gives $f_Z$ the input $W = w'$ together with the previously computed $X_w$. Consequently, $W \to X$ and $W \to Z$ are deleted, while $X \to Z$ remains in $\mc{G}_{\mf{X}_{\ast\ast}}$ (\Cref{fig:ctfscb nested ctfmis_b}). Both targets have a directed edge to $Y$ and receive a fixed input themselves. Thus, $\mf{X}_{\ast\ast}$ is a \ctf-MIS. 

\paragraph{Reducing an action to a CTF-MIS.}
\ctf-\textsc{MISify} (\Cref{alg: ctf-misify}) applies these conditions to a supplied counterfactual set. It first removes all targets blocked from $Y$ by other outer targets. A reverse search from $Y$, stopping at each outer target, identifies exactly the assignments to retain. It then examines each remaining target once, deleting its assignment if the second condition fails. The nested counterfactuals and constant values inside every retained term are unchanged.

Each deletion preserves the reward on every unit. The initial deletions are valid together: every path from an omitted target to $Y$ meets a retained target. Moreover, these deletions preserve the required mechanism evaluations and their fixed inputs, so $\mc{G}_{\mf{X}_{\ast\ast}}$ is unchanged. A target that passes either condition continues to pass after other targets are deleted, because its witnessing path remains available. Thus a single pass suffices, and the returned set satisfies both conditions. 

\begin{algorithm}[t]
\SetAlgoNoEnd
\DontPrintSemicolon

\textbf{function~}\textsc{ctf-MISify}($\mc{G}$, $Y$, $\mf{X}_{\ast\ast}$)\;
\Indp

Compute $\mc{G}_{\mf{X}_{\ast\ast}}$ and keep it fixed throughout. \;

Set $\mf{X} = \mf{V}(\mf{X}_{\ast\ast})$.\;

\tcp{Apply the first condition in \Cref{thm: ctf-mis}.}

Retain only the outer assignments whose targets belong to $\An(Y)_{\mc{G}_{\overline{\mf{X}}}}$. \;

\BlankLine
\tcp{Apply the second condition in \Cref{thm: ctf-mis}.}

Update $\mf{X} = \mf{V}(\mf{X}_{\ast\ast})$.\;

\For{\rm each remaining target $X_i$ in any fixed order}{
    Search backwards along directed edges from $X_i$ in $\mc{G}_{\mf{X}_{\ast\ast}} \setminus (\mf{X} \setminus \{X_i\})$, including $X_i$ itself.\;

    \If{\rm every visited vertex has the same parents in $\mc{G}_{\mf{X}_{\ast\ast}}$ and $\mc{G}$}{
        
        Delete only the outer assignment to $X_i$ from $\mf{X}_{\ast\ast}$.\;
        
        Update $\mf{X} = \mf{V}(\mf{X}_{\ast\ast})$.\;
    }
}

\Return{$\mf{X}_{\ast\ast}$}\;
\Indm
\caption{\small Reducing a counterfactual action to a \ctf-MIS.}
\label{alg: ctf-misify}
\end{algorithm}

\begin{corollary}\label{cor:ctf-misify}
    \ctf-\textsc{MISify} returns an equivalent and realizable \ctf-MIS subset of its input.
\end{corollary}

\begin{proof}[Proof of \Cref{cor:ctf-misify}]
In the first step, every directed path from a discarded target to $Y$ meets a retained target: the last outer target on that path satisfies the first condition. Thus these assignments can be removed together, while keeping every retained nested counterfactual unchanged. 

By the necessity proof of \Cref{thm: ctf-mis}, both deletion steps preserve the reward on every unit, for every compatible SCM and every admissible assignment. They also preserve the required mechanism evaluations and their fixed inputs. Hence realizability and the initially computed graph $\mc{G}_{\mf{X}_{\ast\ast}}$ are preserved.

Deleting other outer targets cannot destroy a path witnessing either condition for a retained target. Therefore, a target that passes a condition continues to pass
it throughout the algorithm. After the single pass, every remaining target satisfies
both conditions. Therefore, \Cref{thm: ctf-mis} gives the conclusion. 
\end{proof}

\paragraph{Computational cost.} Let $n = \lvert \mf{V} \rvert$ and $m$ be the number of edges in $\mc{G}$, and let $k = \lvert \mf{X} \rvert + 1$ for the initial outer target set. Comparing the two graphs' parent sets and performing the initial reverse search take $\mc{O}(n + m)$ time. Each remaining target requires at most one further reverse search, also taking $\mc{O}(n + m)$ time. Therefore \ctf-\textsc{MISify} uses $\mc{O}(k(n + m))$ time and $\mc{O}(n + m)$ auxiliary space. These bounds concern the graphical reduction after the action graph has been formed; retained nested terms are reused without expanding them. 

\paragraph{Equivalent CTF-MISs.} Different \ctf-MISs can still describe the same input action even when neither target set contains the other. Two counterfactual sets are equivalent if every admissible subscript assignment to either set has a corresponding assignment to the other, independent of the SCM, that gives the same reward on every unit. To see this, consider the causal diagram shown in \Cref{fig:ctfscb 6var_a}. The counterfactual sets $\{T_x,Z_{x'}\}$ and $\{W_x, Z_{x'}\}$ satisfy both \ctf-MIS conditions. The first assigns $T_x$ directly to $T$; the second assigns $W_x$ to $W$, which induces the same counterfactual $T_x$ through the mechanism for $T$. In particular, $W$ receives the fixed input $x$, and the retained path $W \to T$ avoids the other target $Z$. This verifies the second condition for $T_x$, and its edge $T \to Y$ verifies the first condition. 

To compare the two actions, first compute $W_x$ on a unit. Feeding this value into the unchanged mechanism for $T$ produces $T_x$. The counterfactual $Z_{x'}$ on
the other branch is identical under both actions, as is the natural value of $X$ received directly by $Y$. Thus the reward mechanism receives the same inputs on the same unit. These substitutions give:
\begin{align}
 \mu_{T_x,Z_{x'}}
    &=\sum_{y,t,z}yP(y_{tz},t_x,z_{x'}) &&\text{CUT}
    \\
    &=\sum_{y,t,z,w}yP(y_{tz},t_x,z_{x'},w_x) &&\text{marg.~over $W_x$}
    \\
    &=\sum_{y,t,z,w}yP(y_{twz},t_{wx},z_{x'},w_x) && \text{\ctf-Rules~1 and~3}
    \\
     &=\sum_{y,t,z,w}yP(y_{twz},t_{w},z_{x'},w_x) && \text{\ctf-Rule~3}
    \\
     &=\sum_{y,t,z,w}yP(y_{twz},t_{wz},z_{x'},w_x) && \text{\ctf-Rule~3}
    \\
     &=\sum_{y,t,z,w}yP(y_{wz},t_{wz},z_{x'},w_x) && \text{\ctf-Rule~1}
    \\
     &=\sum_{y,z,w}yP(y_{wz},z_{x'},w_x) && \text{sum. over $T_{wz}$}
     \\
     &=\mu_{W_x,Z_{x'}}. &&\text{CUT}
\end{align}
Similarly, $\{T_x, R_{x'}\}$ has the same reward. These \ctf-MISs assign the same inputs $x$ on $X \to W$ and $x'$ on $X \to Z$, as shown in \Cref{fig:ctfscb 6var_b}. The action space $\mc{A}_{\ctf}$ identifies these sets through their common input choices. 

\section{Counterfactual Possibly-Optimal Minimal Intervention Sets}

The preceding section identifies redundant outer assignments in a realizable set $\mf{X}_{\ast\ast}$. Removing these assignments preserves the reward for every admissible choice of values. We now compare non-equivalent counterfactual actions after optimizing their subscript values. As in \Cref{chap:scb}, the causal diagram can imply that one optimized counterfactual always performs at least as well as another, regardless of the parameterization of the underlying SCM. These partial-order relations allow the agent to exclude further actions before estimating their rewards. 

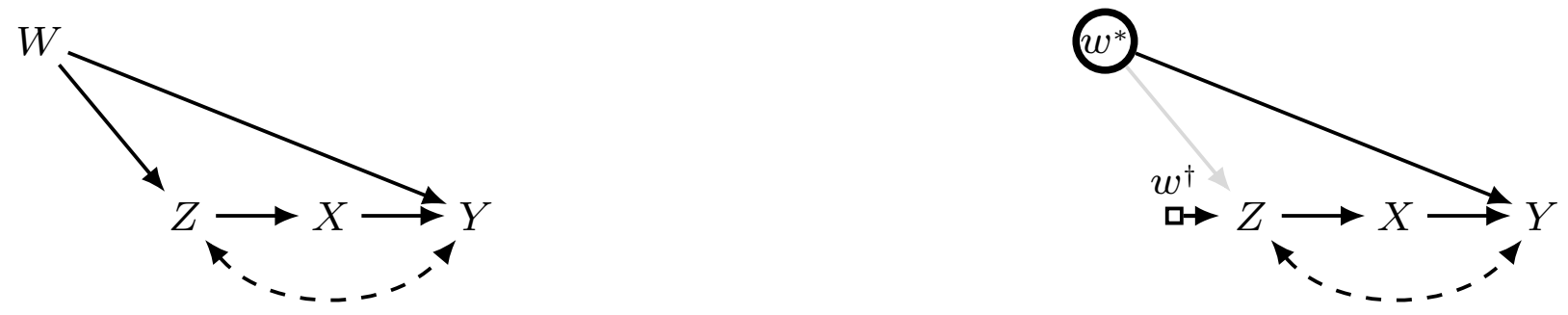
\begin{figure}[t]
    \hfill
    \begin{minipage}[b]{0.43\textwidth}\centering
    \subfloat[$\mc{G}$]{
    \begin{tikzpicture}[x=1cm,y=1cm,>={Latex[width=1.4mm,length=1.7mm]},
        font=\scriptsize,
        line width=0.25mm,
        RR/.style={draw=none,circle,inner sep=0mm, minimum size=4mm, font=\scriptsize}]\centering
        \node[RR] (Z) at (1,0) {$Z$};
        \node[RR] (W) at (0,1.2) {$W$};
        \node[RR] (X) at (2,0) {$X$};
        \node[RR] (Y) at (3,0) {$Y$};
        \draw[->] (Z) -- (X);
        \draw[->] (X) -- (Y);
        \draw[->] (W) -- (Z);
        \draw[->] (W) -- (Y);
        \draw[<->, dashed] (Z) to[bend right = 50] (Y);
    \end{tikzpicture}\label{fig:ctfscb counterfactual improvement_a}}
    \end{minipage}
    \hfill
    \begin{minipage}[b]{0.43\textwidth}\centering
    \subfloat[$\{W_{w^\ast},Z_{w^\dagger}\}$]{
    \begin{tikzpicture}[x=1cm,y=1cm,>={Latex[width=1.4mm,length=1.7mm]},
        font=\scriptsize,
        line width=0.25mm,
        RR/.style={draw=none,circle,inner sep=0mm, minimum size=4mm,font=\scriptsize},
        transp/.style={rectangle,fill=white,minimum size=1mm,draw=black,inner sep=0mm},
        regime/.style={rectangle, draw=betterred, thick,draw=betterred,minimum size=1mm},
        regime line/.style={draw=black,  thick},
        rotate = 0]\centering
        \node[RR] (Z) at (1,0) {$Z$};
        \node[RR,draw,line width = 0.5mm] (W) at (0,1.2){\color{black} $w$\scalebox{0.8}{$^\ast$}};
        \node[RR] (X) at (2,0) {$X$};
        \node[RR] (Y) at (3,0) {$Y$};
        \draw[->] (Z) -- (X);
        \draw[->] (X) -- (Y);
        \draw[->] (W) -- (Y);
        \draw[->, draw = gray!30] (W) -- (Z);
        \draw[<->, dashed] (Z) to[bend right = 50] (Y);
        
        \node[transp, left = 2.5mm of Z] (ZT){};
        \node[above = -0.5mm of ZT] (wd){\color{black} $w$\scalebox{0.8}{$^\dagger$}};
        \draw[->,regime line] (ZT) -- (Z);
    \end{tikzpicture}\label{fig:ctfscb counterfactual improvement_b}}
    \end{minipage}
    \hfill\null
    \caption{\small (a) A causal diagram. (b) A counterfactual action with separately chosen values $w^\ast$ and $w^\dagger$ on the two outgoing edges of $W$. }
    \label{fig:ctfscb counterfactual improvement}
\end{figure}
\paragraph{From ordinary interventions to counterfactual interventions.} First consider ordinary interventions in the causal diagram of \Cref{fig:ctfscb counterfactual improvement_a}. Let $w^\ast$ maximize the reward under $do(W=w)$, and choose $w^\dagger$ to maximize $\mu_{W_{w^\ast},Z_w}$. Choosing $w = w^\ast$ in $\{W_{w^\ast}, Z_w \}$ reproduces the ordinary intervention by effectiveness and \ctf-Rule~1. Hence: 
\begin{align}
    \mu_{w^\ast} = \mu_{W_{w^\ast},Z_{w^\ast}} \leq \mu_{W_{w^\ast},Z_{w^\dagger}}. 
\end{align}
The action $\{W_{w^\ast},Z_{w^\ast}\}$ sends the same value to both children of $W$. In contrast, for $\{W_{w^\ast},Z_{w^\dagger}\}$, the input received directly by $Y$ remains $w^\ast$, while $Z$ may listen to $w^\dagger$. The best reward cannot decrease, since the choice $w^\dagger = w^\ast$ is still available. A different value may improve the reward, depending on the underlying mechanisms. 

For a concrete example, consider the following binary SCM with independent exogenous variables:
\begin{align}
    \mc{M}
= \begin{cases}
        \mf{U} =   \{U_W,U_Z,U_X,U_Y,U_{ZY}\}\\
        \mf{V} = \{W,Z,X,Y\} \\
        \mc{F} = 
        \begin{cases}
            f_{W}(u_W) = u_W\\
            f_{Z}(w,u_Z,u_{ZY}) = u_Z \lor (w \oplus u_{ZY})\\
            f_{X}(z,u_X) = u_X \lor z\\
            f_{Y}(w,x,u_Y,u_{ZY}) = 1-(u_Y \oplus u_{ZY} \oplus x \oplus (1-w))\\
        \end{cases}\\
        P(\mf{U}) = \{U \thicksim \Bern(0.1) : \forall U \in \mf{U}\}.
    \end{cases}
\end{align}
Then $w^\ast = 0$ and $w^\dagger = 1$ give:
\begin{align}
    \mu_{\emptyset} = 0.22464 < \mu_{w^\ast} = 0.2368 < \mu_{W_{w^\ast},Z_{w^\dagger}} = 0.8848.
\end{align}
The first increase is obtained by replacing the natural value of $W$ with its best constant. The much larger second increase is obtained by keeping this zero at the input of $Y$ and sending one to the input of $Z$, as in \Cref{fig:ctfscb counterfactual improvement_b}. 

\paragraph{Counterfactual possibly-optimal minimal intervention sets.} 
We now extend possible optimality from \Cref{chap:scb} to counterfactuals. The comparisons
below concern counterfactual sets with each retained parent input independently selectable; equivalent sets are treated as the same action. 

\begin{definition}[Counterfactual possibly-optimal minimal intervention set]\label{def: ctf-pomis}
    A counterfactual action $\mf{X}_{\ast\ast}$ is a \ctf-POMIS if there exists an SCM conforming to $\mc{G}$ such that $\mu_{\mf{X}_{\ast\ast}^\ast} > \mu_{\mf{Z}_{\ast\ast}^\ast}$ for every non-equivalent counterfactual action $\mf{Z}_{\ast\ast}$. 
\end{definition}
For a retained counterfactual set, one compatible SCM makes its best subscript assignment outperform every assignment to any set that is not equivalent to it. Several counterfactual sets may satisfy this condition, with different SCMs establishing their possible optimality. The agent must still learn their rewards to determine which action is best in the actual environment. 

\subsection{CTF-POMIS under Unconfounded Rewards}

We first consider the case in which the exogenous inputs of $Y$ are independent of those generating its other ancestors. For any upstream intervention, the reward mechanism receives values from its parents. These values may be random. The agent can instead supply a fixed assignment directly to the same parent inputs. If conditioning on the received values does not change the distribution of the reward's exogenous inputs, the resulting mean is a weighted average of the means under these direct assignments. Choosing the best parent assignment is
therefore at least as good as that average. 

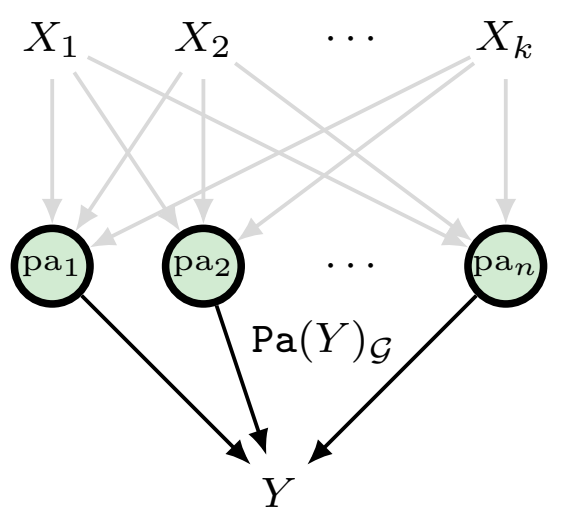
\begin{figure}[t]
    \centering
    \begin{tikzpicture}[x=1cm,y=1cm,font=\scriptsize,
        >={Latex[width=1.4mm,length=1.7mm]},line width=0.25mm,
        RR/.style={draw=none,circle,inner sep=0mm,minimum size=5mm},
        fixed/.style={RR,draw,line width=0.5mm,fill=bettergreen!25}]
        \centering
        \node[RR] (X1) at (-1.5,2.5) {$X_1$};
        \node[RR] (X2) at (-0.5,2.5) {$X_2$};
        \node[RR] (Xk) at (1.5,2.5) {$X_k$};
        \node at (0.5,2.5) {$\cdots$};
        \node[fixed] (P1) at (-1.5,1) {\tiny pa$_1$};
        \node[fixed] (P2) at (-0.5,1) {\tiny pa$_2$};
        \node[fixed] (Pn) at (1.5,1) {\tiny pa$_n$};
        \node at (0.5,1) {$\cdots$};
        \node[RR] (Y) at (0,-0.5) {$Y$};
        \foreach \x in {X1,X2,Xk}
            \foreach \p in {P1,P2,Pn}
                \draw[->,gray!30] (\x) -- (\p);
        \foreach \p in {P1,P2,Pn}
            \draw[->] (\p) -- (Y);
        \node at (0.3,0.5) {$\Pa(Y)_{\mc{G}}$};
    \end{tikzpicture}
    \caption{\small In a Markovian causal diagram, intervening on all parents of $Y$ suffices for optimality. }
    \label{fig:ctfscb markovian}
\end{figure}

\Cref{fig:ctfscb markovian} illustrates this comparison. An upstream intervention changes the distribution of the inputs reaching $Y$. An intervention on $\Pa(Y)_{\mc{G}}$ can instead fix those inputs to the values most favorable to the reward. 

\begin{proposition}\label{prop: ctf-pomis unconfounded}
    If $Y$ is not confounded with $\An(Y)_{\mc{G}} \setminus \{Y\}$, the ordinary intervention set $\Pa(Y)_{\mc{G}}$ is the only \ctf-POMIS up to equivalence. 
\end{proposition}

The proof is given in \Cref{sec: proof of prop: ctf-pomis unconfounded}.

\begin{corollary}\label{cor: ctf-pomis markovian}
    If $\mc{G}$ is Markovian, meaning that it has no bidirected edges, the ordinary intervention set $\Pa(Y)_{\mc{G}}$ is the only \ctf-POMIS up to equivalence. 
\end{corollary}
\begin{proof}[Proof of \Cref{cor: ctf-pomis markovian}]
    \Cref{prop: ctf-pomis unconfounded} applies because a Markovian diagram has no unobserved confounding. 
\end{proof}

Thus, in the Markovian case it suffices to intervene on the parents of $Y$. With unobserved confounding, a parent counterfactual may carry information about the exogenous inputs of the reward. Fixing that parent to a constant removes this dependence, so the preceding averaging argument need not apply. For example, in \Cref{fig:ctfscb counterfactual improvement}, the best ordinary intervention on the parents $\Pa(Y)_{\mc{G}} = \{W, X\}$ has mean $0.82$, below the mean $0.8848$ of $\{W_{w^\ast}, Z_{w^\dagger}\}$. 

\subsection{Graphical Characterization of CTF-POMISs}

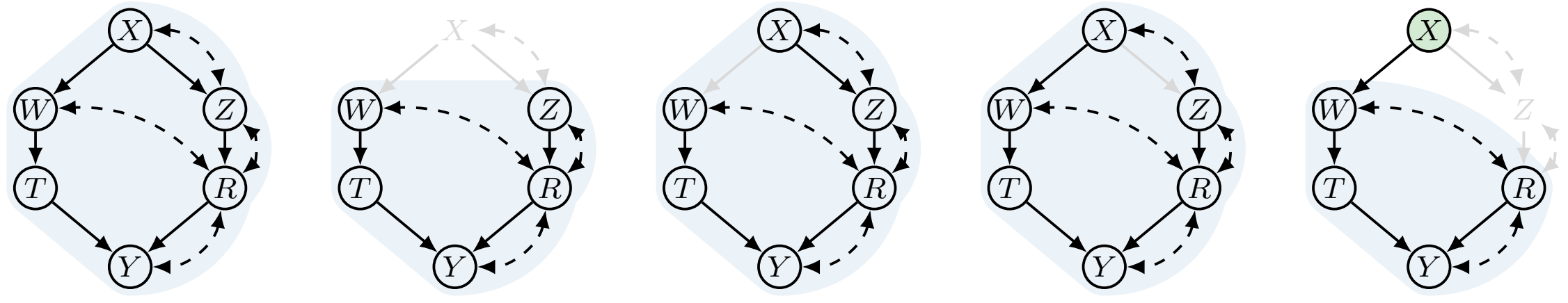
\begin{figure}[t]
    \centering
    \tikzset{ctfpomisarea/.style={overlay,draw=betterblue!10,fill=betterblue!10,
        line width=5.5mm,line cap=round,line join=round}}
    \newcommand{\ctfpomisallterritory}{
        \begin{pgfonlayer}{back}
            \draw[ctfpomisarea]
                (X.center) -- (W.center) -- (T.center) -- (Y.center)
                to[bend right=40] (R.center) to[bend right=50] (Z.center)
                to[bend right=40] (X.center) -- cycle;
        \end{pgfonlayer}
    }
    \newcommand{\ctfpomiswithoutXterritory}{
        \begin{pgfonlayer}{back}
            \draw[ctfpomisarea]
                (W.center) -- (T.center) -- (Y.center)
                to[bend right=40] (R.center) to[bend right=50] (Z.center)
                -- cycle;
        \end{pgfonlayer}
    }
    \newcommand{\ctfpomisRzterritory}{
        \begin{pgfonlayer}{back}
            \draw[ctfpomisarea]
                (W.center) -- (T.center) -- (Y.center)
                to[bend right=40] (R.center) to[bend right=25] (W.center)
                -- cycle;
        \end{pgfonlayer}
    }
    \begin{minipage}[b]{0.19\textwidth}\centering
    \subfloat[$\mc{G}$]{
    \begin{tikzpicture}[x=0.9cm,y=.75cm,>={Latex[width=1.4mm,length=1.7mm]},
        font=\scriptsize,
        line width=0.25mm,
        RR/.style={draw=none,circle,inner sep=0mm, minimum size=4mm, font=\scriptsize},
        transp/.style={regime,fill=betterred,minimum size=1mm,draw=none,inner sep=1mm},
        regime/.style={rectangle,draw=betterred,very thick,draw=betterred,minimum size=1mm},
        regime line/.style={draw=betterred, very thick},
        rotate = 0]\centering
        \node[RR,draw,fill=none] (X) at (0,3) {$X$};
        \node[RR,draw,fill=none] (W) at (-1,2) {$W$};
        \node[RR,draw,fill=none] (Z) at (1,2) {$Z$};
        \node[RR,draw,fill=none] (T) at (-1,1) {$T$};
        \node[RR,draw,fill=none] (R) at (1,1) {$R$};
        \node[RR,draw,fill=none] (Y) at (0,0) {$Y$};
        \draw[->] (X) -- (W);
        \draw[->] (X) -- (Z);
        \draw[->] (W) -- (T);
        \draw[->] (Z) -- (R);
        \draw[->] (T) -- (Y);
        \draw[->] (R) -- (Y);
        \draw[<->, dashed] (R) to[bend left = 40] (Y);
        \draw[<->, dashed] (R) to[bend right = 50] (Z);
        \draw[<->, dashed] (R) to[bend right = 25] (W);
        \draw[<->, dashed] (X) to[bend left = 40] (Z);
            \ctfpomisallterritory
    \end{tikzpicture}\label{fig:ctfscb pomis regimes_a}}
    \end{minipage}
    \hfill
    \begin{minipage}[b]{0.19\textwidth}\centering
    \subfloat[$\mc{H}_{W_{x},Z_{x'}}$]{
    \begin{tikzpicture}[x=0.9cm,y=.75cm,>={Latex[width=1.4mm,length=1.7mm]},
        font=\scriptsize,
        line width=0.25mm,
        RR/.style={draw=none,circle,inner sep=0mm, minimum size=4mm,font=\scriptsize},
        transp/.style={rectangle,fill=white,minimum size=1mm,draw=black,inner sep=0mm},
        regime/.style={rectangle, draw=betterred, thick,draw=betterred,minimum size=1mm},
        regime line/.style={draw=black,  thick},
        rotate = 0]\centering
        \node[RR, text=gray!30] (X) at (0,3) {\color{gray!30} $X$};
        \node[RR,draw,fill=none] (W) at (-1,2) {$W$};
        \node[RR,draw,fill=none] (Z) at (1,2) {$Z$};
        \node[RR,draw,fill=none] (T) at (-1,1) {$T$};
        \node[RR,draw,fill=none] (R) at (1,1) {$R$};
        \node[RR,draw,fill=none] (Y) at (0,0) {$Y$};
        \draw[->, draw = gray!30] (X) -- (W);
        \draw[->, draw = gray!30] (X) -- (Z);
        \draw[->] (W) -- (T);
        \draw[->] (Z) -- (R);
        \draw[->] (T) -- (Y);
        \draw[->] (R) -- (Y);
        \draw[<->, dashed] (R) to[bend left = 40] (Y);
        \draw[<->, dashed] (R) to[bend right = 50] (Z);
        \draw[<->, dashed] (R) to[bend right = 25] (W);
        \draw[<->, dashed, draw = gray!30] (X) to[bend left = 40] (Z);
            \ctfpomiswithoutXterritory
    \end{tikzpicture}\label{fig:ctfscb pomis regimes_b}}
    \end{minipage}
    \hfill
    \begin{minipage}[b]{0.19\textwidth}\centering
    \subfloat[$\mc{H}_{T_{x}}$]{
    \begin{tikzpicture}[x=0.9cm,y=.75cm,>={Latex[width=1.4mm,length=1.7mm]},
        font=\scriptsize,
        line width=0.25mm,
        RR/.style={draw=none,circle,inner sep=0mm, minimum size=4mm,font=\scriptsize},
        transp/.style={rectangle,fill=white,minimum size=1mm,draw=black,inner sep=0mm},
        regime/.style={rectangle, draw=betterred, thick,draw=betterred,minimum size=1mm},
        regime line/.style={draw=black,  thick},
        rotate = 0]\centering
        \node[RR,draw,fill=none] (X) at (0,3) {$X$};
        \node[RR,draw,fill=none] (W) at (-1,2) {$W$};
        \node[RR,draw,fill=none] (Z) at (1,2) {$Z$};
        \node[RR,draw,fill=none] (T) at (-1,1) {$T$};
        \node[RR,draw,fill=none] (R) at (1,1) {$R$};
        \node[RR,draw,fill=none] (Y) at (0,0) {$Y$};
        \draw[->, draw = gray!30] (X) -- (W);
        \draw[->] (X) -- (Z);
        \draw[->] (W) -- (T);
        \draw[->] (Z) -- (R);
        \draw[->] (T) -- (Y);
        \draw[->] (R) -- (Y);
        \draw[<->, dashed] (R) to[bend left = 40] (Y);
        \draw[<->, dashed] (R) to[bend right = 50] (Z);
        \draw[<->, dashed] (R) to[bend right = 25] (W);
        \draw[<->, dashed] (X) to[bend left = 40] (Z);
            \ctfpomisallterritory
    \end{tikzpicture}\label{fig:ctfscb pomis regimes_c}}
    \end{minipage}
    \hfill
    \begin{minipage}[b]{0.19\textwidth}\centering
    \subfloat[$\mc{H}_{R_{x}}$]{
    \begin{tikzpicture}[x=0.9cm,y=.75cm,>={Latex[width=1.4mm,length=1.7mm]},
        font=\scriptsize,
        line width=0.25mm,
        RR/.style={draw=none,circle,inner sep=0mm, minimum size=4mm,font=\scriptsize},
        transp/.style={rectangle,fill=white,minimum size=1mm,draw=black,inner sep=0mm},
        regime/.style={rectangle, draw=betterred, thick,draw=betterred,minimum size=1mm},
        regime line/.style={draw=black,  thick},
        rotate = 0]\centering
        \node[RR,draw,fill=none] (X) at (0,3) {$X$};
        \node[RR,draw,fill=none] (W) at (-1,2) {$W$};
        \node[RR,draw,fill=none] (Z) at (1,2) {$Z$};
        \node[RR,draw,fill=none] (T) at (-1,1) {$T$};
        \node[RR,draw,fill=none] (R) at (1,1) {$R$};
        \node[RR,draw,fill=none] (Y) at (0,0) {$Y$};
        \draw[->] (X) -- (W);
        \draw[->, draw = gray!30] (X) -- (Z);
        \draw[->] (W) -- (T);
        \draw[->] (Z) -- (R);
        \draw[->] (T) -- (Y);
        \draw[->] (R) -- (Y);
        \draw[<->, dashed] (R) to[bend left = 40] (Y);
        \draw[<->, dashed] (R) to[bend right = 50] (Z);
        \draw[<->, dashed] (R) to[bend right = 25] (W);
        \draw[<->, dashed] (X) to[bend left = 40] (Z);
            \ctfpomisallterritory
    \end{tikzpicture}\label{fig:ctfscb pomis regimes_d}}
    \end{minipage}
    \hfill
    \begin{minipage}[b]{0.19\textwidth}\centering
    \subfloat[$\mc{H}_{R_{z}}$]{
    \begin{tikzpicture}[x=0.9cm,y=.75cm,>={Latex[width=1.4mm,length=1.7mm]},
        font=\scriptsize,
        line width=0.25mm,
        RR/.style={draw=none,circle,inner sep=0mm, minimum size=4mm,font=\scriptsize},
        transp/.style={rectangle,fill=white,minimum size=1mm,draw=black,inner sep=0mm},
        regime/.style={rectangle, draw=betterred, thick,draw=betterred,minimum size=1mm},
        regime line/.style={draw=black,  thick},
        rotate = 0]\centering
        \node[RR,draw,fill=bettergreen!25] (X) at (0,3) {$X$};
        \node[RR,draw,fill=none] (W) at (-1,2) {$W$};
        \node[RR, text=gray!30] (Z) at (1,2) {$Z$};
        \node[RR,draw,fill=none] (T) at (-1,1) {$T$};
        \node[RR,draw,fill=none] (R) at (1,1) {$R$};
        \node[RR,draw,fill=none] (Y) at (0,0) {$Y$};
        \draw[->, draw = black] (X) -- (W);
        \draw[->, draw = gray!30] (X) -- (Z);
        \draw[->] (W) -- (T);
        \draw[->, draw = gray!30] (Z) -- (R);
        \draw[->] (T) -- (Y);
        \draw[->] (R) -- (Y);
        \draw[<->, dashed] (R) to[bend left = 40] (Y);
        \draw[<->, dashed, draw = gray!30] (R) to[bend right = 50] (Z);
        \draw[<->, dashed] (R) to[bend right = 25] (W);
        \draw[<->, dashed, draw = gray!30] (X) to[bend left = 40] (Z);
            \ctfpomisRzterritory
    \end{tikzpicture}\label{fig:ctfscb pomis regimes_e}}
    \end{minipage}
    \caption{\small (a) A causal diagram. (b)--(e) Counterfactual regime graphs. Light blue regions mark MUCTs, and green fill marks IB variables. The empty borders in (b)--(d) make these actions \ctf-POMISs; the border $\{X\}$ in (e) excludes that action. }
    \label{fig:ctfscb pomis regimes}
\end{figure}

\paragraph{Counterfactual regime graph.} Recall that $\mc{M}_{\mf{X}_{\ast\ast}}$ fixes the selected parent inputs, and its graph $\mc{G}_{\mf{X}_{\ast\ast}}$ deletes the corresponding directed edges while retaining the bidirected edges. We now define $\mc{H}_{\mf{X}_{\ast\ast}} \triangleq \mc{G}_{\mf{X}_{\ast\ast}} [ \An(Y)_{\mc{G}_{\mf{X}_{\ast\ast}}}]$. 

For concreteness, consider the causal diagram in \Cref{fig:ctfscb pomis regimes_a} and the action $\mf{X}_{\ast\ast} = \{W_x, Z_{x'}\}$. The two counterfactuals assign separate inputs from $X$, giving the substitutions $f_{W}(X,\mf{U}_W) \Rightarrow f_{W}(x,\mf{U}_W)$ and $f_{Z}(X,\mf{U}_Z) \Rightarrow f_{Z}(x',\mf{U}_Z)$, respectively. Consequently, $X \to W$ and $X \to Z$ are deleted. The natural output of $X$ no longer reaches $Y$. Hence, the ancestral restriction removes $X$, as illustrated in \Cref{fig:ctfscb pomis regimes_b}. 

\Cref{fig:ctfscb pomis regimes_b,fig:ctfscb pomis regimes_c,fig:ctfscb pomis regimes_d,fig:ctfscb pomis regimes_e} show how the same construction changes with the action. Under $\mf{X}_{\ast\ast} = \{T_x\}$ in \Cref{fig:ctfscb pomis regimes_c}, only the input $X \to W$ is fixed to produce $T_x$, while the natural value of $X$ still reaches $Y$ through $Z$ and $R$. Under $\mf{X}_{\ast\ast} = \{R_x\}$ in \Cref{fig:ctfscb pomis regimes_d}, only $X \to Z$ is fixed, while $X \to W \to T \to Y$ remains. Thus $X$ is retained in both graphs. Under $\mf{X}_{\ast\ast} = \{R_z\}$ in \Cref{fig:ctfscb pomis regimes_e}, the substitution is instead $f_R(Z, \mf{U}_R) \Rightarrow f_R(z, \mf{U}_R)$. This removes $Z \to R$ and makes the natural output of $Z$ irrelevant to $Y$. Its vertex is omitted, while the exogenous inputs of $R$ remain unchanged.

\paragraph{MUCT and IB for counterfactual actions.} We reuse the c-components, minimal unobserved confounders' territory $\MUCT$, and interventional border $\IB$ from \Cref{def: MUCT for cd,def: IB for cd}. For $\mf{X}_{\ast\ast} = \{W_x,Z_{x'}\}$ in \Cref{fig:ctfscb pomis regimes_b}, the c-component containing $Y$ is $\cc(Y)_{\mc{H}_{\mf{X}_{\ast\ast}}} = \{Y, R, Z, W\}$. Including descendants adds $T$, thus: 
\begin{align}
    \text{For $\mf{X}_{\ast\ast} = \{W_x,Z_{x'}\}$,~} &\MUCT(\mc{H}_{\mf{X}_{\ast\ast}},Y) = \{W,Z,T,R,Y\} \\
    &\Rightarrow \IB(\mc{H}_{\mf{X}_{\ast\ast}},Y) = \emptyset. 
\end{align}
In \Cref{fig:ctfscb pomis regimes_c,fig:ctfscb pomis regimes_d}, the retained bidirected connection $X \bedge Z$ also includes $X$ in the c-component of $Y$. Adding $T$ by descendant closure therefore gives all six vertices, again leaving an empty border. 

In contrast, consider $\mf{X}_{\ast\ast} = \{R_z\}$ in \Cref{fig:ctfscb pomis regimes_e}. After removing $Z$, we obtain $\cc(Y)_{\mc{H}_{\mf{X}_{\ast\ast}}} = \{Y, R, W\}$. Including descendants adds $T$, and neither closure operation adds $X$. Thus: 
\begin{align}
    \text{For $\mf{X}_{\ast\ast} = \{R_z\}$,~} &\MUCT(\mc{H}_{\mf{X}_{\ast\ast}},Y) = \{W,T,R,Y\} \\
    &\Rightarrow \IB(\mc{H}_{\mf{X}_{\ast\ast}},Y) = \{X\}. 
\end{align}
The remaining input from $X$ enters the territory through $W$. Fixing it to $x$ adds the counterfactual $W_x$ to the action. Since the territory's exogenous inputs are independent of those generating $X$, conditioning on $X = x$ gives the same reward mean as supplying this value directly. Hence:
\begin{align}
    \mu_{R_z} 
    & = \sum_{x} \mb{E} [Y_{R_z} \mid X = x] P(x) \\
    &= \sum_{x} \mu_{R_z,W_x} P(x) \leq \mu_{\{R_z,W_x\}^\ast}. 
\end{align}
This shows why a nonempty border provides an additional input choice that weakly improves the best reward. An empty border leaves no such external input to fix. These examples suggest that possible optimality can be characterized by the absence of an interventional border. 

\begin{theorem}[Graphical characterization of \ctf-POMIS]\label{thm: ctf-pomis}
    A counterfactual action $\mf{X}_{\ast\ast}$ is a \ctf-POMIS with respect to $\langle \mc{G},Y\rangle$ if and only if $\IB(\mc{H}_{\mf{X}_{\ast\ast}},Y) = \emptyset$. 
\end{theorem}
The proof is given in \Cref{sec: proof of thm: ctf-pomis}. When the border is nonempty, fixing its inputs gives a counterfactual action whose optimized mean is at least as large. When the border is empty, the retained mechanisms and shared exogenous inputs can make this counterfactual uniquely optimal up to equivalence. 

When $Y$ is unconfounded with its ancestors, its minimal UC-territory is $\{Y\}$. Its border is empty precisely when every incoming reward argument has been fixed. These counterfactuals are equivalent to intervening on $\Pa(Y)_{\mc{G}}$. 

\section{Enumerating the CTF-POMIS Action Space}

The graphical characterization in \Cref{thm: ctf-pomis} identifies counterfactual sets that can be optimal. As in the preceding \unitnames{}, their subscript assignments are then treated as actions. 

\begin{algorithm}[t]
\small
\SetAlgoNoEnd
\DontPrintSemicolon
\SetKwInput{KwInput}{Input}
\SetKwInput{KwOutput}{Output}
\KwInput{A causal diagram $\mc{G}$ and a reward variable $Y$}
\KwOutput{\ctf-POMISs with their selected-input specifications}

\tcp{Retain mechanisms that can affect the reward.}
Set $\mc{H}=\mc{G}[\An(Y)_{\mc{G}}]$.\;
\BlankLine

\tcp{Choose which child mechanisms receive fixed inputs.}
\For{\rm each children choice $(\mf{X}_W)_{W\in\mf{V}(\mc{H})\setminus\{Y\}}\in\bigtimes_{W\in\mf{V}(\mc{H})\setminus\{Y\}}2^{\Ch(W)_{\mc{H}}}$}{
    \BlankLine
    \tcp{Record each selected input at its receiving mechanism.}
    Initialize $\pa[X]=\emptyset$ for every $X\in\mf{V}(\mc{H})$.\;

    \For{\rm each $W$ and each $X\in\mf{X}_W$}{
        Set $\pa[X]\gets\pa[X]\cup\{W\}$.\;
    }
    \BlankLine

    \tcp{Cut selected edges and remove irrelevant receivers.}
    Form $\mc{H}'$ by deleting $W\to X$ from $\mc{H}$ whenever $W\in\pa[X]$.\;
    Replace $\mc{H}'$ by $\mc{H}'[\An(Y)_{\mc{H}'}]$.\;
    Set $\pa[X]=\emptyset$ for each $X\notin\mf{V}(\mc{H}')$.\;
    \BlankLine

    \tcp{Keep exactly the choices satisfying \Cref{thm: ctf-pomis}.}
    \If{\rm $\IB(\mc{H}',Y)\ne\emptyset$}{
        \textbf{continue}\;
    }
    \BlankLine

    \tcp{Leave retained input values separately selectable.}
    Initialize the selected-input specification of $\mf{X}_{\ast\ast}$.\;

    \For{\rm each $X\in\mf{V}(\mc{H}')$ in topological order}{
        \For{\rm each $W\in\pa[X]$}{
            Assign a separately selectable value to the argument $W$ of $f_X$.\;
        }
        Retain every other parent input and every exogenous input of $f_X$.\;
    }
    \BlankLine

    \tcp{Return each retained specification only once.}
    \If{\rm this selected-input specification has not been returned}{
        \textbf{yield} $\mf{X}_{\ast\ast}$ with $\pa[X]$\;
    }
}
\caption{\small Constructing the \ctf-POMIS action space.}
\label{alg: ctf-pomis}
\end{algorithm}

\paragraph{Enumerating counterfactual sets.} \Cref{alg: ctf-pomis} enumerates each choice of children in $\mc{H} = \mc{G}[\An(Y)_{\mc{G}}]$. For a parent $W$, the choice $\mf{X}_W \subseteq \Ch(W)_{\mc{H}}$ lists the mechanisms that will receive a fixed value in place of their input from $W$. The map $\pa[X]$ collects these choices at the receiving mechanism $f_X$. Thus the children choices enumerate the selected edges, and $\pa[X]$ records how to execute the resulting action. Deleting those edges gives the graph of the remaining causal dependence. Restricting this graph to the ancestors of $Y$ then removes mechanisms that cannot affect the reward. 

\paragraph{An accepted choice.} Consider the causal diagram in \Cref{fig:ctfscb pomis regimes_a}. Choose $\mf{X}_X = \{W,Z\}$ and choose no children for the other parents. The resulting map has $\pa[W] = \pa[Z] = \{X\}$, with every other entry empty. The algorithm deletes $X \to W$ and $X \to Z$. Since $X$ then has no directed path to $Y$, the ancestral restriction removes $X$, giving \Cref{fig:ctfscb pomis regimes_b}. Both selected inputs remain in the specification: their receiving mechanisms, $f_W$ and $f_Z$, still affect $Y$. The rule for pruning a selected input therefore concerns its receiving mechanism, even when its source vertex has disappeared from the graph.  

The retained graph has an empty interventional border, so this choice is returned. Assigning $x$ to the input of $f_W$ and $x'$ to the input of $f_Z$ gives the counterfactuals $\{W_x, Z_{x'}\}$. These two values are independently selectable. 

\paragraph{A rejected choice and a redundant specification.} Now choose only $\mf{X}_Z = \{R\}$. This gives $\pa[R] = \{Z\}$ and, after assigning its value, the action $\{R_z\}$. Cutting $Z \to R$ removes $Z$ from the reward ancestry but retains its fixed input at $f_R$. The resulting graph is \Cref{fig:ctfscb pomis regimes_e}. Since its border is not empty, the algorithm rejects this counterfactual. Adding the choice $\mf{X}_X = \{W\}$ fixes the remaining input from this border into the territory. The resulting counterfactual set $\{W_x, R_z\}$ has empty border and is returned. 

If a retained specification passes the condition of \Cref{thm: ctf-pomis}, the algorithm returns it only once, regardless of how many original choices produce it. The returned action $\mf{X}_{\ast\ast}$ retains its selected-input specification and may contain nested counterfactuals. All subscript assignments of these returned counterfactuals form $\mc{A}_{\ctf}^\star$. 

\begin{theorem}\label{thm: enumerating ctf-pomis}
    \Cref{alg: ctf-pomis} returns exactly the \ctf-POMISs with respect to $\langle \mc{G}, Y \rangle$. 
\end{theorem}
\begin{proof}[Proof of \Cref{thm: enumerating ctf-pomis}]
    Every selection of parent inputs occurs in the children-choice loop. Deleting its selected edges and restricting to the ancestors of $Y$ gives exactly $\mc{H}_{\mf{X}_{\ast\ast}}$. Removing fixed inputs at mechanisms outside this ancestral graph preserves the reward counterfactual on every unit, and each remaining selected input can affect $Y$ along an uncut path. The characterization in \Cref{thm: ctf-pomis} therefore accepts exactly the \ctf-POMISs. Identifying repeated specifications changes no available action. 
\end{proof}

\begin{corollary}\label{cor: ctf-pomis returns non equivalent set}
    No two counterfactual sets returned by \Cref{alg: ctf-pomis} are equivalent.
\end{corollary}
\begin{proof}[Proof of \Cref{cor: ctf-pomis returns non equivalent set}]
    The final check compares all retained entries $\pa[X]$ and returns each map once. It remains to show that distinct retained maps are not equivalent.

    Choose an edge selected in one map but not the other. Its receiving child has a retained directed path to $Y$ in the first map. If the other map retains this entire path, the chosen edge is followed by a path retained in both maps. Otherwise, take the last edge of that path selected by the other map. Again, the edge is fixed in one map and retained in the other, and its remaining path to $Y$ is retained in both. 

    Choose a binary SCM in which the source of this edge outputs a private $\Bern(\frac{1}{2})$ variable, and each mechanism along the edge and remaining path copies the preceding variable. Other inputs do not affect these mechanisms. Fixing the edge then gives a constant reward, whereas retaining it gives a non-constant reward on the same units, regardless of the other selected values. Thus the two maps cannot be equivalent. 
\end{proof}

\begin{corollary}\label{cor: ctf-pomis action space preserves optimal reward}
    For every SCM compatible with $\mc{G}$: 
    \begin{align}
        \max_{\mf{X}_{\ast\ast} \in \mc{A}_{\ctf}^\star} \mu_{\mf{X}_{\ast\ast}} = \max_{\mf{X}_{\ast\ast} \in \mc{A}_{\ctf}} \mu_{\mf{X}_{\ast\ast}}.
    \end{align}
\end{corollary}

The proof is given in \Cref{sec: proof of cor: ctf-pomis action space preserves optimal reward}.

\paragraph{Computational cost.} The initial ancestral restriction takes time linear in the size of $\mc{G}$. Let $n$ and $m$ be the numbers of vertices and edges in $\mc{H}$, and let $\hat{m}$ count its directed edges. The children-choice loop examines $2^{\hat{m}}$ choices. For each choice, deleting edges, finding reward ancestors, and computing the territory and border take $\mc{O}(n + m)$ time. 

If $\kappa$ distinct counterfactual sets are returned, comparing a retained specification against all previously returned specifications takes at most $\mc{O}(\kappa (n + m))$ time per choice. Thus a direct implementation takes $\mc{O}(2^{\hat{m}} ( \kappa + 1)(n + m))$ time and $\mc{O}((\kappa + 1)(n + m))$ space, including the returned specifications. These bounds enumerate counterfactual sets; assigning their constant values to form bandit arms is a subsequent step. 

\section{Experiments}\label{sec: ctfscb experiment}

\begin{figure}[t]
    \centering
    \begin{tikzpicture}[x=1cm,y=1cm,font=\small,
            >={Latex[width=1.4mm,length=1.7mm]},line width=0.3mm,
            RR/.style={draw=none,circle,inner sep=0mm,minimum size=4mm}]
            \node[RR] (A) at (0.5,1) {$A$};
            \node[RR] (B) at (0,0) {$B$};
            \node[RR] (Y) at (1,-1) {$Y$};
            \node[RR] (C) at (1.5,0) {$C$};
            \node[RR] (D) at (2,1) {$D$};
            \node[RR] (E) at (2.5,2) {$E$};
            \node[RR] (F) at (-1,1) {$F$};
            \draw[->] (F)--(B);
            \draw[->] (A)--(B);
            \draw[->] (A)--(C);
            \draw[->] (E)--(D);
            \draw[->] (D)--(C);
            \draw[->] (B)--(Y);
            \draw[->] (C)--(Y);
            \draw[<->,dashed] (B) to[bend right=40] (Y);
            \draw[<->,dashed] (B) to[bend left=25] (C);
            \draw[<->,dashed] (Y) to[bend right=50] (D);
        \end{tikzpicture}
    \caption{\small Causal diagram for Tasks 3(a) and 3(b). The two tasks use different structural functions and exogenous distributions on this graph.}
    \label{fig:ctfscb-experiment-graph-task3}
\end{figure}
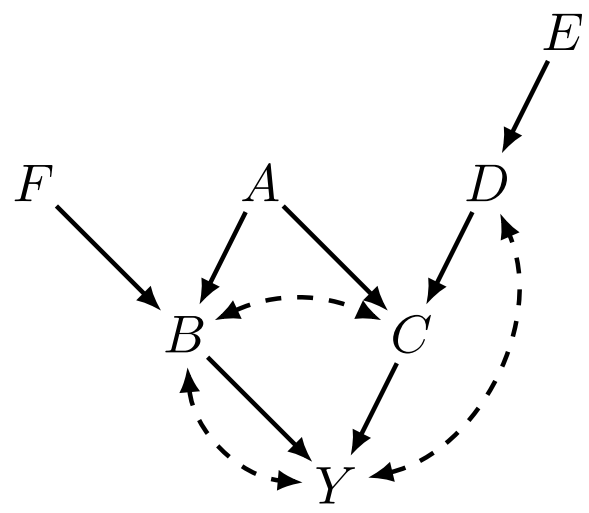

We evaluate how the counterfactual action space and its graphical reduction affect bandit learning. Each task uses a fixed binary SCM with independent Bernoulli exogenous variables, and the agent observes the reward of its selected action on a fresh unit at each round. Here the agent is given the causal diagram and may perform the counterfactual experiments admitted by $\mc{A}_{\ctf}$. Task~1 uses the causal diagram in \Cref{fig:ctfscb counterfactual improvement_a}, and Task~2 uses the diagram in \Cref{fig: realizable_b}. Tasks~3(a) and~3(b) share the diagram in \Cref{fig:ctfscb-experiment-graph-task3}. We use one SCM for each of the first two tasks and two SCMs for Task 3, giving four experimental cases. Their complete specifications appear in \Cref{sec: ctfscb experimental SCMs}. 

\begin{table}[t]
    \centering
    \caption{\small Candidate action spaces. Each strategy column gives the number of sets / arms. For the counterfactual strategies, sets are distinct selections of reward-relevant parent inputs. Tasks 3(a) and 3(b) use the same causal diagram and therefore have the same counts.}
    \label{tab:ctfscb-experiment-design}
    \small
    \setlength{\tabcolsep}{8pt}
    \begin{tabular}{ccrrrr}
        \toprule
        \bf Task & $\lvert\mf{V}\rvert$ & $T$ & \bf CTF-POMIS & \bf CTF-MIS & \bf POMIS \\
        \midrule
        1 & 4 & 10k & 2 / 8 & 8 / 21 & 2 / 6 \\
        2 & 5 & 10k & 6 / 36 & 17 / 67 & 4 / 16 \\
        3(a), 3(b) & 7 & 100k & 5 / 52 & 35 / 187 & 5 / 36 \\
        \bottomrule
    \end{tabular}
\end{table}

\paragraph{Candidate action spaces.} We compare three strategies. The \texttt{CTF-POMIS} strategy uses the actions returned by \Cref{alg: ctf-pomis}; each selected parent argument has its own binary value. Identical choices of fixed parent inputs and their assigned values are counted once.

The \texttt{CTF-MIS} baseline covers all realizable \ctf-MIS actions, including nested counterfactuals. It enumerates the selected parent inputs as in \Cref{alg: ctf-pomis}, but omits the interventional border test. Every selection entering a mechanism required to evaluate $Y$ is retained, together with all binary assignments. This includes ordinary interventions and the empty action. By \Cref{prop: cover realizable nested counterfactuals}, every realizable nested action can be evaluated through these parent-input choices; removing redundant outer assignments preserves its reward. Different expressions giving the same fixed inputs and values are counted once. Both counterfactual strategies count the same choice of fixed inputs and values only once. Their comparison measures the exploration savings from the graphical \ctf-POMIS condition over the complete nested \ctf-MIS baseline.

Finally, \texttt{POMIS} retains all assignments to ordinary POMISs, as in \Cref{chap:scb}. \Cref{tab:ctfscb-experiment-design} reports the numbers of sets and arms. 

\paragraph{Solvers and evaluation.}
Each action space is paired with Thompson sampling (TS) \citep{thompson1933likelihood} and Bernoulli KL-UCB \citep{garivier2011kl,cappe2013kullback}, using the same settings as \Cref{sec: scbmec experiment}. TS uses a $\mathtt{Beta}(1,1)$ prior for each arm. KL-UCB initially selects each arm once and subsequently uses $f(s) = \log s + 3\log\log s$ for $s\geq3$, with $f(s)=1$ otherwise, where $s$ is the number of completed rounds. Thus, after initialization, the decision at round $t$ uses $f(t-1)$. Tasks~1 and~2 use $T=10{,}000$ rounds, and Tasks~3(a) and~3(b) use $T=100{,}000$. Each of the 24 task--strategy--solver configurations is repeated $1{,}000$ times. 

\begin{figure}[t]
    \centering
    \captionsetup[subfigure]{skip=2pt}
    {\color{black}\includegraphics[width=\linewidth]{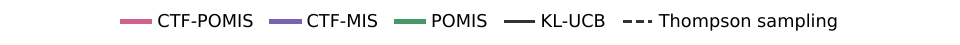}}
    \begin{subfigure}[b]{0.5125\linewidth}
        \centering
        {\color{black}\includegraphics[width=\linewidth]{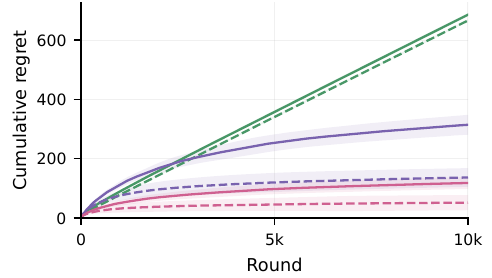}}
        \caption{Task~1}
        \label{fig:ctfscb-experiment-task1}
    \end{subfigure}%
    \begin{subfigure}[b]{0.4875\linewidth}
        \centering
        {\color{black}\includegraphics[width=\linewidth]{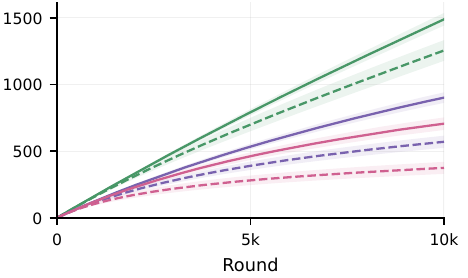}}
        \caption{Task~2}
        \label{fig:ctfscb-experiment-task2}
    \end{subfigure}
    \par\vspace{2pt}
    \begin{subfigure}[b]{0.5125\linewidth}
        \centering
        {\color{black}\includegraphics[width=\linewidth]{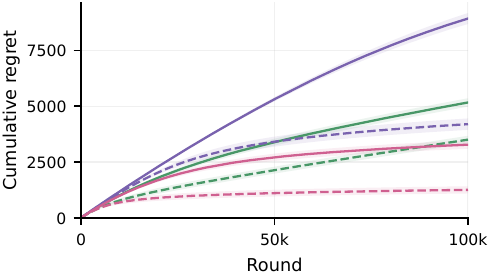}}
        \caption{Task~3(a)}
        \label{fig:ctfscb-experiment-task3a}
    \end{subfigure}%
    \begin{subfigure}[b]{0.4875\linewidth}
        \centering
        {\color{black}\includegraphics[width=\linewidth]{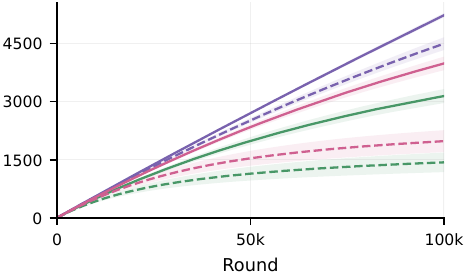}}
        \caption{Task~3(b)}
        \label{fig:ctfscb-experiment-task3b}
    \end{subfigure}
    \caption{\small Curves show mean cumulative regret over $1{,}000$ independent runs; shaded regions show one standard deviation. Every strategy uses the same counterfactual optimum as its regret benchmark.}
    \label{fig:ctfscb-experiment-main}
\end{figure}

The reward means are computed exactly from the fixed SCMs by summing over their exogenous assignments. These means generate Bernoulli rewards and evaluate cumulative regret; the solvers use only the selected arms and observed rewards. All strategies are evaluated against the same optimal counterfactual mean $\mu_{\mf{X}_{\ast\ast}^{\star}}$, using the cumulative regret defined earlier in this \unitname{}. In particular, the ordinary POMIS baseline is compared with the counterfactual optimum even when its own best arm has a smaller mean. Curves in \Cref{fig:ctfscb-experiment-main} show mean cumulative regret, and shaded regions show one population standard deviation across repeated runs. Displayed reward means in \Cref{sec: ctfscb experimental SCMs} are rounded to at most three decimal places. 

\paragraph{Tasks~1 and~2: enlarging and reducing the action space.}
Task~1 corresponds to the causal diagram in \Cref{fig:ctfscb counterfactual improvement}. Its best ordinary intervention has mean $0.82$, while separately choosing the two values supplied by $W$ raises the optimal mean to $0.8848$. The \texttt{CTF-POMIS} strategy retains this optimum with $8$ arms, compared with $21$ for the \texttt{CTF-MIS} baseline. Task~2 uses the causal diagram in \Cref{fig: realizable_b}, where realizability constrains the simultaneous use of counterfactuals. For its fixed SCM, the optimal counterfactual and ordinary means are approximately $0.6759$ and $0.5797$, respectively. Both counterfactual strategies retain the former mean, while \texttt{CTF-POMIS} reduces the number of arms from $67$ to $36$.

\paragraph{Task 3: whether ordinary interventions attain the optimum.}
The two seven-variable SCMs share a causal diagram and therefore the same candidate action spaces: $52$ arms for \texttt{CTF-POMIS}, $187$ for the \texttt{CTF-MIS} baseline, and $36$ for \texttt{POMIS}. Their structural functions determine whether ordinary interventions can attain the counterfactual optimum. In Task~3(a), the optimal counterfactual and ordinary means are approximately $0.6180$ and $0.5932$, respectively, leaving a positive gap of about $0.0248$ per round. In Task~3(b), all three spaces attain the same optimal mean, approximately $0.4365$. The latter case isolates the exploration cost of the larger counterfactual spaces when ordinary interventions already suffice.

For a strategy restricted to ordinary interventions, every selected arm has mean at most $\mu_{\mf{x}^{\star}}$. Consequently, $R_T^{\mc{A}_{\ctf}} \geq T (\mu_{\mf{X}_{\ast\ast}^{\star}}-\mu_{\mf{x}^{\star}})$. The positive gaps in Tasks~1,~2, and~3(a) therefore impose a linear contribution to the ordinary strategy's regret, regardless of how quickly it learns its own best arm. This contribution vanishes in Task~3(b). 

\begin{table}[t]
    \centering
    \caption{\small Final cumulative regret (mean $\pm$ standard deviation across 1,000 runs). Tasks~1 and~2 use $T=10{,}000$; Tasks~3(a) and~3(b) use $T=100{,}000$. Bold entries mark the lowest mean for each task--solver pair.}
    \label{tab:ctfscb-experiment-regret}
    \small
    \setlength{\tabcolsep}{10pt}
    \begin{tabular}{clrrr}
        \toprule
        \bf Task & \bf Solver & \bf CTF-POMIS & \bf CTF-MIS & \bf POMIS \\
        \midrule
        \multirow{2}{*}{1} & TS & \regret{\boldsymbol{51.4}}{25.7} & \regret{136.2}{31.7} & \regret{666.2}{3.5} \\
         & KL-UCB & \regret{\boldsymbol{117.7}}{20.2} & \regret{314.2}{33.6} & \regret{685.9}{5.5} \\
        \addlinespace[3pt]
        \multirow{2}{*}{2} & TS & \regret{\boldsymbol{374.7}}{45.2} & \regret{570.3}{45.7} & \regret{1255.1}{77.4} \\
         & KL-UCB & \regret{\boldsymbol{706.1}}{46.9} & \regret{901.4}{39.7} & \regret{1488.8}{47.2} \\
        \addlinespace[3pt]
        \multirow{2}{*}{3(a)} & TS & \regret{\boldsymbol{1252.4}}{161.5} & \regret{4195.1}{256.2} & \regret{3502.2}{171.9} \\
         & KL-UCB & \regret{\boldsymbol{3275.9}}{185.2} & \regret{8924.4}{246.9} & \regret{5166.2}{187.9} \\
        \addlinespace[3pt]
        \multirow{2}{*}{3(b)} & TS & \regret{1979.8}{282.8} & \regret{4495.5}{164.4} & \regret{\boldsymbol{1432.5}}{252.2} \\
         & KL-UCB & \regret{3985.4}{174.5} & \regret{5229.9}{57.0} & \regret{\boldsymbol{3143.4}}{180.0} \\
        \bottomrule
    \end{tabular}
\end{table}

\paragraph{Observed learning performance.}
\Cref{tab:ctfscb-experiment-regret} reports the simulation results. The \texttt{CTF-POMIS} strategy has lower mean cumulative regret than the \texttt{CTF-MIS} baseline in all eight task--solver comparisons. In Task~1, its regret is $37.7\%$ and $37.5\%$ of the baseline under TS and KL-UCB, respectively. In Task~3(a), these ratios are $29.9\%$ and $36.7\%$. The \texttt{CTF-POMIS} strategy also has lower mean regret than ordinary POMIS under both solvers in Tasks~1,~2, and~3(a). Ordinary POMIS has the lowest mean regret under both solvers in Task~3(b), where it retains an optimal arm with fewer candidates. These results illustrate both the gain from additional counterfactual actions and the exploration savings from graphical reduction on the fixed SCMs considered here. 

\jmlrsection{Conclusion}\label{chap:conclusion}
Across the three settings studied in this \workname{}, causal structure clarifies which decisions merit exploration even when their rewards remain unknown. Under structural uncertainty, graphical criteria support action reduction over compatible models. Information from source environments constrains target rewards through transferable mechanisms and dominance relations. Counterfactual actions expand the available decisions when the required experiments are realizable. These results separate the construction of an action space from learning the rewards of its members, while preserving an optimal action under the stated assumptions.

The scope of these guarantees is determined by their inputs and assumptions. The Markov equivalence experiments use oracle PAGs, and the transportability analysis assumes correct structural information and valid population reward bounds. The counterfactual formulation uses finite endogenous domains and specified experimental capabilities. The numerical studies use fixed finite SCMs and do not establish performance across arbitrary environments. Moreover, graphical action reduction does not eliminate the cost of constructing the reduced space: candidate enumeration and searches over compatible structures can still grow exponentially.

Several questions remain open. A direct graphical criterion for DMIS membership in PAGs and the complexity of deciding that membership deserve further study, as discussed in \Cref{sec:dmis-char}. Another direction is to propagate uncertainty from learned causal structures and estimated source distributions into action pruning and online confidence bounds. Extending the analysis to richer domains and to settings in which counterfactual experiments have unequal costs would help connect these guarantees to a wider range of decision problems.

\clearpage
\appendix
\jmlrsection{Supporting Results for Markov Equivalence}
\section{Background Results}\label{sec: background results}
Throughout this section, $\mc{P}$ is a PAG without selection bias, $\mc{M} \in [\mc{P}]$ is a represented MAG, and $\mc{G}$ is a causal diagram whose corresponding MAG is $\mc{M}$. We denote the observed vertex set by $\mf{V}$ and take $\mf{A} \subseteq \mf{V}$ whenever an induced graph is used. Visibility in an induced graph is interpreted with the full graph's represented model class held fixed, as defined in the main text. A PMG $\mc{Q}$ is assumed to be obtained from $\mc{P}$ by valid local transformations, applying the orientation rules in \Cref{thm: wang2023 thm 1} until no further orientation is possible after each transformation.  

\subsection{Background Results in \texorpdfstring{\citet{zhang2006causal,Zhang2008b,zhang2008a}}{Zhang (2006, 2008a,b)}}

\begin{lemma}[A lemma in the proof of Lemma~5.1.7 in \citet{zhang2006causal}; see also the proof of Lemma~7 in \citet{jaber2018causal}]\label{lem: zhang2006 lem 0}
    If $p = \langle X,\cdots ,Z,V,Y \rangle$ is a discriminating path from $X$ to $Y$ for $V$ in a MAG $\mc{M}$, and the corresponding subpath between $X$ and $V$ in $\mc{P}$ is (also) a collider path, then the path corresponding to $p$ in $\mc{P}$ is also a discriminating path for $V$. 
\end{lemma}

\begin{lemma}[Lemma A.1 in \citet{Zhang2008b} and Lemma 5 in \citet{jaber2018causal}]\label{lem: jaber2018a lem 5}
    For any three nodes $A, B, C \in \mf{A}$, if $A \adedge B \caedge C$ in $\mc{P}[\mf{A}]$, then there is an edge between $A$ and $C$ with an arrowhead at $C$, namely, $A \adedge C$. Furthermore, if the edge between $A$ and $B$ is $A \to B$, then the edge between $A$ and $C$ is either $A \to C$ or $A \cdedge C$ (i.e., it is not $A \bedge C$).
\end{lemma}

\begin{lemma}[Lemma 3.3.2 in \citet{zhang2006causal}]\label{lem: zhang2006b lem 3.3.2}
    For any two distinct vertices $A$ and $B$ connected by a circle path in $\mc{P}$, the following properties hold: 
    \begin{itemize}[leftmargin=10mm]
        \item [1.] If there is an edge between $A$ and $B$, the edge is not into $A$ or $B$;

        \item [2.] For any other node $C$, $C \adedge A$ if and only if $C \adedge B$. Furthermore, $C \bedge A$ if and only if $C \bedge B$.
    \end{itemize}
\end{lemma}
\vspace{0.5em}

\begin{lemma}[Theorem 2 in \citet{Zhang2008b}]\label{lem: zhang2008b thm 2}
    Let $\mc{M}$ be the graph resulting from the following procedure applied to $\mc{P}$.
    \begin{itemize}[leftmargin=17mm]
        \item[Step 1.] Replace all partially directed edges $(~\cdedge)$ in $\mc{P}$ with directed edges $(\to)$.
        
        \item[Step 2.] Orient every circle component of $\mc{P}$ into a DAG with no unshielded colliders.
    \end{itemize}
Then, the resulting graph $\mc{M}$ conforms to $\mc{P}$.
\end{lemma}
\vspace{0.5em}

\begin{lemma}[Lemma B.1 in \citet{Zhang2008b}]\label{lem: zhang2008b lem b.1}
    If $p$ is a possibly directed path from $A$ to $B$ in $\mc{P}$, then some subsequence of $p$ forms an uncovered possibly directed path from $A$ to $B$ in $\mc{P}$.
\end{lemma}

\begin{lemma}[Lemma B.2 in \citet{Zhang2008b}]\label{lem: zhang2008b lem b.2}
    If $p = \langle V_0 (=A),\cdots, V_n(=B) \rangle$ with $n \geq 2$ is an uncovered possibly directed path from $A$ to $B$ in $\mc{P}$, and $V_{i-1} \adedge V_i$ for some $i \in \{1,\cdots, n\}$, then $V_{j-1} \to V_j$ for all $j \in \{i+1, \cdots, n\}$.
\end{lemma}

\begin{lemma}[Lemma B.4 in \citet{Zhang2008b}]\label{lem: zhang2008b lem b.4}
    If there is a possibly directed path from $A$ to $B$ in $\mc{P}$, then the edge between $A$ and $B$, if any, is not into $A$.
\end{lemma}

\begin{lemma}[Lemma B.5 in \citet{Zhang2008b}]\label{lem: zhang2008b lem b.5}
    If there is a possibly directed path from $A$ to $B$ that is into $B$ in $\mc{P}$, then every uncovered possibly directed path from $A$ to $B$ is into $B$.
\end{lemma}

\begin{lemma}[Lemma B.7 in \citet{Zhang2008b}]\label{lem: zhang2008b lem b.7}
     If there is a circle path between two adjacent vertices in $\mc{P}$, then the edge between the two vertices is a circle edge. 
\end{lemma}

\begin{lemma}[Lemma 9 in \citet{zhang2008a}]\label{lem: zhang2008a lem 9}
    Let $\mc{G}$ be a DAG over $\mf{V} \cup \mf{U}$, and $\mc{M}$ be the MAG over $\mf{V}$ that represents the DAG. For any $A,B \in \mf{V}$, if $A \in \An(B)_{\mc{G}}$, and there is an inducing path relative to $\mf{U}$ between $A$ and $B$ that is into $A$ in $\mc{G}$, then there is a directed edge $A \to B$ in $\mc{M}$ that is invisible.
\end{lemma}

\begin{lemma}[Lemma 10 in \citet{zhang2008a}]\label{lem: zhang2008a lem 10}
    Let $\mc{M}$ be any MAG over $\mf{V}$, and $A \to B$ be any directed edge in $\mc{M}$. If $A \to B$ is invisible in $\mc{M}$, then there is a DAG $\mc{G}$ over $\mf{V} \cup \mf{U}$ whose MAG is $\mc{M}$ in which $A$ and $B$ share a latent parent. That is, there is a latent variable $U$ in the DAG such that $A \gets U \to B$ is a subgraph of the DAG $\mc{G}$.
\end{lemma}

\begin{lemma}[Contrapositive of Corollary 13 in \citet{zhang2008a}]\label{lem: zhang2008a cor 13}
    Let $X,Y \in \mf{V}$ be distinct, let $\mf{Z} \subseteq \mf{V} \setminus \{X,Y\}$, and let $\mf{X} \subseteq \mf{V}$. If there is a d-connecting path between $X$ and $Y$ given $\mf{Z}$ in $\mc{G}_{\overline{\mf{X}}}$, then there is an m-connecting path between $X$ and $Y$ given $\mf{Z}$ in $\mc{M}_{\overline{\mf{X}}}$. 
\end{lemma}

\subsection{Background Results in \texorpdfstring{\citet{maathuis2015generalized}}{Maathuis and Colombo (2015)}}

\begin{lemma}[Lemma 7.6 in \citet{maathuis2015generalized}]\label{lem: maathuis2015 lem 7.6}
    Let $\mc{P}$ be a PAG with exactly $k$ incident edges carrying an arrowhead at $X$, where $k \geq 0$. There exists a MAG $\mc{M}\in[\mc{P}]$ with exactly the same $k$ arrowheads at $X$; in particular, no circle at $X$ is replaced by an arrowhead.
\end{lemma}

\subsection{Background Results in \texorpdfstring{\citet{perkovic2018complete}}{Perkovic et al. (2018)}}

\begin{lemma}[Lemma 48 in \citet{perkovic2018complete}]\label{lem: perkovic 2018 lem48}
    Fix a vertex $X$ in a PAG $\mc{P}$. Construct $\mc{M} \in [\mc{P}]$ by replacing all $\cdedge$ edges with directed edges and orienting every circle component as a DAG without unshielded colliders, with every circle edge incident to $X$ oriented out of $X$. Then every edge that is $X \cedge Y$, $X \cdedge Y$, or an invisible $X \to Y$ in $\mc{P}$ becomes an invisible $X \to Y$ in $\mc{M}$.
\end{lemma}

\subsection{Background Results in \texorpdfstring{\citet{jaber2018causal,jaber2022causal}}{Jaber et al. (2018, 2022)}}
\begin{lemma}[Proposition 1 in \citet{jaber2018causal}]\label{lem: jaber2018a prop 1}
    Let $\mc{G}$ be any causal diagram consistent with $\mc{P}$, and let $X,Y \in \mf{A}$. If $X \in \An(Y)_{\mc{G}[\mf{A}]}$, then $X \in \Poss\An(Y)_{\mc{P}[\mf{A}]}$. 
\end{lemma}

\begin{lemma}[Lemma 4 in \citet{jaber2018causal}]\label{lem: jaber2018a lem 4}
    For every directed edge $X \to Y$ in $\mc{P}[\mf{A}]$, if it is visible in $\mc{P}$, then it remains visible in $\mc{P}[\mf{A}]$.
\end{lemma}

\begin{lemma}[Proposition 2 in \citet{jaber2018causal}]\label{lem: jaber2018a prop 2}
    Let $\mc{G}$ be any causal diagram consistent with $\mc{P}$, and let $X,Y \in \mf{A}$. If $X \in  \cc(Y)_{\mc{G}[\mf{A}]}$, then $X \in \pc(Y)_{\mc{P}[\mf{A}]}$.
\end{lemma}

\begin{algorithm}[t]
\DontPrintSemicolon
\SetKwInput{KwInput}{Input}
\SetKwInput{KwOutput}{Output}
\KwInput{$\mc{P}$, $\mf{A}\subseteq \mf{V}(\mc{P})$}
\KwOutput{Partial topological order over $\mc{P}[\mf{A}]$}

\setcounter{AlgoLine}{0}

\BlankLine 
\While{\rm there exists a circle component $\mf{B}$ in $\mc{P}[\mf{A}]$ such that every edge between $\mf{B}$ and $\mf{A}\setminus\mf{B}$ has an arrowhead at its endpoint in $\mf{B}$}
{

Extract $\mf{B}$ from $\mc{P}[\mf{A}]$\;

$\mf{A} \gets \mf{A} \setminus \mf{B}$
}
{\rm The partial order is $\mf{B}^1 \prec \cdots \prec \mf{B}^m$ in reverse order of the circle component extraction, i.e., $\mf{B}^1$ is the last circle component extracted and $\mf{B}^m$ is the first.}
\caption{\small Partial topological order (\textsc{pto}) \citep{jaber2018causal}. }\label{alg: pto}
\end{algorithm}

\begin{lemma}[Proposition 4 in \citet{jaber2018causal}]\label{lem: jaber2018 prop 4}
   \textsc{pto} (\Cref{alg: pto}) is sound over $\mc{P}[\mf{A}]$, in the sense that the partial order is valid with respect to $\mc{G}[\mf{A}]$, for any causal diagram $\mc{G}$ represented by $\mc{P}$. 
\end{lemma}

\begin{lemma}[Lemma~6 in \citet{jaber2018causal}]\label{lem: lem 6 in jaber2018}
    If there is a bidirected path $p$ between $X$ and $Y$ in $\mc{G}$, then there is a path $p'$ between $X$ and $Y$ in $\mc{M}$ over a subsequence of $p$ such that: 
    \begin{itemize}
            \item[1.] all non-endpoint nodes are colliders; and
            \item[2.] all directed edges on $p'$ are invisible.
        \end{itemize}
\end{lemma}

\begin{lemma}[Lemma~7 in \citet{jaber2018causal}]\label{lem: lem 7 in jaber2018}
If there is a path $p$ between $X$ and $Y$ in $\mc{M}$ such that: 
    \begin{itemize}
        \item[1.] all non-endpoint nodes are colliders; and
        \item[2.] all directed edges, if any, are invisible,
    \end{itemize}
then there is a path $p^\ast$ between $X$ and $Y$ in $\mc{P}$ over a subsequence of $p$ such that: 
    \begin{itemize}
        \item[1.] all non-endpoint nodes along the path are definite colliders; and
        \item[2.] none of the edges are visible. 
    \end{itemize}
\end{lemma}

\begin{lemma}[Lemma 6 in \citet{jaber2019causal}]\label{lem: jaber2019 lem 6}
    In $\mc{M}[\mf{A}]$, the following property holds:
    \begin{itemize}
        \item[] For any three vertices $A,B,C \in \mf{A}$, if $A \adedge B \to C$ and both edges are invisible, then we have $A \adedge C$ and the edge is invisible.
    \end{itemize}
\end{lemma}

\begin{lemma}[Lemma 18 in \citet{jaber2022causal}]\label{lem: jaber2022 lem 18}
    In $\mc{P}[\mf{A}]$, the following property holds:
    \begin{itemize}
    \item[] For any three vertices $A,B,C \in \mf{A}$, if $A \adedge B \qdedge C$ and both edges are invisible, then we have $A \adedge C$ and the edge is invisible. 
    \end{itemize}
\end{lemma}

\subsection{Background Results in \texorpdfstring{\citet{wang2023sound}}{Wang et al. (2023)}}

\begin{proposition}[Lemma 4 in \citet{wang2023sound}]\label{prop: wang2023 lem 4}
    Consider a MAG $\mc{M}$ consistent with both $\mc{Q}$ and a valid local transformation $\mf{C}_{X}^{\mc{Q}}$. Then $Y \in \Poss\De(X)_{\mc{Q}\setminus \mf{C}_{X}^{\mc{Q}}}$ \textit{if and only if} $Y\in \De(X)_\mc{M}$. 
\end{proposition}


\begin{lemma}[Lemma 2 in \citet{wang2023sound}]\label{lem: wang2023 lem 2}
    Let $\mc{Q}$ be obtained from a PAG by valid local transformations, applying the orientation rules in \Cref{thm: wang2023 thm 1} to closure after each transformation. Every possibly directed path in $\mc{Q}$ has an uncovered possibly directed subsequence with the same endpoints. 
\end{lemma}



\begin{lemma}[Lemma 16.1 in \citet{wang2023sound}]\label{lem: wang2023 lem 16.1}
    Let $\mc{Q}$ be obtained from a PAG by valid local transformations and orientation closure, as specified above. Replace every $\cdedge$ edge in $\mc{Q}$ with a directed edge and orient each circle component into a DAG without unshielded colliders. Every graph obtained by these simultaneous choices is a MAG in $[\mc{Q}]$. 
\end{lemma}


\section{Omitted Proofs}

\subsection{Constructing a Binary SCM}
\begin{lemma}[Binary SCM construction]\label{lem: binary scm construction} Let $\mc{G}$ be a causal diagram. There exists a binary SCM $\mc{S}^\ast$ conforming to $\mc{G}$ with the following property. If $\mf{Z} \subsetneq \mf{X} \subseteq \mf{V} \setminus \{Y\}$ and some $W \in \mf{X} \setminus \mf{Z}$ has a proper directed path to $Y$ with respect to $\mf{X}$, then $\mu_{\mf{X} = \boldsymbol{1}}^{\mc{S}^\ast} > \mu_{\mf{Z} = \boldsymbol{1}}^{\mc{S}^\ast}$. 
\end{lemma}
\begin{proof}[Proof of \Cref{lem: binary scm construction}]
    We first specify one SCM using only the causal diagram $\mc{G}$; none of its functions or distributions will depend on the intervention sets being compared. Here, binary refers to the observed variables; the exogenous variables need not be binary. Give each observed vertex $V$ a private exogenous variable $U_V \sim \Unif(0,1)$, which appears directly only in the equation for $V$. For each bidirected edge $e$, introduce a separate exogenous variable $L_e \sim \Bern(\frac{1}{2})$, which appears in the equations for both endpoints of $e$. Thus each $U_V$ is used only for $V$, whereas the $L_e$ variables represent the unobserved confounders shown by the bidirected edges. Choose all these exogenous variables to be mutually independent. 

    In the following sums, $e$ ranges over the bidirected edges of $\mc{G}$. For each $V$, choose strictly positive coefficients $\alpha_V$, $\beta_{PV}$, $\gamma_{eV}$ such that $\alpha_V + \sum_{P \in \Pa(V)_{\mc{G}}} \beta_{PV} + \sum_{e: V \in e} \gamma_{eV} <  1$. For example, assign each coefficient in this inequality the value $\frac{1}{2(1 + \lvert \Pa(V)_{\mc{G}} \rvert + \lvert \{e: V \in e\}  \rvert )}$. Their sum is then $\frac{1}{2}$. Define the structural equation for $V$ by: 
    \begin{align}
        V &= f_V(\Pa(V)_{\mc{G}} \cup \{U_V\} \cup \{L_e\}_{e: V \in e}) \\
            &= \indicate \big\{ 
            U_V \leq \alpha_V + \sum_{P \in \Pa(V)_{\mc{G}}} \beta_{PV} P + \sum_{e: V \in e} \gamma_{eV} L_e
        \big\}. 
    \end{align}
    The argument set in the first line specifies the variables whose values are supplied to $f_V$. Every parent value and every $L_e$ is either zero or one. Consequently, the threshold on the right-hand side is at least $\alpha_V > 0$ and at most the sum of the coefficients, which is strictly less than one. Acyclicity lets us evaluate these equations in a topological order, with each parent evaluated before its child, and every resulting observed variable is binary. 

    The equation for $V$ contains exactly its directed parents in $\mc{G}$, each with a positive coefficient. The only exogenous variables shared by different equations are the $L_e$ variables, one for each specified bidirected edge. Equivalently, each bidirected edge can be represented by an unobserved common parent of its two endpoints. Adding the directed edges $U_V \to V$ for the separate exogenous variables gives the DAG underlying this causal diagram. The SCM therefore conforms to $\mc{G}$. Fix this SCM as $\mc{S}^\ast$ for all comparisons below. 

    \textbf{Expected values under an intervention.} For any intervention set $\mf{S}$, write $\eta_V^{\mf{S}} = \mb{E}[V \mid do(\mf{S} = \boldsymbol{1})]$ in the fixed SCM. If $V \in \mf{S}$, its equation is replaced by the constant one, so $\eta_{V}^{\mf{S}} = 1$. If $V \notin \mf{S}$, its original mechanism remains. The private variable $U_V$ is independent of both the parent values and the shared variables appearing in that equation: it is independent of all other exogenous variables, and acyclicity prevents any parent of $V$ from depending on $U_V$. Conditional on those parent values and shared variables, the probability that $V = 1$ thus equals its threshold. 

    An intervention changes structural equations but leaves the exogenous distributions unchanged. Hence each $L_e$ still has mean $\frac{1}{2}$. For $V \notin \mf{S}$, taking the expectation of the threshold gives: 
    \begin{align}
        \eta_V^{\mf{S}} = \alpha_V + \sum_{P \in \Pa(V)_{\mc{G}}} \beta_{PV} \eta_P^{\mf{S}} + \frac{1}{2}\sum_{e: V \in e} \gamma_{eV}, \qquad 0 < \eta_V^{\mf{S}} < 1.
    \end{align}
    Only the independence of the private variable $U_V$ was needed for the conditional probability calculation. In particular, a parent of $V$ may depend on a shared $L_e$ and therefore be correlated with it. Linearity of expectation still gives the displayed formula; it does not require independence between the terms in the threshold. The two strict inequalities hold because the threshold stays strictly between zero and one for every assignment of the parents and shared variables. 

    \textbf{Comparing two intervention sets.} Now choose any $\mf{Z} \subsetneq \mf{X}$ satisfying the condition in the lemma, and let $\delta_V = \eta_{V}^{\mf{X}} - \eta_{V}^{\mf{Z}}$. There are three cases. If $V \in \mf{Z}$, both interventions set $V$ to one, thus $\delta_V = 0$. If $V \in \mf{X} \setminus \mf{Z}$, only the first intervention sets $V$ to one, whereas its mean under the second is strictly below one; thus $\delta_V = 1 - \eta_{V}^{\mf{Z}} > 0$. Finally, if $V \notin \mf{X}$, its structural equation is unchanged under both interventions. Subtracting the two expectation formulas cancels the constant and the terms involving the shared exogenous variables, leaving: 
    \begin{align}
        \delta_V = \sum_{P \in \Pa(V)_{\mc{G}}} \beta_{PV} \delta_P.
    \end{align}
    In a topological order, the differences for the parents have already been evaluated when this last formula is used. The first two cases give nonnegative differences directly, and the last case is a sum of nonnegative differences for the parents with positive coefficients. Induction therefore gives $\delta_V \geq 0$ at every vertex. 

    By the condition in the lemma, some $W \in \mf{X} \setminus \mf{Z}$ has a proper directed path $W(= V_0) \to V_1 \to \cdots \to V_k(=Y)$ with respect to $\mf{X}$. Every vertex after $W$ on this path lies outside $\mf{X}$ and hence also outside $\mf{Z}$. The displayed equation for the difference in expected values therefore applies at each such vertex. The terms for its other parents are nonnegative, thus $\delta_{V_i} \geq \beta_{V_{i-1}V_i} \delta_{V_{i-1}}$ for $i = 1, \cdots, k$. Applying these inequalities successively yields the explicit bound: 
    \begin{align}
        \delta_{Y} \geq \delta_W \prod_{i=1}^k  \beta_{V_{i-1}V_i} = (1 - \eta_{W}^{\mf{Z}})\prod_{i=1}^k  \beta_{V_{i-1}V_i} > 0.
    \end{align}
    The final inequality is strict because $\eta_{W}^{\mf{Z}} < 1$ and every coefficient on the path is positive. Since $\delta_Y = \mu_{\mf{X} = \boldsymbol{1}}^{\mc{S}^\ast} - \mu_{\mf{Z} = \boldsymbol{1}}^{\mc{S}^\ast}$, this is the required reward comparison. The argument applies to every pair $\mf{X}, \mf{Z}$ satisfying the condition in the lemma within the same fixed SCM, and all reward means are bounded because $Y$ is binary. 
\end{proof}

\subsection{Constructing a Causal Diagram from a MAG}
\begin{lemma}[Causal diagram with the same MUCT and IB]\label{lem: causal diagram construction}  
    Let $\mc{M}$ be a MAG and let $\mf{X} \subseteq \mf{V} \setminus \{Y\}$. There exists a causal diagram $\mc{G}$ whose corresponding MAG is $\mc{M}$ and whose directed edges are exactly those of $\mc{M}$ such that $\MUCT(\mc{G}_{\overline{\mf{X}}}, Y) = \MUCT(\mc{M}, Y, \mf{X})$ and $\IB(\mc{G}_{\overline{\mf{X}}}, Y) = \IB(\mc{M}, Y, \mf{X})$. 
\end{lemma}
\begin{proof}[Proof of \Cref{lem: causal diagram construction}]
    Let $\mf{H} = \An(Y)_{\mc{M} \setminus \mf{X}}$ and $\mf{R} = \MUCT(\mc{M},Y,\mf{X})$. The set $\mf{H}$ contains exactly the vertices that can reach $Y$ by a directed path avoiding $\mf{X}$. The set $\mf{R}$ is the MUCT of the MAG that we want to obtain as the MUCT of a causal diagram $\mc{G}$. We will retain every edge of $\mc{M}$ and add selected bidirected edges. The proof has two tasks: 
    \begin{itemize}
        \item showing that these additions give the desired territory; and

        \item showing that they preserve the corresponding MAG.
    \end{itemize}
    The second task requires considering all the added edges together. 

    All decisions about visibility refer to the original MAG. For a directed edge retained in $\mc{M}[\mf{H}]$, we use the causal model class represented by the full $\mc{M}$ and retain its latent variables, as in the definition of visibility for induced subgraphs. A visible edge in $\mc{M}$ remains visible when the observed vertices are restricted to $\mf{H}$. An invisible directed edge remains invisible because \Cref{lem: zhang2008a lem 10} provides a corresponding DAG in which its endpoints share a latent parent. This parent and its two edges remain when observed vertices outside $\mf{H}$ are removed. 

    \textbf{An observation about invisible directed edges.} We first establish a fact that will be used when checking preservation of the MAG. Suppose that $A \to B$ is invisible and that there is a bidirected path from $C$ to $A$ in $\mc{M}$ whose vertices are all parents of $B$. Then $C \to B$ is also invisible. 

    To see this, suppose instead that $C \to B$ were visible. By the graphical visibility criterion in \Cref{lem: visible}, there would be a vertex $V$ nonadjacent to $B$ and a collider path from $V$ into $C$ whose non-endpoint vertices are parents of $B$. This includes the possibility of a single edge from $V$ into $C$. Starting from $V$, follow this path up to its first intersection with the bidirected path, and then follow the latter path to $A$. The resulting path has no repeated vertices: the part before the first intersection is disjoint from the bidirected path.

    If the intersection is not $A$, both incident edges have arrowheads there; if it is $A$, the resulting path ends with an arrowhead at $A$. Every non-endpoint vertex of the resulting path is a parent of $B$. Thus it is a collider path from $V$ into $A$ satisfying the visibility criterion for $A \to B$. This contradicts the invisibility of $A \to B$ and proves the observation. 

    \textbf{Constructing the causal diagram.} Begin with $\mc{G} = \mc{M}$, viewed as a causal diagram. For the current $\mc{G}$, start with $Y$ and repeatedly include directed descendants and vertices connected by bidirected paths in $\mc{G}[\mf{H}]$, until no more vertices can be included. Call the resulting set $\mf{T}$. Equivalently, $\mf{T}$ is the least subset of $\mf{H}$ containing $Y$ that is closed under these two operations. As shown below, after intervention on $\mf{X}$ this is precisely the MUCT of the current causal diagram. 

    If $\mf{T} \neq \mf{R}$, choose a vertex $B \in \mf{T}$ for which there is an invisible directed edge $C \to B$ with $C \in \mf{H} \setminus \mf{T}$. For this one chosen $B$, add the bidirected edge $C \bedge B$ for \textbf{every} such $C$ at the same time, retaining the original directed edges $C \to B$ and all other edges. Recompute $\mf{T}$ and repeat this step until $\mf{T} = \mf{R}$. 

    The requirement to include every eligible parent of the chosen $B$ is part of the construction. Each such parent becomes connected to a vertex already in $\mf{T}$ and therefore enters $\mf{T}$ in this step. Once a vertex has entered $\mf{T}$, it remains there, because edges are only added. Consequently, a vertex $C$ can occur as the parent in an addition of $C \bedge B$ only once. In a later step, it cannot qualify as a parent outside the current territory. 

    For example, \Cref{fig:scbmec_mag_construction} illustrates this step with $\mf{X} = \{A,B\}$ and $\mf{H} = \{C,D,Y\}$. Initially the current territory is $\mf{T} = \{Y\}$. The directed edges $C \to Y$ and $D \to Y$ are invisible, thus choosing $Y$ requires adding both $C \bedge Y$ and $D \bedge Y$ before recomputing the territory. It then becomes $\mf{T} = \{C,D,Y\}$, whose parents outside the set are $\{A,B\}$. All original directed edges are retained in the constructed causal diagram; only the intervention removes edges into $\{A,B\}$ in the manipulated panels. 

    \textbf{The construction reaches the required territory.} We verify both that $\mf{T}$ never exceeds $\mf{R}$ and that another addition is possible whenever $\mf{T}$ is smaller than $\mf{R}$. Initially $\mf{T} \subseteq \mf{R}$, because $\mf{R}$ contains $Y$ and is closed under directed descendants and original bidirected connections in $\mc{M}[\mf{H}]$. Suppose this inclusion holds before a step choosing $B \in \mf{T}$. For every added $C \bedge B$, the original directed edge $C \to B$ is invisible, which gives:
    \begin{align}
        C \in \pc(B)_{\mc{M}[\mf{H}]} \subseteq \mf{R}.
    \end{align}
    Thus both endpoints of every added edge belong to $\mf{R}$. The directed descendant relation is unchanged, and the original bidirected connections also remain within $\mf{R}$ whenever they start there. Recomputing $\mf{T}$ therefore preserves $\mf{T} \subseteq \mf{R}$. 

    Now suppose that $\mf{T} \neq \mf{R}$. By construction, $\mf{T}$ is already closed under directed descendants and original bidirected connections in $\mc{M}[\mf{H}]$. It cannot also satisfy $\pc(\mf{T})_{\mc{M}[\mf{H}]}= \mf{T}$. Otherwise it would be a UC-territory in $\mc{M}[\mf{H}]$ containing $Y$ and properly contained in $\mf{R}$, contrary to the minimality of $\mf{R}$. Hence, by \Cref{def: pc}, there is a collider path in $\mc{M}[\mf{H}]$ from a vertex of $\mf{T}$ to a vertex outside $\mf{T}$, and none of its edges are visible. Traverse it from the vertex in $\mf{T}$ and consider the first edge crossing to the complement of $\mf{T}$. This edge cannot be bidirected, because $\mf{T}$ is closed under original bidirected connections. It cannot be directed from $\mf{T}$ to its complement, because $\mf{T}$ is closed under directed descendants. It must therefore be an invisible directed edge from a vertex outside $\mf{T}$ into a vertex $B \in \mf{T}$. This supplies the next choice required by the construction. 

    Every step adds at least one vertex to $\mf{T}$, and $Y \in \mf{T} \subseteq \mf{R} \subseteq \mf{H}$ throughout. The procedure therefore terminates with $\mf{T} = \mf{R}$ after at most $\lvert \mf{H} \rvert - 1$ steps. Since each vertex outside the initial territory can occur as $C$ in an added edge $C \bedge B$ with original directed edge $C \to B$ at most once, at most $\lvert \mf{H} \rvert - 1$ bidirected edges are added. 

    \textbf{The corresponding MAG is preserved.} We next check that the final $\mc{G}$ still has corresponding MAG $\mc{M}$. The following argument extends the single-edge argument in the proof of \Cref{lem: zhang2008a lem 10} to all edges added by the construction. All original adjacencies and ancestral relations among observed vertices are retained because no original edge is removed and no directed edge is added. It remains to show that the added bidirected edges do not create an adjacency between vertices that were nonadjacent in $\mc{M}$. Such a new adjacency would require an inducing path between those vertices in $\mc{G}$. 

    For the sake of contradiction, suppose that $p$ is an inducing path between vertices $S,T$ nonadjacent in $\mc{M}$. For a causal diagram, such a path has only colliders as non-endpoint vertices, each of which is an ancestor of $S$ or $T$. Among all these inducing paths, choose one with the fewest vertices; among paths with that number of vertices, choose one using the fewest added bidirected edges. Both parts of this choice will be used: a path with fewer vertices contradicts the first part, and a path with the same vertices but fewer added edges contradicts the second. The chosen path uses an added edge. Otherwise the same path, with every non-endpoint collider still an ancestor of $S$ or $T$, would already be an inducing path in $\mc{M}$, contradicting maximality. 

    We first rule out a particular arrangement of two consecutive added edges on $p$. Here consecutive means that no other added edge occurs between them; original edges may occur between them. Since all vertices between these two edges are non-endpoint colliders, the intervening original edges must be bidirected. The arrangement we rule out is:
    \begin{align}
        B \bedge A (=U_0) \bedge U_1 \bedge \cdots \bedge C (= U_k) \bedge D, \qquad A \to B \text{~and~} C \to D \text{~in~} \mc{M},
    \end{align}
    where the first and last bidirected edges were added and the middle ones were already in $\mc{M}$. Thus the two original directed edges point toward $B$ and $D$, away from the intervening bidirected segment. The case $k = 0$, with $A = C$, is included. The vertices $B$ and $D$ may be endpoints of $p$, but they are distinct because $p$ has no repeated vertices. 

    If $k = 0$, the two added edges have the same original parent $A = C$. They cannot occur because each parent participates in at most one addition, as shown above. Hence assume $k \geq 1$. Suppose first that $A \bedge B$ was added before $C \bedge D$. We show that every $U_i$ is a parent of $B$ in $\mc{M}$. The first vertex $U_0 (=A)$ is a parent by assumption. For the induction step, suppose $U_0, \cdots, U_{i-1}$ are parents of $B$. If $U_i$ were nonadjacent to $B$, the original collider path $U_i \bedge \cdots \bedge A$, whose non-endpoint vertices are parents of $B$, would make $A \to B$ visible. Therefore, $U_i$ is adjacent to $B$. 
    
    The edge cannot be $B \to U_i$: together with $U_{i-1} \to B$, it would make $U_{i-1}$ an ancestor of $U_i$ despite the original edge $U_{i-1} \bedge U_i$. The edge cannot be $U_i \bedge B$ either: replacing the segment from $B$ through $A$ to $U_i$ by this original bidirected edge would shorten $p$. All retained non-endpoint vertices would still be colliders and ancestors of $S$ or $T$, thus the shortened path would still be inducing. The only remaining possibility is $U_i \to B$, completing the induction. 

    In particular, $C$ and $A$ are joined by an original bidirected path whose vertices are all parents of $B$. The observation at the beginning of the proof therefore implies that $C \to B$ is invisible. Consider the step in which $A \bedge B$ was added. If $C$ was already in the territory at that time, it could not be eligible for a later addition of $C \bedge D$. If $C$ was outside the territory, then $C \in \mf{H} \setminus \mf{T}$ and its invisible directed edge into the chosen $B$ required $C \bedge B$ to be added in that same step. This would place $C$ in the territory, again preventing the later addition $C \bedge D$. Both possibilities contradict the construction. If $C \bedge D$ was added first, the same argument read from $D$ towards $B$ gives the symmetric contradiction. Finally, the two edges cannot have been added in the same step, because that step chooses a single vertex, whereas $B \neq D$. Thus the displayed arrangement cannot occur on $p$. 

    Now list the added edges in their order from $S$ to $T$, and consider the direction of each one's original directed edge in $\mc{M}$. The arrangement just excluded is exactly a left-pointing original directed edge followed by a right-pointing original directed edge at consecutive positions in this list. Consequently, if the first directed edge points to the left, the last directed edge must also point to the left; otherwise such a change would occur somewhere in the list. We can therefore read $p$ from one of its endpoints so that the first added edge is traversed from $A$ to $B$ with $A \to B$ in $\mc{M}$. Indeed, use the endpoint $S$ if its first directed edge points to the right, and otherwise read the path from $T$. Rename the chosen starting endpoint $S$ and the other endpoint $T$. All edges preceding this first added $A \bedge B$ are original edges. 

    If $A = S$, replace $A \bedge B$ on the path by the original directed edge $A \to B$. The mark at $A$ is irrelevant because $A$ is an endpoint, and the arrowhead at $B$ is unchanged. Hence every non-endpoint vertex remains a collider and an ancestor of $S$ or $T$. The result is an inducing path with the same number of vertices but one fewer added edge, contradicting the second part of our choice of $p$. This replacement is used only when the original parent $A$ is the path endpoint; it does not assume that every added edge touching an endpoint can be replaced in this way. 

    It remains to consider $A \neq S$. Write the original path segment from $S$ to $A$ as $\langle S, O_1, \cdots, O_k (=A) \rangle$. It is a collider path into $A$, and all edges between the $O_i$ are original bidirected edges. We show backwards that every $O_i$ is a parent of $B$. Starting with $O_k = A$, suppose that $O_{i+1}, \cdots, O_k$ are parents of $B$. The invisibility of $A \to B$ makes $O_i$ adjacent to $B$, since otherwise $O_i \bedge \cdots \bedge A$ would satisfy the visibility criterion for $A \to B$. The edge $B \to O_i$ would create an almost directed cycle consisting of $O_{i+1} \to B \to O_i$ and $O_i \bedge O_{i+1}$. The edge $O_i \bedge B$ would shorten $p$, and all remaining non-endpoint vertices would still be colliders and ancestors of $S$ or $T$. Therefore $O_i \to B$. This proves the claim for all vertices $O_1, \cdots, O_k$. 

    The same visibility criterion implies that $S$ is adjacent to $B$. Otherwise, the path $\langle S, O_1, \cdots, O_k \rangle$ would establish visibility of $A \to B$. This adjacency cannot be $B \to S$. If the first original edge is $S \to O_1$, then $S \to O_1 \to B \to S$ would be a directed cycle. If it is $S \bedge O_1$, then $O_1 \to B \to S$ together with that bidirected edge would form an almost directed cycle. Thus the edge between $S$ and $B$ has an arrowhead at $B$. If $B = T$, this is already a contradiction, since $S$ and $T$ were assumed nonadjacent in $\mc{M}$. Otherwise replace the path segment $\langle S, O_1, \cdots, O_k(=A), B \rangle$ by this original edge from $S$ into $B$. The arrowhead at $B$ is preserved, and all other retained non-endpoint vertices remain colliders and ancestors of $S$ or $T$. We have obtained a shorter inducing path, again a contradiction. 

    We conclude that no inducing path in $\mc{G}$ creates a new adjacency. Its corresponding MAG therefore has exactly the adjacencies of $\mc{M}$. It also has exactly the same endpoint marks, because the ancestral relations used to determine those marks have not changed. Hence its corresponding MAG is $\mc{M}$. 

    \textbf{Identifying the resulting territory and border.} Finally, we connect the set $\mf{T}$ used in the construction with the ordinary MUCT after intervention on $\mf{X}$. The directed edges of $\mc{G}$ and $\mc{M}$ coincide, and a directed path from a vertex outside $\mf{X}$ to $Y$ survives this intervention exactly when it avoids $\mf{X}$. Thus:
    \begin{align}
        \An(Y)_{\mc{G}_{\overline{\mf{X}}}} \setminus \mf{X} = 
        \An(Y)_{\mc{M} \setminus \mf{X}} = \mf{H}.
    \end{align}
    Some intervened vertices may themselves remain ancestors of $Y$. However, they cannot enter the territory grown from $Y$ since the intervention removes every edge into them, so neither a bidirected connection nor a directed path from the current territory can reach a vertex in $\mf{X}$. The ordinary MUCT computation in the ancestral subgraph of $\mc{G}_{\overline{\mf{X}}}$ therefore takes place entirely within $\mf{H}$ and uses exactly the two closure operations defining $\mf{T}$. Since the construction ended with $\mf{T} = \mf{R}$, this proves $\MUCT(\mc{G}_{\overline{\mf{X}}}, Y) = \MUCT(\mc{M}, Y, \mf{X})$. 

    Every vertex of $\mf{R}$ is outside $\mf{X}$, so intervention on $\mf{X}$ removes none of the directed edges entering $\mf{R}$. Together with the fact that no directed edge was changed during the construction, this shows that the parents of $\mf{R}$ outside $\mf{R}$ are the same in $\mc{G}_{\overline{\mf{X}}}$ and $\mc{M}$. Consequently, $\IB(\mc{G}_{\overline{\mf{X}}}, Y) = \Pa(\mf{R})_{\mc{M}} \setminus \mf{R} = \IB(\mc{M}, Y, \mf{X})$. 
\end{proof}

\subsection{Proof of Theorem~\ref{thm: IB = POMIS for MAG}}\label{sec: proof of thm: IB = POMIS for MAG}
\begin{proof}[Proof of \Cref{thm: IB = POMIS for MAG}]
    \textbf{Necessity.} Fix $\mf{X}$ and write: 
    \begin{align*}
         \mc{H}&=\mc{M}[\An(Y)_{\mc{M}\setminus\mf{X}}],&
         \mf{T}&=\MUCT(\mc{M},Y,\mf{X}),\\
         \mf{W}&=\Pa(\mf{T})_{\mc{M}}\setminus\mf{T},&
         \mf{W}'&=\mf{W}\setminus\mf{X},\qquad
         \mf{X}'=\mf{X}\setminus\mf{W}.
    \end{align*}
    Thus $\mf{W} = \IB(\mc{M},Y,\mf{X})$. The defining closures are $\De(\mf{T})_{\mc{H}}=\mf{T}$ and $\pc(\mf{T})_{\mc{H}}=\mf{T}$; these are closures in $\mc{H}$, not in the full MAG. Also, $\mf{T}\cap\mf{X}=\emptyset$.

    Our aim is to show that intervening on $\mf{W}$ can achieve at least the optimal reward attainable by intervening on $\mf{X}$, in every conforming SCM. If $\mf{W} \neq \mf{X}$, this will provide a competing DMIS that prevents $\mf{X}$ from being strictly optimal. The set $\mf{W}'$ consists of the border variables that are missing from the current intervention, whereas $\mf{X}'$ consists of intervention variables that are not on the border. The two separation claims below will allow us first to intervene on $\mf{W}'$ and then to remove $\mf{X}'$ from the intervention. 

    First, $\mf{W}$ is an MIS for $\mc{M}$. For each $W \in \mf{W}$, choose $W \to T$ with $T \in \mf{T}$. Since $T$ is a vertex of $\mc{H}$, it has a directed path to $Y$ within $\mc{H}$. Descendant closure keeps that path entirely in $\mf{T}$, so the path from $W$ to $Y$ avoids every other member of $\mf{W}$. Now apply \Cref{thm: MIS for MAG iff An}. Thus, $\mf{W}$ is also a DMIS by \Cref{prop: if MIS then DMIS for cd and mag}. This includes $\mf{W} = \emptyset$. 

    For MAGs, lower manipulation at $\mf{W}'$ deletes visible directed edges out of $\mf{W}'$ and replaces invisible directed edges out of $\mf{W}'$ by bidirected edges. Apply this operation before upper manipulation at $\mf{X}$, as required by the MAG manipulation convention \citep{zhang2008a}. 

    \begin{claim}\label{claim1: POMIS for MAG}
        $(Y \Perp \mf{W}' \mid \mf{X})_{\mc{M}_{\overline{\mf{X}}\underline{\mf{W}'}}}$. 
    \end{claim}
    \begin{proof}[Proof of \Cref{claim1: POMIS for MAG}]
        Suppose that an m-connecting path $p$ in $\mc{M}_{\overline{\mf{X}}\underline{\mf{W}'}}$ joins $W \in \mf{W}'$ to $Y$ given $\mf{X}$. All incoming edges to $\mf{X}$ have been removed, so a collider on $p$ could neither belong to $\mf{X}$ nor be an ancestor of a vertex in $\mf{X}$. It would therefore violate the condition for m-connection. Thus, $p$ has no colliders, and every non-endpoint vertex lies outside $\mf{X}$. Lower manipulation leaves only arrowheads at vertices in $\mf{W}'$. A non-endpoint vertex in $\mf{W}'$ would therefore be a collider. Consequently, $p$ has no non-endpoint vertex in $\mf{W}$ and its first edge has an arrowhead at $W$. 

        Every member of $\mf{W}'$ is a vertex of $\mc{H}$: it is outside $\mf{X}$ and has a child in $\mf{T}$ followed by a directed path to $Y$ in $\mc{H}$. Consider first a path whose first edge $W \bedge V$ was created from an invisible $W \to V$. If $V \neq Y$, the next edge must point away from $V$; otherwise, $V$ would be a collider. Applying the same argument at each subsequent vertex shows that the remaining path is directed from $V$ to $Y$. The case $V = Y$ is included. Starting at $Y \in \mf{T}$ and considering each preceding vertex on this directed path, any parent outside $\mf{T}$ would lie in $\mf{W}$. Since there is no non-endpoint vertex of $p$ in $\mf{W}$, $V$ must belong to $\mf{T}$. 

        The edge $W \to V$ is invisible in $\mc{H}$ under the inherited interpretation of visibility. Indeed, \Cref{lem: zhang2008a lem 10} gives a DAG whose corresponding MAG is $\mc{M}$ and which contains $W \gets U \to V$ for a latent vertex $U$. This path remains when only observed vertices outside $\mc{H}$ are removed. Hence $W \in \pc(V)_{\mc{H}} \subseteq \mf{T}$, contradicting $W \in \mf{W}$. 

        Otherwise, all edges of $p$ are original MAG edges. Any added bidirected edge would have to meet a member of $\mf{W}'$, but no non-endpoint vertex of $p$ belongs to that set. Since $p$ has no colliders and its first edge has an arrowhead at $W$, it has one of the following forms:
        \begin{align}
            & W\gets \cdots \gets L \to \cdots \to Y,\\
            &W \gets \cdots \gets L \bedge R \to \cdots \to Y.
        \end{align}
        To see this, start at $W$ and follow the path. The edges may initially point towards $W$. Once an edge points towards the next vertex, or a bidirected edge is encountered, every subsequent edge must point towards $Y$ to avoid a collider. In either form, the directed path ending at $Y$ may have length zero and consist only of $Y$. In the second form, the directed path from $L$ to $W$ may also have length zero, with $L = W$. In the first form, however, $L \neq W$, because the first edge has an arrowhead at $W$. 

        All vertices of $p$ are in the vertex set of $\mc{H}$. The directed path ending at $Y$ consists of ancestors of $Y$ and avoids $\mf{X}$. The directed path from $L$ to $W$ also avoids $\mf{X}$, and $W \in \mf{V}(\mc{H})$ has a directed path to $Y$ avoiding $\mf{X}$. Its vertices are therefore also ancestors of $Y$ in $\mc{M} \setminus \mf{X}$. 

        We now show that both forms imply $W \in \mf{T}$. In the first form, start at $Y \in \mf{T}$ and consider each preceding vertex on the directed path from $L$ to $Y$. As above, a parent outside $\mf{T}$ would belong to $\mf{W}$, which is impossible for a non-endpoint vertex of $p$. Thus, $L \in \mf{T}$. In the second form, the same argument along the directed path from $R$ to $Y$ gives $R \in \mf{T}$. The bidirected edge $L \bedge R$ is invisible, so $L \in \pc(R)_{\mc{H}} \subseteq \mf{T}$. In both forms, $W$ is a descendant of $L$ along a directed path in $\mc{H}$. Since $\De(\mf{T})_{\mc{H}} = \mf{T}$, we obtain $W \in \mf{T}$, contradicting $W \in \mf{W}$.
        \end{proof}
        
        \begin{claim}\label{claim2: POMIS for MAG}
            $(Y\Perp\mf{X}'\mid\mf{W})_{\mc{M}_{\overline{\mf{W} \cup \mf{X}'}}}$.
        \end{claim}
        \begin{proof}[Proof of \Cref{claim2: POMIS for MAG}]
        An m-connecting path from $X \in \mf{X}'$ would start with an outgoing edge. No collider can be activated by conditioning on $\mf{W}$, whose vertices have no incoming edges after manipulation. The path must therefore be directed towards $Y$ and avoid $\mf{W}$. Since $Y\in\mf{T}$ and $X \notin \mf{T}$, this directed path has a first edge entering $\mf{T}$ from outside. The vertex just before that first member of $\mf{T}$ lies in $\Pa(\mf{T})_{\mc{M}}\setminus\mf{T}=\mf{W}$, a contradiction.
        \end{proof}
        For any assignment $\mf{x}$ in any conforming SCM, marginalization and Rules~2 and~3 for MAGs yield: 
        \begin{align}
        \mu_{\mf{x}}
        &=\sum_{\mf{w}'}\mb{E}[Y\mid do(\mf{x}),\mf{w}']
            P(\mf{w}'\mid do(\mf{x}))\nonumber\\
        &=\sum_{\mf{w}'}\mb{E}[Y\mid do(\mf{x},\mf{w}')]
            P(\mf{w}'\mid do(\mf{x}))\label{eq: corresponding to claim1 POMIS for MAG}\\
        &=\sum_{\mf{w}'}\mb{E}[Y\mid do(\mf{x}\cap\mf{W},\mf{w}')]
            P(\mf{w}'\mid do(\mf{x}))\label{eq: corresponding to claim2 POMIS for MAG}\\
     &\leq\mu_{\mf{w}^\ast}.\nonumber
        \end{align}
        The first rule application exchanges observation of $\mf{W}'$ for intervention, using \Cref{claim1: POMIS for MAG}. The second removes the interventions on $\mf{X}'$, using \Cref{claim2: POMIS for MAG}. Both rule applications are within the same fixed SCM. What remains is an average of rewards under assignments to $\mf{W}=(\mf{X}\cap\mf{W})\cup\mf{W}'$. Each term is at most the best reward attainable on $\mf{W}$, and the nonnegative averaging weights sum to one. This gives the last inequality. For general domains, replace the sum by the corresponding integral. The bound holds for every $\mf{x}$, so it also holds after maximizing over assignments to $\mf{X}$. If $\mf{W} \neq \mf{X}$, the DMIS $\mf{W}$ is therefore a distinct competitor whose optimal reward is never smaller. Thus $\mf{X}$ cannot be a POMIS, proving necessity. 

    \textbf{Sufficiency.} Suppose $\mf{X} = \IB(\mc{M}, Y, \mf{X})$. By \Cref{lem: causal diagram construction}, there is a causal diagram $\mc{G}$ with corresponding MAG $\mc{M}$, exactly the same directed edges, and $\IB(\mc{G}_{\overline{\mf{X}}},Y) = \mf{X}$. By \Cref{thm: IB = POMIS for CD}, $\mf{X}$ is a POMIS, and hence an MIS, for $\mc{G}$. Thus there is one SCM $\mc{S}^\ast$ conforming to $\mc{G}$ in which the optimal reward attainable on $\mf{X}$ is strictly greater than that attainable on every other MIS of $\mc{G}$. 

    Because $\mc{G}$ and $\mc{M}$ have exactly the same directed edges, their proper directed paths agree for every intervention set. Hence \Cref{prop: MIS} and \Cref{thm: MIS for MAG iff An} imply:
    \begin{align}
        \mb{M}_{\mc{G},Y} =\mb{M}_{\mc{M},Y} =\mb{D}_{\mc{M},Y}.
    \end{align}
    The final equality follows from \Cref{prop: if DMIS then MIS,prop: if MIS then DMIS for cd and mag}. The same $\mc{S}^\ast$ conforms to $\mc{M}$ and supplies all strict inequalities required by \Cref{def: POMIS for PAG}. Therefore, $\mf{X}$ is a POMIS for $\mc{M}$. 
\end{proof}

\subsection{Terminal PMG--MAG Correspondence}
Let $\mc{P}$ be a PAG on $\mf{V}$, and let $\mf{X} \subseteq \mf{V} \setminus \{Y\}$. The PMG $\mc{Q}$ is obtained from $\mc{P}$ by valid local transformations and the orientation rules in \Cref{thm: wang2023 thm 1} applied to closure after each transformation. We first explain how to compare the territories of $\mc{Q}$ and its represented MAGs. We then show how to choose the remaining orientations so that one represented MAG has the required interventional border. Visibility in every induced graph retains the model class of its full PMG. 

\paragraph{Visibility in induced graphs.} Consider a directed edge $A \to B$ in $\mc{Q}$ and an observed vertex set $\mf{A}$ containing both endpoints. Visibility relative to the inherited model class is the same in $\mc{Q}$ and $\mc{Q}[\mf{A}]$. We verify both directions because the ancestor sets used later depend on the graph. 

First suppose that $A \to B$ is visible in the full PMG. If a represented DAG restricted to $\mf{A}$ and its latent vertices had an inducing path into $A$ between $A$ and $B$, the same path would be present in the full DAG. The directed paths that establish the ancestry of any colliders would also remain present in the full DAG. This contradicts visibility in $\mc{Q}$, so the edge is visible after restriction. 

Conversely, suppose that $A \to B$ is invisible in the full PMG. There is then a represented DAG with an inducing path between $A$ and $B$ that is into $A$. Its corresponding MAG has an invisible $A \to B$ by \Cref{lem: zhang2008a lem 9}. By \Cref{lem: zhang2008a lem 10}, a DAG represented by this same MAG can be chosen to contain a latent common parent $A \gets U \to B$. The vertices $A,B,U$ and these two edges survive restriction to $\mf{A}$ and the latent vertices. They therefore establish invisibility after restriction as well. Non-directed edges are invisible by definition. For each fixed full graph and its represented model class, these observations allow us to use the same visibility status when an edge appears in different induced ancestor graphs. 

\paragraph{Expressing the territory using individual edges.} Let $\mc{D}$ be $\mc{Q}$ or a MAG in $[\mc{Q}]$, let $\mf{A}_{\mc{D}} = \Poss\An(Y)_{\mc{D}\setminus \mf{X}}$ and let $\mf{T}_{\mc{D}} = \MUCT(\mc{D},Y,\mf{X})$. Then $\mf{T}_{\mc{D}}$ is the set reachable from $Y$ in $\mc{D}[\mf{A}_\mc{D}]$ when invisible edges can be traversed in both directions and visible edges can be traversed from tail to arrowhead. 

An invisible edge is itself a path satisfying the possible c-component (pc) condition: it has no internal vertices, so its internal-collider condition is vacuous. Consequently, if a set is closed under $\pc$ and contains one endpoint of an invisible edge, it must also contain the other endpoint. 

Conversely, suppose that a set is closed under this operation on individual invisible edges. A pc path consists entirely of invisible edges. Starting from an endpoint in the set, we can follow its edges one at a time and conclude that every vertex of the path, including the other endpoint, is in the set. Thus closure under $\pc$ is equivalent to closure under traversing individual invisible edges in either direction. This argument does not require the pc relation itself to be transitive.

Possible descendant closure additionally permits movement along a possibly directed path. Each invisible edge on such a path is already covered by the preceding operation, $\pc$. Any visible edge is directed, and a possibly directed path can use it only from tail to arrowhead. It follows that the set reached from $Y$ by the movements in the statement is closed under both operations. Every set containing $Y$ and closed under both operations must contain all vertices reached in this way. Hence the reached set is precisely the minimal such set, $\mf{T}_{\mc{D}}$. 

\begin{lemma}\label{lem: PMG IB equal X iff border equal ancestry}
    Let $\mc{D}$ be $\mc{Q}$ or a MAG in $[\mc{Q}]$, let $\mf{A}_{\mc{D}} = \Poss\An(Y)_{\mc{D}\setminus \mf{X}}$ and let $\mf{T}_{\mc{D}} = \MUCT(\mc{D},Y,\mf{X})$. If every $X \in \mf{X}$ has a proper directed path to $Y$ in $\mc{D}$, then $\IB(\mc{D},Y,\mf{X}) = \mf{X}$ if and only if $\mf{T}_{\mc{D}} = \mf{A}_{\mc{D}}$. 
\end{lemma}
\begin{proof}[Proof of \Cref{lem: PMG IB equal X iff border equal ancestry}]
    \textbf{An unchanged border requires the entire ancestor set.} Suppose that $\IB(\mc{D},Y,\mf{X}) = \mf{X}$, and assume for a contradiction that some $W$ belongs to $\mf{A}_{\mc{D}} \setminus \mf{T}_{\mc{D}}$. By the definition of $\mf{A}_{\mc{D}}$, there is a possibly directed path from $W$ to $Y$ avoiding $\mf{X}$. Every vertex on this path belongs to $\mf{A}_{\mc{D}}$, since the remaining part of the path is also a possibly directed path to $Y$ avoiding $\mf{X}$. 

    The path starts outside $\mf{T}_{\mc{D}}$ and ends at $Y \in \mf{T}_{\mc{D}}$. Let $U \aedge V$ be the first edge with $U \notin \mf{T}_{\mc{D}}$ and $V \in \mf{T}_{\mc{D}}$. This edge cannot be invisible. Otherwise, $U$ belongs to $\mf{T}_{\mc{D}}$. It must therefore be visible. Because the path is possibly directed from $U$ toward $V$, its orientation is $U \to V$. Thus $U$ is a parent outside the territory and belongs to $\IB(\mc{D},Y,\mf{X})$. But $U$ lies on a path avoiding $\mf{X}$, so $U \notin \mf{X}$. This contradicts the assumed border equality $\IB(\mc{D},Y,\mf{X}) = \mf{X}$. Therefore, $\mf{T}_{\mc{D}} = \mf{A}_{\mc{D}}$. 
    
    \textbf{Recover the border from the entire ancestor set.} Now suppose that $\mf{T}_{\mc{D}} = \mf{A}_{\mc{D}}$. Let $W \to V$ be any edge with $V \in \mf{A}_{\mc{D}}$ and $W \notin \mf{A}_{\mc{D}}$. If $W \notin \mf{X}$, this edge can be followed by a possibly directed path from $V$ to $Y$ avoiding $\mf{X}$. Such a path cannot already contain $W$, since its remaining part from $W$ would make $W$ a member of $\mf{A}_{\mc{D}}$. The concatenation is therefore a possibly directed path from $W$ to $Y$ avoiding $\mf{X}$, again a contradiction. Hence every parent outside the territory belongs to $\mf{X}$, giving $\IB(\mc{D},Y,\mf{X}) \subseteq \mf{X}$. 

    For the reverse inclusion, fix $X \in \mf{X}$ and take its proper directed path to $Y$ with respect to $\mf{X}$. The first child on this path is an ancestor of $Y$ along a path avoiding $\mf{X}$, and thus belongs to $\mf{A}_{\mc{D}} = \mf{T}_{\mc{D}}$. Since $X$ itself is excluded from $\mf{A}_{\mc{D}}$, it is a parent outside the territory. This proves $\mf{X} \subseteq \IB(\mc{D},Y,\mf{X})$ and completes the equivalence. 
\end{proof}

\paragraph{Choosing the remaining orientations.} We will obtain a MAG from $\mc{Q}$ by replacing every $\cdedge$ edge with a directed edge and orienting each circle component as a DAG without unshielded colliders. \Cref{lem: wang2023 lem 16.1} applies to the simultaneous orientation of all these components and ensures that the resulting graph belongs to $[\mc{Q}]$. 

Within a circle component, we can require any specified vertex $R$ to have no incoming edges from that component. To see this, choose $R$ first, then repeatedly choose a remaining vertex with the largest number of already chosen neighbors, breaking ties arbitrarily. Direct every edge from the earlier vertex to the later vertex in this order. For a chordal graph, each vertex's earlier neighbors form a clique \citep{tarjan1984simple}. They therefore cannot form an unshielded collider at that vertex. The ordering also excludes a directed cycle. 

Moreover, every vertex of the component is a descendant of $R$ in this orientation. Indeed, while a connected component is only partly visited, some unvisited vertex has a visited neighbor. The stated choice rule therefore selects a vertex with at least one earlier neighbor. Its edge from that neighbor is directed into the newly visited vertex. Starting with $R$ and repeating this argument gives a directed path from $R$ to every vertex of the component. The choices of $R$ made below can differ between components; the same cited lemma (\Cref{lem: wang2023 lem 16.1}) ensures that the resulting orientations together give one compatible MAG. 

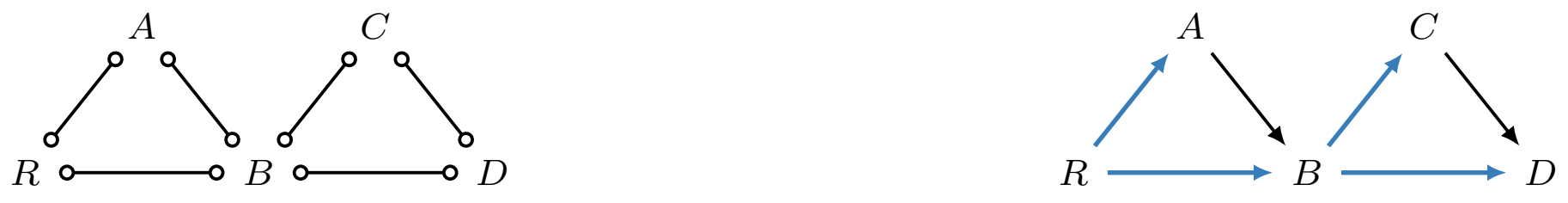
\begin{figure}[t]
    \centering
    \tikzset{
        crgraph/.style={x=0.9cm,y=0.9cm,font=\scriptsize,
            text=black,draw=black,
            >={Latex[width=1.3mm,length=1.5mm]},line width=0.25mm},
        crcircle/.tip={Circle[open,length=3.5pt,width=3.5pt]},
        crvertex/.style={draw=none,circle,fill=white,
            inner sep=0mm,minimum size=5mm},
        crpath/.style={->,draw=betterblue,line width=0.35mm}
    }
    \newcommand{\crvertices}{
        \path[use as bounding box] (-0.4,-0.55) rectangle (4.4,1.75);
        \node[crvertex] (R) at (0,0) {$R$};
        \node[crvertex] (A) at (1,1.25) {$A$};
        \node[crvertex] (B) at (2,0) {$B$};
        \node[crvertex] (C) at (3,1.25) {$C$};
        \node[crvertex] (D) at (4,0) {$D$};
    }
    \begin{minipage}[b]{0.47\textwidth}\centering
    \subfloat[A circle component]{
        \begin{tikzpicture}[crgraph]
            \crvertices
            \foreach \u/\v in {R/A,R/B,A/B,B/C,B/D,C/D}
                \draw[{crcircle}-{crcircle},black] (\u) -- (\v);
        \end{tikzpicture}\label{fig: component root a}}
    \end{minipage}
    \hfill
    \begin{minipage}[b]{0.47\textwidth}\centering
    \subfloat[]{
        \begin{tikzpicture}[crgraph]
            \crvertices
            \foreach \u/\v in {R/A,R/B,B/C,B/D}
                \draw[crpath] (\u) -- (\v);
            \foreach \u/\v in {A/B,C/D}
                \draw[->,black] (\u) -- (\v);
        \end{tikzpicture}\label{fig: component root b}}
    \end{minipage}
    \caption{\small Orienting a circle component starting at $R$. In (b), each edge points from the earlier vertex to the later vertex in the order $R \prec A \prec B \prec C \prec D$. The blue edges give directed paths from $R$ to every other vertex.} \label{fig: component root}
\end{figure}

For example, consider the circle component in \Cref{fig: component root a}, formed by the two triangles on $\{R,A,B\}$ and $\{B,C,D\}$. Choose $R$ first. Both $A$ and $B$ then have one already chosen neighbor, so we may choose $A$. Now $B$ has two already chosen neighbors, $R$ and $A$, whereas $C$ and $D$ have none; hence $B$ is chosen next. Both $C$ and $D$ now have $B$ as an already chosen neighbor. We may choose $C$ and then $D$, giving the order $R \prec A \prec B \prec C \prec D$. Directing each edge from the earlier vertex to the later vertex gives \Cref{fig: component root b}. The blue edges provide the paths: $R \to A$, $R \to B$, $R \to B \to C$, and $R \to B \to D$. Thus even $C$ and $D$, which are not adjacent to $R$, are descendants of $R$. Each newly chosen vertex receives an edge from an earlier vertex that is already reachable from $R$. This is the step repeated in the general argument above. The two incoming edges at $B$ have adjacent parents $R, A$, and those at $D$ have adjacent parents $B, C$; the resulting orientation therefore has no unshielded colliders. 

For an edge $R \to B$, a \textit{visibility witness} is a path into $R$ whose first vertex is not adjacent to $B$ and whose internal vertices are colliders and parents of $B$. A path consisting of one edge is allowed. These are the conditions in \Cref{lem: visible}, which we use below to establish visibility. 

\begin{lemma}\label{lem: visibility after orienting a circle component}
    Let $\mc{M}$ be obtained from $\mc{Q}$ by replacing every $\cdedge$ edge with a directed edge and orienting each circle component as a DAG without unshielded colliders. Suppose that $R$ has no incoming edges within its circle component in this orientation. If $R \to B$ is visible in $\mc{M}$, then the edge between $R$ and $B$ is already oriented as $R \to B$ in $\mc{Q}$ and has a graphical visibility witness there. In particular, an invisible edge of $\mc{Q}$ that becomes $R \to B$ remains invisible in $\mc{M}$. 
\end{lemma}
\begin{proof}[Proof of \Cref{lem: visibility after orienting a circle component}]
    \textbf{Step~1: Find a path that establishes visibility in the MAG.} By \Cref{lem: visible}, visibility of $R \to B$ supplies a path $p = \langle C_0, C_1, \cdots, C_k (=R) \rangle$ such that $C_0$ is not adjacent to $B$, the path has an arrowhead at $R$, and all internal vertices are colliders on the path and parents of $B$. The endpoint $R$ is also a parent of $B$ by the edge under consideration. Thus $C_i \to B$ in $\mc{M}$ for $1 \leq i \leq k$. All these vertices are distinct from $B$. When $k \geq 2$, every edge from $C_1$ through $R$ has arrowheads at both endpoints, so this part of the path is bidirected. 

    \textbf{Step~2: The path's arrowheads were already in the PMG.} The procedure used to obtain $\mc{M}$ creates no bidirected edges. Consequently, the bidirected part from $C_1$ through $R$ was already present in $\mc{Q}$. If the first edge is bidirected, it too was already present. If the first edge is directed into $C_1$, its arrowhead at $C_1$ could be new only when $C_0 \cedge C_1$ was oriented as $C_0 \to C_1$. For $k=1$, such a new arrowhead would give $R$ an incoming edge from another vertex in its circle component, contrary to the assumption on $R$. For $k \geq 2$, suppose that this possibility does occur. We then have $C_0 \cedge C_1 \bedge C_2$ in $\mc{Q}$. The bidirected edge equivalence in \Cref{lem: PMG circle balanced} gives $C_0 \bedge C_2$. We may therefore replace the beginning $\langle C_0, C_1, C_2\rangle$ by $\langle C_0, C_2\rangle$. This deletes one vertex and preserves the collider condition, the arrowhead at $R$, and all the remaining parent conditions. It also leaves the same endpoint $C_0$, which is still not adjacent to $B$. After this possible deletion, every arrowhead required by the resulting path is already present in $\mc{Q}$. 

    \textbf{Step~3: Every required parent mark was already fixed.} Relabel the resulting path as $\langle C_0, C_1, \cdots, C_k (=R) \rangle$. We prove in order, for $i = 1, \cdots, k$, that the edge is already oriented as $C_i \to B$ in $\mc{Q}$. Because the corresponding edge is directed from $C_i$ to $B$ in $\mc{M}$, the possibilities in $\mc{Q}$ are $C_i \to B$, $C_i \cdedge B$, or $C_i \cedge B$. 

    For $i = 1$, the circle edge possibility $C_1 \cedge B$ is excluded by balance: together with the already present $C_0 \adedge C_1$, it would make $C_0$ adjacent to $B$. If the remaining possibility is $C_1 \cdedge B$, the triple $C_0 \adedge C_1 \cdedge B$ is unshielded, and $\mc{R}_1$ orients $C_1 \to B$. Since $\mc{Q}$ is closed under this rule, the tail at $C_1$ must already be displayed in $\mc{Q}$. 
    
    Now take $i > 1$ and assume that $C_1, \cdots, C_{i-1}$ have already been shown to be parents of $B$ in $\mc{Q}$. If $C_i \cedge B$, the edge $C_{i-1} \bedge C_i$ and circle balance would imply $C_{i-1} \bedge B$. This contradicts the orientation $C_{i-1} \to B$ in $\mc{Q}$ just established. Hence only $C_i \to B$ or $C_i \cdedge B$ remains. 

    To exclude the latter possibility, consider $\langle C_0, C_1, \cdots,C_i, B \rangle$. Its first endpoint $C_0$ is not adjacent to $B$, while $C_i$ is adjacent to $B$. Every intermediate vertex $C_1, \cdots, C_{i-1}$ is a collider on this path, by Step~2, and a parent of $B$ in $\mc{Q}$, by the induction hypothesis. The vertices are distinct. Thus this is a discriminating path for $C_i$, and $\mc{R}_4'$ would orient $C_i \cdedge B$ as $C_i \to B$. Closure again implies that this tail is already fixed in $\mc{Q}$. This completes the induction. 

    \textbf{Step~4: Apply the visibility criterion in the PMG.} The path now has all its required arrowheads in $\mc{Q}$, and all its internal vertices are parents of $B$ in $\mc{Q}$. Its endpoint $C_0$ remains nonadjacent to $B$. It therefore establishes visibility of the edge $R \to B$, already oriented in $\mc{Q}$, in every represented MAG and in the inherited causal model class of $\mc{Q}$. The last assertion of the lemma is the contrapositive of this conclusion. 
\end{proof}

\begin{corollary}\label{cor: graphical characterization of visibility in a PMG}
    A directed edge $R \to B$ is visible in $\mc{Q}$ if and only if it has a graphical visibility witness in the full graph $\mc{Q}$. 
\end{corollary}
\begin{proof}[Proof of \Cref{cor: graphical characterization of visibility in a PMG}]
    Suppose first that $R \to B$ is visible in $\mc{Q}$. Choose $R$ first when orienting its circle component as described above, and orient all other circle components as DAGs without unshielded colliders. Replace every $\cdedge$ edge by a directed edge. By \Cref{lem: wang2023 lem 16.1}, these simultaneous choices give a MAG $\mc{M} \in [\mc{Q}]$ in which $R$ has no incoming edges from its circle component. Every DAG represented by $\mc{M}$ belongs to the model class specified by $\mc{Q}$, so $R \to B$ is visible in $\mc{M}$. By \Cref{lem: visibility after orienting a circle component}, it has a graphical visibility witness in $\mc{Q}$. 

    Conversely, suppose that such a witness is present in $\mc{Q}$. Its nonadjacency, required arrowheads, and directed parent edges are preserved in every MAG in $[\mc{Q}]$. The criterion in \Cref{lem: visible} therefore makes $R \to B$ visible in every such MAG, and hence in the full causal model class specified by $\mc{Q}$. 
\end{proof}

\begin{lemma}\label{lem: visibility for edges into a circle component}
    Let $B,C$ lie in the same circle component of $\mc{Q}$ and let $R$ lie outside that component. If $R \to B$ has a graphical visibility witness in $\mc{Q}$, then the edge between $R$ and $C$ is already oriented as $R \to C$ in $\mc{Q}$ and has a graphical visibility witness there. 
\end{lemma}
\begin{proof}[Proof of \Cref{lem: visibility for edges into a circle component}]
    If $B = C$, the conclusion is immediate. Otherwise, it suffices to prove the result for adjacent vertices $B \cedge C$ and repeat it along a circle path from $B$ to $C$. 

    \textbf{Step~1: Obtain the possible parent edges into $C$.} Choose a visibility witness $\langle C_0, C_1, \cdots , C_k (=R) \rangle$ for $R \to B$. For every $1 \leq i \leq k$, $\mc{Q}$ already contains the directed edge $C_i \to B$. The stronger balanced property applied to $C_i \to B \cedge C$ therefore gives $C_i \to C$ or $C_i \cdedge C$. In particular, each $C_i$ is adjacent to $C$, with an arrowhead at $C$. Each $C_i$ lies outside the circle component of $B$: if it lay in that component, its adjacency to $B$ would have to be a circle edge by \Cref{lem: PMG circle balanced}(1), contradicting $C_i \to B$. Thus none of these vertices equals $C$. 

    \textbf{Step~2: The first endpoint stays nonadjacent to $C$.} For the sake of contradiction, suppose that $C_0$ is adjacent to $C$. We consider all possible forms of this edge. 

    If $C_0 \cedge C$, then $C_0$ lies in the same circle component as $B$. Since $C_1$ is outside that component, the arrowhead equivalence in \Cref{lem: PMG circle balanced}(2), applied to $C_1 \to B$, gives an arrowhead at $C_0$ on the $C_1 \aedge C_0$ edge. Its bidirected edge equivalence also excludes $C_1 \bedge C_0$, because $C_1 \to B$ is not bidirected. The edge must therefore be $C_1 \to C_0$ or $C_1 \cdedge C_0$. Both possibilities contradict the witness arrowhead at $C_1$. 

    Next suppose that the edge has no arrowhead at $C$ and is not a circle edge. In any MAG obtained by replacing all $\cdedge$ edges with directed edges and orienting the circle components as DAGs without unshielded colliders, this edge is $C \to C_0$. Step~1 also gives $C_1 \to C$ in that MAG. The first edge of the witness still points into $C_1$. If it is $C_0 \to C_1$, these three edges form $C \to C_0 \to C_1 \to C$. If it is $C_0 \bedge C_1$, then $C_1 \to C \to C_0$ together with $C_0 \bedge C_1$ is an almost directed cycle. Both conclusions contradict the ancestral property of the MAG, whose existence follows from \Cref{lem: wang2023 lem 16.1}.

    The remaining possibility is an arrowhead at $C$ on the $C_0 \aedge C$ edge. The balanced property applied to $C_0 \adedge C \cedge B$ then gives an adjacency between $C_0$ and $B$. This contradicts the defining nonadjacency of the visibility witness. All possibilities lead to contradictions; therefore, $C_0$ is not adjacent to $C$. 

    \textbf{Step~3: Fix the parent marks into $C$ in order.} For $C_1$, Step~1 leaves $C_1 \to C$ or $C_1 \cdedge C$. Since $C_0$ is not adjacent to $C$, the triple $C_0 \adedge C_1 \cdedge C$ would trigger $\mc{R}_1$. Since $\mc{Q}$ is closed under the rule, $C_1 \to C$ is already oriented in $\mc{Q}$. 

    For $i > 1$, assume that the vertices $C_1, \cdots, C_{i-1}$ have already been shown to be parents of $C$ in $\mc{Q}$. The path $\langle C_0, \cdots, C_i, C\rangle$ has distinct vertices, its first endpoint is nonadjacent to $C$, and its intermediate vertices are colliders and parents of $C$ in $\mc{Q}$. Thus it is a 
    discriminating path for $C_i$. If $C_i \cdedge C$ remained, $\mc{R}_4'$ would orient $C_i \to C$. Closure shows that this tail is already fixed. The induction reaches $C_k = R$. 

    The edge $R \to C$ is therefore already oriented in $\mc{Q}$. The same collider path into $R$ now has all its internal vertices as parents of $C$ in $\mc{Q}$. Step~2 supplies the required nonadjacency of its first endpoint. This is the desired visibility witness for $R \to C$. 
\end{proof}

\begin{lemma}\label{lem: terminal pmg-mag correspondence}
    Suppose that every circle mark at a vertex in $\mf{X} \cup \{Y\}$ has been resolved in $\mc{Q}$ and that every $X \in \mf{X}$ has a proper directed path to $Y$ with respect to $\mf{X}$ in $\mc{Q}$. Then $\IB(\mc{Q},Y,\mf{X}) = \mf{X}$ if and only if there exists a MAG $\mc{M} \in [\mc{Q}]$ satisfying $\IB(\mc{M},Y,\mf{X}) = \mf{X}$. 
\end{lemma}
\begin{proof}[Proof of \Cref{lem: terminal pmg-mag correspondence}]
    Let $\mf{A}_{\mc{M}} = \An(Y)_{\mc{M}\setminus \mf{X}}$ and let $\mf{T}_{\mc{M}} = \MUCT(\mc{M},Y,\mf{X})$. Similarly, let $\mf{A}_{\mc{Q}} = \Poss\An(Y)_{\mc{Q}\setminus \mf{X}}$ and let $\mf{T}_{\mc{Q}} = \MUCT(\mc{Q},Y,\mf{X})$. For a represented MAG, $\mf{A}_{\mc{M}}$ consists of the actual ancestors of $Y$ after $\mf{X}$ is deleted. For the PMG, $\mf{A}_{\mc{Q}}$ consists of possible ancestors under the same deletion. We always have $\mf{A}_{\mc{M}} \subseteq \mf{A}_{\mc{Q}}$, since a directed path in $\mc{M}$ is possibly directed in $\mc{Q}$. This inclusion can be strict. The argument below keeps the two sets separate. 

    \begin{center}
        \textbf{From a Represented MAG to the PMG}
    \end{center}
    \textbf{Step~1: Transfer membership in the MAG territory to the PMG.}
    Suppose that $\mc{M} \in [\mc{Q}]$ satisfies $\IB(\mc{M},Y,\mf{X}) = \mf{X}$. The proper directed paths from $\mf{X}$ in $\mc{Q}$ remain directed in $\mc{M}$. Thus \Cref{lem: PMG IB equal X iff border equal ancestry} gives $\mf{T}_{\mc{M}} = \mf{A}_{\mc{M}}$. We next show that $\mf{T}_{\mc{M}} \subseteq \mf{T}_{\mc{Q}}$. 

    Every vertex used to reach $\mf{T}_{\mc{M}}$ lies in $\mf{A}_{\mc{M}}$ and hence in $\mf{A}_{\mc{Q}}$. If an edge traversed in $\mc{M}$ is invisible, its counterpart in $\mc{Q}$ is invisible as well. If its counterpart is not directed, it is invisible by definition. If its counterpart is directed, the same edge is directed in $\mc{M}$, and a DAG represented by $\mc{M}$ that establishes its invisibility also belongs to the causal model class specified by $\mc{Q}$; the visibility-under-restriction observation at the beginning of this subsection applies to both induced graphs. Such an edge can therefore also be traversed in $\mc{Q}$. If a traversed edge in $\mc{M}$ is visible, it is traversed in its directed direction. Its counterpart in $\mc{Q}$ is either invisible, allowing either direction, or is a visible directed edge with the same orientation, allowing that forward direction. Thus every step used to reach a vertex of $\mf{T}_{\mc{M}}$ can also be made in $\mc{Q}[\mf{A}_{\mc{Q}}]$. This proves the claimed inclusion $\mf{T}_{\mc{M}} \subseteq \mf{T}_{\mc{Q}}$. 

    \textbf{Step~2: Include the remaining possible ancestors in the PMG territory.} Fix $W \in \mf{A}_{\mc{Q}}$. The case $W = Y$ is immediate. Otherwise, \Cref{lem: wang2023 lem 2} supplies an uncovered possibly directed path from $W$ to $Y$ avoiding $\mf{X}$. Before its first arrowhead, all edges are circle edges. Once an arrowhead appears, each following uncovered triple triggers $\mc{R}_1$ if the next edge still has a circle at its starting vertex. Closure therefore makes the rest of the path directed. The path has the form $W \cedge \cdots \cedge V \qdedge U \to \cdots \to Y$, where $V \qdedge U$ denotes either $V \to U$ or $V \cdedge U$, and the initial circle segment or the final directed segment may have length zero. A nontrivial path consisting entirely of circle edges is excluded because the marks at $Y$ are resolved. 

    The vertex $U$ has a directed path to $Y$ avoiding $\mf{X}$ in every represented MAG. In particular: 
    \begin{align}
        U \in \mf{A}_{\mc{M}} = \mf{T}_{\mc{M}} \subseteq \mf{T}_{\mc{Q}}. 
    \end{align}
    If $V \to U$, then $V$ also has a directed path to $Y$ avoiding $\mf{X}$, so the same reasoning puts $V$ in $\mf{T}_{\mc{Q}}$. If $V \cdedge U$, this edge is invisible in $\mc{Q}$. Since $U \in \mf{T}_{\mc{Q}}$ and both endpoints lie in $\mf{A}_{\mc{Q}}$, closure along invisible edges also puts $V$ in $\mf{T}_{\mc{Q}}$. Each preceding circle edge is also invisible. Following them back toward $W$ gives $W \in \mf{T}_{\mc{Q}}$. 

    Since $W$ was arbitrary, $\mf{A}_{\mc{Q}} \subseteq \mf{T}_{\mc{Q}}$. The reverse inclusion is part of the definition of the territory, so these sets are equal. \Cref{lem: PMG IB equal X iff border equal ancestry} gives $\IB(\mc{Q},Y,\mf{X}) = \mf{X}$. 

    \begin{center}
        \textbf{From the PMG to One Represented MAG}
    \end{center}
    \textbf{Step~1: Choose one vertex in each relevant circle component.} Suppose that $\IB(\mc{Q},Y,\mf{X}) = \mf{X}$. By \Cref{lem: PMG IB equal X iff border equal ancestry}, $\mf{T}_{\mc{Q}} = \mf{A}_{\mc{Q}}$. 
    
    The set $\mf{A}_{\mc{Q}}$ is a union of entire circle components. To see this, suppose a component contains a possible ancestor of $Y$ in $\mc{Q} \setminus \mf{X}$. Any other vertex in the component can follow a circle path to that vertex and then its possibly directed path to $Y$. Removing repeated vertices, if necessary, leaves a possibly directed path. This path avoids $\mf{X}$: every vertex in $\mf{X}$ has all its marks resolved and therefore lies in a singleton circle component. 

    Consider the components contained in $\mf{A}_{\mc{Q}}$. By the closure characterization, every vertex in them can be reached from $Y$ using invisible edges in either direction and visible edges in their directed direction. Start with the singleton component $\{Y\}$. If some components remain unvisited, choose an allowed path from $Y$ to one of them and take its first edge leaving the already visited components. Let $T$ be its endpoint in a visited component and $R$ its endpoint in the new component. Record this edge and choose $R$ as the vertex from which to orient the new component. Mark the new component as visited and repeat. Every repetition adds one component, so eventually every component contained in $\mf{A}_{\mc{Q}}$ has been visited. Each receives exactly one such choice. 

    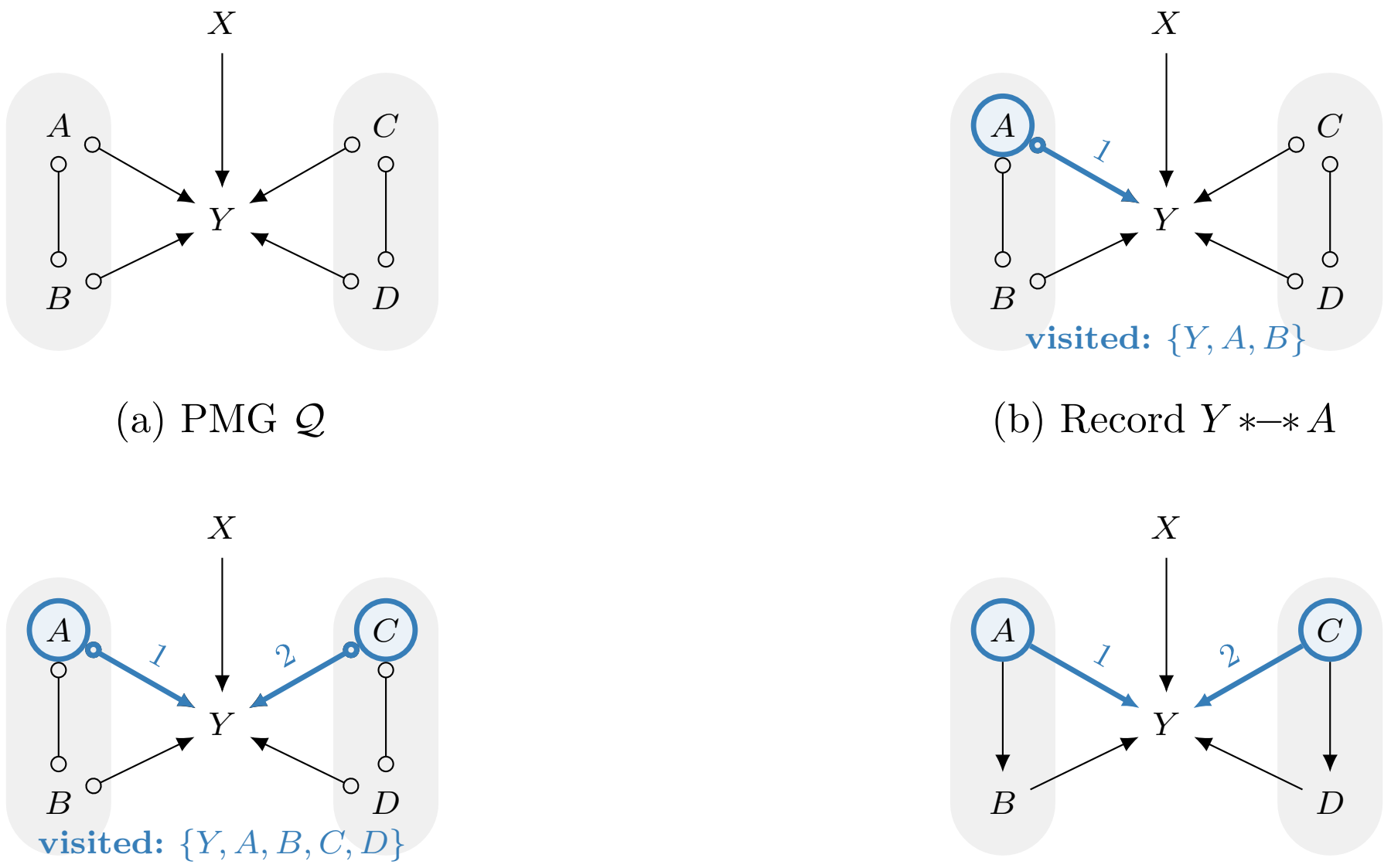
\begin{figure}[t]
    \centering
    \pgfdeclarelayer{recordedbackground}
    \pgfsetlayers{recordedbackground,main}
    \tikzset{
        regraph/.style={x=0.80cm,y=0.80cm,font=\scriptsize,
            text=black,draw=black,
            >={Latex[width=1.3mm,length=1.5mm]},line width=0.25mm},
        recircle/.tip={Circle[open,length=4pt,width=4pt]},
        revertex/.style={draw=none,circle,
            inner sep=0mm,minimum size=5mm},
        reroot/.style={draw=betterblue,line width=0.45mm,fill=betterblue!10},
        rerecorded/.style={draw=betterblue,line width=0.45mm}
    }
    \newcommand{\revertices}{%
        \path[use as bounding box] (-2.4,-1.65) rectangle (2.4,2.65);
        \node[revertex] (X) at (0,2.1) {$X$};
        \node[revertex] (A) at (-1.75,1.0) {$A$};
        \node[revertex] (B) at (-1.75,-0.85) {$B$};
        \node[revertex] (C) at (1.75,1.0) {$C$};
        \node[revertex] (D) at (1.75,-0.85) {$D$};
        \node[revertex] (Y) at (0,0) {$Y$};
    }
    \newcommand{\repmgedges}{%
        \draw[->] (X) -- (Y);
        \draw[{recircle}-{recircle}] (A) -- (B);
        \draw[{recircle}-{recircle}] (C) -- (D);
        \foreach \v in {A,B,C,D}
            \draw[{recircle}->] (\v) -- (Y);
    }
    \newcommand{\recomponents}{%
        \begin{pgfonlayer}{recordedbackground}
            \draw[draw=gray!12,line width=9mm,line cap=round] (A.center) -- (B.center);
            \draw[draw=gray!12,line width=9mm,line cap=round] (C.center) -- (D.center);
        \end{pgfonlayer}
    }
    \begin{minipage}[t]{0.47\textwidth}\centering
        \subfloat[PMG $\mc{Q}$]{%
            \begin{tikzpicture}[regraph]
                \revertices\recomponents\repmgedges
            \end{tikzpicture}\label{fig: recorded edges input}}
    \end{minipage}\hfill
    \begin{minipage}[t]{0.47\textwidth}\centering
        \subfloat[Record $Y \aedge A$]{%
            \begin{tikzpicture}[regraph]
                \revertices\recomponents
                \node[revertex,reroot] (A) at (A.center) {$A$};
                \repmgedges
                \draw[{recircle}->,rerecorded] (A) -- (Y)
                    node[pos=0.54,above,sloped,text=betterblue] {$1$};
                \node[below=5mm,text=betterblue] at (Y.south) {$\textbf{visited: }\{Y,A,B\}$};
            \end{tikzpicture}\label{fig: recorded edges first}}
    \end{minipage}

    \medskip
    \begin{minipage}[t]{0.47\textwidth}\centering
        \subfloat[Record $Y \aedge C$]{%
            \begin{tikzpicture}[regraph]
                \revertices\recomponents
                \node[revertex,reroot] (A) at (A.center) {$A$};
                \node[revertex,reroot] (C) at (C.center) {$C$};
                \repmgedges
                \draw[{recircle}->,rerecorded] (A) -- (Y)
                    node[pos=0.54,above,sloped,text=betterblue] {$1$};
                \draw[{recircle}->,rerecorded] (C) -- (Y)
                    node[pos=0.54,above,sloped,text=betterblue] {$2$};
                \node[below=5mm,text=betterblue] at (Y.south) {$\textbf{visited: }\{Y,A,B,C,D\}$};
            \end{tikzpicture}\label{fig: recorded edges second}}
    \end{minipage}\hfill
    \begin{minipage}[t]{0.47\textwidth}\centering
        \subfloat[MAG $\mc{M}$]{%
            \begin{tikzpicture}[regraph]
                \revertices\recomponents
                \node[revertex,reroot] (A) at (A.center) {$A$};
                \node[revertex,reroot] (C) at (C.center) {$C$};
                \foreach \u/\v in {X/Y,A/B,C/D,B/Y,D/Y}
                    \draw[->] (\u) -- (\v);
                \draw[->,rerecorded] (A) -- (Y)
                    node[pos=0.54,above,sloped,text=betterblue] {$1$};
                \draw[->,rerecorded] (C) -- (Y)
                    node[pos=0.54,above,sloped,text=betterblue] {$2$};
            \end{tikzpicture}\label{fig: recorded edges mag}}
    \end{minipage}
    \caption{\small Recording edges and completing the PMG in (a), with $\mf{X}=\{X\}$. Gray backgrounds group the two nontrivial circle components of $\mc{Q}$; $X$ lies outside the ancestor set after deleting $\mf{X}$. Panels (b)--(c) record existing edges in blue, numbered in visit order, and outline the chosen vertices $A$ and $C$. These panels retain the PMG's endpoint marks. A recorded edge only specifies how a new component was reached. Panel (d) orients each component from its chosen vertex, giving one compatible MAG.}
    \label{fig: recorded edges}
\end{figure}

    \textbf{Step~2: Make all the choices into one MAG.} Within each component, orient its circle edges as a DAG without unshielded colliders, with no incoming component edges at its chosen vertex $R$. The preceding discussion of chordal components shows that this can be done and that $R$ then has a directed path to every vertex of its component. The component of $Y$ is a singleton, so there is no internal choice there. Orient any components outside $\mf{A}_{\mc{Q}}$ as DAGs without unshielded colliders as well, and replace every $\cdedge$ edge by a directed edge. 

    These choices specify the orientation of every remaining circle mark. \Cref{lem: wang2023 lem 16.1} applies to the simultaneous choices and gives a single MAG $\mc{M} \in [\mc{Q}]$. We use this same MAG throughout the rest of the proof. 

    For concreteness, consider the PMG in \Cref{fig: recorded edges input}, with $\mf{X} = \{X\}$. Its relevant ancestor set and territory are both $\mf{A}_{\mc{Q}} = \mf{T}_{\mc{Q}} = \{A,B,C,D,Y\}$, and its relevant circle components are $\{Y\}$, $\{A,B\}$, and $\{C,D\}$. The singleton $\{X\}$ is outside this set because $X$ has been deleted when computing ancestors. 

    Start with $\{Y\}$ visited. To visit $\{A,B\}$, use the invisible edge $A \cdedge Y$ in the direction from $Y$ to $A$. Record this edge with $T =Y$ and $R = A$, and choose $A$ as the vertex from which to orient the new component (\Cref{fig: recorded edges first}). The visited vertices are now $\{Y,A,B\}$. Next use $C \cdedge Y$ in the direction from $Y$ to $C$, record it with $T = Y$ and $R = C$, and choose $C$ for the component $\{C,D\}$ (\Cref{fig: recorded edges second}). All relevant components have now been visited. Recording these two edges does not change their endpoint marks; it specifies the connection and the chosen vertex for each new component. 

    Make both choices in the same graph: orient $A \to B$ and $C \to D$, and replace $V \cdedge Y$ by $V \to Y$ for every $V \in \{A,B,C,D\}$, while keeping $X \to Y$. These choices give the single MAG $\mc{M}$ in \Cref{fig: recorded edges mag}. The selected vertex $A$ reaches both vertices of its component, and $C$ does the same for its component. Every vertex other than $X$ has a directed path to $Y$ avoiding $X$, so in this example $\mf{A}_{\mc{M}} = \mf{A}_{\mc{Q}}$. 

    The recorded edges become $A \to Y$ and $C \to Y$ and remain invisible in $\mc{M}$. Indeed, $Y$ is adjacent to every other vertex, so no visibility witness can have the required first vertex nonadjacent to $Y$. Starting from $Y$, invisible edge closure therefore includes $A$ and $C$; descendant closure along $A \to B$ and $C \to D$ then includes $B$ and $D$. Thus $\mf{T}_{\mc{M}} = \{A,B,C,D,Y\}$ and its only outside parent is $X$, giving $\IB(\mc{M},Y,\{X\}) = \{X\}$. This illustrates the reverse traversal of invisible edges in Step~4 below. The remaining argument also covers completions in which $\mf{A}_{\mc{M}}$ is strictly smaller than $\mf{A}_{\mc{Q}}$. 

    \textbf{Step~3: Verify ancestry for the chosen vertices.} For every component contained in $\mf{A}_{\mc{Q}}$, its chosen vertex $R$ belongs to $\mf{A}_{\mc{M}}$. This is immediate for $R = Y$. For any other $R$, take an uncovered possibly directed path from $R$ to $Y$ avoiding $\mf{X}$. As in the first direction of this proof, it consists of circle edges, then one directed or partially directed edge, followed by directed edges to $Y$. The initial circle segment stays in the component of $R$. In $\mc{M}$, there is a directed path from $R$ to its last vertex because of the orientation chosen in Step~2. The next edge is directed in $\mc{M}$, and the remaining part of the path is already directed in $\mc{Q}$. Since $\mc{M}$ has no directed cycle, concatenating these edges gives a directed path from $R$ to $Y$. 
    
    It avoids $\mf{X}$ because the circle component is disjoint from $\mf{X}$ and the remaining path was chosen to avoid $\mf{X}$. Hence $R \in \mf{A}_{\mc{M}}$.

    A component may also contain vertices that are in $\mf{A}_{\mc{Q}}$ but not in $\mf{A}_{\mc{M}}$. The following induction will require territory membership only for its vertices in $\mf{A}_{\mc{M}}$. 

    \textbf{Step~4: Include the actual ancestors component by component.}
    Consider the components in the order recorded in Step~1. We prove that every vertex in an already considered component that belongs to $\mf{A}_{\mc{M}}$ also belongs to $\mf{T}_{\mc{M}}$. The first component is $\{Y\}$, and $Y \in \mf{T}_{\mc{M}}$ by definition.

    For the next component, use its recorded edge $T \aedge R$, where $T$ is in an earlier component and $R$ is the chosen vertex of the new component. Let $S$ be the chosen vertex of the component containing $T$. Step~3 gives $R,S \in \mf{A}_{\mc{M}}$. The induction hypothesis in particular gives $S \in \mf{T}_{\mc{M}}$. We show that $R$ also belongs to $\mf{T}_{\mc{M}}$ by considering the recorded edge. 
    \begin{itemize}[leftmargin=5mm]
        \item \textbf{The edge points from the earlier component to the new one.} Suppose that the edge in $\mc{Q}$ is $T \to R$ or $T \cdedge R$. It is $T \to R$ in $\mc{M}$. Because $R$ has a directed path to $Y$ avoiding $\mf{X}$, so does $T$. Thus $T \in \mf{A}_{\mc{M}}$, and the induction hypothesis puts $T$ in $\mf{T}_{\mc{M}}$. The edge $T \to R$ lies in the induced graph on $\mf{A}_{\mc{M}}$, so descendant closure puts $R$ in $\mf{T}_{\mc{M}}$. 

        \item \textbf{The edge is bidirected.} Suppose that the recorded edge is $R \bedge T$. If $T = S$, it directly connects $R$ to the known territory vertex $S$. Otherwise, $T$ and $S$ are in the same circle component, and \Cref{lem: PMG circle balanced} gives $R \bedge S$. This edge is unchanged in $\mc{M}$. Both endpoints belong to $\mf{A}_{\mc{M}}$, and $S$ already belongs to $\mf{T}_{\mc{M}}$. Closure along this invisible edge therefore includes $R$. 

        \item \textbf{The edge points toward the earlier component and is invisible.} The remaining possibility is an invisible $R \to T$ or $R \cdedge T$, used in the reverse direction when visiting the new component. If $T = S$, retain this edge. Otherwise, the arrowhead equivalence in \Cref{lem: PMG circle balanced}(2) gives an arrowhead at $S$ on the $R \aedge S$ edge. The bidirected edge equivalence excludes $R \bedge S$, since it would imply $R \bedge T$. Thus the edge to $S$ is $R \to S$ or $R \cdedge S$, and in either case it becomes $R \to S$ in $\mc{M}$. 

        We must also show that $R \to S$ is invisible in $\mc{M}$. Suppose instead that it is visible. The choice of $R$ in Step~2 satisfies \Cref{lem: visibility after orienting a circle component}, which then shows that $R \to S$ is already oriented in $\mc{Q}$ and has a graphical visibility witness there. If $T \neq S$, \Cref{lem: visibility for edges into a circle component} shows that $R \to T$ is also already oriented in $\mc{Q}$ and has such a witness. If $T = S$, the same conclusion already holds. 

        In both cases the edge originally used to visit the component would be visible in $\mc{Q}$, contradicting its selection as an invisible edge. Therefore $R \to S$ is invisible in $\mc{M}$. Since both endpoints are in $\mf{A}_{\mc{M}}$ and $S \in \mf{T}_{\mc{M}}$, invisible edge closure includes $R$. 
    \end{itemize}
    These cases exhaust the possibilities. Circle edges cannot join different circle components, and a visible edge was allowed only in its directed direction. We have shown that the chosen vertex $R$ of the new component belongs to $\mf{T}_{\mc{M}}$. 

    Now let $V$ be any other vertex of this component with $V \in \mf{A}_{\mc{M}}$. Step~2 supplies a directed path from $R$ to $V$ inside the component. Every vertex on that path is also in $\mf{A}_{\mc{M}}$: it reaches $V$, and $V$ reaches $Y$ along a directed path avoiding $\mf{X}$. The component itself contains no vertex of $\mf{X}$. Thus the path from $R$ to $V$ lies entirely in the induced ancestor graph. Descendant closure puts $V$ in $\mf{T}_{\mc{M}}$, completing the induction step. 

    All vertices of $\mf{A}_{\mc{M}}$ lie in $\mf{A}_{\mc{Q}}$, whose components have all been considered. The induction therefore proves $\mf{A}_{\mc{M}} \subseteq \mf{T}_{\mc{M}}$. The reverse inclusion is definitional, giving $\mf{T}_{\mc{M}} = \mf{A}_{\mc{M}}$. Finally, the proper directed paths from $\mf{X}$ present in $\mc{Q}$ are preserved in this same $\mc{M}$. \Cref{lem: PMG IB equal X iff border equal ancestry} now gives $\IB(\mc{M},Y,\mf{X}) = \mf{X}$, which proves the required existence statement.
\end{proof}

\subsection{Proof of Theorem~\ref{thm: isPOMIS sound and complete}} \label{sec: proof of thm: isPOMIS sound and complete}

\begin{proof}[Proof of \Cref{thm: isPOMIS sound and complete}]
    \textbf{Termination.} A recursive call considers finitely many local transformations at its chosen vertex. Each recursive descent removes that vertex from the set still to be processed, so the depth is at most $\lvert \mf{X} \cup \{Y\} \rvert$. The orientation closure within each call also terminates: each effective rule application resolves at least one circle mark, and there are only finitely many marks. Thus every branch and the entire search terminate.

    \textbf{The terminal correspondence applies.} By the time a branch reaches its terminal PMG, it has processed every vertex in $\mf{X} \cup \{Y\}$. The local transformations have therefore resolved all circle marks at these vertices. 

    Furthermore, at each processed $X \in \mf{X}$ the branch passed the proper path check. By \Cref{prop: local proper path}, the resulting PMG contains a proper directed path from $X$ to $Y$ with respect to $\mf{X}$. Later valid transformations preserve the tails and arrowheads already displayed along this path, so it remains directed and proper in the terminal PMG. Consequently, each terminal PMG satisfies both conditions of \Cref{lem: terminal pmg-mag correspondence}. 

    \textbf{Soundness.} Suppose that \textsf{IsPOMIS} returns \textsf{True}. Then there is an accepting terminal PMG $\mc{Q}$ for which the tested equality is $\IB(\mc{Q},Y,\mf{X}) = \mf{X}$. Every local transformation on this branch was valid. It therefore retains at least one compatible MAG, and sound orientation closure preserves the MAGs consistent with the chosen transformations. Since the branch started from $\mc{P}$, we have $\emptyset \neq [\mc{Q}] \subseteq [\mc{P}]$. The terminal correspondence supplies a single $\mc{M} \in [\mc{Q}]$ with $\IB(\mc{M},Y, \mf{X}) = \mf{X}$. By \Cref{thm: IB = POMIS for MAG}, $\mf{X}$ is a POMIS for this MAG. Because $\mc{M} \in [\mc{P}]$, the set $\mf{X}$ belongs to $\widehat{\mb{P}}_{\mc{P},Y}$. Thus every \textsf{True} output has the claimed meaning. 

    \textbf{Completeness.} Suppose that $\mf{X} \in \widehat{\mb{P}}_{\mc{P},Y}$. Fix one MAG $\mc{M} \in [\mc{P}]$ for which $\mf{X}$ is a POMIS. In particular, $\mf{X}$ is an MIS for $\mc{M}$, so each of its vertices has a proper directed path to $Y$ in $\mc{M}$. The corresponding paths in $\mc{P}$ are possibly directed and remain proper. Hence $\mf{X}$ passes the algorithm's initial MIS screen. 

    Now follow choices that agree with this fixed $\mc{M}$ at every processed vertex. We verify that these choices define an admissible branch. Initially $\mc{M} \in [\mc{P}]$. Whenever the current PMG represents $\mc{M}$, the local choice agreeing with $\mc{M}$ is valid, since this same MAG realizes all the chosen marks. Sound orientation closure preserves it; thus, the resulting PMG again represents $\mc{M}$. This argument applies successively throughout the branch, as in \Cref{prop: complete PMG}. 

    It remains to check that the possible ancestor test never rejects this branch. At a processed $X \in \mf{X}$, the proper directed path already present in $\mc{M}$ avoids $\mf{X} \setminus \{X\}$. Since $\mc{M}$ also agrees with the local choice, the reverse implication in \Cref{prop: local proper path} shows that the required possible ancestor test holds in the current PMG. At the processed vertex $Y$, the test holds by reflexivity: neither $\mf{X}$ nor the local transformation set contains $Y$. 

    Thus the search either returns \textsf{True} on an earlier branch or reaches a terminal PMG $\mc{Q}$ representing this same $\mc{M}$. The necessary direction of \Cref{thm: IB = POMIS for MAG} gives $\IB(\mc{M},Y,\mf{X}) = \mf{X}$. Applying \Cref{lem: terminal pmg-mag correspondence} to the terminal PMG then gives $\IB(\mc{Q},Y,\mf{X}) = \mf{X}$. 

This branch therefore returns \textsf{True}. Every member of $\widehat{\mb{P}}_{\mc{P},Y}$ is accepted, which completes the proof. 
\end{proof}
\section{Experimental Structural Causal Models}\label{sec: scbmec experimental SCMs}

This section specifies the six fixed SCMs used in \Cref{sec: scbmec experiment}. All observed and exogenous variables are binary, and the reward is $Y$. Within each task, all distinct exogenous variables are mutually independent, with the Bernoulli distributions specified in each model. A variable $U_V$ appears only in the equation for $V$. Each variable $U_{VW}$ appears in two structural functions. Sharing an exogenous variable in this way allows dependence between observed variables, even though distinct exogenous variables are independent. 

\paragraph{Task~1.}
Let $\mf{V}=\{A,B,C,D,Y\}$ and $\mf{U}=\{U_V:V\in\mf{V}\}\cup\{U_{BY},U_{AC}\}$. The structural functions are:
\begingroup
\begin{align}
\begin{aligned}
 f_A &= u_{AC} \land u_A\\
 f_C &= \bigl(u_C \lor \lnot a\bigr) \lor u_{AC}\\
 f_B &= \lnot u_{BY} \oplus \bigl(c \lor u_B\bigr)\\
 f_Y &= u_Y \lor \bigl(\lnot u_{BY} \land \bigl(\lnot b \lor c\bigr)\bigr)\\
 f_D &= \lnot u_D \land y.
\end{aligned}
\label{eq:scbmec-task1-scm}
\end{align}
\endgroup
The exogenous distributions are:
\begingroup
\begin{align}
\begin{aligned}
 U_A&\sim\Bern(0.57) & U_B&\sim\Bern(0.57) & U_C&\sim\Bern(0.77)\\
 U_D&\sim\Bern(0.61) & U_Y&\sim\Bern(0.42) & U_{BY}&\sim\Bern(0.46)\\
 U_{AC}&\sim\Bern(0.62).
\end{aligned}
\end{align}
\endgroup
\paragraph{Task~2.}
Let $\mf{V}=\{A,B,C,D,E,Y\}$ and $\mf{U}=\{U_V:V\in\mf{V}\}\cup\{U_{BY},U_{AY}\}$. The structural functions are:
\begingroup
\begin{align}
\begin{aligned}
 f_A &= u_A \oplus u_{AY}\\
 f_B &= u_{BY} \lor u_B\\
 f_C &= \lnot u_C \lor b\\
 f_D &= c \land u_D\\
 f_E &= \bigl(\lnot a \oplus u_E\bigr) \land c\\
 f_Y &= u_{AY} \land \bigl(\bigl(\bigl(\lnot u_{BY} \lor \lnot d\bigr) \land e\bigr) \oplus u_Y\bigr).
\end{aligned}
\label{eq:scbmec-task2-scm}
\end{align}
\endgroup
The exogenous distributions are:
\begingroup
\begin{align}
\begin{aligned}
 U_A&\sim\Bern(0.44) & U_B&\sim\Bern(0.48) & U_C&\sim\Bern(0.36)\\
 U_D&\sim\Bern(0.38) & U_E&\sim\Bern(0.47) & U_Y&\sim\Bern(0.72)\\
 U_{BY}&\sim\Bern(0.55) & U_{AY}&\sim\Bern(0.37).
\end{aligned}
\end{align}
\endgroup
\paragraph{Task~3.}
Let $\mf{V}=\{A,B,C,D,E,F,G,H,I,Y\}$ and $\mf{U}=\{U_V:V\in\mf{V}\}\cup\{U_{FG},U_{HI},U_{AB},U_{BE},U_{IY}\}$. The structural functions are:
\begingroup
\begin{align}
\begin{aligned}
 f_A &= u_A \oplus u_{AB}\\
 f_B &= \lnot u_B \lor \bigl(a \lor \bigl(u_{BE} \oplus u_{AB}\bigr)\bigr)\\
 f_C &= \lnot b \land u_C\\
 f_D &= c \land u_D\\
 f_E &= \lnot d \land \bigl(u_{BE} \lor u_E\bigr)\\
 f_F &= u_{FG} \land \bigl(e \lor u_F\bigr)\\
 f_H &= u_H \oplus \bigl(u_{HI} \land e\bigr)\\
 f_G &= \lnot h \lor \bigl(e \lor \bigl(u_G \lor u_{FG}\bigr)\bigr)\\
 f_I &= \bigl(\lnot u_{HI} \land \bigl(\lnot u_{IY} \oplus u_I\bigr)\bigr) \land f\\
 f_Y &= \lnot i \lor \bigl(u_{IY} \land \bigl(\bigl(\lnot f \land \lnot u_Y\bigr) \lor h\bigr)\bigr).
\end{aligned}
\label{eq:scbmec-task3-scm}
\end{align}
\endgroup
The exogenous distributions are:
\begingroup
\begin{align}
\begin{aligned}
 U_A&\sim\Bern(0.31) & U_B&\sim\Bern(0.79) & U_C&\sim\Bern(0.62)\\
 U_D&\sim\Bern(0.65) & U_E&\sim\Bern(0.27) & U_F&\sim\Bern(0.25)\\
 U_G&\sim\Bern(0.60) & U_H&\sim\Bern(0.40) & U_I&\sim\Bern(0.42)\\
 U_Y&\sim\Bern(0.52) & U_{FG}&\sim\Bern(0.45) & U_{HI}&\sim\Bern(0.69)\\
 U_{AB}&\sim\Bern(0.54) & U_{BE}&\sim\Bern(0.73) & U_{IY}&\sim\Bern(0.68).
\end{aligned}
\end{align}
\endgroup
\paragraph{Task~4.}
Let $\mf{V}=\{A,B,C,D,E,Y\}$. The exogenous variables are: 
\begin{align}
    \mf{U}=\{U_V:V\in\mf{V}\}\cup\{U_{AY},U_{AC},U_{BY}\}.
\end{align} 
The structural functions are:
\begingroup
\begin{align}
\begin{aligned}
 f_B &= u_B \oplus u_{BY}\\
 f_D &= u_D \lor b\\
 f_A &= u_A \lor \bigl(\bigl(\lnot d \oplus \bigl(u_{AY} \land u_{AC}\bigr)\bigr) \land b\bigr)\\
 f_E &= \lnot d \lor \bigl(\lnot u_E \oplus b\bigr)\\
 f_C &= \lnot b \land \bigl(\lnot u_{AC} \oplus \bigl(u_C \oplus e\bigr)\bigr)\\
 f_Y &= \bigl(\lnot u_Y \lor \bigl(b \lor \bigl(u_{BY} \lor u_{AY}\bigr)\bigr)\bigr) \oplus c.
\end{aligned}
\label{eq:scbmec-task4-scm}
\end{align}
\endgroup
The exogenous distributions are:
\begingroup
\begin{align}
\begin{aligned}
 U_A&\sim\Bern(0.37) & U_B&\sim\Bern(0.37) & U_C&\sim\Bern(0.47)\\
 U_D&\sim\Bern(0.32) & U_E&\sim\Bern(0.32) & U_Y&\sim\Bern(0.51)\\
 U_{AY}&\sim\Bern(0.25) & U_{AC}&\sim\Bern(0.49) & U_{BY}&\sim\Bern(0.42).
\end{aligned}
\end{align}
\endgroup
\paragraph{Task~5.}
Let $\mf{V}=\{A,B,C,D,E,F,G,Y\}$. The exogenous variables are: 
\begin{align}
    \mf{U}=\{U_V:V\in\mf{V}\}\cup\{U_{FG},U_{AB},U_{BY},U_{CD}\}.
\end{align}
The structural functions are:
\begingroup
\begin{align}
\begin{aligned}
 f_C &= \lnot u_{CD} \lor u_C\\
 f_D &= \bigl(u_{CD} \lor u_D\bigr) \oplus c\\
 f_E &= d \oplus u_E\\
 f_A &= \lnot u_A \lor \bigl(\lnot u_{AB} \oplus e\bigr)\\
 f_F &= e \lor \bigl(u_F \land u_{FG}\bigr)\\
 f_B &= \bigl(f \lor \bigl(\lnot u_{BY} \land u_{AB}\bigr)\bigr) \lor u_B\\
 f_G &= u_G \oplus \bigl(\bigl(\lnot u_{FG} \lor a\bigr) \lor e\bigr)\\
 f_Y &= u_{BY} \land \bigl(\bigl(f \lor b\bigr) \oplus \bigl(u_Y \lor a\bigr)\bigr).
\end{aligned}
\label{eq:scbmec-task5-scm}
\end{align}
\endgroup
The exogenous distributions are:
\begingroup
\begin{align}
\begin{aligned}
 U_A&\sim\Bern(0.75) & U_B&\sim\Bern(0.69) & U_C&\sim\Bern(0.26)\\
 U_D&\sim\Bern(0.42) & U_E&\sim\Bern(0.22) & U_F&\sim\Bern(0.53)\\
 U_G&\sim\Bern(0.68) & U_Y&\sim\Bern(0.23) & U_{FG}&\sim\Bern(0.31)\\
 U_{AB}&\sim\Bern(0.42) & U_{BY}&\sim\Bern(0.35) & U_{CD}&\sim\Bern(0.68).
\end{aligned}
\end{align}
\endgroup
\paragraph{Task~6.}
Let $\mf{V}=\{A,B,C,D,E,F,G,Y\}$. The exogenous variables are: 
\begin{align}
    \mf{U}=\{U_V:V\in\mf{V}\}\cup\{U_{AY},U_{EF},U_{FG}\}. 
\end{align}
The structural functions are:
\begingroup
\begin{align}
\begin{aligned}
 f_A &= u_{AY} \land u_A\\
 f_B &= u_B\\
 f_F &= u_{FG} \oplus \bigl(u_{EF} \oplus u_F\bigr)\\
 f_C &= \bigl(f \land \lnot u_C\bigr) \oplus a\\
 f_D &= u_D \lor f\\
 f_E &= \lnot u_{EF} \oplus \bigl(\lnot d \oplus \bigl(\lnot b \land u_E\bigr)\bigr)\\
 f_G &= \bigl(u_G \land c\bigr) \land u_{FG}\\
 f_Y &= \bigl(e \oplus u_Y\bigr) \oplus \bigl(d \lor \bigl(u_{AY} \lor g\bigr)\bigr).
\end{aligned}
\label{eq:scbmec-task6-scm}
\end{align}
\endgroup
The exogenous distributions are:
\begingroup
\begin{align}
\begin{aligned}
 U_A&\sim\Bern(0.48) & U_B&\sim\Bern(0.57) & U_C&\sim\Bern(0.40)\\
 U_D&\sim\Bern(0.60) & U_E&\sim\Bern(0.39) & U_F&\sim\Bern(0.67)\\
 U_G&\sim\Bern(0.77) & U_Y&\sim\Bern(0.60) & U_{AY}&\sim\Bern(0.21)\\
 U_{EF}&\sim\Bern(0.57) & U_{FG}&\sim\Bern(0.60).
\end{aligned}
\end{align}
\endgroup

\jmlrsection{Supporting Results for Transportability}
\section{Omitted Proofs}

This section proves the soundness and regret guarantees from \Cref{chap:scb tr}. Source distributions and the resulting causal bounds are population quantities. All fixed arguments of a causal factor remain fixed during its marginalization or extraction. 

\subsection{Proof of Theorem~\ref{tr:thm: sound patr}}\label{sec: proof of tr:thm: sound patr}
\label{tr:sec: proof sound patr}

The source-selection test has a useful equivalent form once the
graphical reductions in \textsc{paTR} have been applied.

\begin{lemma}[Proposition~1 in \citet{lee2020general}]
\label{tr:lem: prop1 in lee2020}
    Suppose that $\mf{V}\setminus \mf{X}=\An(\mf{Y})_{\mc{G}\setminus\mf{X}}$. Then $(\mf{S}^i\Perp\mf{Y})_{\mc{G}^{\boldsymbol{\Delta}}\setminus\mf{X}}$ if and only if $\Delta^i\cap(\mf{V}\setminus\mf{X})=\emptyset$. 
\end{lemma}
\begin{proof}[Proof of \Cref{tr:lem: prop1 in lee2020}]
If $\Delta^i\cap(\mf{V}\setminus\mf{X})=\emptyset$, removing $\mf{X}$ removes every child of $\mf{S}^i$, giving the separation. Otherwise, a remaining child $V\in\Delta^i\setminus\mf{X}$ has a directed path to $\mf{Y}$ by the assumed ancestry condition. Its selection edge followed by that path is unblocked, so separation fails. 
\end{proof}

\begin{proof}[Proof of \Cref{tr:thm: sound patr}]
We first check the transformations in \textsc{paTR}, then the expressions supplied by \textsc{paID}. The direct-source branch, ancestral restriction, addition of irrelevant interventions, and c-component decomposition follow the sound transformations of general transportability \citep{lee2020general}. 

The first branch uses the source experiment only when its selection nodes are separated from the query. Removing non-ancestors leaves its distribution unchanged, as does fixing the remaining non-ancestors by Rule~3 of do-calculus. Decomposition expresses the query as a sum of products of causal factors. Source experiments may differ between factors because each expression bounds the corresponding target factor.

Each graphical reduction forwards its recursive expressions and ends the current call. Consequently, when the source loop is reached:
\begin{align}
    \mf{C}=\mf{V}\setminus\mf{X}
    =\An(\mf{Y})_{\mc{G}\setminus\mf{X}},
    \qquad
    \cc_{\mc{G}\setminus\mf{X}}=\{\mf{C}\}.
    \label{tr:eq: patr invariant}
\end{align}
The source test ensures $\Delta^i\cap\mf{C}=\emptyset$ and hence $Q^\ast[\mf{C}]=Q^i[\mf{C}]$. \Cref{tr:lem: prop1 in lee2020} also shows why this test agrees with the separation test at this stage. The selected experiment fixes only variables in $\mf{X}$ among the current vertices. Removing those variables therefore leaves $\mf{C}$, its induced graph, and \Cref{tr:eq: patr invariant} unchanged. 

Consider the resulting call to \textsc{paID}. Its input $P$ is an available source distribution marginalized to its current vertices, with the experimental arguments fixed. Subsequent ancestral marginalization and c-component extraction make each input distribution or causal factor an identifiable source factor $Q^i[\mf{V}_l]$, where the subscript $l$ denotes the variables and graph local to that call. Ancestral marginalization and the products of topologically ordered conditionals are the causal factor identities of \citet{tian:02}. 

The set of non-intervened vertices remains $\mf{C}$ throughout these calls. Indeed, each vertex in $\mf{C}$ has a directed path to $\mf{Y}$ entirely inside $\mf{C}$, so an ancestral restriction keeps all of $\mf{C}$. An enclosing c-component $\mf{C}'$ contains $\mf{C}$, so restriction to $\mf{C}'$ also keeps these paths and the same induced graph on $\mf{C}$. Thus the singleton component specified by \textsc{paID} is always well defined. The source invariance of $Q[\mf{C}]$ is preserved as well. In particular, the query at every such call is the marginal
$\sum_{\mf{c}\setminus\mf{Y}}Q^i[\mf{C}] =\sum_{\mf{c}\setminus\mf{Y}}Q^\ast[\mf{C}]$ with its inherited arguments fixed.

With no remaining intervention, the marginal of $P$ is exact. Extraction and marginalization are also exact when $\mf{C}$ is a c-component of the current graph. It remains to check the branch where the whole current graph is one c-component. Write $P_l = Q^i[\mf{V}_l]$ and $\mf{C}=\mf{V}_l\setminus\mf{X}_l$. Apply \Cref{tr:prop: Q trans bound} with both smaller sets equal to $\mf{C}$ and the enclosing set equal to $\mf{V}_l$. Therefore, the following pointwise inequality holds:
\begin{align}
    P_l(\mf{v}_l)
    \leq Q^\ast[\mf{C}]
      =Q^i[\mf{C}] 
    \leq P_l(\mf{v}_l)+1
       -\sum_{\mf{c}}P_l(\mf{v}_l).
    \label{tr:eq: paid factor bound}
\end{align}
The inner sum varies only $\mf{C}$, holding the local intervention and all external arguments fixed. Summing over
$\mf{C}\setminus\mf{Y}$ gives exactly the lower or upper expression yielded by the fallback branch.

Every yielded bound is nonnegative: the consistency upper expression adds to $P_l$ the complement of a marginal probability. Products of nonnegative lower bounds, or of nonnegative upper bounds, therefore preserve their inequality, as does summation. Induction through the c-component products and the preceding exact transformations proves soundness for every expression yielded by the original call. If a recursive call returns no expression, it contributes no candidate bound. Only expressions whose required source conditionals are defined are evaluated. These conventions can remove a candidate formula without invalidating any retained bound. 
\end{proof}

\subsection{Proof of Theorem~\ref{tr:thm: regret}}\label{tr:sec: proof regret}

\begin{proof}[Proof of \Cref{tr:thm: regret}]
Let $\mc{A}^\ast$ be the retained action space, let $\kappa=\lvert \mc{A}^\ast \rvert$, and assume $T \geq \kappa$. The causal and dominance bounds ensure that an optimal target action $\mf{x}^\star$ is retained and that, for every retained action:
\begin{align}
    \ell_{\mf{x}}
    \leq\mu^\ast_{\mf{x}}
    \leq b_{\mf{x}}=\min\{u_{\mf{x}},u^\star\}.
\end{align}
Since $0\leq Y\leq1$ implies $Y^2\leq Y$:
\begin{align}
    \operatorname{Var}[Y\mid do(\mf{x})]
    \leq \mu^\ast_{\mf{x}}(1-\mu^\ast_{\mf{x}})
    \leq \sigma_{\mf{x}}^2.
\end{align}

Expose for each action an independent sequence of rewards with its target distribution. Each time that action is selected, the policy reveals the next element of its sequence. Fresh independent exogenous draws at each round justify this representation. Write $\widehat m_{\mf{x},n}$ for the average of the first $n$ rewards in the sequence and set:
\begin{align}
    \zeta_T=\log(2\kappa T^2),\qquad
    r_{\mf{x}}(n)
    =\sqrt{\frac{2\sigma_{\mf{x}}^2\zeta_T}{n}}
      +\frac{2\zeta_T}{3n}.
\end{align}
Bernstein's inequality gives:
\begin{align}
    \Pr\left(
      \left\lvert \widehat m_{\mf{x},n}-\mu^\ast_{\mf{x}}\right\rvert
      > r_{\mf{x}}(n)\right)
      \leq 
      2e^{-\zeta_T}.
\end{align}
Taking a union bound over all $\kappa$ actions and $1\leq n \leq T$ yields:
\begin{align}
    \mc{E}
    &=\bigcap_{\mf{x}\in\mc{A}^\ast}\bigcap_{n=1}^T
      \left\{
       \left|\widehat m_{\mf{x},n}-\mu^\ast_{\mf{x}}\right|
       \leq r_{\mf{x}}(n)\right\},\\
    \Pr(\mc{E}^c)&\leq
    2\kappa T e^{-\zeta_T}=\frac{1}{T}.
    \label{tr:eq: simultaneous confidence}
\end{align}
This simultaneous event applies to the adaptively chosen counts $N_{t-1}(\mf{x})$, at which $\widehat m_{\mf{x},N_{t-1}(\mf{x})} =\widehat\mu^\ast_{\mf{x},t-1}$. The linear term in $r_{\mf{x}}$ is needed because the variance bound alone is not a sub-Gaussian variance proxy.

On $\mc{E}$, both the empirical UCB and the clipping level of an optimal action are at least its mean. Thus: 
\begin{align}
    \bar U_{\mf{x}^\star}(t)
    \geq\mu^\ast_{\mf{x}^\star}\qquad(t>\kappa).
\end{align}
A retained action with $b_{\mf{x}} < \mu^\ast_{\mf{x}^\star}$ has a strictly smaller clipped index at every such round, so it is played only once during initialization on $\mc{E}$. An action pruned before initialization is never played.

Now take a retained action with $b_{\mf{x}}\geq\mu^\ast_{\mf{x}^\star}$ and gap $\Delta_{\mf{x}}>0$. If $A_t=\mf{x}$ after initialization, index maximization and the confidence event imply, with $n=N_{t-1}(\mf{x})$:
\begin{align}
    \mu^\ast_{\mf{x}^\star}
    \leq \bar U_{\mf{x}}(t)
    \leq U_{\mf{x}}(t)
    \leq \mu^\ast_{\mf{x}}+2r_{\mf{x}}(n). 
\end{align}
Selection therefore requires $\Delta_{\mf{x}}\leq 2r_{\mf{x}}(n)$.

Define:
\begin{align}
    q_{\mf{x}}
    =\frac{32\sigma_{\mf{x}}^2\zeta_T}{\Delta_{\mf{x}}^2}
      +\frac{8\zeta_T}{3\Delta_{\mf{x}}}.
\end{align}
If $n > q_{\mf{x}}$, the square-root and linear terms in $2r_{\mf{x}}(n)$ are each strictly smaller than $\frac{\Delta_{\mf{x}}}{2}$, so selection is impossible. Any selection after initialization has a preceding count of at most $\lfloor q_{\mf{x}}\rfloor$. Accounting also for the initialization pull therefore gives:
\begin{align}
    N_T(\mf{x}) \leq 1+q_{\mf{x}} \quad \text{on~} \mc{E}.
\end{align}
On $\mc{E}^c$, $N_T(\mf{x}) \leq T$. Consequently, the expected number of pulls is at most two for a retained action with $b_{\mf{x}} < \mu^\ast_{\mf{x}^\star}$, and at most $2+q_{\mf{x}}$ for any other retained suboptimal action.

For regret, charge failure once for the entire process. On $\mc{E}$, initialization contributes $\sum_{\mf{x}\in\mc{A}^\ast}\Delta_{\mf{x}}$, and only retained suboptimal actions with $b_{\mf{x}} \geq \mu^\ast_{\mf{x}^\star}$ can contribute further. Since $R_T\leq T$ and $\Pr(\mc{E}^c) \leq \frac{1}{T}$, the failure contribution is at most one. The regret decomposition therefore gives:
\begin{align}
    \mb{E}R_T
    \leq & \sum_{\mf{x}\in\mc{A}^\ast}\Delta_{\mf{x}}
    +\zeta_T\sum_{\substack{\mf{x}\in\mc{A}^\ast:\\
              b_{\mf{x}} \geq \mu^\ast_{\mf{x}^\star},\
              \Delta_{\mf{x}}>0}}
      \left(\frac{32\sigma_{\mf{x}}^2}{\Delta_{\mf{x}}}
                       +\frac{8}{3}\right)+1.
\end{align}
This is the claimed bound. Since
$u^\star\geq\mu^\ast_{\mf{x}^\star}$, the condition
$b_{\mf{x}}\geq\mu^\ast_{\mf{x}^\star}$ is equivalent to
$u_{\mf{x}}\geq\mu^\ast_{\mf{x}^\star}$.
\end{proof}

\jmlrsection{Supporting Results for Counterfactual Structural Causal Bandits}
\section{Omitted Proofs}

\subsection{Proof of Theorem~\ref{thm: ctf-mis}}\label{sec: proof of thm: ctf-mis}

We assume that $\mf{X}_{\ast\ast}$ is realizable and every assigned counterfactual is nonconstant, i.e., after effectiveness, it still requires its target mechanism. 

The proof follows the two parts of the graphical condition. If either part fails, we identify an assignment that can be removed without changing the reward. If both hold, we use the binary SCM of \Cref{lem: binary scm construction} to make every proper subset give a strictly smaller mean reward. One such SCM is enough to rule out equality in every compatible SCM. 

\begin{proof}[Proof of \Cref{thm: ctf-mis}]
    \textbf{(Only if)} We show that failure of either condition permits deleting one counterfactual from $\mf{X}_{\ast\ast}$.

    \textbf{Failure of the first condition.} Suppose the first condition fails for $X_i$. Let $\mf{X}'_{\ast\ast}$ omit the counterfactual assigned to $X_i$, keeping all other terms. Every directed path from $X_i$ to $Y$ then meets a target whose counterfactual is still assigned. Consequently, changing the value assigned to $X_i$ cannot affect the reward once the values at those other targets have been supplied.

    To apply \ctf-Rule~3 explicitly, a path starting at $X_i$ cannot use an edge entering $X_i$. The failed ancestor condition therefore gives $X_i \notin \An(Y)_{\mc{G}_{\overline{\mf{X} \setminus\{X_i\} }}}$. Therefore, we have: 
    \begin{align}
        P(Y_{\mf{X}_{\ast\ast}} = y) &= \sum_{\mf{x}} P(y_{\mf{x}}, \mf{X}_{\ast\ast} = \mf{x}) && \text{CUT} 
        \\
        & = \sum_{\mf{x}} P(y_{\mf{x} \setminus \{X_i\}}, \mf{X}_{\ast\ast} = \mf{x}) && \text{\ctf-Rule~3} 
        \\
        & = \sum_{\mf{x} \setminus \{X_i\}} P(y_{\mf{x} \setminus \{X_i\}}, \mf{X}'_{\ast\ast} = \mf{x} \setminus \{X_i\}) && \text{marg. over $X_i$} 
        \\
        &= P(Y_{\mf{X}'_{\ast\ast}} = y). && \text{CUT} 
    \end{align}
    Counterfactuals inside the remaining terms are unchanged. Only the separate assignment to $X_i$ has disappeared. Thus the deletion preserves the reward on every unit and introduces no new required evaluation. The resulting action is realizable and has the same mean reward. If no assignment remains, its reward is simply $Y$.

    \textbf{Failure of the second condition.} Suppose the first condition holds but the second fails for $X_i$. Again, let $\mf{X}'_{\ast\ast}$ remove only the counterfactual assigned to $X_i$. The question is now whether computing $X_i$ through its mechanism, with the remaining assignments in place, reproduces that counterfactual. 

    For example, consider $\{X_w,Z_w\}$ in \Cref{fig: realizable_b}. The mechanism of $Z$ still receives $X_w$, together with the same natural value of $T$ and the same exogenous realization. It therefore produces $Z_w$ without a separate assignment to $Z$.
    As explained in the main text, $Z_w = Z_{X_w}$ and thus $Y_{X_w, Z_w} = Y_{X_w}$. In this example, the second condition fails for $Z$ because the fixed value $w$ reaches its computation only through the other target $X$. 

    For the general case, compare the computations in three steps:
    \begin{itemize}
        \item[1.] Starting at $X_i$, follow parent edges backwards in $\mc{G}_{\mf{X}_{\ast\ast}}$. Stop a branch when another target in $\mf{X} \setminus \{X_i\}$ is reached. Its assigned counterfactual is still supplied under $\mf{X}'_{\ast\ast}$. 

        \item[2.] At each vertex $V$ visited before these stopping points, failure of the second condition gives $\Pa(V)_{\mc{G}_{\mf{X}_{\ast\ast}}} = \Pa(V)_{\mc{G}}$. Thus no parent value received by $f_V$ was replaced by a constant. Each such value is calculated by its parent mechanism or supplied by the counterfactual at a stopping point. At a stopping point, realizability ensures that the counterfactual needed by this calculation is the one assigned to that remaining target.

        \item[3.]  Evaluate the visited mechanisms in topological order. The values supplied at the stopping points are unchanged. Every other parent value is obtained from an earlier mechanism whose value already agrees, and the exogenous realization is the same. Hence each mechanism returns the same value, including $f_{X_i}$. 
    \end{itemize}
    Consequently, removing the assignment to $X_i$ reproduces its assigned counterfactual through the same mechanisms. \ctf-Rule~1 gives $Y_{\mf{X}_{\ast\ast}} = Y_{\mf{X}'_{\ast\ast}}$. The deletion preserves realizability because it reuses evaluations already needed by the original action. Both cases establish equality for every compatible SCM and every admissible subscript assignment, proving necessity. 

    \textbf{(If)} Suppose both conditions hold. We will use the second condition to obtain a strict difference at a removed target, and the first condition to carry that difference to $Y$. 

    \textbf{Step~1: Choose the binary SCM and set the subscript values.} We first take $\mc{M}$ from \Cref{lem: binary scm construction} and set every constant subscript value to one. This choice respects the equalities required by realizability. Recall three properties of the construction: 
    \begin{itemize}
        \item Every observed variable is binary. Increasing a parent value while keeping the exogenous realization fixed cannot decrease the child's value.

        \item Every parent has a strictly positive coefficient $\beta_{PV}$ in the expectation formula. A positive difference in a parent's mean therefore increases the child's mean when the other differences are nonnegative.

        \item Every value computed by a mechanism has mean strictly between zero and one. In particular, it does not equal one with probability one, even when all its parent values are one.
    \end{itemize}
    Fix any proper subset $\mf{X}'_{\ast\ast} \subsetneq \mf{X}_{\ast\ast}$. We compare it with the original set in this same SCM, using the same exogenous realization and leaving every remaining counterfactual unchanged. 

    \textbf{Step~2: Compare the values used in the two calculations.} First consider a target whose counterfactual remains in $\mf{X}'_{\ast\ast}$. Its assigned value is identical in the two rewards. For every other mechanism $f_V$ required by the original action, $V$ is evaluated under the subset through its structural mechanism. Compare the values received from each parent as follows:
    \begin{itemize}
        \item If $\mf{X}_{\ast\ast}$ fixes the value received from $P$, that value is one. The corresponding value under the subset is $P_{\mf{X}'_{\ast\ast}}$, which is either zero or one. 

        \item If the value received from $P$ is calculated by its mechanism, compare that calculation with $P_{\mf{X}'_{\ast\ast}}$ first. In topological order, this parent comparison has already been made before evaluating its child. 
    \end{itemize}
    Starting with the earliest mechanisms, monotonicity therefore makes every value required by the original action at least as large as its counterpart in the subset's reward calculation. In particular, for each target:
    \begin{align}
        X_{i[\mf{X}_{\ast\ast}]}(\mf{u}) \geq X_{i[\mf{X}'_{\ast\ast}]}(\mf{u}). \label{eq: proof of ctf-mis}
    \end{align}
    For the next step, the notation $\mb{E}_{\mc{M}_{\mf{X}_{\ast\ast}}} V$ denotes the mean of the value returned by the mechanism for $V$ when computing the original action. The notation $\mb{E}_{\mc{M}} V_{\mf{X}'_{\ast\ast}}$ denotes the mean of $V$ in the subset's reward calculation. These may refer to different counterfactuals. For instance, computing $Z_w$ may require $X_w$ even when the reward after a deletion also receives the natural $X$. The comparison above keeps these calculations distinct. It remains valid when the subset is not realizable, because both counterfactuals are defined in the same SCM.

    \textbf{Step~3: Find a strict difference where a parent value was fixed.} Consider a required mechanism $f_V$ for which the subset supplies no assigned counterfactual at $V$. There are two contributions to the difference of its means. A parent edge deleted from $\mc{G}_{\mf{X}_{\ast\ast}}$ contributes the difference between the fixed value one and the calculated value $P_{\mf{X}'_{\ast\ast}}$. A parent edge retained in that graph contributes the difference between the two calculated values of $P$. The expectation formula from \Cref{lem: binary scm construction} gives:
    \begin{align}
        \mb{E}_{\mc{M}_{\mf{X}_{\ast\ast}}} V &- \mb{E}_{\mc{M}} V_{\mf{X}'_{\ast\ast}} \\
        &=  \sum_{P \in \Pa(V)_{\mc{G}} \setminus \Pa(V)_{\mc{G}_{\mf{X}_{\ast\ast}}}} \beta_{PV} (1 - \mb{E}_{\mc{M}} P_{\mf{X}'_{\ast\ast}}) \label{eq1: proof ctf-mis}
        \\
        &+ 
        \sum_{P \in \Pa(V)_{\mc{G}_{\mf{X}_{\ast\ast}}}} \beta_{PV} (\mb{E}_{\mc{M}_{\mf{X}_{\ast\ast}}} P - \mb{E}_{\mc{M}} P_{\mf{X}'_{\ast\ast}}). \label{eq2: proof ctf-mis}
    \end{align}
    The first sum is over the parent values replaced by one; the second is over the parent values still computed by their mechanisms. The constant and shared latent terms cancel because the structural function and exogenous distribution are unchanged. Every summand is nonnegative by Step~2. 

    Moreover, every $P_{\mf{X}'_{\ast\ast}}$ has mean strictly below one. If $P$ is a remaining target, its counterfactual still requires its mechanism by assumption. Otherwise, its mechanism is evaluated with the values produced under the subset. In both cases, the threshold of the binary construction stays strictly between zero and one. Therefore, whenever $f_V$ receives a fixed parent value in the original action, the corresponding summand in the first sum is strictly positive.

    The expectation calculation remains valid for nested counterfactuals. The private noise $U_V$ is independent of all values supplied by parent mechanisms, since these use only ancestral mechanisms and their exogenous variables. Parents may be correlated with shared latent variables; the calculation uses independence only of $U_V$. 

    \textbf{Step~4: Use the second condition to reach a removed target.} Choose any target $X_i$ whose counterfactual was removed. The second condition provides a vertex $V$ that receives a fixed parent value and a retained directed path from $V$ to $X_i$. This path, including $V$, avoids every other target in $\mf{X}$, so none of its mechanisms is replaced by a remaining assignment.

    Step~3 gives a strictly positive difference at $V$. At each next vertex on the path, the contribution of the preceding parent appears in \Cref{eq2: proof ctf-mis}, multiplied by a strictly positive coefficient. All other contributions are nonnegative. The difference therefore stays strictly positive until it reaches $X_i$. If $V = X_i$, no propagation is needed. 

    \textbf{Step~5: Use the first condition to reach the reward.} The first condition provides a directed path from $X_i$ to $Y$ that avoids every other target in $\mf{X}$. After $X_i$, these mechanisms are evaluated in both rewards with their original structural mechanisms. The strict difference at $X_i$ therefore propagates along this path to $Y$ by the same positive coefficient calculation. The other parent differences are nonnegative by \Cref{eq: proof of ctf-mis} and monotonicity. Hence $\mu_{\mf{X}_{\ast\ast}} > \mu_{\mf{X}'_{\ast\ast}}$. 

    The proper subset was arbitrary: it may remove several assignments, and the resulting counterfactual need not be realizable. The same SCM and the same subscript values, all equal to one, distinguish the original set from every such subset. Thus no proper subset preserves its mean reward for every compatible SCM and every admissible assignment. This proves that $\mf{X}_{\ast\ast}$ is a \ctf-MIS. 
\end{proof}

\subsection{Proof of Proposition~\ref{prop: ctf-pomis unconfounded}}\label{sec: proof of prop: ctf-pomis unconfounded}
\begin{proof}[Proof of \Cref{prop: ctf-pomis unconfounded}]
    Fix an assigned counterfactual $\mf{X}_{\ast\ast}$ and condition on the values received by the parent inputs of $f_Y$. These values are determined by upstream exogenous variables and fixed constants, independently of the exogenous inputs of $Y$. The conditional reward mean is therefore $\mb{E}Y_{\mf{pa}_Y}$. Averaging over the received values gives:
    \begin{align}
        \mu_{\mf{X}_{\ast\ast}} \leq \max_{\mf{pa}_Y} \mb{E} Y_{\mf{pa}_Y} = \mu_{\mf{pa}_Y^\ast}. 
    \end{align}
    Consequently, no non-equivalent counterfactual can be strictly better than the best parent intervention. To establish that the parent intervention set itself is a \ctf-POMIS, we also need one SCM in which the comparison is strict against every non-equivalent counterfactual.

    Choose binary domains and independent private noises $U_V \sim \Bern(\epsilon)$ for $V \neq Y$ and $U_Y \sim \Bern(1-\epsilon)$, where $0 < \epsilon < 1$. Set $f_V  = U_V \prod_{P \in \Pa(V)_{\mc{G}}}P$ for $V \in \mf{V}$, where an empty product equals one. Fixing every parent input of $Y$ to one gives mean $1 - \epsilon$. Any non-equivalent counterfactual leaves some input $P \to Y$ natural: if all these inputs are fixed, every upstream assignment is irrelevant to the reward. The natural output supplied by $P$ is at most $U_P$, whatever its other inputs. Hence every choice of subscript values has mean at most $\epsilon(1-\epsilon) < 1 - \epsilon$. This single SCM establishes strictness against all non-equivalent counterfactuals. When $\Pa(Y)_{\mc{G}} = \emptyset$, every input assignment is irrelevant and every counterfactual is equivalent to the empty action. 
\end{proof}

\subsection{Proof of Theorem~\ref{thm: ctf-pomis}}\label{sec: proof of thm: ctf-pomis}

The two proof directions of \Cref{thm: ctf-pomis} use different properties of the minimal UC-territory. When its border is empty, we construct an SCM in which the given counterfactual preserves useful dependence between the reward and its ancestors. Replacing a retained input by a constant breaks this dependence. When the border is nonempty, the inputs entering the territory are independent of its exogenous variables. Their random values can then be replaced by the best fixed values. 

\paragraph{An SCM construction.} For each shared binary exogenous variable, {\blue} and {\red} vertices carry its value along the two sides of the corresponding bidirected edge. {\Purple} vertices compare these values after the directed paths meet. The mechanisms retain a separate coordinate for each confounder, so all these comparisons take place in one SCM. Thus the colors specify the structural functions used in the proof. 

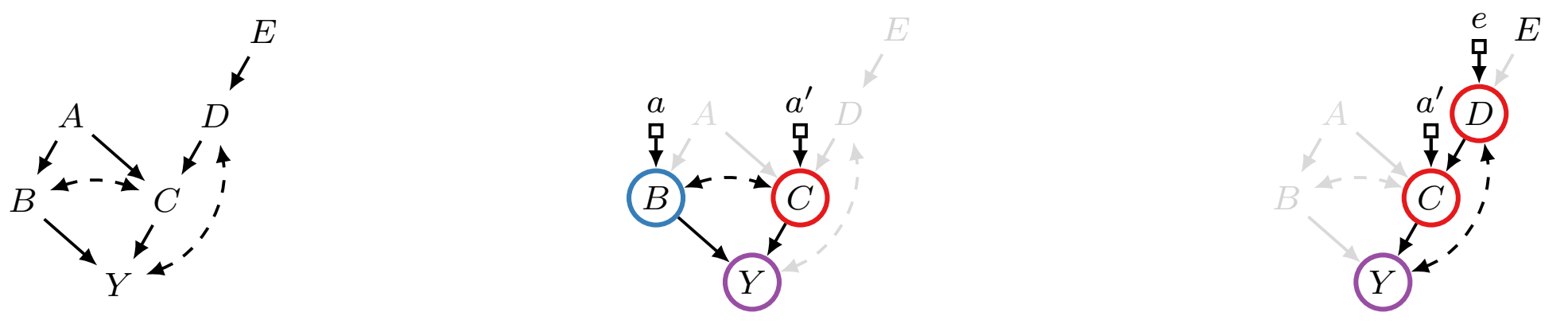
\begin{figure}[t]
    \centering
    \tikzset{
        ctfproofgraph/.style={x=0.8cm,y=0.7cm,font=\scriptsize,
            >={Latex[width=1.3mm,length=1.5mm]},line width=0.25mm},
        ctfproofnode/.style={draw=none,circle,inner sep=0mm,minimum size=4.5mm},
        ctfproofblue/.style={draw=betterblue,line width=0.4mm},
        ctfproofred/.style={draw=betterred,line width=0.4mm},
        ctfproofpurple/.style={draw=betterpurple,line width=0.4mm},
        ctfproofinactive/.style={text=gray!35},
        ctfprooffixed/.style={->,dashed,draw=gray!60}.
    }
    \begin{minipage}[b]{0.31\textwidth}\centering
    \subfloat[$\mc{G}$]{
        \begin{tikzpicture}[ctfproofgraph]
            \node[ctfproofnode] (A) at (0.5,1) {$A$};
            \node[ctfproofnode] (B) at (0,0) {$B$};
            \node[ctfproofnode] (Y) at (1,-1) {$Y$};
            \node[ctfproofnode] (C) at (1.5,0) {$C$};
            \node[ctfproofnode] (D) at (2,1) {$D$};
            \node[ctfproofnode] (E) at (2.5,2) {$E$};
            \draw[->] (A) -- (B);
            \draw[->] (B) -- (Y);
            \draw[->] (A) -- (C);
            \draw[->] (C) -- (Y);
            \draw[->] (E) -- (D);
            \draw[->] (D) -- (C);
            \draw[<->,dashed] (B) to[bend left=20] (C);
            \draw[<->,dashed] (Y) to[bend right=40] (D);
        \end{tikzpicture}\label{fig:ctfscb pomis proof_a}}
    \end{minipage}
    \hfill
    \begin{minipage}[b]{0.31\textwidth}\centering
    \subfloat[$\mc{H}_1$]{
        \begin{tikzpicture}[ctfproofgraph, 
        transp/.style={rectangle,fill=white,minimum size=1mm,draw=black,inner sep=0mm},
        regime line/.style={draw=black,  thick}]
            \node[ctfproofnode, gray!30] (A) at (0.5,1) {$A$};
            \node[ctfproofnode,ctfproofblue] (B) at (0,0) {$B$};
            \node[ctfproofnode,ctfproofpurple] (Y) at (1,-1) {$Y$};
            \node[ctfproofnode,ctfproofred] (C) at (1.5,0) {$C$};
            \node[ctfproofnode,ctfproofinactive] (D) at (2,1) {$D$};
            \node[ctfproofnode,ctfproofinactive] (E) at (2.5,2) {$E$};
            \draw[->,draw=gray!30] (A) -- (B);
            \draw[->] (B) -- (Y);
            \draw[->,draw=gray!30] (A) -- (C);
            \draw[->] (C) -- (Y);
            \draw[->,draw=gray!30] (E) -- (D);
            \draw[->,draw=gray!30] (D) -- (C);
            \draw[<->,dashed] (B) to[bend left=20] (C);
            \draw[<->,dashed,draw=gray!30] (Y) to[bend right=40] (D);

            \node[transp, above = 2.5mm of B] (BT){};
            \node[above = -0.5mm of BT] (wd){\color{black} $a$};
            \draw[->,regime line] (BT) -- (B);

            \node[transp, above = 2.5mm of C] (CT){};
            \node[above = -0.5mm of CT] (wd){\color{black} $a'$};
            \draw[->,regime line] (CT) -- (C);
        \end{tikzpicture}\label{fig:ctfscb pomis proof_b}}
    \end{minipage}
    \hfill
    \begin{minipage}[b]{0.31\textwidth}\centering
    \subfloat[$\mc{H}_2$]{
        \begin{tikzpicture}[ctfproofgraph, 
        transp/.style={rectangle,fill=white,minimum size=1mm,draw=black,inner sep=0mm},
        regime line/.style={draw=black,  thick}]
            \node[ctfproofnode, gray!30] (A) at (0.5,1) {$A$};
            \node[ctfproofnode,ctfproofinactive] (B) at (0,0) {$B$};
            \node[ctfproofnode,ctfproofpurple] (Y) at (1,-1) {$Y$};
            \node[ctfproofnode,ctfproofred] (C) at (1.5,0) {$C$};
            \node[ctfproofnode,ctfproofred] (D) at (2,1) {$D$};
            \node[ctfproofnode] (E) at (2.5,2) {$E$};
            \draw[->,draw=gray!30] (A) -- (B);
            \draw[->,draw=gray!30] (B) -- (Y);
            \draw[->,draw=gray!30] (A) -- (C);
            \draw[->] (C) -- (Y);
            \draw[->,draw=gray!30] (E) -- (D);
            \draw[->] (D) -- (C);
            \draw[<->,dashed,draw=gray!30] (B) to[bend left=20] (C);
            \draw[<->,dashed] (Y) to[bend right=40] (D);

            \node[transp, above = 2.5mm of C] (CT){};
            \node[above = -0.5mm of CT] (wd){\color{black} $a'$};
            \draw[->,regime line] (CT) -- (C);

            \node[transp, above = 2.5mm of D] (DT){};
            \node[above = -0.5mm of DT] (wd){\color{black} $e$};
            \draw[->,regime line] (DT) -- (D);
        \end{tikzpicture}\label{fig:ctfscb pomis proof_c}}
    \end{minipage}
    \caption{\small The SCM construction for \ctf-POMIS. (a) The original causal diagram. (b) $B$ carries the {\blue} value $u_1$, $C$ carries the {\red} value $1 - u_1$, and the {\purple} vertex $Y$ checks the two values. (c) $D$ and $C$ carry the {\red} value $1-u_2$; $Y$ compares it with its {\blue} exogenous input $u_2$. }
    \label{fig:ctfscb pomis proof}
\end{figure}
\begin{table}[t]
    \centering
    \small
    \setlength{\tabcolsep}{3pt}
    \begin{tabular}{cc ccc ccc ccccc}
        \toprule
        \multicolumn{2}{c}{} & \multicolumn{3}{c}{$\mc{M}_1$}
        & \multicolumn{3}{c}{$\mc{M}_2$} & \multicolumn{5}{c}{$\mc{M}$} \\
        \cmidrule(lr){3-5}\cmidrule(lr){6-8}\cmidrule(lr){9-13}
        $U_1$ & $U_2$
        & {\color{betterblue}$B^{(1)}$} & {\color{betterred}$C^{(1)}$}
        & {\color{betterpurple}$Y^{(1)}$}
        & {\color{betterred}$D^{(2)}$} & {\color{betterred}$C^{(2)}$}
        & {\color{betterpurple}$Y^{(2)}$}
        & {\color{betterblue}$B$} & {\color{betterred}$D$}
        & {\color{betterred}$C$} & {\color{betterpurple}$Y'$} & $Y$ \\
        \midrule
        0 & 0 & 0 & 1 & 2 & 1 & 1 & 2
        & $00\,00$ & $00\,01$ & $01\,01$ & $10\,10$ & $1$ \\
        0 & 1 & 0 & 1 & 2 & 0 & 0 & 1
        & $00\,00$ & $00\,00$ & $01\,00$ & $10\,01$ & $1$ \\
        1 & 0 & 1 & 0 & 1 & 1 & 1 & 2
        & $01\,00$ & $00\,01$ & $00\,01$ & $01\,10$ & $1$ \\
        1 & 1 & 1 & 0 & 1 & 0 & 0 & 1
        & $01\,00$ & $00\,00$ & $00\,00$ & $01\,01$ & $1$ \\
        \bottomrule
    \end{tabular}
    \caption{\small Values under $\{B_a,C_{a',e}\}$ in \Cref{fig:ctfscb pomis proof}, with $a\neq a'$. Each coordinate is written using two bits, with the first component
    followed by the second; an inactive coordinate is $00$. For example, $C=01\,01$ represents $(C^{(1)},C^{(2)})=(1,1)$. The intermediate value $Y'$ records $(Y^{(1)},Y^{(2)})$. Both coordinates are $1$ or $2$ in every row, so the binary reward is $Y=1$ on every unit.} 
    \label{table:ctfscb pomis construction}
\end{table}

Consider \Cref{fig:ctfscb pomis proof_a}. Fix the inputs $A \to B$, $A \to C$, and $E \to D$ to $a$, $a'$, and $e$, respectively. Choose $a \neq a'$ and reserve these assigned values separately from the values naturally output by the mechanisms. Deleting the three selected edges and restricting to the ancestors of $Y$ leaves $\{B,C,D,Y\}$. Starting from $Y$, the bidirected edge $Y \bedge D$ includes $D$, descendant closure includes $C$, and $C \bedge B$ includes $B$. The retained graph is therefore its own minimal UC-territory, with empty border. 

Let $U_1$ be the fair bit shared by $B$ and $C$, and let $U_2$ be the independent fair bit shared by $D$ and $Y$. In \Cref{fig:ctfscb pomis proof_b}, the {\blue} vertex $B$ carries $U_1$ and the {\red} vertex $C$ carries $1 - U_1$. The {\purple} vertex $Y$ checks that these two values are complementary, recording $2$ for $(0,1)$ and $1$ for $(1,0)$. In \Cref{fig:ctfscb pomis proof_c}, the {\red} vertices $D$ and $C$ carry $1 - U_2$. At $Y$, this value is checked against the genuine $U_2$ available to its own mechanism. The same {\purple} encoding records $2 - U_2$. \Cref{table:ctfscb pomis construction} shows the calculations for $U_1$ and $U_2$ separately as $\mc{M}_1$ and $\mc{M}_2$, then combines them into $\mc{M}$. The two coordinates of $B$, $D$, and $C$ are $(U_1,0)$, $(0, 1 - U_2)$, and $(1 - U_1, 1 - U_2)$, respectively; zero fills a coordinate unused at that vertex. The intermediate value $Y'$ contains the two comparison results, and $Y = 1$ exactly when both are valid. Before making these calculations, the mechanisms of $B$, $C$, and $D$ check their selected inputs against $a$, $a'$, and $e$. Under the specified counterfactual, all checks succeed for every $(u_1, u_2)$, giving reward one. 

This example also explains why fixing a retained input can be harmful. If $B \to Y$ is fixed while $C \to Y$ remains natural, its first coordinate must agree with a fair bit carried through $C$. If $D \to C$ is fixed while $C \to Y$ remains natural, its second coordinate must agree with the genuine $U_2$ at $Y$. If $C \to Y$ is fixed, its second coordinate must again agree with that $U_2$, even when the other inputs are fixed as well. Each comparison succeeds with probability at most $\frac{1}{2}$. If all retained inputs remain natural, omitting a selected input fails its assigned value check. The following construction applies these comparisons to arbitrary retained paths, including paths that merge. 

\begin{proof}[Proof of \Cref{thm: ctf-pomis}]
    \textbf{(If)} Suppose $\IB(\mc{H}_{\mf{X}_{\ast\ast}},Y) = \emptyset$. Every vertex of the graph is ancestral to $Y$. If a vertex lay outside the minimal UC-territory, a directed path from it to $Y$ would have a first edge entering the territory. The parent of this edge would belong to the border, a contradiction. Therefore:
    \begin{align}
       \mf{V}^\dagger = \mf{V}(\mc{H}_{\mf{X}_{\ast\ast}}) = \MUCT(\mc{H}_{\mf{X}_{\ast\ast}}, Y).  
    \end{align}
    Selected inputs whose children lie outside $\mf{V}^\dagger$ cannot affect $Y$ and may be discarded. We compare the remaining selections of fixed inputs, allowing each selected value to vary freely and identifying equivalent counterfactual sets. 

    \textbf{Choosing the assigned values.} Choose the subscript values of $\mf{X}_{\ast\ast}$ so that distinct selected occurrences of the same parent variable receive distinct values. Reserve separate values for the coordinate tuples used below and for an invalid output. 

    Each mechanism checks every input selected by $\mf{X}_{\ast\ast}$ against its chosen value. No natural output will take one of these reserved assigned values. Consequently, a competing counterfactual cannot pass a selected-input check merely by leaving that input natural. 

    \paragraph{Constructing the mechanisms for one confounder.} Let $U_1, \cdots, U_k$ be independent fair binary confounders, one for each bidirected edge of $\mc{H}_{\mf{X}_{\ast\ast}}$. Denote the endpoints of $U_j$ by $B^{(j)}$ and $R^{(j)}$. Descendants in the following coloring are computed in the retained graph:
    \begin{align}
        \blue: \quad& \De(B^{(j)}) \setminus \De(R^{(j)}),\\
        \red: \quad& \De(R^{(j)}) \setminus \De(B^{(j)}),\\
        \purple: \quad& \De(B^{(j)}) \cap \De(R^{(j)}).
    \end{align}
    The {\blue} endpoint supplies $u_j$ as an additional {\blue} input to its mechanism, and the {\red} endpoint supplies $1-u_j$ as an additional {\red} input. This convention also applies when an endpoint is itself {\purple} because it is a descendant of the other endpoint.

    At a {\blue} or {\red} vertex, the mechanism returns the common value of its inputs of that color, provided they agree and belong to $\{0,1\}$. At a {\purple} vertex, the valid input combinations and outputs are: 
    \begin{align}
    \begin{array}{ccc|c}
        \blue & \red & \purple &\textbf{output}\\
        0&1&2&2\\
        1&0&1&1
    \end{array}
    \label{eq: ctf-pomis colored mechanisms}
    \end{align}
    Each entry specifies the common value required of all inputs of that color; a color with no input imposes no condition. An unequal or malformed input, a failed selected-input check, or an invalid parent output gives the invalid value $3$. Invalid outputs propagate along every retained directed edge. A vertex outside the two descendant sets has an unused zero coordinate.
    
    These rules have a simple interpretation. {\Blue} values encode $u_j$, {\red} values encode $1-u_j$, and {\purple} values encode $2-u_j$. Thus every mechanism compares copies of the same bit, with the encoding determined by the sender's color. For example, a {\purple} input equal to $2$ is consistent with a {\blue} input equal to $0$ and a {\red} input equal to $1$. Keeping the {\purple} value allows subsequent mechanisms to make the same comparison after paths have already merged.
    
    \paragraph{Combining the mechanisms into one SCM.}
    Represent each variable in $\mf{V}^\dagger\setminus\{Y\}$ by the $k$ coordinates just constructed, using two bits per coordinate as in \Cref{table:ctfscb pomis construction}. Its mechanism applies all selected-input checks and all coordinate comparisons. A coordinate tuple with an incorrect number of coordinates or a nonzero unused coordinate is invalid. Any failure makes the whole output invalid. The mechanism of $Y$ makes the corresponding comparisons, including any exogenous bits it receives directly, and returns one exactly when every check succeeds. Natural mechanisms outside $\mf{V}^\dagger$ also avoid the reserved assigned values. Other allowed exogenous inputs may be ignored.
    
    Under the chosen subscript values, every coordinate carries the genuine shared bit in its designated encoding, and every selected-input check succeeds. Hence:
    \begin{align}
        \mu_{\mf{X}_{\ast\ast}} = \mu_{\mf{X}_{\ast\ast}^\ast} = 1.
    \end{align}
    If $\mf{V}^\dagger=\{Y\}$, there are no colored coordinates: checking the selected parent inputs of $Y$ is sufficient. In either case, this fixes one SCM before any competing counterfactual is considered.
    
    \paragraph{Comparing every competing counterfactual.}
    Consider an assigned counterfactual $\mf{Z}_{\ast\ast}$ with mean greater than $\frac{1}{2}$. We first show that it fixes no input retained in $\mc{H}_{\mf{X}_{\ast\ast}}$.
    
    Start at $Y$, whose length-zero path to itself is unchanged. Suppose every directed path from a vertex to $Y$ is unchanged, and that vertex shares $U_j$ with another endpoint. On every accepted outcome, its genuine copy of $U_j$ reaches $Y$. If a path from the other endpoint to $Y$ contains an input fixed by $\mf{Z}_{\ast\ast}$, take the last such input on that path. This input supplies a constant coordinate tuple. Its $j$th coordinate must propagate along the unchanged remainder of the path and agree with the genuine copy, either at an earlier merge or at $Y$. A malformed coordinate is rejected. Otherwise, the constant can agree with the fair bit in only one of its two states, so the competing reward mean is at most $\frac{1}{2}$, a contradiction. 
    
    Every directed path from the other endpoint to $Y$ must therefore be unchanged. The same conclusion holds for each of its descendants: any fixed input on a descendant's path would also lie on a path from that endpoint. Repeatedly applying these bidirected and descendant steps reaches precisely the minimal UC-territory. Since that territory is $\mf{V}^\dagger$, all its retained inputs must remain natural under an assignment with mean greater than $\frac{1}{2}$.
    
    Every input selected by $\mf{X}_{\ast\ast}$ must also retain its designated value. Omitting its assignment supplies a natural output, which never takes that value and therefore fails its check. Since all retained paths are unchanged, the resulting invalid output reaches $Y$. If the competing counterfactual keeps every selected input, any further selections outside $\mf{V}^\dagger$ are irrelevant: their paths to $Y$ already pass through fixed inputs. Removing them leaves the same selected inputs with the same unrestricted choices of values. The two counterfactuals are then equivalent. We have proved, in one SCM, that:
    \begin{align}
        \mu_{\mf{Z}_{\ast\ast}^\ast}\leq\tfrac12
        <1=\mu_{\mf{X}_{\ast\ast}^\ast}
    \end{align}
    for every non-equivalent counterfactual. This establishes possible optimality. 
    
    \textbf{(Only if)} Suppose $\mf{T} = \IB(\mc{H}_{\mf{X}_{\ast\ast}},Y) \neq \emptyset$, and fix the best assignment $\mf{X}_{\ast\ast}^\ast$ in an arbitrary compatible SCM. The minimal UC-territory is closed under descendants and bidirected connections. Its exogenous inputs are therefore independent of those generating $\mf{T}$. Indeed, if an ancestor of a border vertex shared an exogenous input with the territory, bidirected closure would include that ancestor, and descendant closure would then include the border vertex. This contradicts the definition of the border. All parents entering the territory from outside belong to $\mf{T}$.

    Construct $\mf{X}'_{\ast\ast}$ by retaining the original input assignments and fixing every remaining input from $\mf{T}$ into the territory. For any value $\mf{t}$, supply its component for $T$ to all newly selected inputs from that border variable, while retaining the original optimizing values. The resulting reward has the same mean as the original reward conditional on $\mf{T}=\mf{t}$: the inputs received by the territory are the same, and conditioning does not change its exogenous distribution.
    
    The border values here are evaluated in the modified SCM $\mc{M}_{\mf{X}_{\ast\ast}^{\ast}}$. In the summand below, $\mf{X}'_{\ast\ast}$ denotes the assignment corresponding to $\mf{t}$. Conditioning gives:
    \begin{align}
        \mu_{\mf{X}_{\ast\ast}^\ast}
        &=\sum_{\mf{t}}\mb{E}_{\mc{M}_{\mf{X}_{\ast\ast}^{\ast}}}
          [Y\mid\mf{T}=\mf{t}]
          P_{\mc{M}_{\mf{X}_{\ast\ast}^{\ast}}}(\mf{T}=\mf{t})\\
        &=\sum_{\mf{t}}\mu_{\mf{X}'_{\ast\ast}}
          P_{\mc{M}_{\mf{X}_{\ast\ast}^{\ast}}}(\mf{T}=\mf{t})
          \leq\mu_{{\mf{X}'_{\ast\ast}}^\ast}.
        \label{eq: ctf-pomis border mixture}
    \end{align}
    The last optimization allows independent values at the newly selected inputs, including the common-value assignments appearing in the sum. The best such assignment is at least as good as their average. Thus $\mf{X}'_{\ast\ast}$ dominates $\mf{X}_{\ast\ast}$ in every compatible SCM. 

    The two counterfactuals are non-equivalent. Choose $T \in \mf{T}$ and a directed path from $T$ into the territory and onward to $Y$. Let $T$ be an independent fair bit and let the mechanisms along this path copy it, ignoring their other inputs. Every assignment of $\mf{X}_{\ast\ast}$ then has mean $\frac{1}{2}$, whereas fixing the new input from $T$ to one gives mean one. Finally, discard selected inputs in $\mf{X}'_{\ast\ast}$ whose children are no longer ancestral to $Y$. This preserves the reward and leaves the corresponding counterfactual action. The original counterfactual therefore cannot be strictly better than every non-equivalent counterfactual in any compatible SCM, so it is not a \ctf-POMIS. 
\end{proof}

\subsection{Proof of Corollary~\ref{cor: ctf-pomis action space preserves optimal reward}}\label{sec: proof of cor: ctf-pomis action space preserves optimal reward}
\begin{proof}[Proof of \Cref{cor: ctf-pomis action space preserves optimal reward}]
    For a rejected counterfactual set, fix the remaining inputs from its interventional border into its minimal UC-territory, retaining the originally selected inputs and their values. The argument in \Cref{eq: ctf-pomis border mixture} shows that optimizing the new inputs weakly improves the best mean. The remaining graph is the subgraph induced by that territory and has empty border, so the algorithm returns the resulting counterfactual set. Every omitted action is thus weakly dominated by an action in $\mc{A}_{\ctf}^{\star}$; the reverse inequality follows from $\mc{A}_{\ctf}^{\star} \subseteq \mc{A}_{\ctf}$. 
\end{proof}

\section{Experimental Structural Causal Models}\label{sec: ctfscb experimental SCMs}

This section specifies the fixed SCMs used in \Cref{sec: ctfscb experiment}. All observed and exogenous variables are binary. Distinct exogenous variables are mutually independent. A variable $U_V$ is specific to the mechanism of $V$, while a shared variable $U_{VW}$ appears in the two mechanisms connected by the corresponding bidirected edge. The symbols $\land$, $\lor$, $\oplus$, and $\neg$ denote AND, OR, exclusive OR, and negation. Parentheses specify the order of the logical operations.

\paragraph{Task 1.}
The graph is shown in \Cref{fig:ctfscb counterfactual improvement_a}. The structural functions are:
\begingroup
\begin{align}
\begin{aligned}
 f_W &= u_W & f_X &= z\lor u_X\\
 f_Z &= u_Z\lor(w\oplus u_{ZY}) &
 f_Y &= \neg\bigl(u_Y\oplus x\oplus(\neg w)\oplus u_{ZY}\bigr).
\end{aligned}
\end{align}
\endgroup
Each of $U_W,U_X,U_Z,U_Y,U_{ZY}$ follows $\Bern(0.1)$. The action supplying $w=0$ to $f_Y$ and $w=1$ to $f_Z$ has mean approximately $0.885$, whereas the optimal ordinary intervention has mean $0.82$.

\paragraph{Task 2.}
The graph is shown in \Cref{fig: realizable_b}. The structural functions are:
\begingroup
\begin{align}
\begin{aligned}
 f_T &= u_T \qquad f_W=u_W\\
 f_X &= (\neg u_{XY})\oplus(u_X\lor w)\\
 f_Z &= \bigl(\neg(t\oplus(\neg u_Z))\lor x\bigr)\oplus u_{ZY}\\
 f_Y &= \neg\bigl(u_{ZY}\lor(\neg u_{XY})\bigr)\oplus u_Y\oplus(x\lor z).
\end{aligned}
\end{align}
\endgroup
The exogenous distributions are:
\begingroup
\begin{align}
\begin{aligned}
 U_T&\sim\Bern(0.65) & U_W&\sim\Bern(0.65) & U_X&\sim\Bern(0.71)\\
 U_Y&\sim\Bern(0.73) & U_Z&\sim\Bern(0.43) & U_{XY}&\sim\Bern(0.76)\\
 U_{ZY}&\sim\Bern(0.43).
\end{aligned}
\end{align}
\endgroup
Exact summation gives optimal counterfactual and ordinary means approximately $0.676$ and $0.58$, respectively.

\paragraph{Task 3(a).}
The graph is shown in \Cref{fig:ctfscb-experiment-graph-task3}. The structural functions are:
\begingroup
\begin{align}
\begin{aligned}
 f_A &= u_A \qquad f_E=u_E \qquad f_F=u_F\\
 f_B &= \neg\bigl((\neg a)\land u_B\bigr)\oplus u_{BC}\oplus u_{BY}\oplus f\\
 f_C &= \neg\bigl((\neg u_C)\oplus(\neg a)\bigr)\land(d\oplus u_{BC})\\
 f_D &= (e\oplus u_{DY})\land u_D\\
 f_Y &= \neg\bigl((\neg b)\land u_Y\bigr)\oplus u_{DY}\oplus u_{BY}\oplus c.
\end{aligned}
\end{align}
\endgroup
The exogenous distributions are:
\begingroup
\begin{align}
\begin{aligned}
 U_A&\sim\Bern(0.2) & U_B&\sim\Bern(0.3) & U_C&\sim\Bern(0.44)\\
 U_D&\sim\Bern(0.52) & U_E&\sim\Bern(0.55) & U_F&\sim\Bern(0.24)\\
 U_Y&\sim\Bern(0.67) & U_{BY}&\sim\Bern(0.74) & U_{DY}&\sim\Bern(0.35)\\
 U_{BC}&\sim\Bern(0.37).
\end{aligned}
\end{align}
\endgroup
Here $f_Y$ shares $U_{BY}$ with $f_B$, as required by the bidirected edge $B\leftrightarrow Y$; $U_{BC}$ is shared only by $f_B$ and $f_C$. The optimal counterfactual and ordinary means are approximately $0.618$ and $0.593$, respectively.

\paragraph{Task 3(b).}
The graph is unchanged, but the structural functions are:
\begingroup
\begin{align}
\begin{aligned}
 f_A &= u_A \qquad f_E=u_E \qquad f_F=u_F\\
 f_B &= (\neg u_{BY})\lor\bigl(\neg(u_{BC}\lor u_B)\land f\land a\bigr)\\
 f_C &= (\neg u_{BC})\lor\bigl((\neg a)\land(\neg u_C)\land d\bigr)\\
 f_D &= \bigl(u_D\lor(\neg u_{DY})\bigr)\land e\\
 f_Y &= (\neg u_{BY})\lor\bigl(\neg(u_{DY}\lor u_Y)\land c\land b\bigr).
\end{aligned}
\end{align}
\endgroup
The exogenous distributions are:
\begingroup
\begin{align}
\begin{aligned}
 U_A&\sim\Bern(0.34) & U_B&\sim\Bern(0.52) & U_C&\sim\Bern(0.75)\\
 U_D&\sim\Bern(0.47) & U_E&\sim\Bern(0.46) & U_F&\sim\Bern(0.76)\\
 U_Y&\sim\Bern(0.67) & U_{BY}&\sim\Bern(0.63) & U_{DY}&\sim\Bern(0.68)\\
 U_{BC}&\sim\Bern(0.26).
\end{aligned}
\end{align}
\endgroup
Assigning $b=c=1$ maximizes $f_Y$ for every exogenous realization, giving:
\begingroup
\begin{align}
\mu_{b=1,c=1}
 = 1 - 0.63 + 0.63 (1-0.68) (1-0.67)
 \approx 0.437.
\end{align}
\endgroup
This ordinary intervention attains the counterfactual optimum as well.

\clearpage
\bibliography{references}
\end{document}